\documentclass[acmtog]{acmart}
\pdfoutput=1
\usepackage{xr}

\usepackage{color}

\usepackage[export]{adjustbox}
\usepackage{array}
\usepackage{arydshln}
\usepackage{booktabs}
\usepackage{colortbl}
\usepackage{float,wrapfig}
\usepackage{hhline}
\usepackage{multirow}
\usepackage{tabularx}
\usepackage{newfloat}

\usepackage{bm}
\usepackage{nicefrac}
\usepackage{microtype}
\usepackage{xfrac}

\usepackage{changepage}
\usepackage{extramarks}
\usepackage{fancyhdr}
\usepackage{setspace}
\usepackage{soul}
\usepackage{xspace}

\usepackage{etoolbox,siunitx}
\usepackage{afterpage}
\usepackage{subcaption}

\usepackage{calc}
\usepackage[title]{appendix}

\usepackage{extarrows}

\usepackage{lipsum}  
\usepackage{pifont}
\usepackage{enumitem}
\usepackage[ruled]{algorithm2e} %

\usepackage{physics}
\usepackage{makecell}
\usepackage{cleveref}

\newcommand{\figref}[1]{\figurename~\ref{#1}}
\newcommand{\secref}[1]{Section~\ref{#1}}
\newcommand{\subsecref}[1]{Subsection~\ref{#1}}
\newcommand{\eqnref}[1]{Eqn.~\eqref{#1}}
\newcommand{\tblref}[1]{Table~\ref{#1}}

\newcommand{\ignorethis}[1]{}

\DeclareMathOperator*{\argmin}{arg\,min}

\makeatletter
\DeclareRobustCommand\onedot{\futurelet\@let@token\@onedot}
\def\@onedot{\ifx\@let@token.\else.\null\fi\xspace}

\def\eg{\emph{e.g}\onedot} 
\def\ie{\emph{i.e}\onedot} \def\Ie{\emph{I.e}\onedot}

\def\etal{et~al\onedot}
\makeatother

\newcommand*{\rom}[1]{\expandafter\romannumeral #1}

\definecolor{mydarkblue}{rgb}{0,0.08,1}
\definecolor{mydarkgreen}{rgb}{0.02,0.6,0.02}
\definecolor{mydarkred}{rgb}{0.8,0.02,0.02}
\definecolor{mydarkorange}{rgb}{0.40,0.2,0.02}
\definecolor{mypurple}{RGB}{111,0,255}
\definecolor{myred}{rgb}{1.0,0.0,0.0}
\definecolor{mygold}{rgb}{0.75,0.6,0.12}
\definecolor{myblue}{rgb}{0,0.2,0.8}
\definecolor{mydarkgray}{rgb}{0.66,0.66,0.66}

\newif\ifcolor
\colortrue

\newif\ifdraft
\draftfalse

\ifdraft
    \newcommand{\kac}[1]{{\color{magenta}\textbf{Kfir:} #1}}
    \newcommand{\ync}[1]{{\color{blue}\textbf{Yotam:} #1}}
    \newcommand{\dcc}[1]{{\color{red}\textbf{Danny:} #1}}
    \newcommand{\ygc}[1]{{\color{cyan}\textbf{Yossi:} #1}}
    \newcommand{\imc}[1]{{\color{green}\textbf{Inbar:} #1}}
    \newcommand{\qhc}[1]{{\color{teal}\textbf{Charles:} #1}}
    \newcommand{\myc}[1]{{\color{teal}\textbf{Michal:} #1}}
    \newcommand{\olc}[1]{{\color{violet}\textbf{Orly:} #1}}
    \newcommand{\ypc}[1]{{\color{red}\textbf{Yael:} #1}}

    \newcommand{\yn}[1]{{\color{blue}#1}}

    \newcommand{\nuke}[1]{{\color{red}#1}} %
    \newcommand{\move}[1]{{\color{orange}#1}} %
    
\else
    \newcommand{\kac}[1]{}
    \newcommand{\ync}[1]{}
    \newcommand{\dcc}[1]{}
    \newcommand{\ygc}[1]{}
    \newcommand{\imc}[1]{}
    \newcommand{\qhc}[1]{}
    \newcommand{\olc}[1]{}
    \newcommand{\ypc}[1]{}
    \newcommand{\myc}[1]{}
    \newcommand{\nuke}[1]{} %
    \newcommand{\move}[1]{} %

    \newcommand{\yn}[1]{{\color{black}#1}}

\fi

\newif\ifcamera
\camerafalse

\ifcamera
    \newcommand{\camera}[1]{#1}
\else
    \newcommand{\camera}[1]{}
\fi

\newcommand{\vect}[1]{\boldsymbol{\mathbf{#1}}}

\newcommand{\w}{$\mathcal{W}$\xspace}
\newcommand{\wplus}{$\mathcal{W}^+$\xspace}
\newcommand{\wpplus}{$\mathcal{W}_{p}^+$\xspace}
\newcommand{\domainw}{$\mathcal{W}_d$\xspace}

\newcommand{\personw}{$\mathcal{W}_p$\xspace}

\newcommand{\p}{$\mathcal{P}$\xspace}
\newcommand{\pbeta}{$\mathcal{P}_{\beta}$\xspace}
\newcommand{\pbetaplus}{$\mathcal{P}^{+}_{\beta}$\xspace}

\newcommand{\aspace}{$\alpha$-space\xspace}
\newcommand{\apspace}{$\alpha^{+}$-space\xspace}

\newlength{\imwidth}

\newlist{todolist}{itemize}{2}
\setlist[todolist]{label=$\square$}

\usepackage{titletoc}

\SetAlFnt{\small}
\SetAlCapFnt{\small}
\SetAlCapNameFnt{\small}
\SetAlCapHSkip{0pt}

\makeatletter
\let\@authorsaddresses\@empty
\makeatother

\setcopyright{acmcopyright}\acmJournal{TOG}
\acmYear{2022}\acmVolume{41}\acmNumber{6}\acmArticle{206}\acmMonth{12} \acmDOI{10.1145/3550454.3555436}

\begin{document}
\title{MyStyle: A Personalized Generative Prior}

\author{Yotam Nitzan}
\affiliation{
    \institution{Google Research}
    \country{USA}
}
\affiliation{
    \institution{Tel-Aviv University}
    \country{Israel}
}

\author{Kfir Aberman}
\author{Qiurui He}
\author{Orly Liba}
\author{Michal Yarom}
\author{Yossi Gandelsman}
\author{Inbar Mosseri}
\author{Yael Pritch}
\affiliation{
    \institution{Google Research}
    \country{USA}
}
\author{Daniel Cohen-Or}
\affiliation{
    \institution{Google Research}
    \country{USA}
}
\affiliation{
    \institution{Tel-Aviv University}
    \country{Israel}
}

\begin{abstract}
	We introduce MyStyle, a personalized deep generative prior trained with a few shots of an individual. MyStyle allows to reconstruct, enhance and edit images of a specific person, such that the output is faithful to the person's key facial characteristics. 
Given a small reference set of portrait images of a person ($\sim 100$), we tune the weights of a pretrained StyleGAN face generator to form a local, low-dimensional, personalized manifold in the latent space. 
We show that this manifold constitutes a personalized region that spans latent codes associated with diverse portrait images of the individual.
Moreover, we demonstrate that we obtain a personalized generative prior, and propose a unified approach to apply it 
to various ill-posed image enhancement problems, such as inpainting and super-resolution, as well as semantic editing. Using the personalized generative prior we obtain outputs that exhibit high-fidelity to the input images and are also faithful to the key facial characteristics of the individual in the reference set.
We demonstrate our method with fair-use images of numerous widely recognizable individuals for whom we have the prior knowledge for a qualitative evaluation of the expected outcome.
We evaluate our approach against few-shots baselines and show that our personalized prior, quantitatively and qualitatively, outperforms state-of-the-art alternatives.
\end{abstract}



\begin{teaserfigure}
    \centering
    \includegraphics[width=0.98\linewidth]{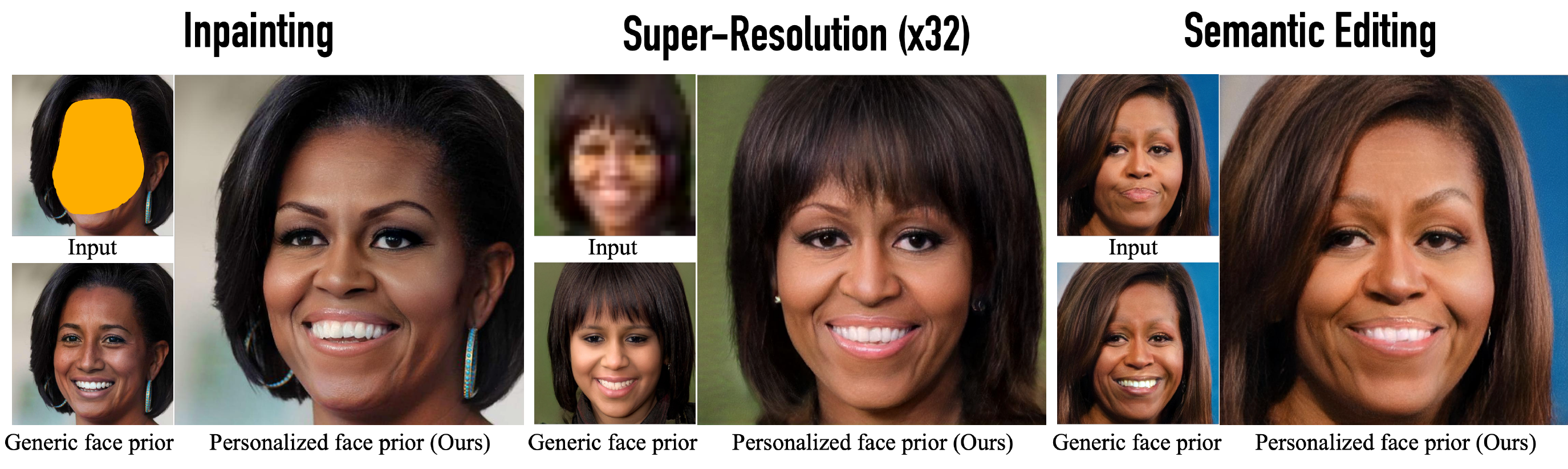} 
    \caption{Using our personalized prior trained for Michelle Obama, we solve various challenging tasks while faithfully preserving her key facial characteristics.
    Each example shows the original input image of Obama (top left), which may be corrupted, and the output based on our personalized face prior (right), compared to a generic face prior (bottom left). The generic face prior is learned from a diverse set of people and produces results that do not preserve Michelle Obama's key facial characteristics. 
    Conversely, our personalized prior is learned from a few images of Michelle Obama. \camera{Left and middle blocks \copyright U.S. Government, right block \copyright Tim Pierce.}
    }
    \label{fig:teaser}
\end{teaserfigure}

\maketitle

\vspace{-2mm}
\section{Introduction}
\label{sec:intro}

Our personal digital album contains a myriad of images depicting ourselves in different scenery, poses, expressions, and lighting conditions. Although each image can be very different from the others, they all contain our unique facial characteristics. Leveraging this property of one’s photo collection may enable the development of editing operations that are tailored to a specific individual, providing faithful reconstruction of their key facial characteristics in various scenarios. In particular, such an approach may be useful in image enhancement applications where only partial cues related to perceived identity of the captured subject are present, such as super-resolution, deblurring, inpainting, and more.

In recent years, the domain of image editing and enhancement in general, and face editing in particular, has experienced a significant shift. From pixel-level editing approaches \cite{averbuch2017bringing}, the field gradually shifted to latent-space editing methods that essentially interpret operators that are applied in a latent space of a generative model \cite{goodfellow2014generative} as explicit image editing operations. This latent-space-based editing enables new capabilities and demonstrates state-of-the-art performance. In particular, StyleGAN \cite{karras2019style} became the gold-standard and core component for semantic editing of face images \cite{wu2020stylespace,abdal2020styleflow}.
While results are impressive, all methods operate by hallucinating information from a general \textit{domain prior} that is learned from a large and diverse dataset containing many identities. Hence, when editing an image of a recognizable person, such as Michelle Obama, the result of using a generic face prior may be a person that only resembles Obama and does not preserve her key facial features (see Figure~\ref{fig:teaser}).  

In this work, we propose an approach to a new problem - \emph{personalization} of a face prior, which addresses the following question: Given a few portrait images of an individual, can we learn a \emph{personalized-prior} that facilitates face-editing and enhancement operations while being faithful to the unique facial characteristics of that individual?

Our key idea is to learn a low-dimensional, local, manifold in the latent space of a generative model, where the manifold embeds the individual's images and constitutes a personalized prior that enables generation and enchantment of identity features that are faithful to the individual depicted in the images.

In practice, we begin with a generator pretrained on a dataset of general, diverse, faces \cite{karras2019style}. Such a generator constitutes a \textit{domain prior} which encapsulates understanding of high-quality face imagery and semantic facial features. We aim to preserve these properties while tailoring the prior for a specific person. Given a small set of images of an individual, we next tune the generator's weights, such that each image in this set is reconstructed with a particular fixed code, coined \textit{anchor}. The anchor is calculated by projecting the image onto the latent space using a dedicated encoder.
Because our tuning is applied to specific regions in the latent space, it only affects a low-rank local manifold, enabling preservation of the essential properties of StyleGAN. 

Owing to the latent space being smooth and disentangled \cite{karras2020analyzing},
linear combinations of anchors also become faithful to the individual's identity and exhibit fusion of low-level and high-level attributes that appear in the reference set. In particular, we show that the convex hull defined by the anchors constitutes a \textit{personalized prior}. 
Finally, we leverage this newly created personalized prior and propose a dedicated projection method for various image enhancement tasks, as well as a novel method for identity preserving semantic editing.

Our contribution is threefold: (i) we introduce the notion of a personalized face prior; (ii) we present a few-shot tuning technique to obtain a personalized subspace in the StyleGAN latent space, and (iii) we propose a unified approach that leverages it for personalized image enhancement and editing applications and achieves state-of-the-art identity preserving results.

We extensively evaluate our framework on several applications and against multiple baselines. Specifically, we evaluate the identity preservation of our generator when used to synthesize images, and when it serves as prior that regularizes image enhancement and semantic editing tasks.
Additionally, we perform a thorough analysis of the personalized space, enabling detailed  understanding of its key qualities.
With qualitative and quantitative evaluation, and a user study, we demonstrate that our approach excels in capturing the characteristic features of the face identity and outperforms alternatives in all the aforementioned applications.

\vspace{-2mm}
\section{Related Work}

\subsection{Generative Prior}

Recently, pre-trained GANs \cite{goodfellow2014generative} have become the image prior of choice for many popular tasks such as image enhancement \cite{menon2020pulse, yang2021gan, chan2021glean, richardson2021encoding,wang2021towards,luo2020time,pan2021exploiting, gu2020image} and semantic editing \cite{shen2020interfacegan, patashnik2021styleclip, voynov2020unsupervised, abdal2020styleflow, harkonen2020ganspace, nitzan2020dis}. These methods project an image to the latent space of a GAN and leverage its prior to perform enhancement or editing. For a thorough introduction, we refer the reader to a recent survey \cite{Bermano2022STAR}.

A substantial issue with this approach is that the GAN is evidently unable to faithfully represent all images of the domain, often causing inaccurate projection. This problem is often mitigated by using an enlarged latent space \cite{abdal2019image2stylegan}. However, this is a mixed blessing, as it is also known to weaken the prior, presenting a trade-off \cite{tov2021designing}. Recently, \citet{pan2021exploiting, bau2020semantic, roich2021pivotal}, and \citet{alaluf2021hyperstyle} have proposed to overcome the trade-off by further tuning the generator, at ``test time'', to facilitate an accurate projection of a specific image.

Our method involves a similar fine-tuning technique. However, its purpose is entirely different. We aim to form a personalized generative prior that faithfully captures a rich distribution of appearances of an individual.
One can then ordinarily leverage the prior to perform image enhancement and semantic editing. Differing from previous works, as the prior is personalized, our results are faithful to the identity of the individual.

\vspace{-1mm}
\subsection{Few-Shot Generative Models}

Forming a personalized generative prior first requires training a generative model.
Many methods have been proposed to train generative models with limited data.
These methods focus on preventing over-fitting using augmentation \cite{zhao2020differentiable, Karras2020ada}, an auxiliary task \cite{yang2021data}, regularizing training with a dedicated loss \cite{ojha2021few, li2020few}, or by limiting which weights are trained \cite{mo2020freeze, gal2021stylegan}.

We experiment with several few-shot methods and find that they fail to adequately preserve the identity of an individual depicted in a few images. We speculate that one key reason is that such methods map a distribution of a small set of images into the entire latent space, only enabling preservation of coarse features. We therefore propose a different training method based on Pivotal Tuning \cite{roich2021pivotal}, which proves to be superior for our application.

\vspace{-1mm}
\subsection{Personalization}

In recent years, personalization has become an important factor in some fields of Machine Learning research such as recommendation systems \cite{netflix} and language models \cite{relate}.
Compared to these fields, personalization has not yet made as strong of an impact on Computer Vision and Graphics. 

Recently, several facial enhancement works \cite{dolhansky2018eye, ge2020occluded, wang2020multiple, zhao2018identity, li2020enhanced} have provided a reference image of the subject at test time to serve as a visual cue. However, an accurate prior of a person's appearance cannot be learned from a single image. Accordingly, these methods are limited to relatively simple tasks and settings and their performance is not on par with their no-reference counterparts. 

In this work, we overcome this gap and bring personalization to the forefront of image enhancement and editing. By forming a personalized generative prior, we can produce highly identity-preserving results for a wide range of applications with comparable quality to state-of-the-art non-personalized methods.

\section{Method}

Our objective is to create a \textit{personalized generative prior} for an individual given relatively few photos of them, roughly 100. To this end, we seek to adapt a StyleGAN \cite{karras2020analyzing} model, trained on a large scale face dataset, such as FFHQ \cite{karras2019style}, to capture a personalized prior on top of the domain prior.

Our approach can be divided into two parts. First, we use an intra-domain adaption method based on Pivotal Tuning \cite{roich2021pivotal} (\subsecref{subsec:training}). 
Tuning is applied to specific regions in the latent space, and hence, does not transform it uniformly.
Once applied, we identify a low-rank, local, manifold that has been personalized while preserving the high quality typical to StyleGAN (\subsecref{subsec:obtaining}).
Second, we leverage this newly created personalized prior and propose a generic method for various image enhancement tasks based on latent space projection (\subsecref{subsec:applying}), as well as a novel method for identity preserving semantic editing (\subsecref{subsec:editing}).

\begin{figure}
\setlength{\tabcolsep}{1pt}
    \centering
    \includegraphics[width=\linewidth]{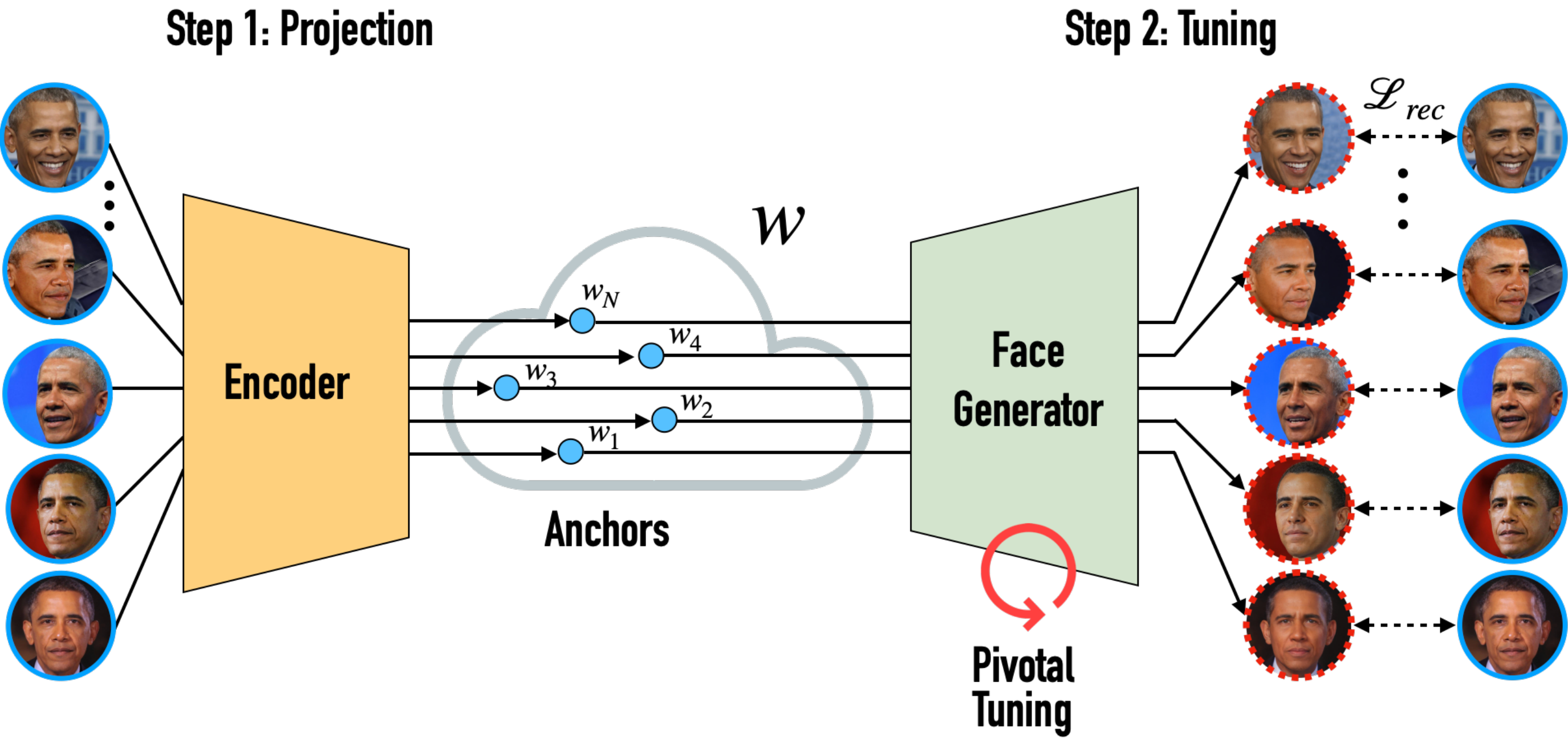}
    \vspace{-4mm}
    \label{fig:tuning}
    \caption{An illustration of our tuning method. We project a set of $N$ portrait images of an individual into StyleGAN's $\mathcal{W}$ space, resulting in a set of \textit{anchors} that are the nearest possible neighbors. We then tune the generator reconstruct the input images from their corresponding anchors.
    }
    \vspace{-4mm}
    \label{fig:method_tuning}
\end{figure}

\subsection{Adaptation}
\label{subsec:training}

Given a reference set of $N$ images depicting an individual, $\mathcal{D}_p = \{x_i\}_{i=1}^N$, we adapt a pre-trained StyleGAN generator, $G_d$, constituting a domain prior into one constituting a personalized prior, $G_p$. 
We denote \domainw, \personw $\subseteq \mathbb{R}^k$, to be the learned latent spaces of $G_d$ and $G_p$, respectively.

To obtain $G_p$, we propose a training scheme inspired by Pivotal Tuning \cite{roich2021pivotal}, shown in \figref{fig:method_tuning}.
We aim to change \domainw as little as possible so it would encode the personalized reference set $\mathcal{D}_p$, without harming the rich semantic domain prior previously learned.
To this end, we invert an image $x_i$ into \domainw using a pretrained inversion encoder \cite{richardson2021encoding}, yielding a code $w_i$, coined \textit{anchor}.
$G_d(w_i)$ is not a perfect reconstruction of $x_i$ but rather an estimate to its nearest neighbor possible in StyleGAN's \domainw space.
We thus tune $G_d$ with a simple reconstruction objective -- reconstruct $x_i$ given the latent code $w_i$. 
We use a combination of a pixel loss, $L_2$, and an LPIPS loss \cite{zhang2018unreasonable}, $\mathcal{L}_{lpips}$.
Formally, the reconstruction loss over the entire set is
\begin{equation}
\label{eq:training_loss}
    \mathcal{L}_{rec}(G,x_i, w_i) = \mathbb{E}_i \; [\mathcal{L}_{lpips}(G(w_i), x_i) + \lambda_{L_2} \norm{G(w_i) - x_i}_2].
\end{equation}
where $\lambda_{L_2}$ is a hyperpatameter balancing the two losses. We optimize the weights of the generator, initialized with $G_d$, and take the result to be $G_p$.

\subsection{Obtaining a Personalized Sub-Space}
\label{subsec:obtaining}

We have now obtained a generator $G_p$ that reconstructs the reference set from the anchors. Naturally, a question arises -- \emph{what is encoded in other latent codes?} In \Cref{sec:dch_analysis,subsec:supp_dch_analysis,subsec:beta_tradeoff_appendix}, we extensively investigate and answer this question. In the subsection hereafter, we  first describe the key observations arising from the said investigation and provide intuitions to their origin. We then leverage the observations to model a personalized subspace.

We first observe that latent codes "close`` to an anchor have been personalized, \ie, possess the identity of the individual. This effect is expected since a GAN's generator is often smooth with respect to its input. StyleGAN's generator, in particular, was explicitly trained for this purpose \cite{karras2020analyzing}. 
While our tuning optimizes only on a finite set of anchors, it empirically appears easier for the generator to maintain smoothness and propagate the effect to neighboring latents than doing otherwise. This "ripple effect`` was considered an undesired side effect by Roich \etal \shortcite{roich2021pivotal}, and they regularized training to prevent it. On the contrary, we willingly embrace it. 
Such neighboring latents are \textbf{implicitly} trained to portray the individual's identity but not to reconstruct any image. Therefore, these latents expand the personalized prior and increase its expressive power. 
Conversely, latents that are "far`` from all anchors are not constrained by an explicit loss nor by smoothness constrains, and are therefore not personalized.

We additionally find that linear interpolations between pairs of anchors have been personalized. 
This phenomenon can be attributed StyleGAN's highly disentangled latent space. It is known that linear interpolation between two latent codes in \w rarely depict features that are absent from both endpoints \cite{karras2019style}.
While identity is a relatively ambiguous feature, it is not inherently different.
Therefore, since the anchors depict the same identity, so does the interpolation between them.
A simple induction then suggests that every convex combination of anchors is identity preserving.

Given both aforementioned findings - we consider a \textit{"Dilated`` Convex Hull} defined by the anchors to be a personalized subspace within \personw. By dilated we refer to expanding the convex hull outwards to capture more neighboring latent codes.
Throughout this work, we represent the convex hull using normalized generalized Barycentric coordinates \cite{floater2015generalized}. Simply put, let us consider $\textit{V} = span(\{w_i\}_{i=1}^N)$. Then, generalized Barycentric coordinates are the coordinates with respect to the set of anchors as basis, whose sum is $1$.
Using normalized generalized Barycentric coordinated, the $\beta$-dilated convex hull is easily defined by
\begin{equation}
\label{eq:a_b}
    \mathcal{A}_\beta = \{ \vect{\alpha} \in \mathbb{R}^N \vert \, \sum_i \alpha_i = 1, \forall i: \alpha_i \geq -\beta \},
\end{equation}
where $\beta \in [0,\infty)$ controls the amount of dilation. 
Let $M$ $\in\mathbb{R}^{k\times N}$ be the matrix with the anchors along its columns. The dilated convex hull in standard coordinates used for \personw is given by
\begin{equation}
\label{eq:p_b}
    \mathcal{P}_{\beta} = \{ M \vect{\alpha} \vert \, \vect{\alpha} \in \mathcal{A}_{\beta} \}
    = \{ \sum_i \alpha_i w_i \vert \, \sum_i \alpha_i = 1, \forall i: \alpha_i \geq -\beta \}.
\end{equation}

From \eqnref{eq:p_b} we note that, $\mathcal{P}_{0}$ is the set of all convex combinations of the anchors and is therefore by definition the convex hull of the anchors.
Intuitively, one can consider the \aspace to be the space of the coefficients of those convex combinations. On the other hand, $\mathcal{P}_{\infty} = \textit{V}$.
We often omit $\beta$ from the notation and informally note \pbeta as \p and $\mathcal{A}_\beta$ as \aspace. 

The distinction made between "close`` and "far`` points when considering anchors' neighborhoods is a simplification. Due to the space being smooth, the vanishment of the prior is gradual (see \figref{fig:analysis}).
As $\beta$ increases, further latent codes are included in \pbeta, weakening the prior. However, we observe that such further codes exhibit greater expressivity. Therefore, $\beta$ controls a tradeoff between the prior -- image quality and personalization -- and expressiveness. The value of $\beta$ could be determined empirically, according to the user's preference between the prior and expressiveness.

\subsection{Personalized Image Enhancement}
\label{subsec:applying}

\begin{figure}
\setlength{\tabcolsep}{1pt}
    \centering
    \includegraphics[width=\linewidth]{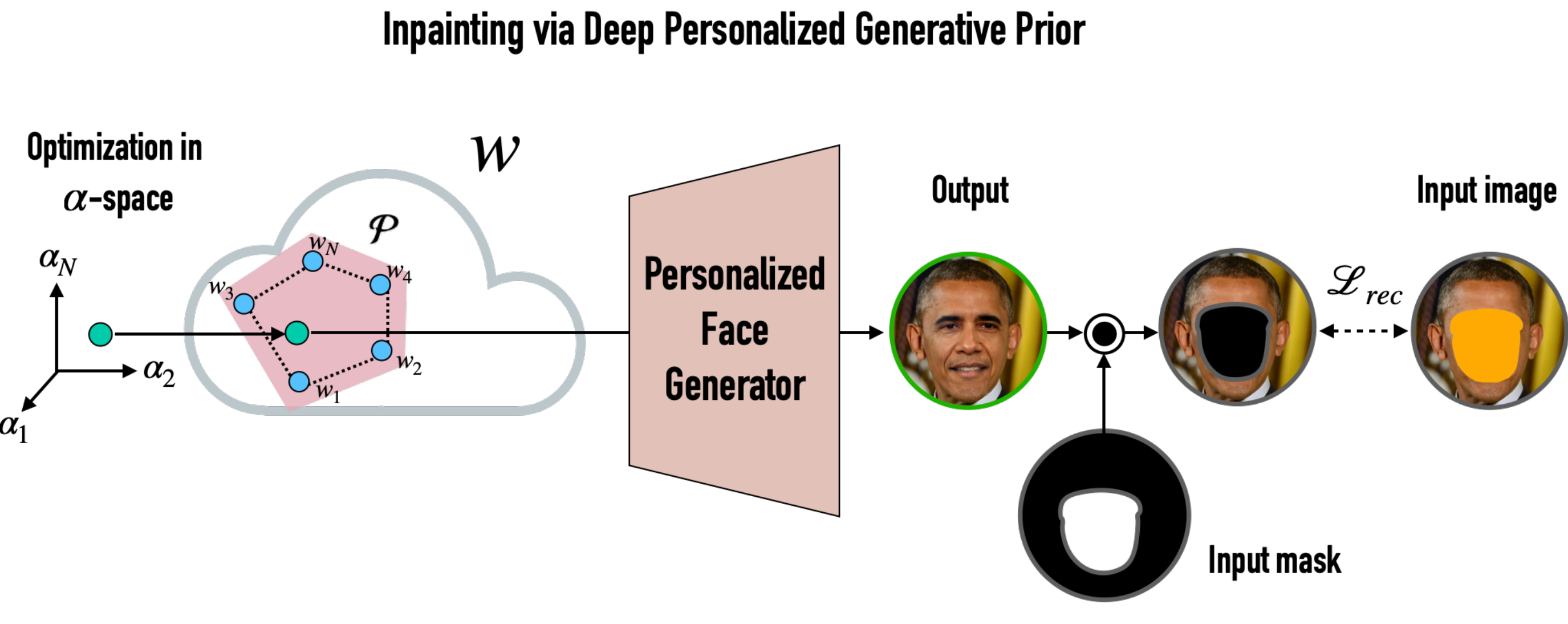} \\ 
    \vspace{-3mm}
    \caption{
    Illustration of our enhancement method for Inpainting. We optimize a latent code in \aspace to find the latent code in \p that most accurately reconstruct the corrupted input image, outside of the mask.
    }
    \vspace{-3mm}
    \label{fig:method_inpainting}
\end{figure}

Having modeled a personalized prior, we now wish to leverage it for image enhancement. To this end, we adapt a common projection-based method used with a domain prior \cite{menon2020pulse, Luo-Rephotography-2020} to use the personalized prior instead.
The method's core idea is to find the latent code that best reconstructs the degraded input image, $I$, with respect to a know degradation function $\phi$. See \Cref{subsec:domain_enhancement_bg} for additional details.
An illustration of the method is displayed in \figref{fig:method_inpainting}.

In order to leverage the prior, one must restrict the optimization solution to remain on the manifold. 
This restriction can be efficiently implemented in \aspace, and we therefore project to \aspace, rather than to $\mathcal{P} \subset \mathcal{W}$.
Adopting the reconstruction loss defined in \eqnref{eq:training_loss}, the new projection problem can be described as
\begin{equation}
\label{eq:a_projection}
    \vect{\alpha}^* = \argmin_{\vect{\alpha}_\beta \in \mathcal{A}_{\beta}} \mathcal{L}_{rec} (\phi \circ G, I, M \vect{\alpha}_\beta)
\end{equation}

The restriction for $\vect{\alpha}_\beta$ to remain in $\mathcal{A}_{\beta}$ is implemented intuitively and efficiently using the constraints in \aspace's definition. To bound the minimal negative value of $\vect{\alpha}_\beta$ to $-\beta$, we pass an unrestricted $\vect{\alpha}$ through a softplus function shifted by $\beta$.
Formally,
\begin{equation}
\label{eq:softplus}
 \vect{\alpha}_{\beta} = \frac{1}{s}log(1+e ^ {s (\vect{\alpha} + \beta)}) - \beta.
\end{equation}
Note that $\vect{\alpha}$ is a vector and all operations in \eqnref{eq:softplus} are element-wise. $s$ is a "sharpness`` hyper-parameter.
Restricting the sum of $\vect{\alpha}_{\beta}$ to $1$, is done by adding both a soft penalty term to the objective --
\begin{equation}
    \label{eq:sum_reg}
    \mathcal{L}_{sum}(\vect{\alpha}_\beta) = (\sum_i \alpha_i - 1 ) ^2,
\end{equation}
and explicitly normalizing the optimization result.

We have thus far described a projection method to \aspace. Next, inspired by \wplus's extended expressiveness, we similarly define a new latent spaces $\mathcal{A}_{\beta}^{+}$, coined \apspace, that may use a different $\vect{\alpha}_{\beta}$ for each layer of the generator. The latent space $\mathcal{P}_{\beta}^{+}$, is defined similarly as before (see \eqnref{eq:p_b}).
The projection described for \aspace is generalized to \apspace, where $\mathcal{L}_{sum}$ is calculated per-layer and then averaged.

To regularize the solution in \apspace, we adopt the regularization proposed by \citet{tov2021designing} -- to minimize variation between the latents in different layers. Technically, instead of optimizing a different latent code in each layer, we hold a single latent $\vect{\alpha}_{\beta}$ and an additive offset for each layer, \ie $\vect{\alpha}_{\beta}^{+} = (\vect{\alpha}_{\beta} + \vect{\Delta}_0, ..., \vect{\alpha}_{\beta} + \vect{\Delta}_N)$. We then regularize the norm of the offsets $\Delta = \{\vect{\Delta_i}\}_{i=1}^{N}$, via
\begin{equation}
\label{eq:delta_reg}
    \mathcal{L}_{reg}(\Delta) = \sum_{i} \norm{\vect{\Delta_i}}_{2}.
\end{equation}
Formally, our final objective is given by 
\begin{equation}
\begin{split}
    \mathcal{L}_{final} = \mathcal{L}_{rec} (\phi \circ G, I_d, M\vect{\alpha}^+_\beta) +  \lambda_{reg}\mathcal{L}_{reg}(\Delta) + \mathcal{L}_{sum}(\vect{\alpha}^+_\beta).
\end{split}
\end{equation}
We optimize the objective for $\vect{\alpha}$ and $\Delta$, and note the results as $\vect{\alpha}^*, \Delta^*$.
The final enhanced image is taken to be $G(M\vect{\alpha}^{*^+}_\beta)$.

\subsection{Personalized Semantic Editing}
\label{subsec:editing}

Semantic editing is commonly performed by traversing from an initial latent code, $w$, along a direction, $\vect{n}$, that controls one semantic property in a disentangled manner.
The edited latent code is then given by $w^{edit} = w + \theta \vect{n}$, where $\theta$ is a scalar that determines the magnitude of the edit.
To perform personalized semantic editing, we adapt this common domain prior approach to use the \p space.

Due to model alignment \cite{wu2021stylealign} and similarly to previous works \cite{gal2021stylegan, pinkney2020resolution, roich2021pivotal}, we find that $G_p$ inherits $G_d$'s editing directions.
However, as such directions were learned from the entire domain, they are clearly not personalized. Considering the geometric shape of \pbeta as the personalized space, the reasons are clear. The direction $\vect{n}$ may not reside within $\textit{V}$ and regardless, $\vect{n}$ is infinite. Both factors inevitably cause the editing to exit \pbeta's bounds and hence seize being personalized.
The second factor, in fact, is not unique to our setting and generally exists with other generative models, where traversing "too far`` arrives at regions with low density and hence artifacts \cite{spingarn2020gan}.

\begin{figure*}[h]
    \centering
    \includegraphics[width=0.95\linewidth]{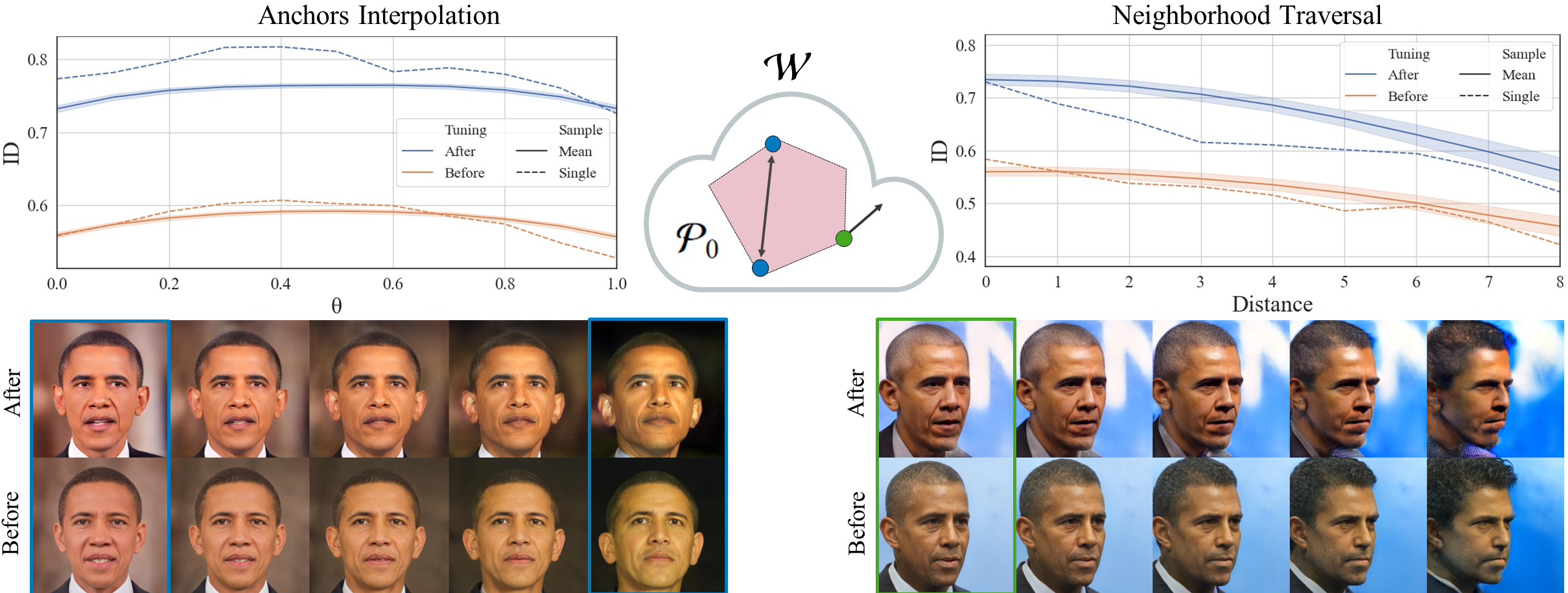}
    \caption{
    Evaluating the identity preservation of Barack Obama, using our \textit{ID} metric, for two types of latent paths -- interpolation between anchor pairs and random walk in anchor's neighborhood. We compare the results {\color{orange} before} and {\color{blue} after} generator tuning.
    Solid lines indicate the mean over a large set and dashed lines represent a single sample, for which visual results are also provided. Images with colored frames are anchors.
    }
    \vspace{-1mm}
    \label{fig:analysis}
\end{figure*}

We now propose a simple two-step method to personalize any linear editing direction by resolving the reasons discussed. First, constraining the editing to $\textit{V}$ is trivially done by projecting $\vect{n}$ upon $\textit{V}$, which yields $\vect{\hat{n}}$.
Second, In order to know if an edit remains in \pbeta, one needs to verify that the minimal dilation required for the edited code is smaller than the permitted dilation, $\beta$. This is could easily be done in \aspace.
Therefore, we express $\vect{\hat{n}}$ with \aspace coordinates, given by $\gamma = (M^T M)^{-1} M^T \vect{\hat{n}}$. The initial latent code, $w$, is obtained using the projection method described in \subsecref{subsec:applying}, to invert a real image. Therefore, it is already given in \aspace coordinates -- $\alpha_{w}$.
Now, we can easily transform the linear editing in \personw to a linear editing in \aspace,
\begin{equation}
    \hat{w}^{edit} = w + \theta \vect{\hat{n}} =  M(\alpha_w + \theta \gamma) = M(\alpha^{edit}).
\end{equation}
Last, the minimal dilation that holds $w^{edit}$, is by definition given by
$\beta^{\,edit} = \abs{\min( \{ \alpha^{edit}_i\}_i \cup \{0\}) }$ and could then be compared with the maximal allowed dilation.

Once the maximal allowed dilation is reached, the user is presented with a choice. They can stop editing, giving the editing direction endpoints, as advocated by Spingarn-Eliezer \etal \shortcite{spingarn2020gan}, or continue editing but project $w_{edit}$ back to \pbeta. The effect of this choice is further discussed in \subsecref{subsec:editing_projection}.

\section{Analysis of the Personalized Prior}
In this section, we analyze the prior captured by \pbeta as a result of the tuning process.
We start by demonstrating that \pbeta has indeed been personalized and analyze the effect of the parameter $\beta$ (\secref{sec:dch_analysis}). We additionally study the effect of the reference set's size and diversity on the prior (\secref{subsec:ablation_size_diversity}). 
Further analysis of personalization, as well as other properties of the personalized generator and prior, is provided in \Cref{sec:more_ablation}.

\subsection{Personalization of \pbeta}
\label{sec:dch_analysis}

We demonstrate that the interpolations and neighborhoods of anchors, have been personalized as a result of tuning.

We sample latent codes along the interpolations of 1000 random anchor pairs of Barack Obama. To sample from neighborhoods, we start from each anchor and gradually traverse a randomly sampled latent direction. We then qualitatively and quantitatively evaluate the identity preservation along both types of traversals.

The quantitative \emph{ID} score is inspired by the top-1 face identification task \cite{cao2018pose}. Under this setting, the identity of an image is determined to be the nearest identity from a reference set, where distance is calculated on deep features extracted from a classifier trained on the domain. Accordingly, our \textit{ID} score reports the cosine similarity to the nearest image in the personalized set. 
We omit the anchors used for the traversal from the ranking to prevent a perfect self-similarity.
We compare the results before and after tuning, \ie on the FFHQ-trained StyleGAN and our generator, respectively.

The results are provided in \figref{fig:analysis}. As can be seen, the tuning results not only in personalization of the anchors themselves, but with a consistent and significant personalization of the anchors' interpolations and neighborhoods, \ie the space captured by \pbeta. 
Interestingly, on average, ID scores of interpolated codes are higher than those of anchors. We speculate this is due to interpolation averaging out extreme properties in the anchor endpoints (see \Cref{subsec:supp_dch_analysis} for in-depth analysis). 

Conversely, traversals outwards from the manifold cause a gradual vanishment of personalization. As greater $\beta$ values are required for containing further traversals, one might be tempted to mitigate this effect by setting $\beta=0$. However, we next demonstrate allowing greater $\beta$ values have the advantage of increasing the expressiveness of the prior.

\begin{figure}
    \centering
    \includegraphics[width=\linewidth]{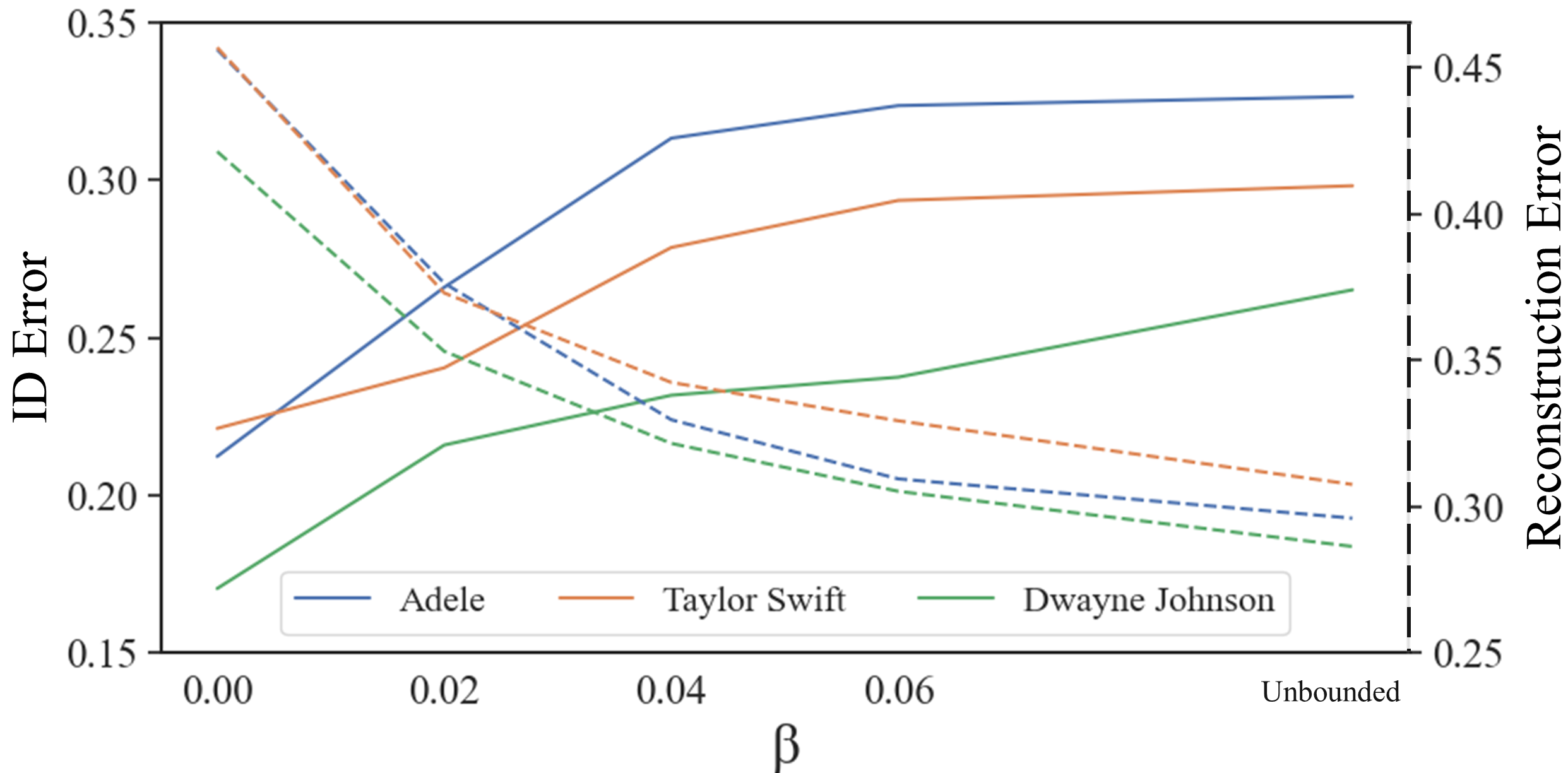}
    \caption{
    Measuring identity preservation and image reconstruction for images inverted into \pbetaplus, while varying the value of $\beta$, the independent variable. At the extreme, $\beta$ is unbounded (\eqnref{eq:softplus} is not used).
    ID errors are depicted using solid lines and the left and solid $y$ axis. Reconstruction (LPIPS) errors are reported using dashed lines and the right and dashed $y$ axis.
    As can be seen, increasing $\beta$ value leads to greater ID error, while reducing reconstruction error. 
    }
    \label{fig:beta_tradeoff_quant}
    \vspace{-2mm}
\end{figure}

To this end, we invert test images into \pbetaplus, using the method described in \secref{subsec:applying}, while varying the values of $\beta$.
We evaluate two different properties of the inverted images -- their reconstruction of the input, and their faithfulness to the prior, \ie preservation of the individual's identity, as defined by their reference set. Reconstruction is measured using LPIPS distance \cite{zhang2018unreasonable} between input and inverted image, and identity preservation using the \textit{ID} score. For convenience, we report the \textit{ID Error} which is the 1-complement of the usual ID metric, measuring distance instead of similarity.
Quantitative results for three individuals are given in \figref{fig:beta_tradeoff_quant}. 
As can be seen, increasing the value of $\beta$ is accompanied with an increase of ID error and decrease of reconstruction error. 

We call this tradeoff, controlled by $\beta$, the prior-expressiveness tradeoff. Allowing greater dilation (greater $\beta$) allows the projection to result in further latent codes. Further latent codes in \pbeta were already demonstrated to be less personalized. The new result, reaffirms that the same phenomenon exists for projection into \pbetaplus.
At the same time, the greater dilation, enlarges the space of possible solutions to the projection. As the objective for optimization is minimizing reconstruction, it is not surprising that allowing more solutions causes smaller reconstruction errors.
We note that this tradeoff is analogous to the distortion-perception tradeoffs discussed by Blau and Michaeli \shortcite{Blau2018ThePT} and then by Tov \etal \shortcite{tov2021designing}, who demonstrated this phenomenon for StyleGAN's common latent spaces - \w and \wplus. 

Additional analysis of the prior-expressiveness tradeoff and its effect on different applications are provided in \Cref{subsec:beta_tradeoff_appendix}.

\subsection{Effect of Dataset Size and Diversity}
\label{subsec:ablation_size_diversity}
We now study the effect the reference set's size and diversity has on the quality of the personalized prior and generator. 
To this end, we measure the inversion accuracy of unseen test images into $\mathcal{P}_{0}$. Accurate inversions exists even in a non-personalized \wplus. However, since we restrain the inversion to $\mathcal{P}_{0}$, its accuracy provides an estimate to the expressive power of the personalized prior.

We sample several subsets of images from the reference sets of three individuals: Joe Biden, Emilia Clarke and Michelle Obama. The subsets are of sizes $10$, $50$, $100$ and $200$.
For each subset, we estimate the diversity by computing the average pair-wise LPIPS, as suggested by \citet{ojha2021few}.
Additionally, we tune $G_p$ and invert $20$ test images following the projection protocol in \secref{subsec:applying}. We then measure the reconstruction accuracy using LPIPS \cite{zhang2018unreasonable}.
We repeat this experiment 5 times for all set sizes other than $200$, which represents the size of the entire set in this experiment.

\begin{figure}
\setlength{\tabcolsep}{1pt}
    \centering
     \begin{subfigure}[b]{\linewidth}
         \centering
         \includegraphics[width=\linewidth]{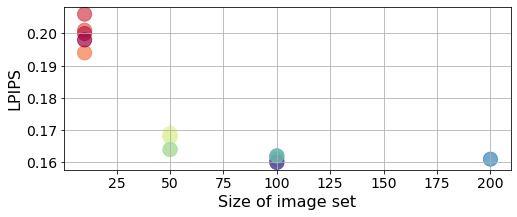}
         \caption{}
         \label{fig:ablation_set_size_graph}
    \end{subfigure}
    \hfill
    \begin{subfigure}[b]{\linewidth}
        \centering
        \setlength{\imwidth}{0.19\linewidth}
        \begin{tabular}{*5c}
            Real & $N = 10$ & $N = 50$ & $N = 100$ & $N=200$ \\ 
            \includegraphics[width=\imwidth]{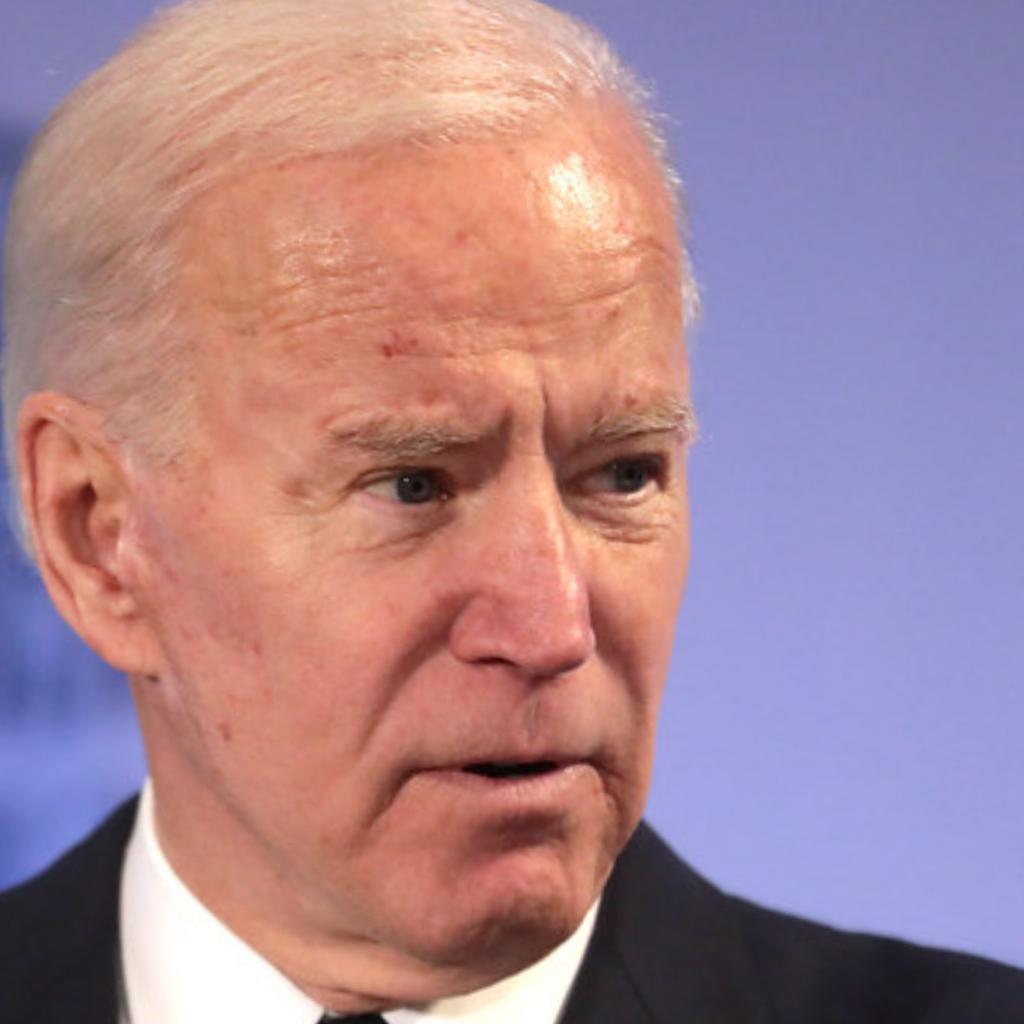} &
            \includegraphics[width=\imwidth]{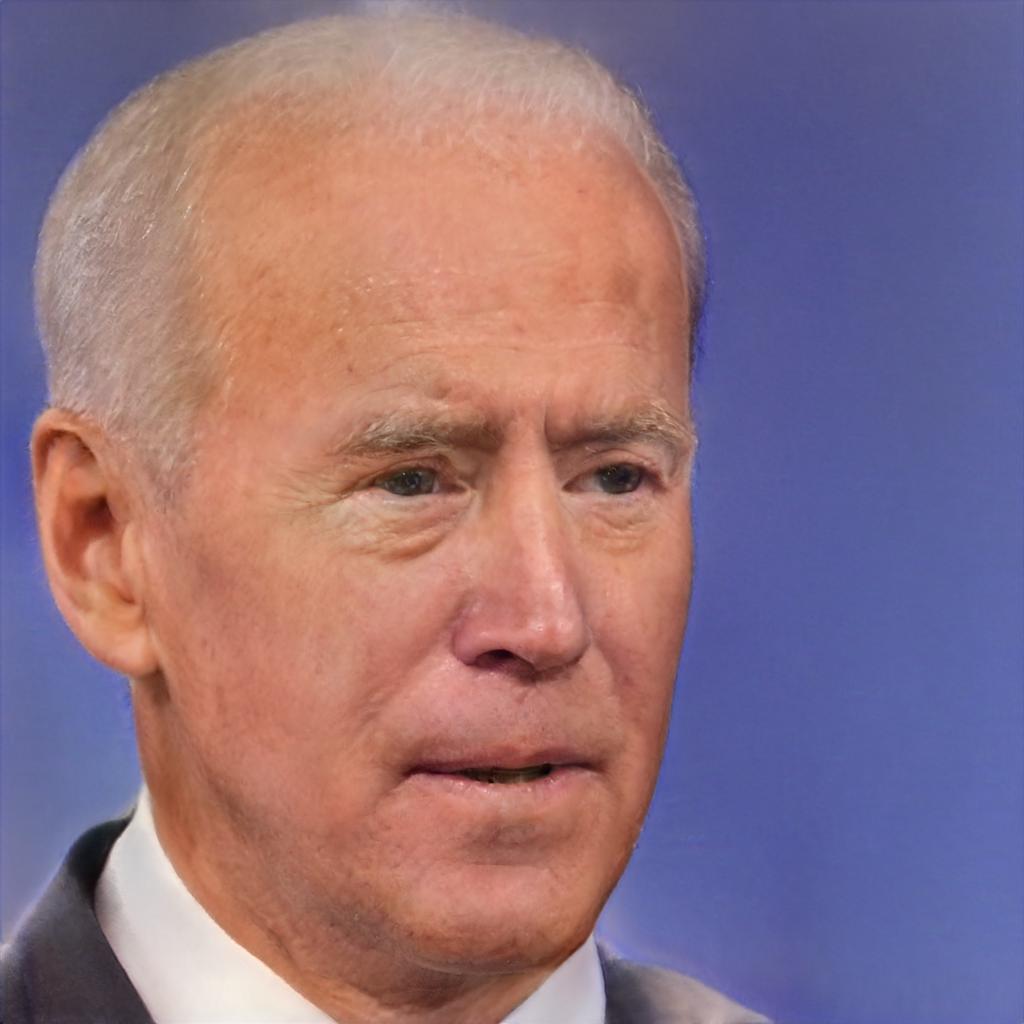} &
            \includegraphics[width=\imwidth]{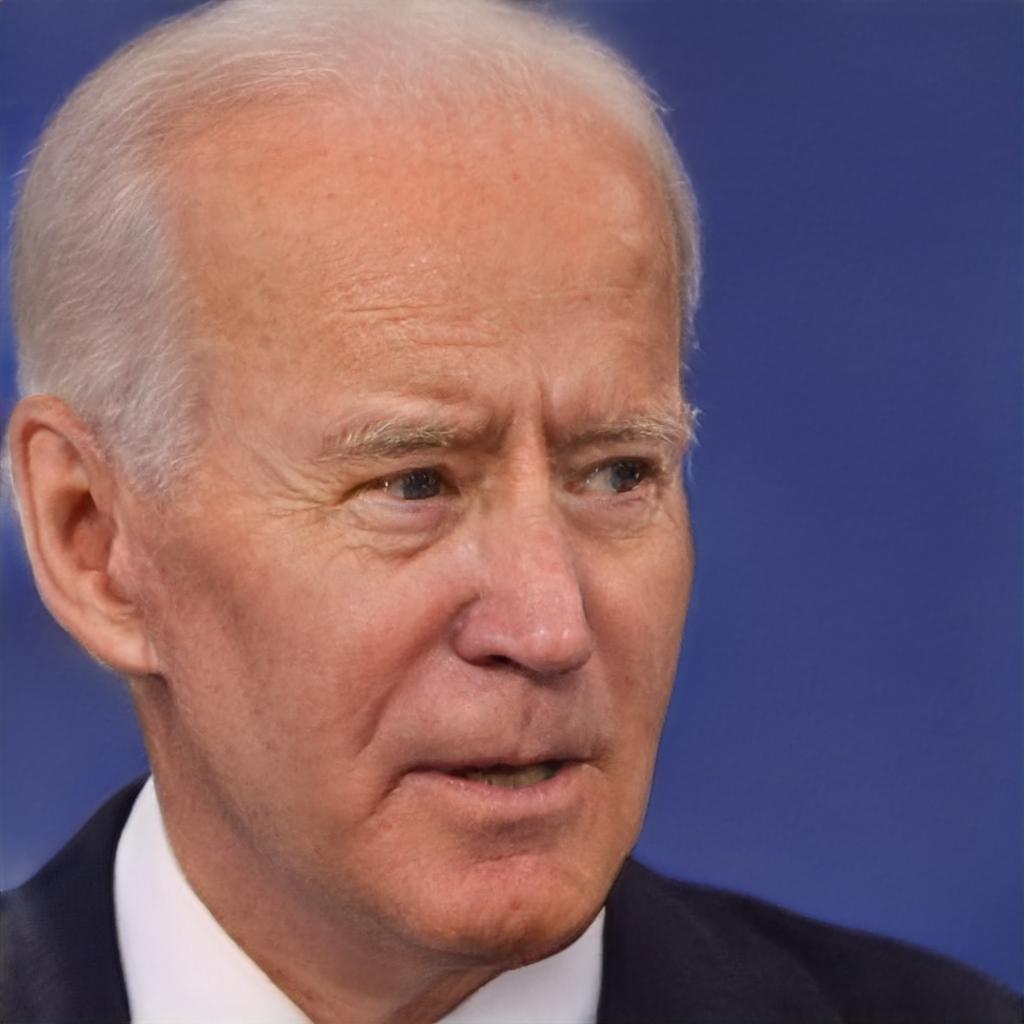} &
            \includegraphics[width=\imwidth]{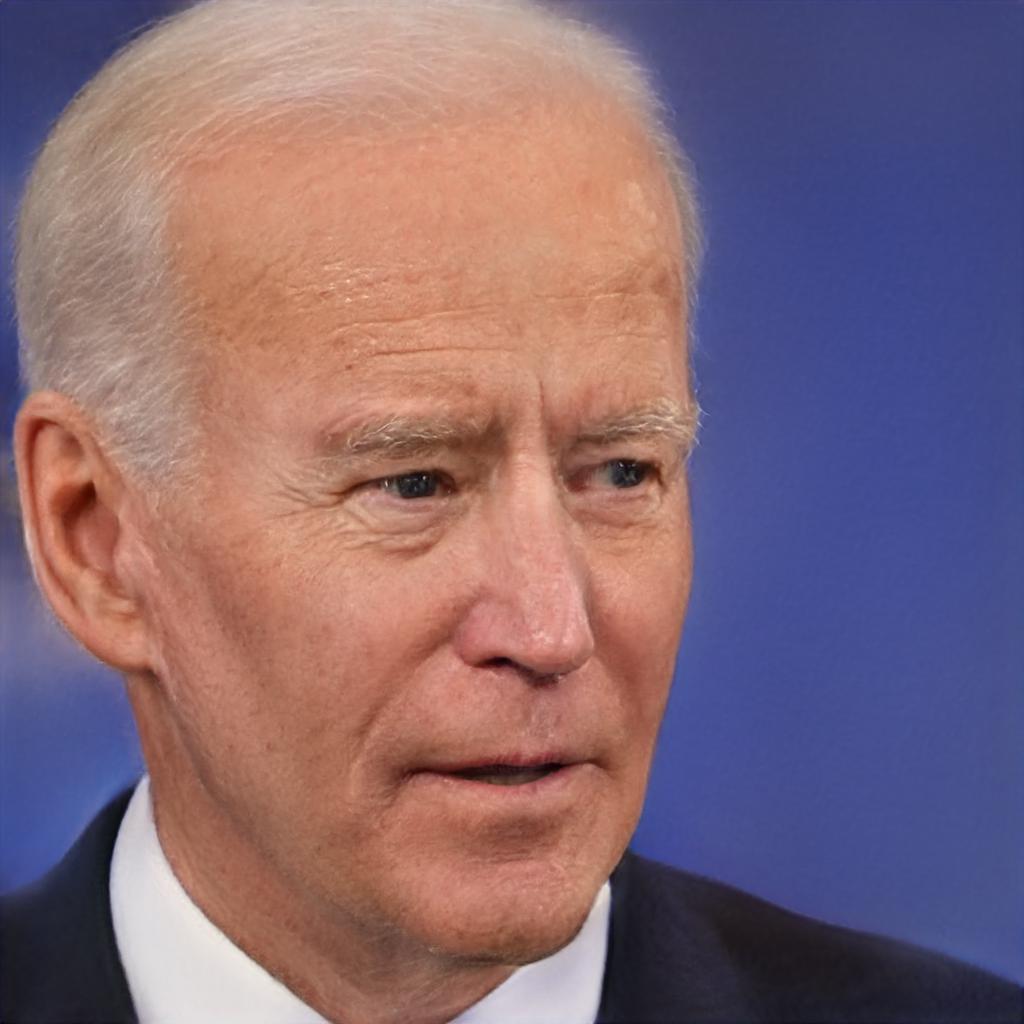} &
            \includegraphics[width=\imwidth]{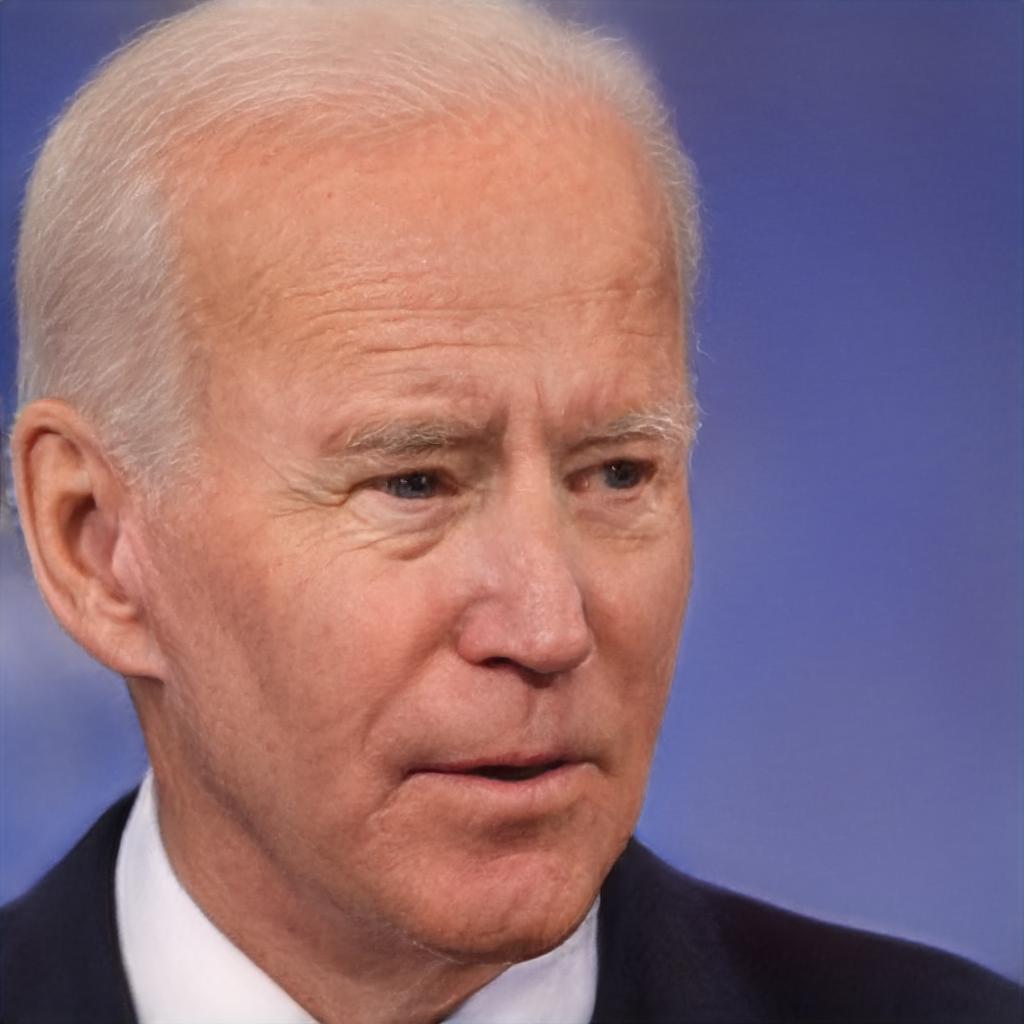}
    	\end{tabular}
         \caption{}
         \label{fig:ablation_set_size_visual}
    \end{subfigure}
     
    \caption{
        The effect of reference set size and diversity on the prior's expressiveness.
        We sample subsets of different sizes from the reference set of Joe Biden.
        For each subset we additionally tune a model, $G_p$, and invert a set of test images to its \p space. (a) Reports the inversion error using average LPIPS distance as a function of set size and diversity. Diversity is computed using average pair-wise LPIPS distances and is reported as color in the spectrum between red (low) and purple (high).
        (b) Visual examples of inverting a given real image with various set sizes. \camera{\copyright Gage Skidmore.}}
    \label{fig:ablation_set_size}
    \vspace{-1mm}
\end{figure}

As can be seen in \figref{fig:ablation_set_size}(a) for Joe Biden, at first, increasing the set size improves both the set's diversity and inversion accuracy. 
However, this is not the case for $100$ and $200$ images where an interesting phenomenon occurs. Although differences are relatively minor, performance correlates to the diversity of the set, regardless of the set size. While further experiments are required, we speculate that adding an image that does not contribute to diversity might add a burden to tuning and increase the dimension of \aspace, thus hurting results.
Visual sample of results is provided in \figref{fig:ablation_set_size}(b). One can observe the improvement moving from $10$ to $50$ images and again with $100$ images, but no major difference with $200$ images.

\section{Applications}
\begin{figure*}
\setlength{\tabcolsep}{1pt}
    \centering
    \vspace{-3mm}
    \includegraphics[width=\linewidth]{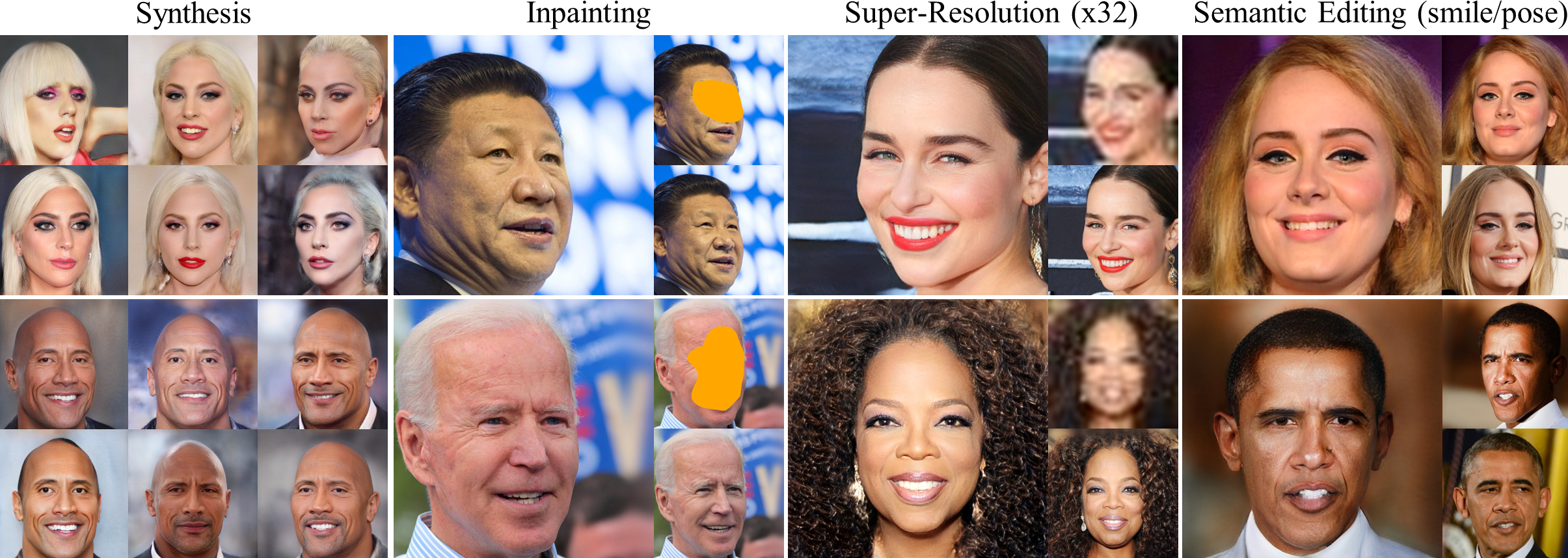}  
    \caption{
    Additional results of our personalized prior for widely recognizable individuals.
    Synthesis results are randomly sampled images. For inpainting and super-resolution we provide triples of inputs, our outputs and ground-truths. There are no ground truths available for semantic editing. We thus provide reference images instead.
    We suggest zooming-in to better view fine details. 
    \camera{Top to bottom and left to right - \copyright Gian Ehrenzeller, \copyright Gage Skidmore, \copyright Gordon Correll, \copyright Canticle, \copyright Kristopher Harris, \copyright U.S. Government.}
    }
    \vspace{-2mm}
    \label{fig:main_results_SIGA_figure}
\end{figure*}

We now turn to demonstrate the application of our personalized prior for popular generative tasks -
image synthesis (\secref{subsec:synth_comparison}), image enhancement (\secref{subsec:app_enhance}) and semantic editing (\secref{subsec:app_edit}). 
In all experiments, we tune the generator from a pretrained FFHQ StyleGAN2 \cite{karras2020analyzing} on a personalized reference set. See \secref{sec:dataset_curation} for information regarding the datasets. 

We note that our personalization approach could be considered as solving a few-shot domain adaption task from the domain of all faces to the domain of the face of a specific person.
While existing domain adaptation methods mostly focus on synthesis, their obtained generator can similarly be leveraged as a prior for all discussed applications.
We therefore consider such methods as the most direct baseline and compare to them for all applications.
For image enhancement, we additionally compare to state-of-the-art methods designed for this specific task.

\subsection{Image Synthesis}
\label{subsec:synth_comparison}

\begin{figure}
	\centering
	\setlength{\tabcolsep}{1pt}
	\begin{center}
        \begin{subfigure}{\linewidth}
        	\setlength{\imwidth}{0.17\linewidth}
    	    \begin{tabular}{*6c}
    		
    		\rotatebox{90}{\phantom{kk} \shortstack{Ojha \\ \etal}} &
            \includegraphics[width=\imwidth]{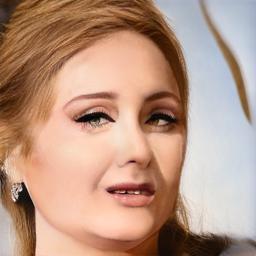} &
            \includegraphics[width=\imwidth]{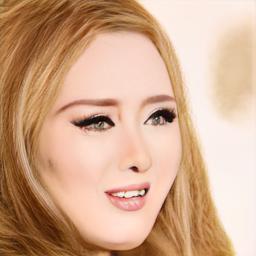} &
    		\includegraphics[width=\imwidth]{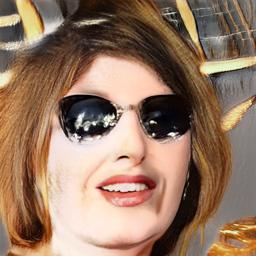} &
    		\includegraphics[width=\imwidth]{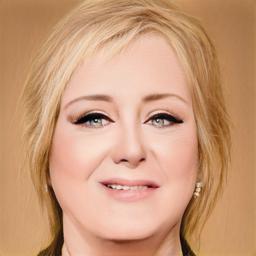} &
    		\includegraphics[width=\imwidth]{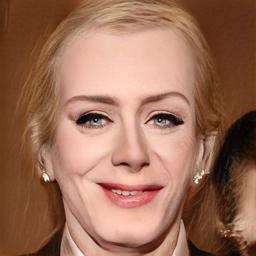}
    		\\
    		
            \rotatebox{90}{\phantom{k} \shortstack{Diff- \\ Augment}} &
            \includegraphics[width=\imwidth]{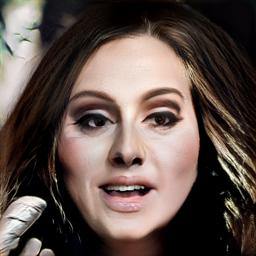} &
            \includegraphics[width=\imwidth]{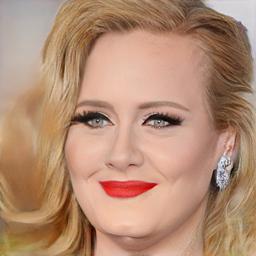} &
    		\includegraphics[width=\imwidth]{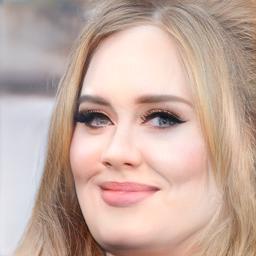} &
    		\includegraphics[width=\imwidth]{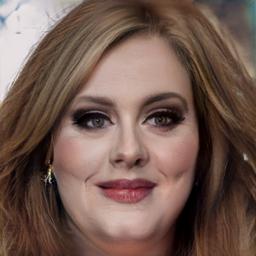} &
    		\includegraphics[width=\imwidth]{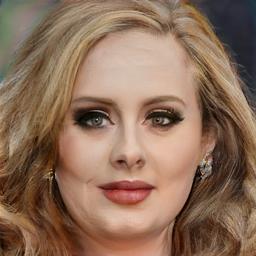}
    
    		 \\
            \rotatebox{90}{\phantom{k} \shortstack{MyStyle \\ (ours)}} &
            \includegraphics[width=\imwidth]{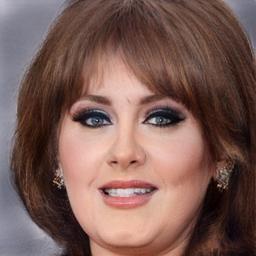} &
            \includegraphics[width=\imwidth]{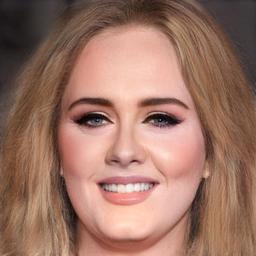} &
            \includegraphics[width=\imwidth]{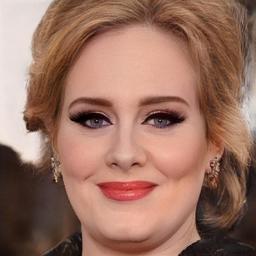} &
            \includegraphics[width=\imwidth]{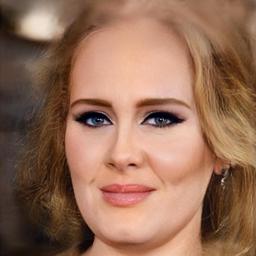} &
            \includegraphics[width=\imwidth]{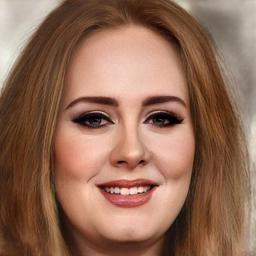} 
    		 \\
    		 
    	    \end{tabular}
        \end{subfigure}%
        
        \begin{subtable}{\linewidth}
            \centering
            \begin{tabular}{l >{\centering\arraybackslash}p{2cm} >{\centering\arraybackslash}p{2cm} >{\centering\arraybackslash}p{2cm} >{\centering\arraybackslash}p{2cm}}
                \toprule
                Method &  User \% ($\uparrow$) & ID ($\uparrow$) & Diversity ($\uparrow$) \\
                \midrule
                Ojha \etal  &  1.4 & 0.53 $\pm$ 0.08  & 2.42 $\pm$ 0.16 \\
                DiffAugment & 31.7 & 0.76 $\pm$ 0.05 & 2.99 $\pm$ 0.17 \\
                MyStyle (Ours)  &  \textbf{68.9} & \textbf{0.79 $\pm$ 0.04}  & \textbf{3.44 $\pm$ 0.16}  \\
                \midrule 
                Real Images  &  83.1  & 1  & 0 \\
                \bottomrule
            \end{tabular} 
        \end{subtable}	    
	\end{center}

	\caption{
	Comparing the synthesis of our generator to few-shots training approaches -- Ojha~\etal~\shortcite{ojha2021few}, and DiffAugment~\cite{zhao2020differentiable}. The user study values reflect the percentages of images that appeared real to the users. Generated images of Adele are provided for visual inspection.
	Our results exhibit diverse appearances of Adele, which are faithful to her actual different appearances over several years.
	}
	\vspace{-2mm}
	\label{fig:synthesis_united_comparison}
\end{figure}

We compare our synthesis results with existing few-shot domain adaptation methods - Ojha~\etal~\shortcite{ojha2021few} and  DiffAugment~\cite{zhao2020differentiable}.
We initialize the generator to the same StyleGAN2~\cite{karras2020analyzing} model trained on FFHQ~\cite{karras2019style} and run each method's training approach on three personalized datasets: Adele (109 images), Kamala Harris (110 images), and Joe Biden (206 images). We randomly synthesize images from each generator and compare the results based on identity preservation and diversity. To synthesize an image using our method, we sample a latent code from $\mathcal{P}_{0}$ (see protocol in \secref{subsec:sampling}) and forward it through $G_p$.

Generative models are often evaluated based on synthesis quality. We note that the term \emph{quality} refers not to the visual quality of the images, but whether their distribution is faithful to the distributions of the training dataset. Therefore, in our setting, identity preservation is a strong indication to quality. To measure identity preservation we first use the ID metric presented in \secref{sec:dch_analysis}. We note that it is reported with respect to the training set, similarly to other quality metrics (\eg FID \cite{heusel2017gans}). Second, we
conduct a user study. Users were presented with an image which was either real or synthesized by one of the methods and were asked to choose whether it looks like a real image of the specific person. We asked users to respond
only if they were familiar with
what the person looked like.
We gathered 1674 responses from 45 users. 

Diversity of synthesized images is evaluated using the 
protocol suggested by Ojha~\etal \shortcite{ojha2021few}. We synthesize 10,000 images from each model and cluster them according to their nearest neighbor in the training set. We then compute the mean and standard deviation of the intra-cluster LPIPS~\cite{zhang2018unreasonable} distances. 

Aforementioned metrics are averaged across all individuals, and reported in \figref{fig:synthesis_united_comparison} alongside a sample of qualitative results.
As can be seen, our model outperforms both alternatives on all metrics, and visual results are significantly more realistic and identity-preserving. Additional qualitative results are provided in the supplementary.

\subsection{Image Enhancement}
\label{subsec:app_enhance}

We choose the tasks of image inpainting and super-resolution as representative examples for image enhancement. We follow the same evaluation protocol for both. As baseline methods, we use the generator obtained by DiffAugment \cite{zhao2020differentiable} as an alternative personalized prior, a state-of-the-art Domain Prior method 
and a version of it fine-tuned on a personalized reference set, noted as Domain Prior+FT. The Domain Prior method serves only as a reference to represent current, non-personalized, state-of-the-art.

We compare the methods according to identity preservation using the ID metric and user study results that reflect both identity preservation and fidelity.
In the study, users were presented with an input image and two results, one of ours and one of a baseline. They were then asked to pick the result which better resembles the person and has higher fidelity to the input. Results are reported as the percentage of responses that preferred a different method over ours. The images used in the user study are a randomly sampled subset of those used for quantitative evaluation.

We explicitly note that, in line with previous works \cite{zhao2021comodgan}, we do not evaluate quality based on reconstruction to a ground truth. By definition, image enhancement hallucinates missing details, which might be valid despite differing from a specific ground truth image. Therefore, quantitative evaluation of reconstruction is not an indication of a successful or failed enhancement. Ideally, qualitative evaluation should be based on the input-output alone, based on the reader's familiarity with the well-known individuals in the experiments. Nevertheless, we include ground truth images to benefit readers who are not familiar with the individuals.

\subsubsection{Inpainting (IP)}

The degradation transformation $\phi$ is modeled as element-wise multiplication of the original image with a binary mask $m$, \ie $\phi(x) = x \odot m$.
we use CoModGAN \cite{zhao2021comodgan} as the state-of-the-art Domain Prior.
All Outputs are post-processed by blending the network's output in the missing regions with the input elsewhere.
We compare all methods on images of Barack Obama (192 images), Lady Gaga (133 images), and Jeff Bezos (114 images). We use $\beta=0.02$ in all experiments.

\begin{figure}
    \small
    \begin{center}
        \label{fig:enhance_quant_and_qual}
        \centering
        \begin{subfigure}{\linewidth}
            \setlength{\tabcolsep}{0.8pt}
            \setlength{\imwidth}{0.15\linewidth}
        	\begin{tabular}{*7c}
    		&
    		Original & Input & \shortstack{Domain \\ Prior} & \shortstack{Domain \\ Prior+FT} & \shortstack{Diff- \\ Augment} & \shortstack{MyStyle \\ (ours)} 
    		\\
    		
    		\raisebox{5mm}{\multirow{2}{*}{\rotatebox{90}{Inpainting (IP)}}} &
            \includegraphics[width=\imwidth]{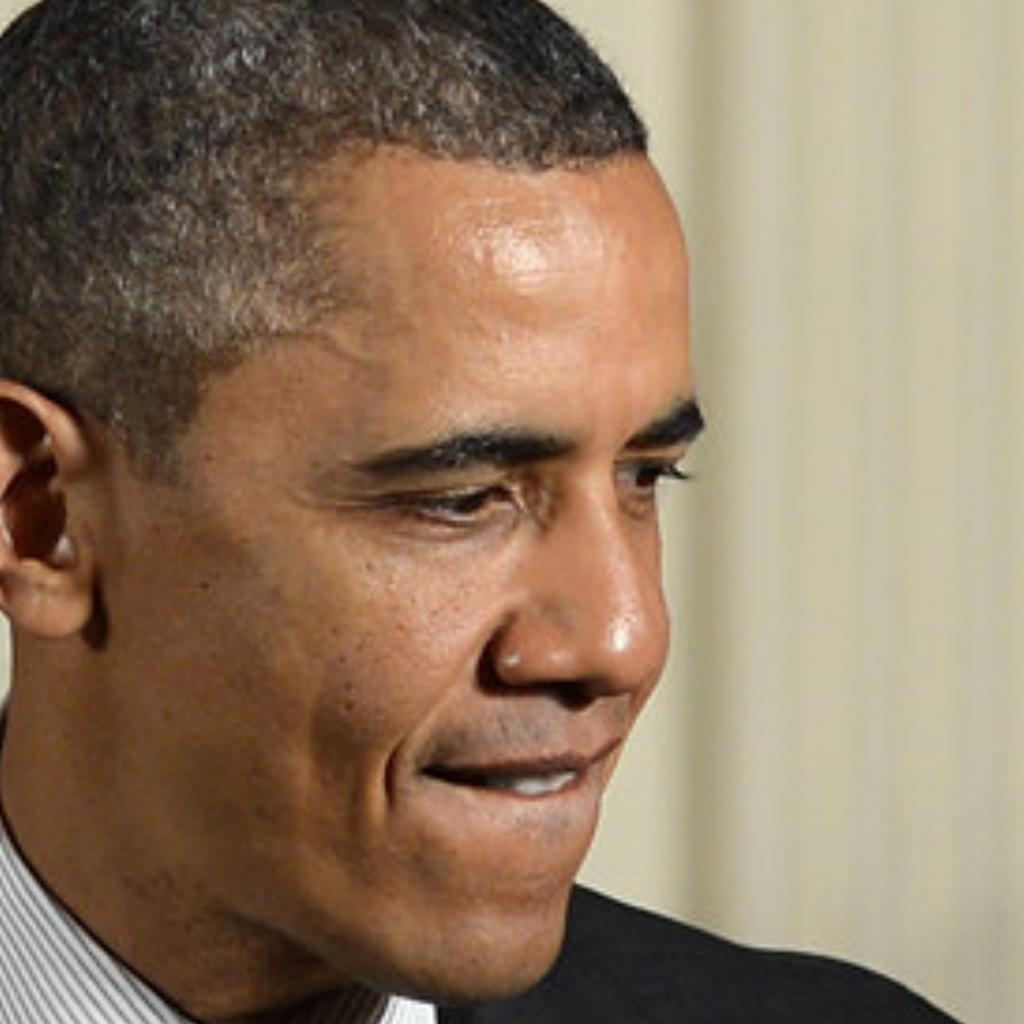} &
            \includegraphics[width=\imwidth]{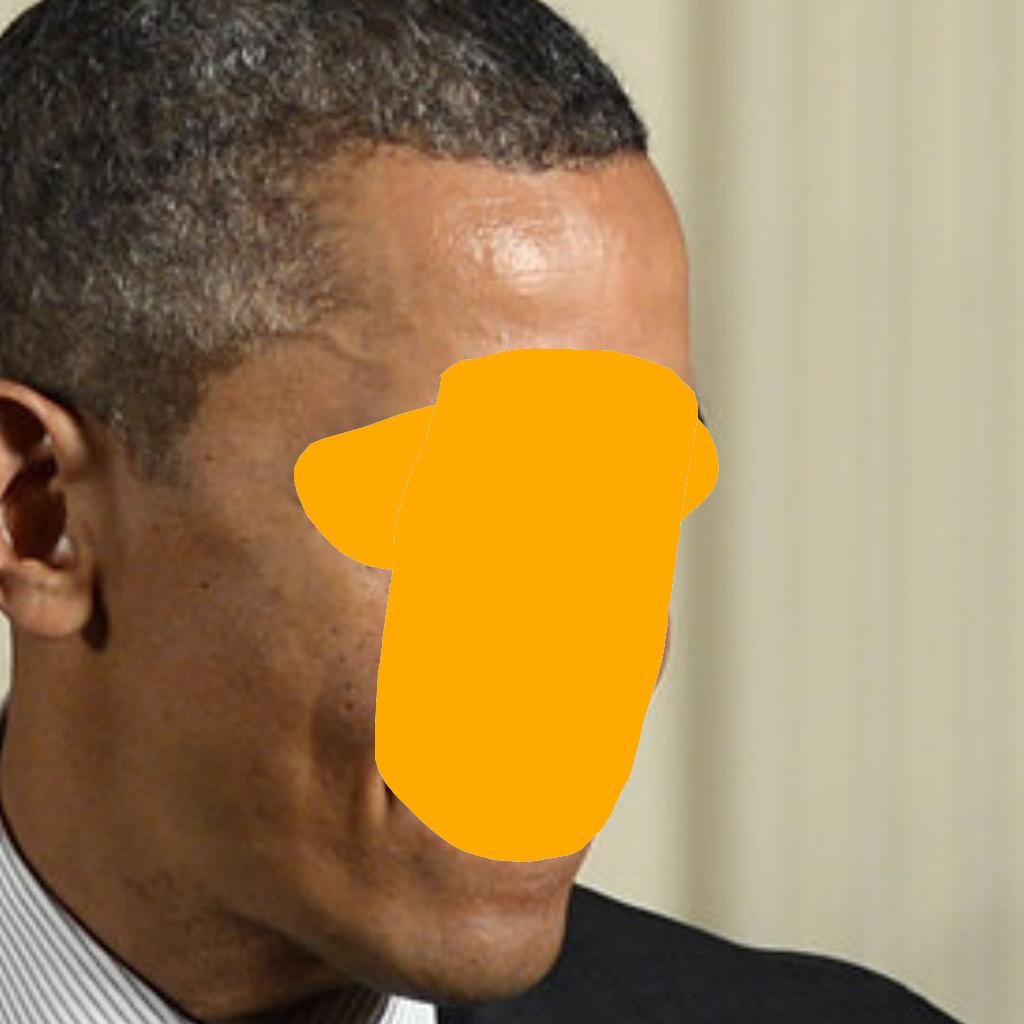} &
            \includegraphics[width=\imwidth]{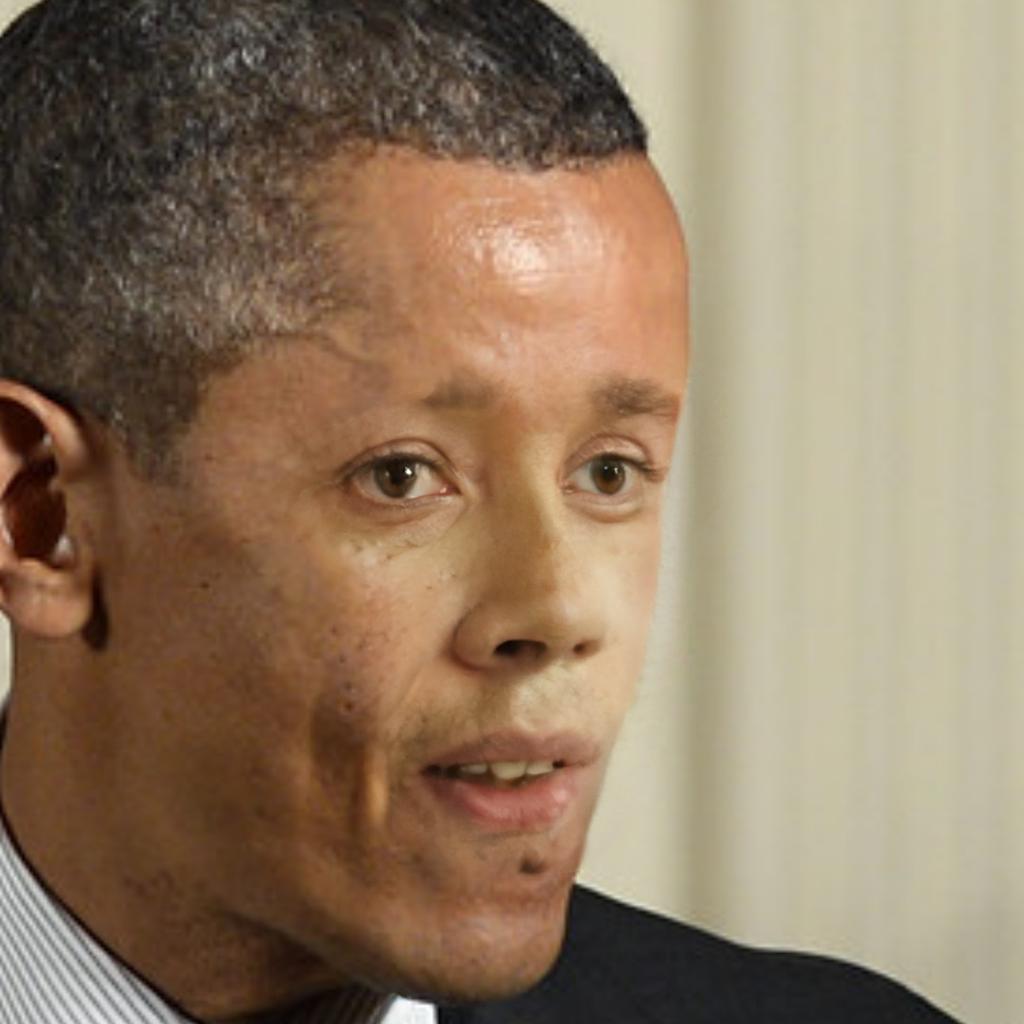} &
            \includegraphics[width=\imwidth]{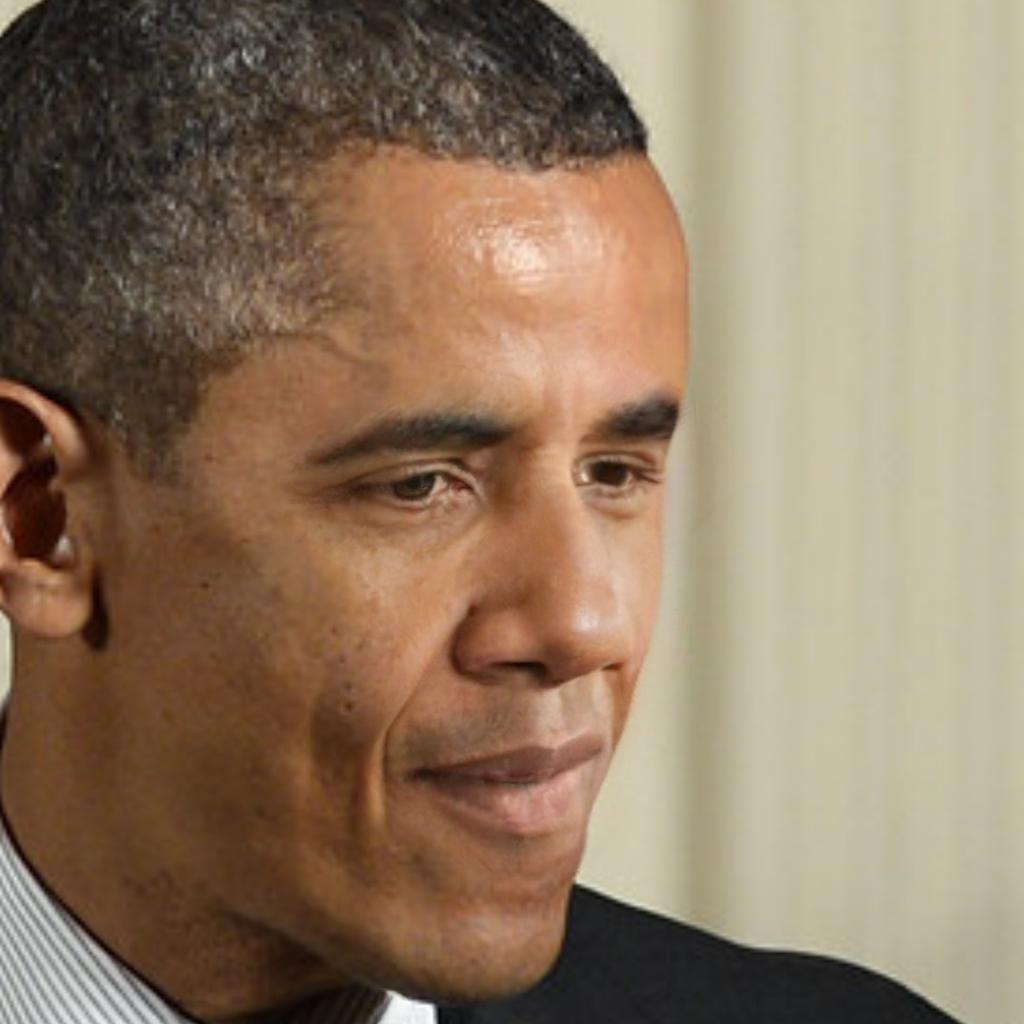} &
            \includegraphics[width=\imwidth]{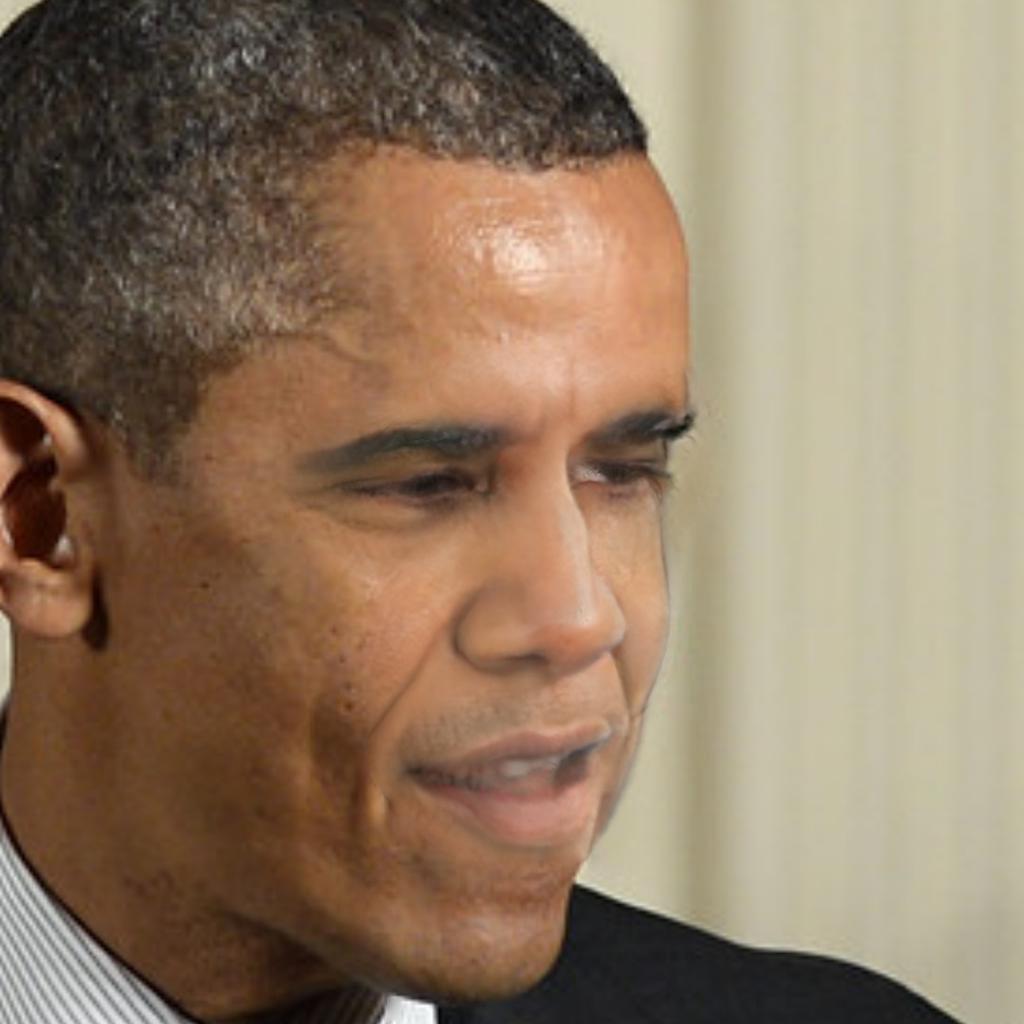} &
            \includegraphics[width=\imwidth]{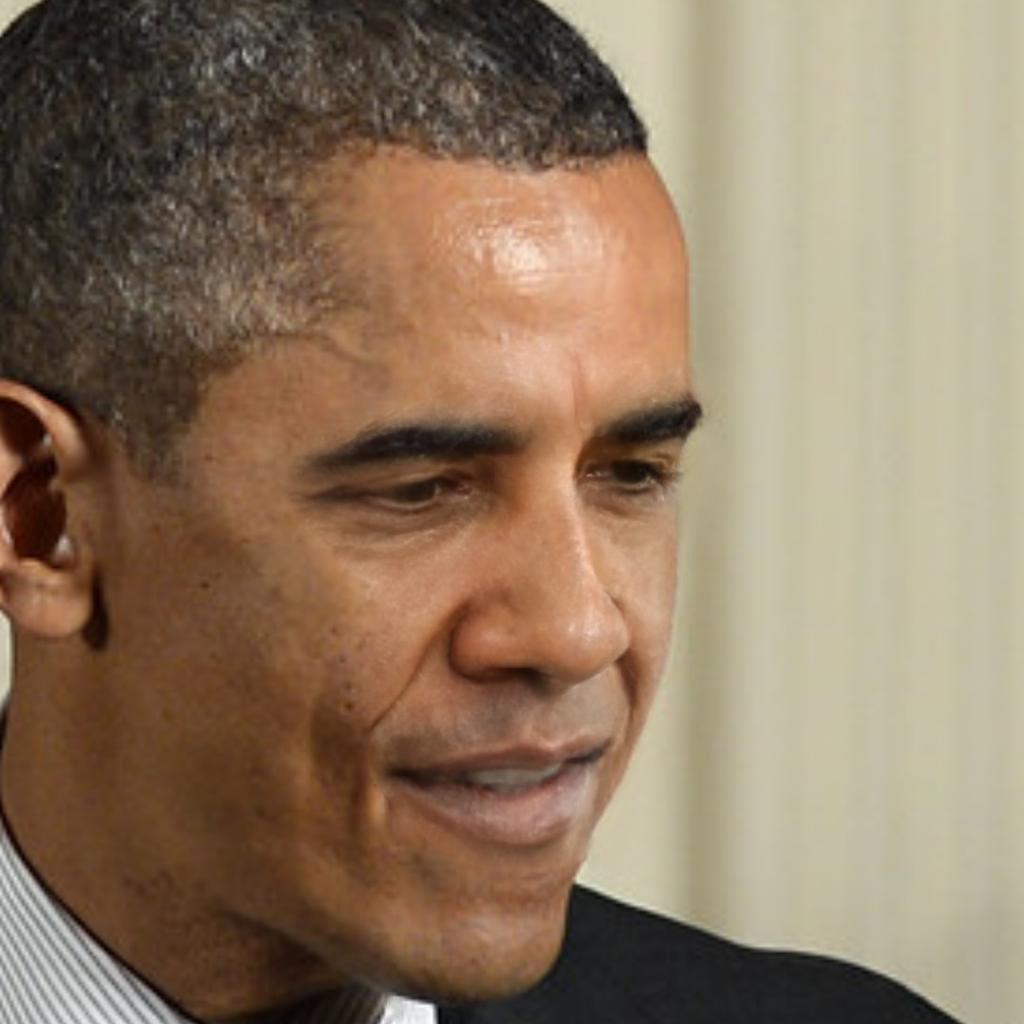}
            \\
            
            &
            \includegraphics[width=\imwidth]{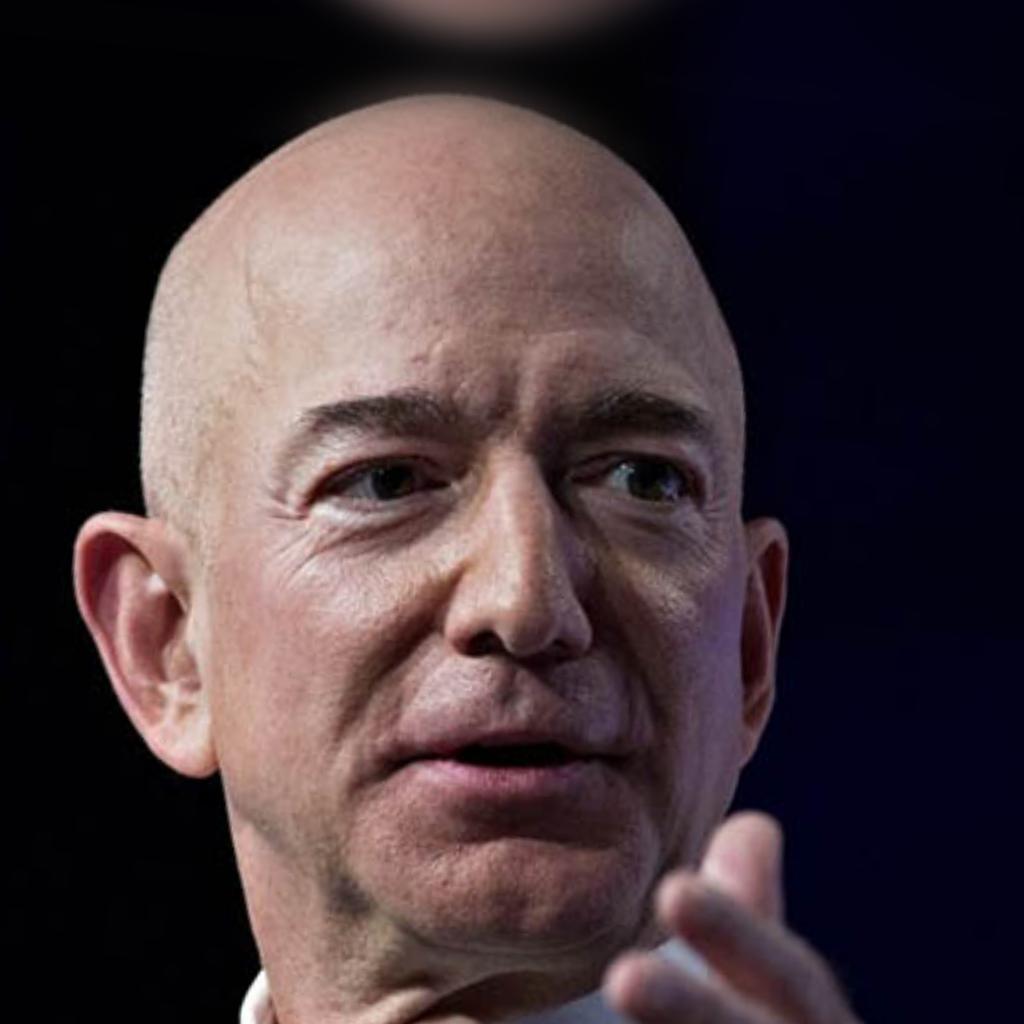} &
            \includegraphics[width=\imwidth]{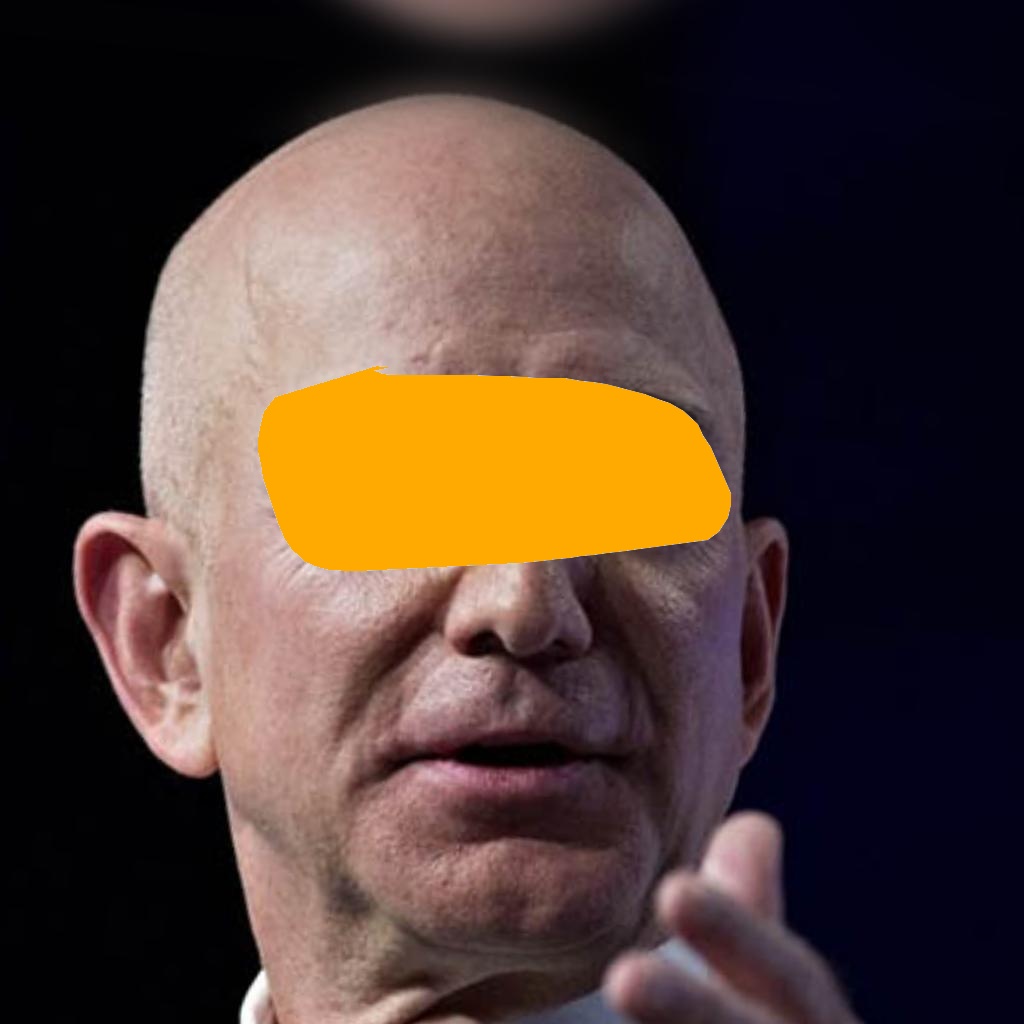} &
            \includegraphics[width=\imwidth]{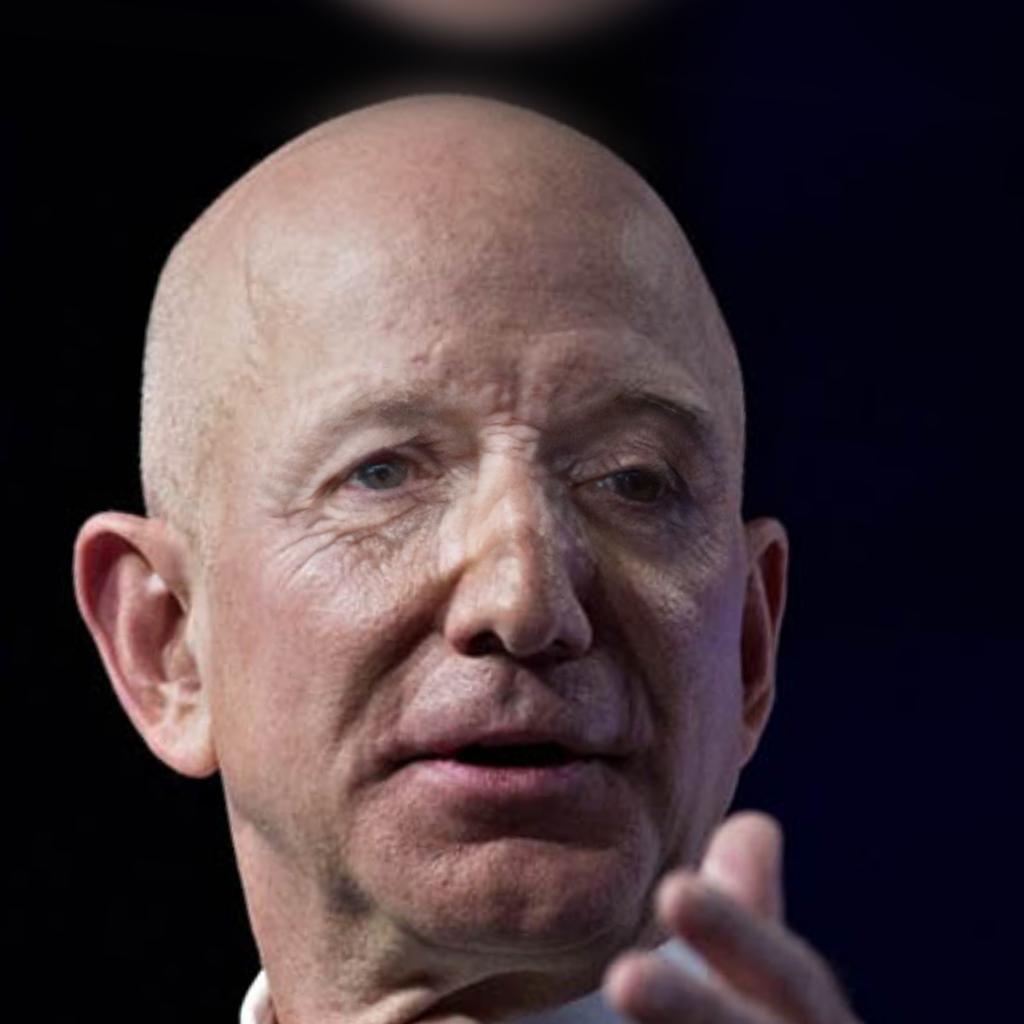} &
            \includegraphics[width=\imwidth]{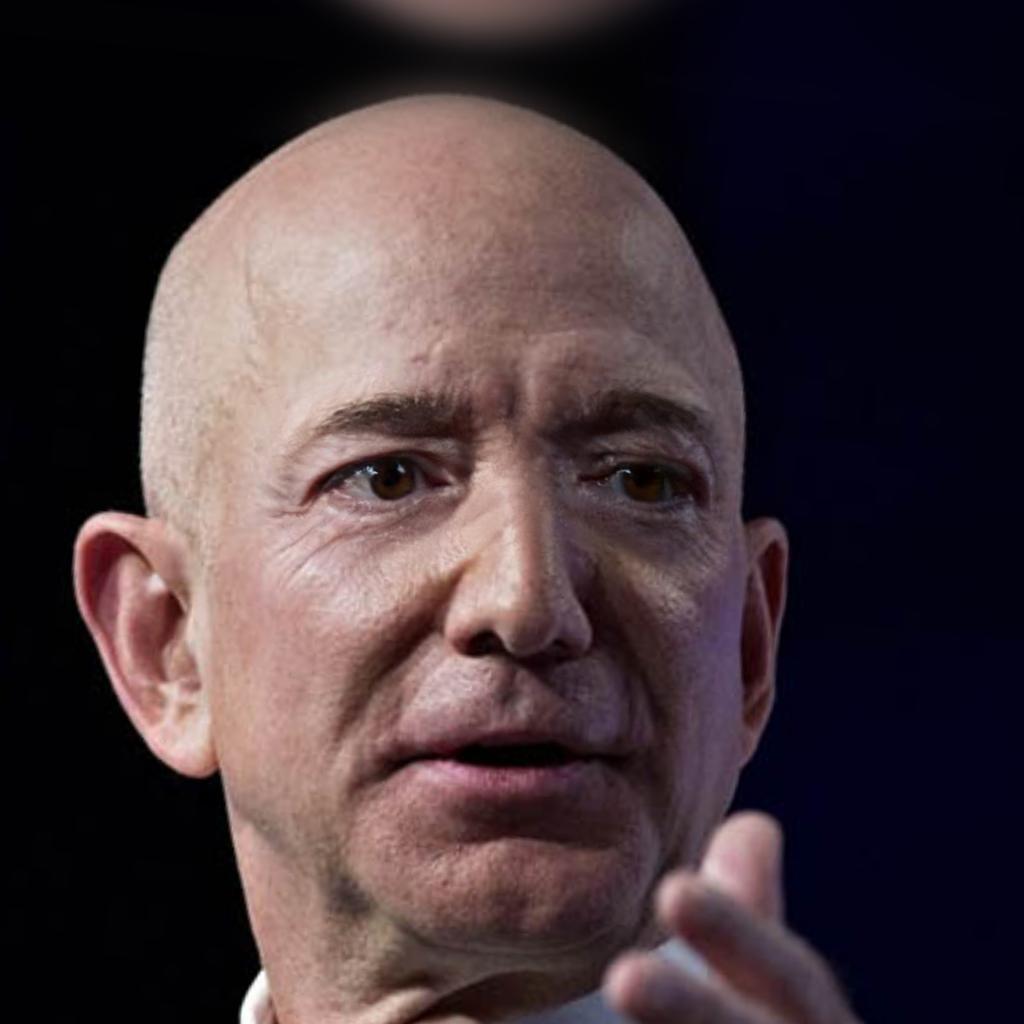} &
            \includegraphics[width=\imwidth]{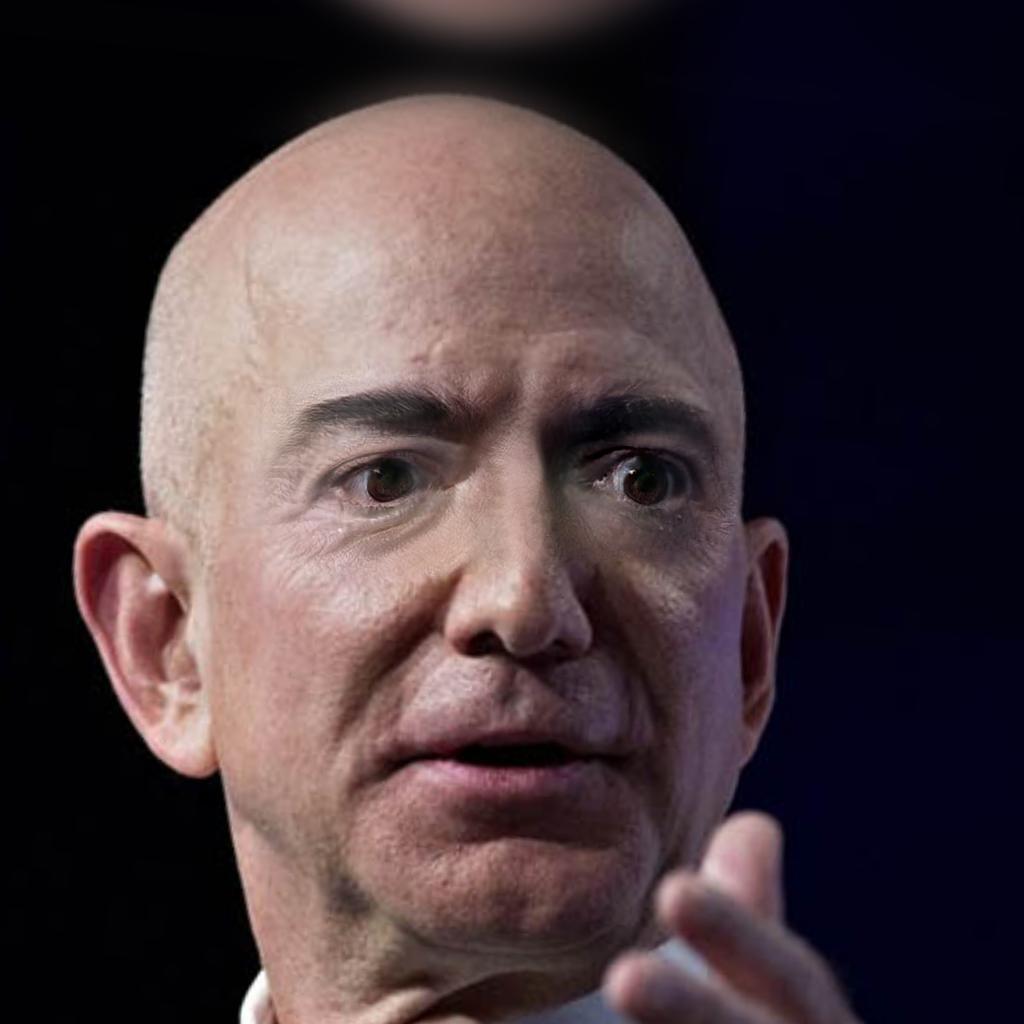} &
            \includegraphics[width=\imwidth]{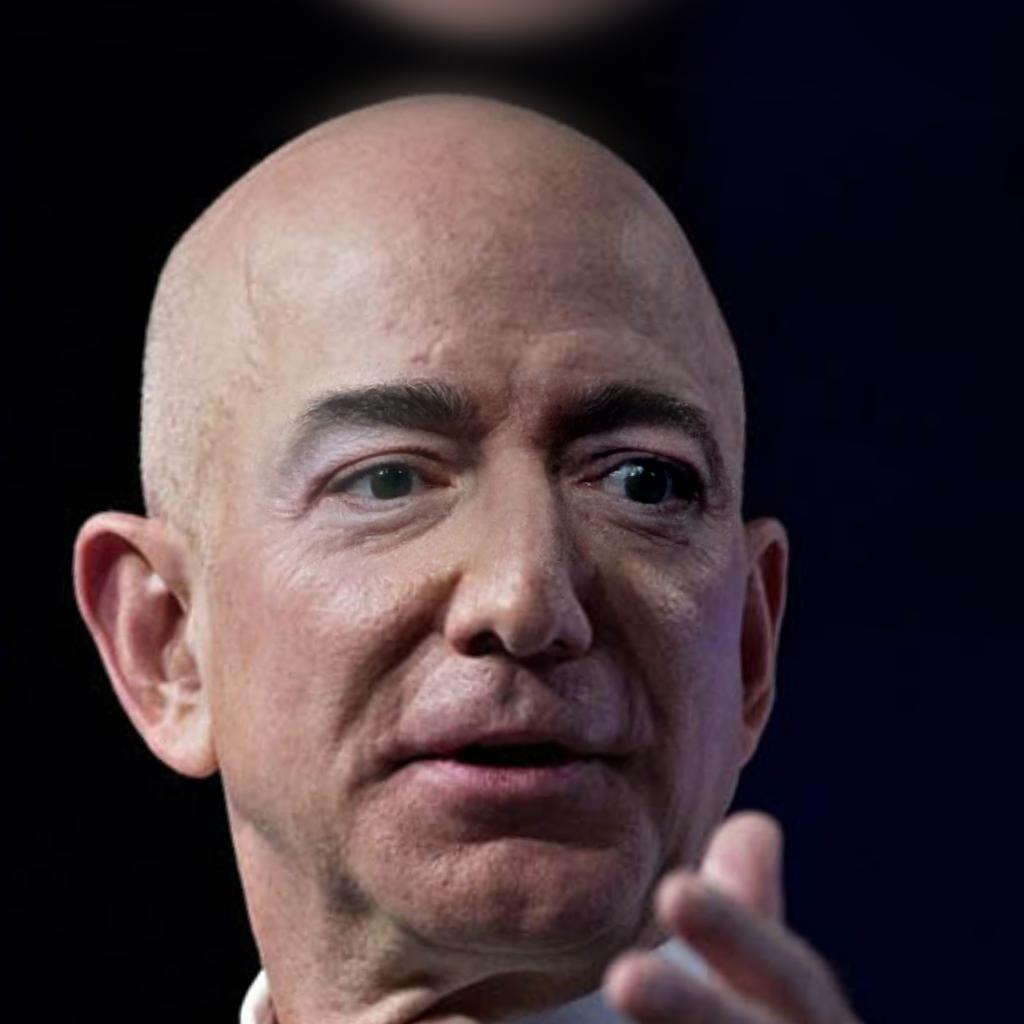}
            \\
            
            \midrule 
            
            \raisebox{10mm}{\multirow{2}{*}{\rotatebox{90}{Super-Resolution (SR)}}} &
            \includegraphics[width=\imwidth]{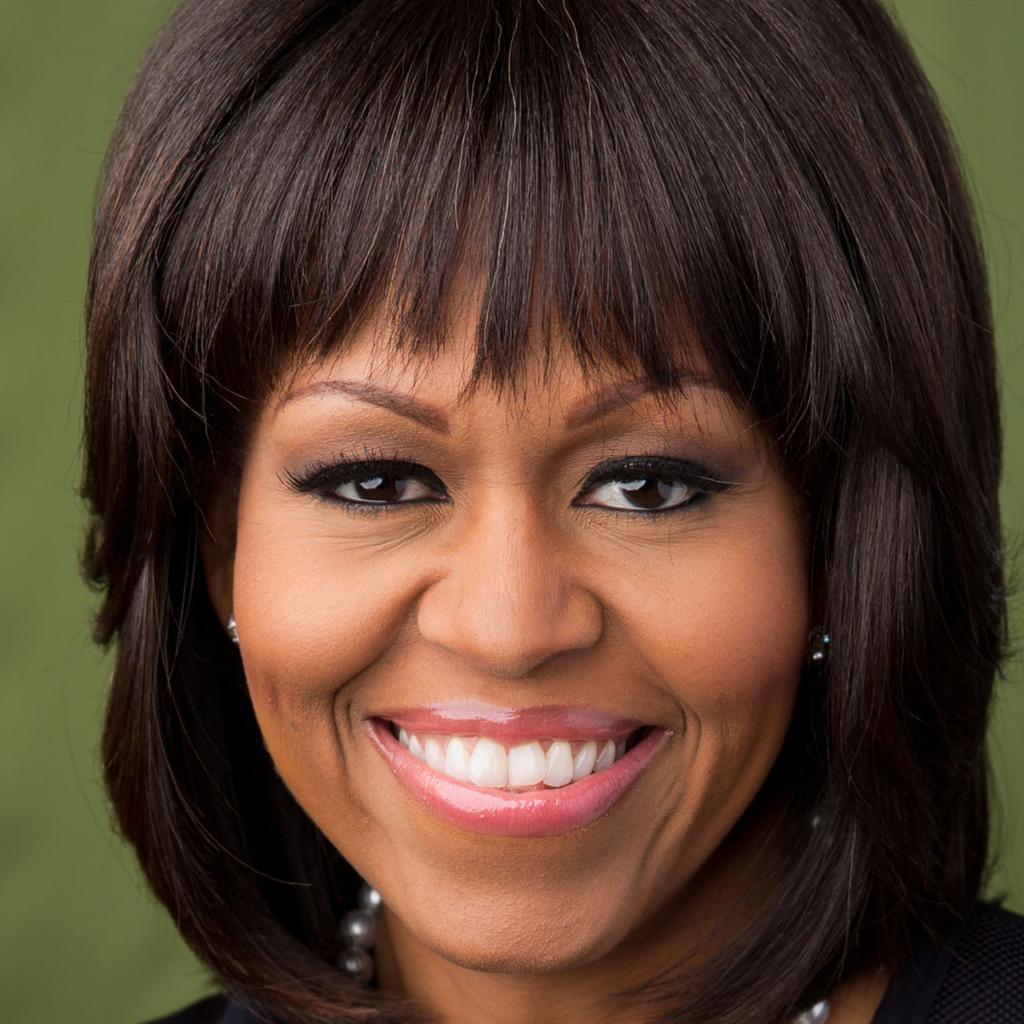} &
            \includegraphics[width=\imwidth]{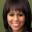} &
    		\includegraphics[width=\imwidth]{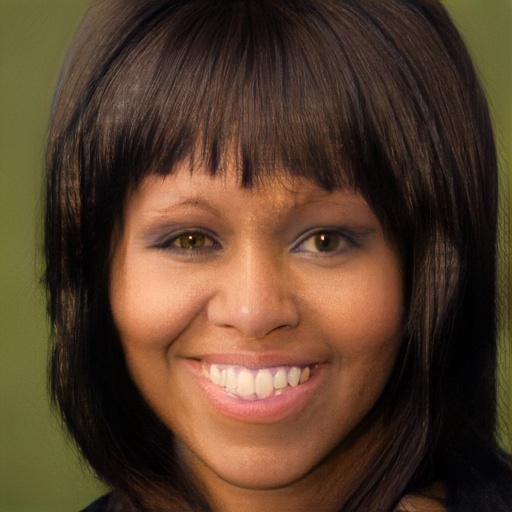} &
    		\includegraphics[width=\imwidth]{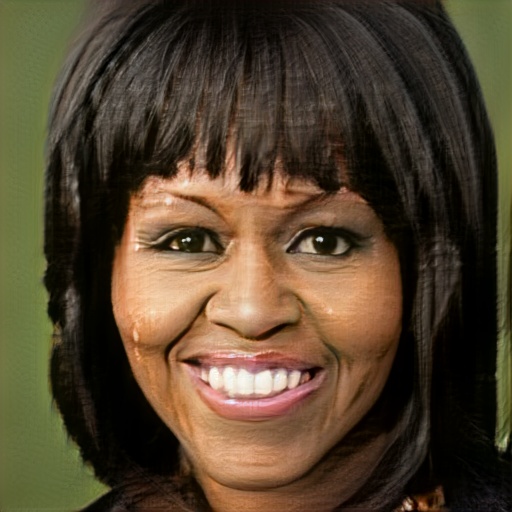} &
    		\includegraphics[width=\imwidth]{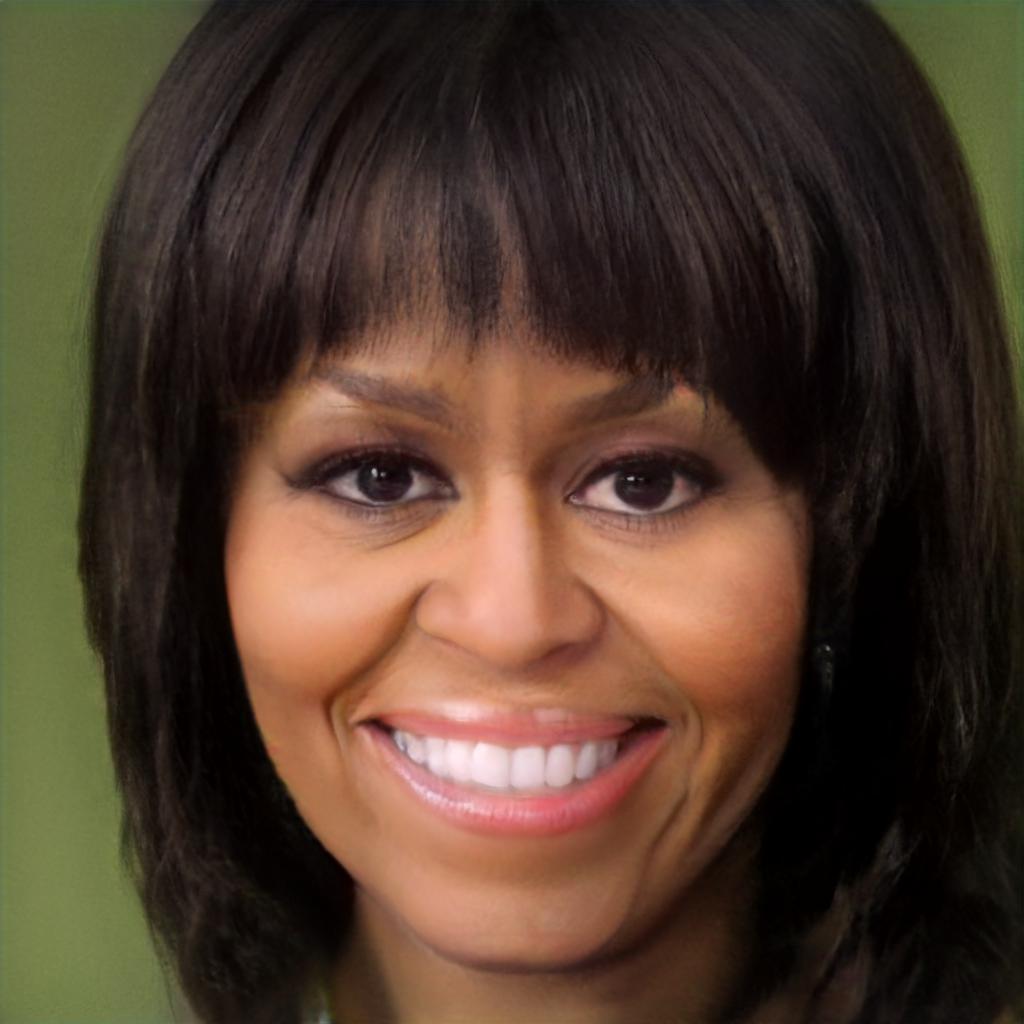} &
    		\includegraphics[width=\imwidth]{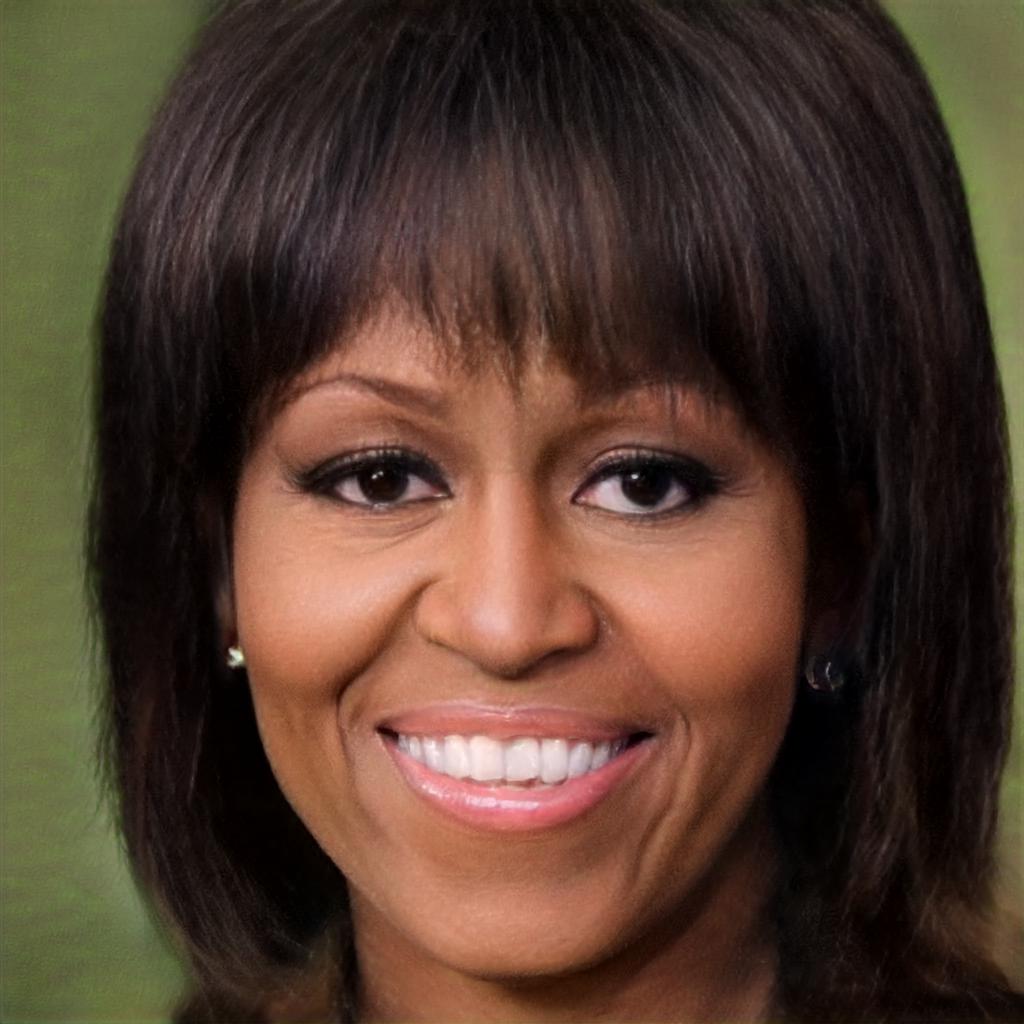}
            
            \\
            &
            \includegraphics[width=\imwidth]{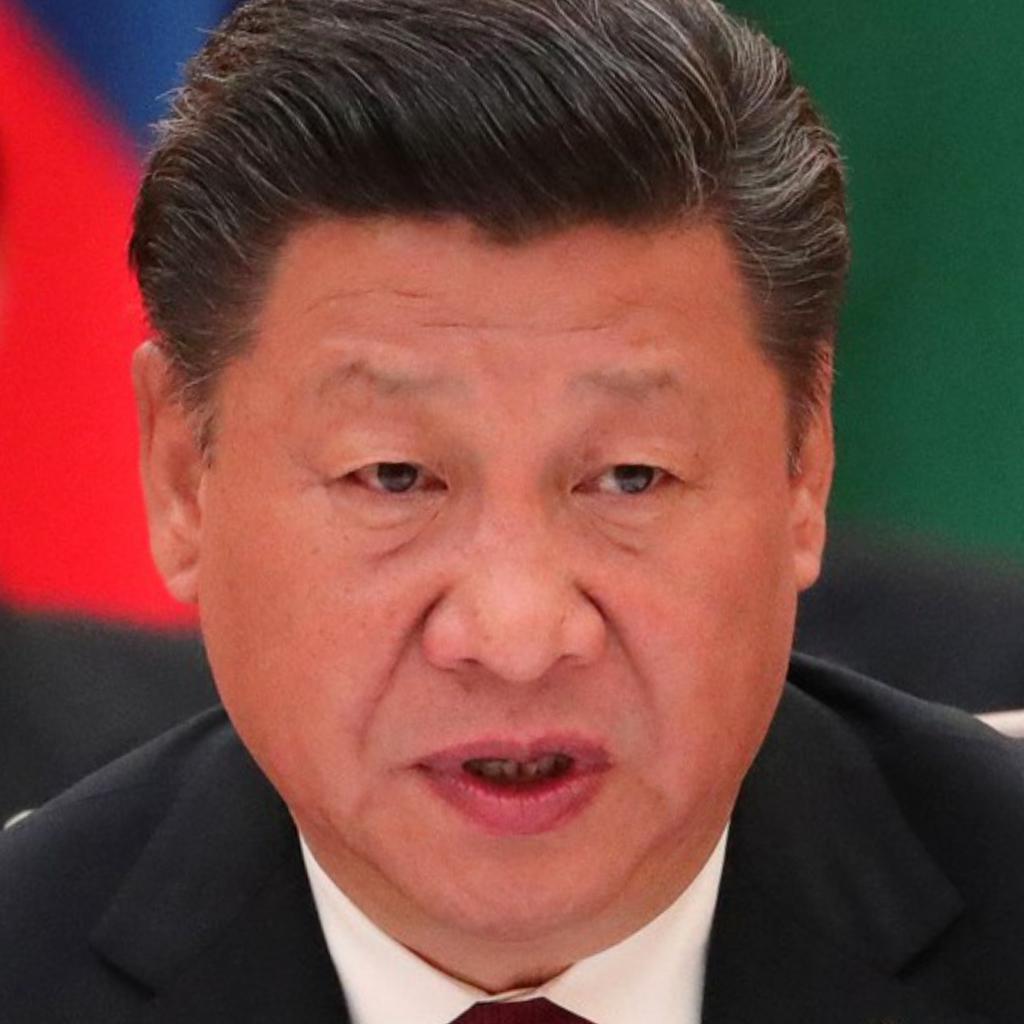} &
            \includegraphics[width=\imwidth]{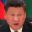} &
    		\includegraphics[width=\imwidth]{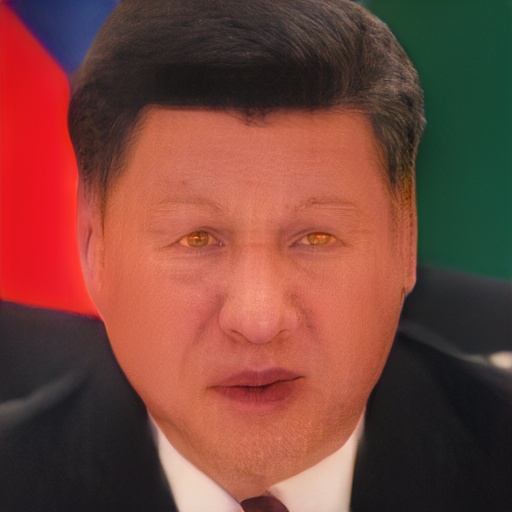} &
    		\includegraphics[width=\imwidth]{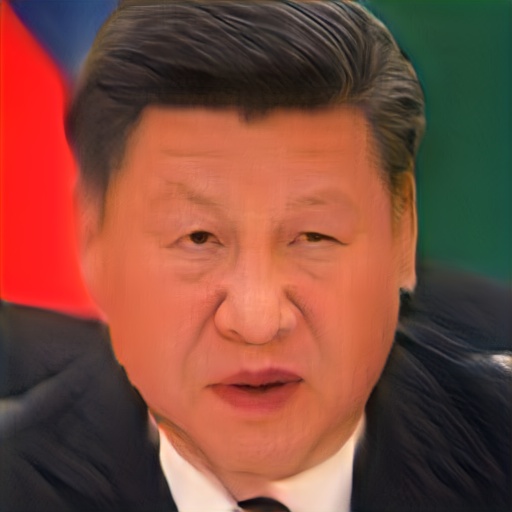} &
    		\includegraphics[width=\imwidth]{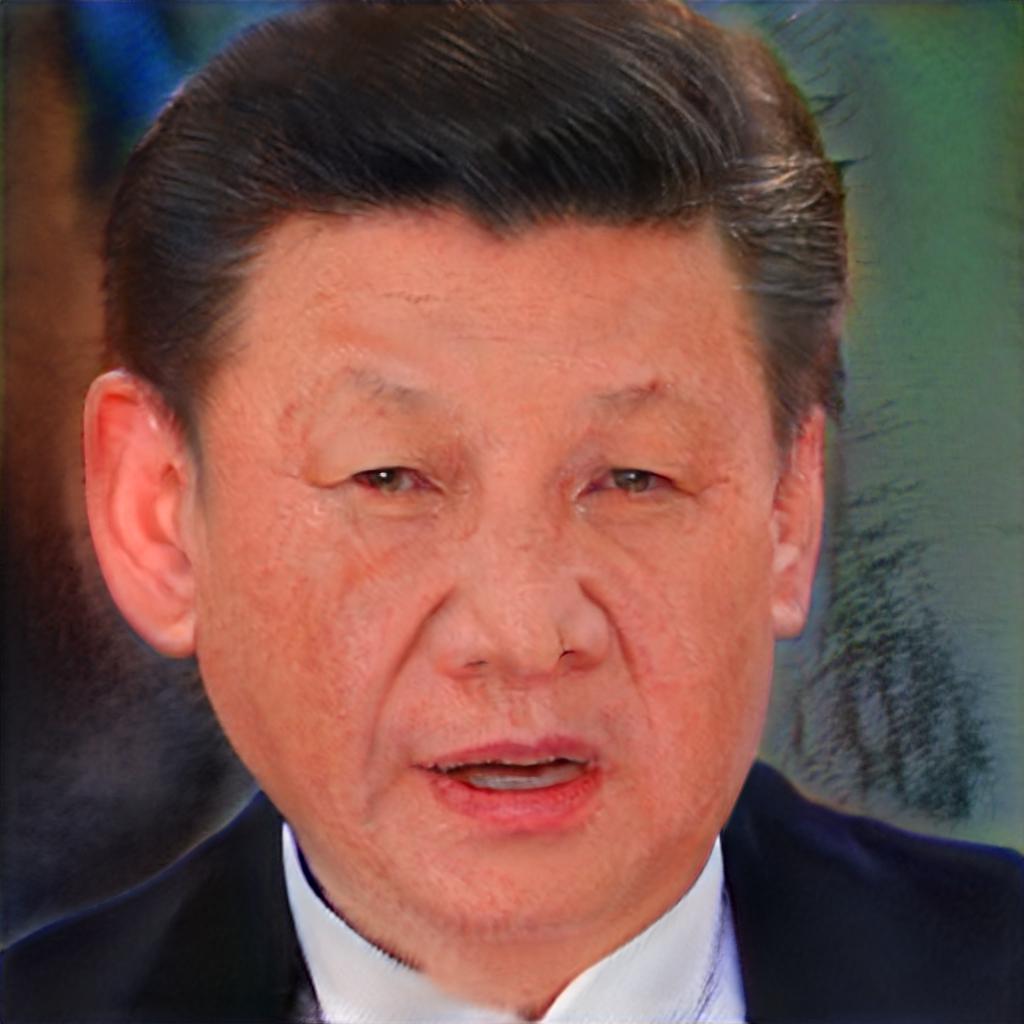} &
    		\includegraphics[width=\imwidth]{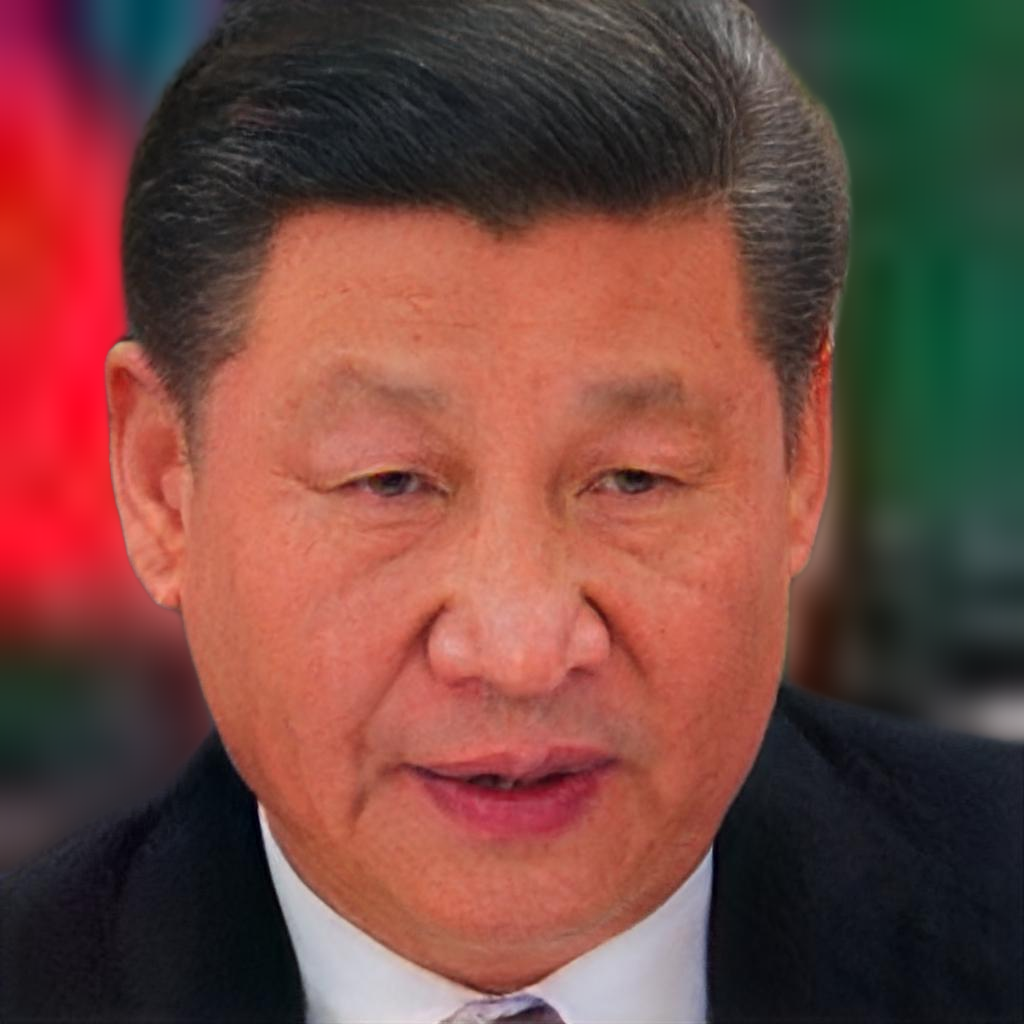}
    		
	        \end{tabular}
        \end{subfigure}
        
        \begin{subtable}{\linewidth}
            \begin{tabular}{clcccc}
                    \toprule
                    & Metric & \shortstack{Domain \\ Prior} & \shortstack{Domain \\ Prior+FT} & \shortstack{Diff-\\Augment} &  \shortstack{MyStyle \\ (Ours)} \\
                    \midrule
                    \parbox[t]{2mm}{\multirow{2}{*}{IP}} &
                    User \% ($\uparrow$) & 0.9 & 24.9 & 9.3 & - \\
                    & ID ($\uparrow$) & 0.55 $\pm$ 0.08 & 0.71 $\pm$ 0.08 & 0.68 $\pm$ 0.09 & \textbf{0.72 $\pm$ 0.08} \\
                    \midrule
                    \parbox[t]{2mm}{\multirow{2}{*}{SR}} &
                    User \% ($\uparrow$) & 0.3 & 2.2 & 4.6 & - \\
                    & ID ($\uparrow$) & 0.65 $\pm$ 0.09 & 0.75 $\pm$ 0.09 & 0.75 $\pm$ 0.05 & \textbf{0.81 $\pm$ 0.04} \\
                \bottomrule
            \end{tabular}
        \end{subtable}
    \end{center}
    \caption{Comparison of our personalized enhancement with alternative approaches. "Domain Prior`` refers to ComodGAN \cite{zhao2021comodgan} for inpainting and GPEN \cite{yang2021gan} for super-resolution. "+FT`` refers to fine-tuning these  methods on the personalized set. Zoom-in to better view fine details. 
    User study values reflect the percentages of responses in which the compared method was preferred over MyStyle.
    \camera{Top to bottom - \copyright EJ Hersom, \copyright Andrew Harrer, \copyright U.S. Government, \copyright Kremlin.}
    }
    \label{fig:enhancement_united_compare}
\end{figure}

\subsubsection{Super-Resolution (SR)}

The degradation transform $\phi$ is modeled by downsampling the input image by an $f \times f$ area kernel. 
We use GPEN~\cite{yang2021gan} as the state-of-the-art Domain Prior.
The size of input images is $32 \times 32$ pixels. 
DiffAugment and our method perform 32x upsampling while GPEN baselines perform an easier 16x upsampling in order to use the official model made available. 
We post-process the model's output by replacing its background regions, segmented using \citet{portraitmode}, with a Lanczos-upsampled version of the input image.
We compare the methods on images of Michelle Obama (279 images), Emilia Clarke (258 images) and Xi Jinping (92 images). We use $\beta=0.05$ in all experiments.

For both applications, quantitative and qualitative results are provided in \figref{fig:enhancement_united_compare}. Additional qualitative results are available in the supplementary. As can be seen, our results are significantly more faithful to the person's identity, have comparable fidelity and superior visual quality.

\subsection{Semantic Editing}
\label{subsec:app_edit}

We compare the performance of our personalized editing to the alternative prior of DiffAugment and a domain prior from an FFHQ-trained StyleGAN, as done by current state-of-the-art method, PTI \cite{roich2021pivotal}. 
Note that PTI has no trained weights, other than those of the GAN, and therefore there is no "Domain Prior+FT" baseline.
We compare only the process of applying an editing operator, with other factors held constant. We use the same pose and smile InterFaceGAN \cite{shen2020interfacegan} directions, and the same inversion method, PTI, with all generators.

\begin{figure}
	\begin{center}
    	\setlength{\tabcolsep}{1pt}
        \begin{subfigure}{\linewidth}
        	\setlength{\imwidth}{0.24\linewidth}
        	\begin{tabular}{*5c}
                \raisebox{3mm}{\rotatebox{90}{Head Pose}} &
        		\includegraphics[width=\imwidth]{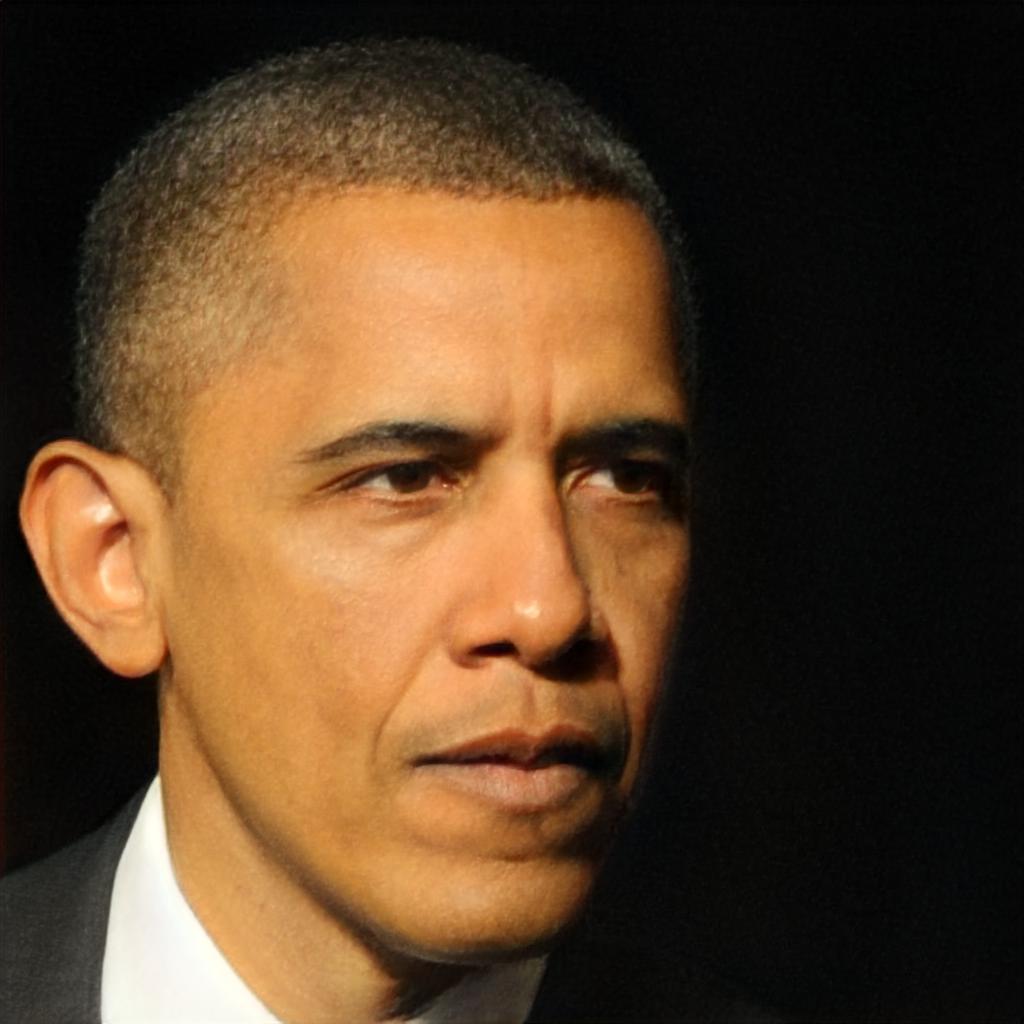} &
        		\includegraphics[width=\imwidth]{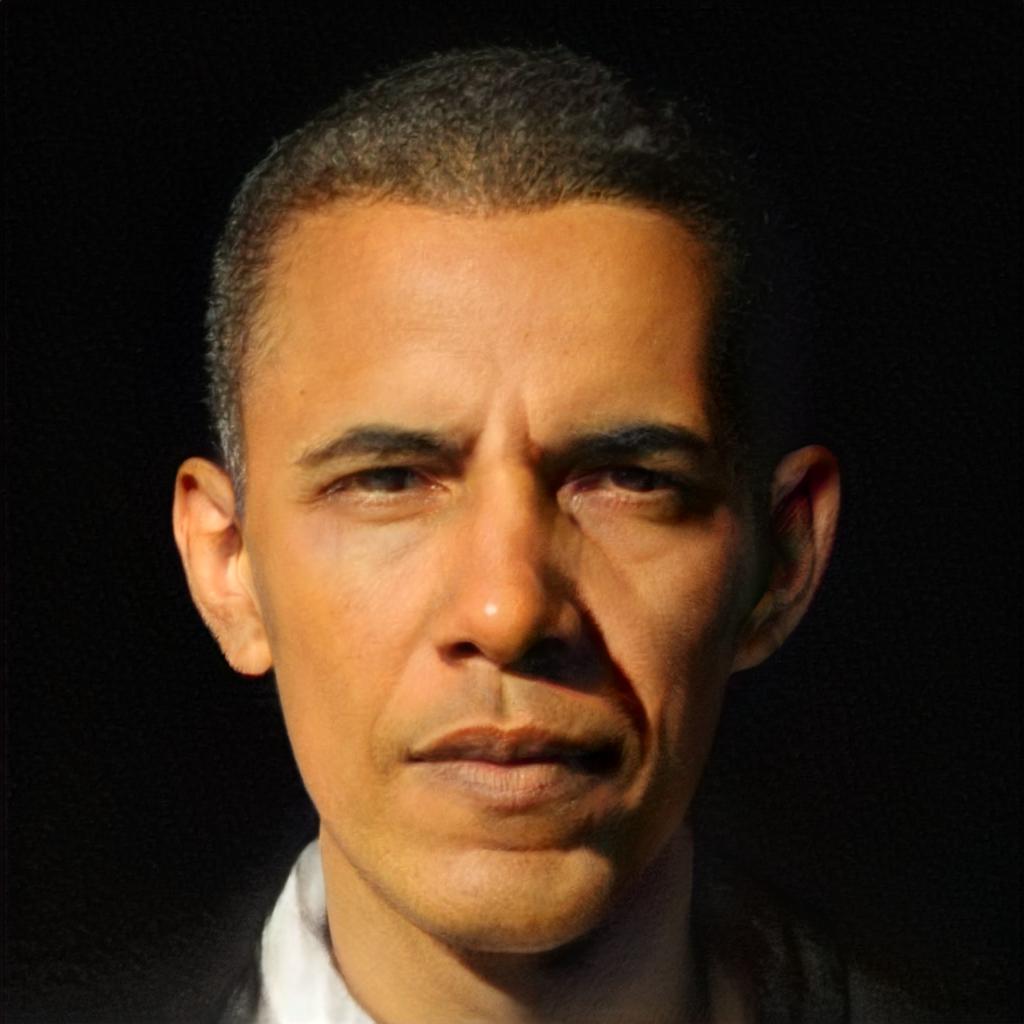} &
        		\includegraphics[width=\imwidth]{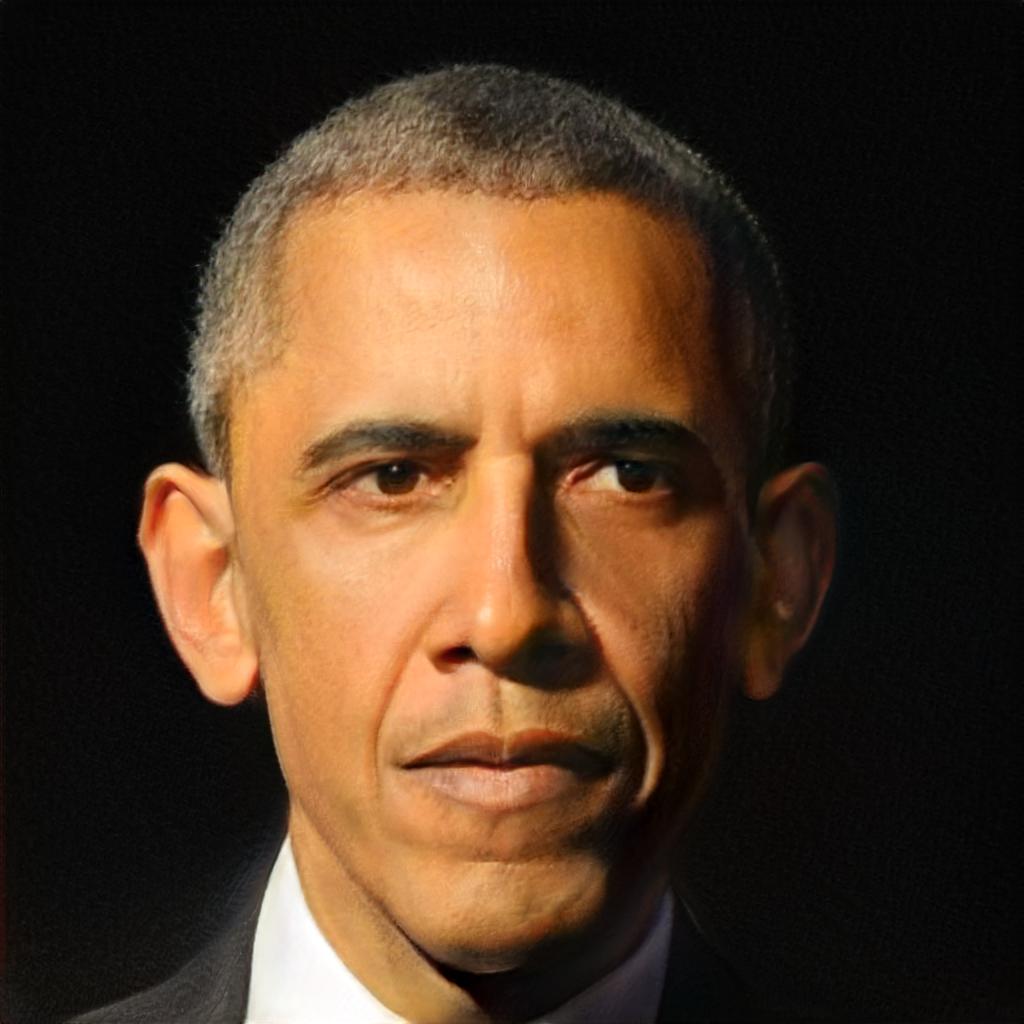} &
        		\includegraphics[width=\imwidth]{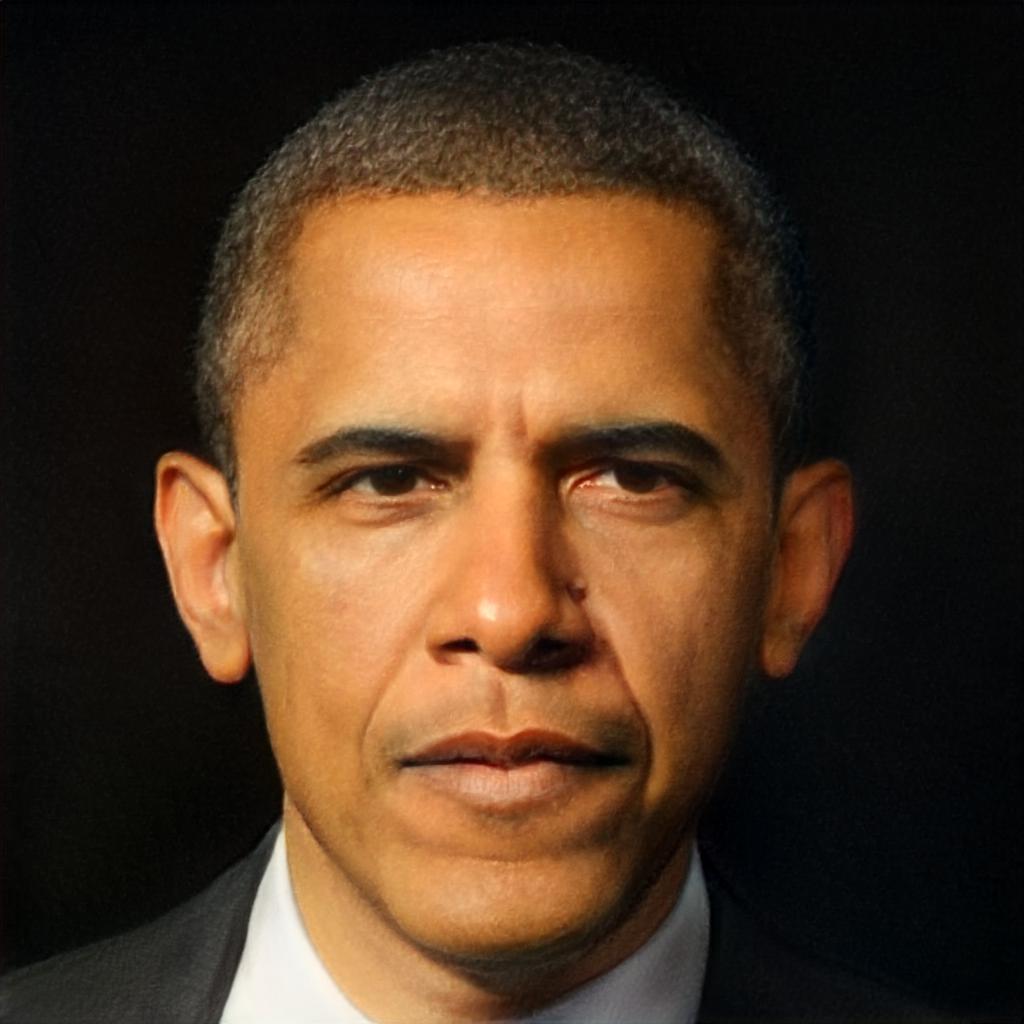} 
                \\
                
                \raisebox{7mm}{\rotatebox{90}{Smile}} &
        		\includegraphics[width=\imwidth]{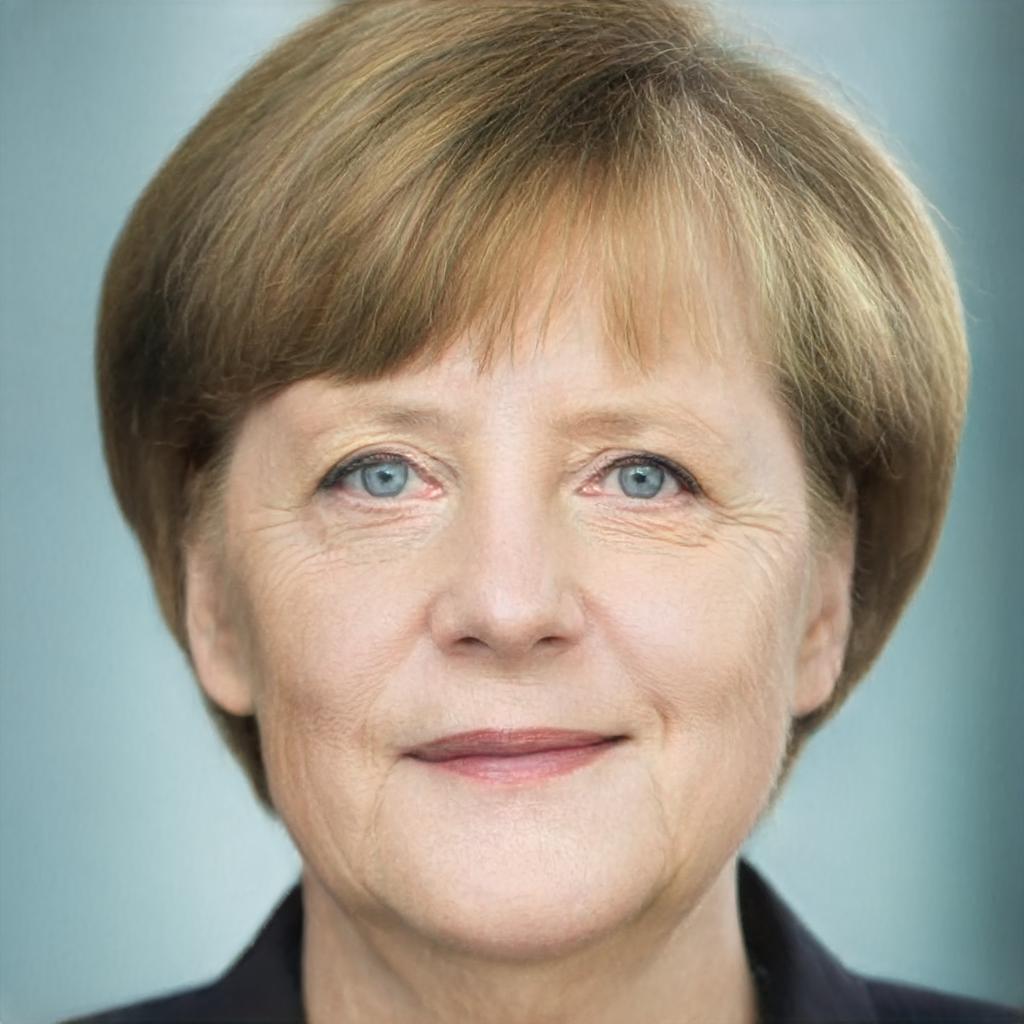} &
        		\includegraphics[width=\imwidth]{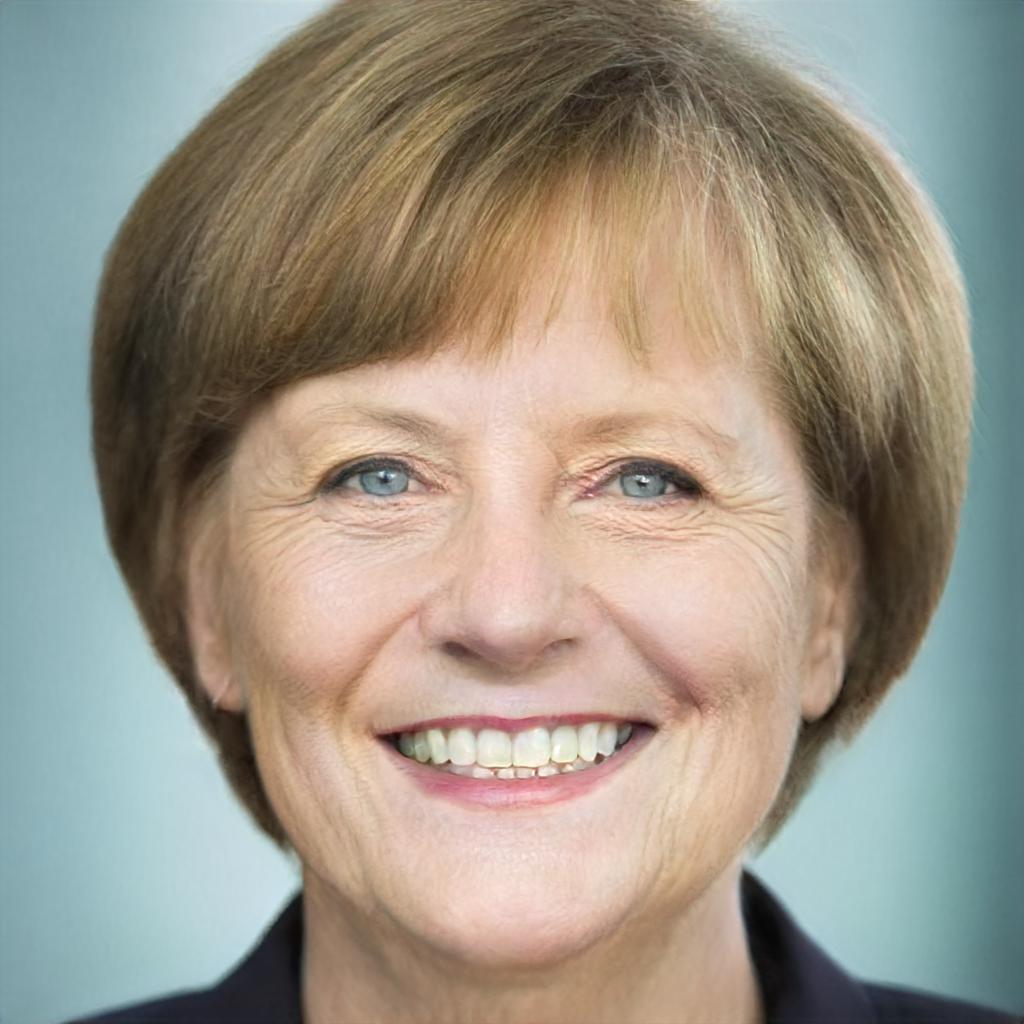} &
        		\includegraphics[width=\imwidth]{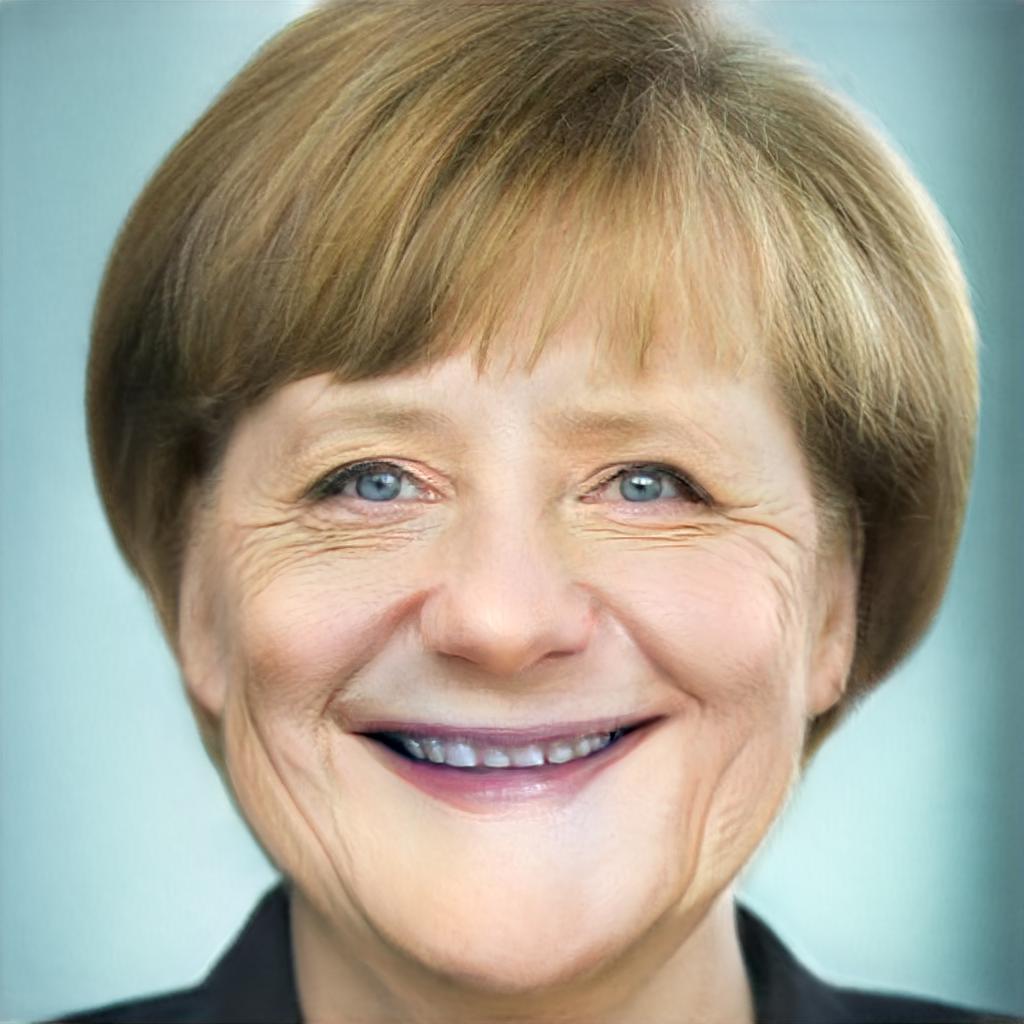} &
        		\includegraphics[width=\imwidth]{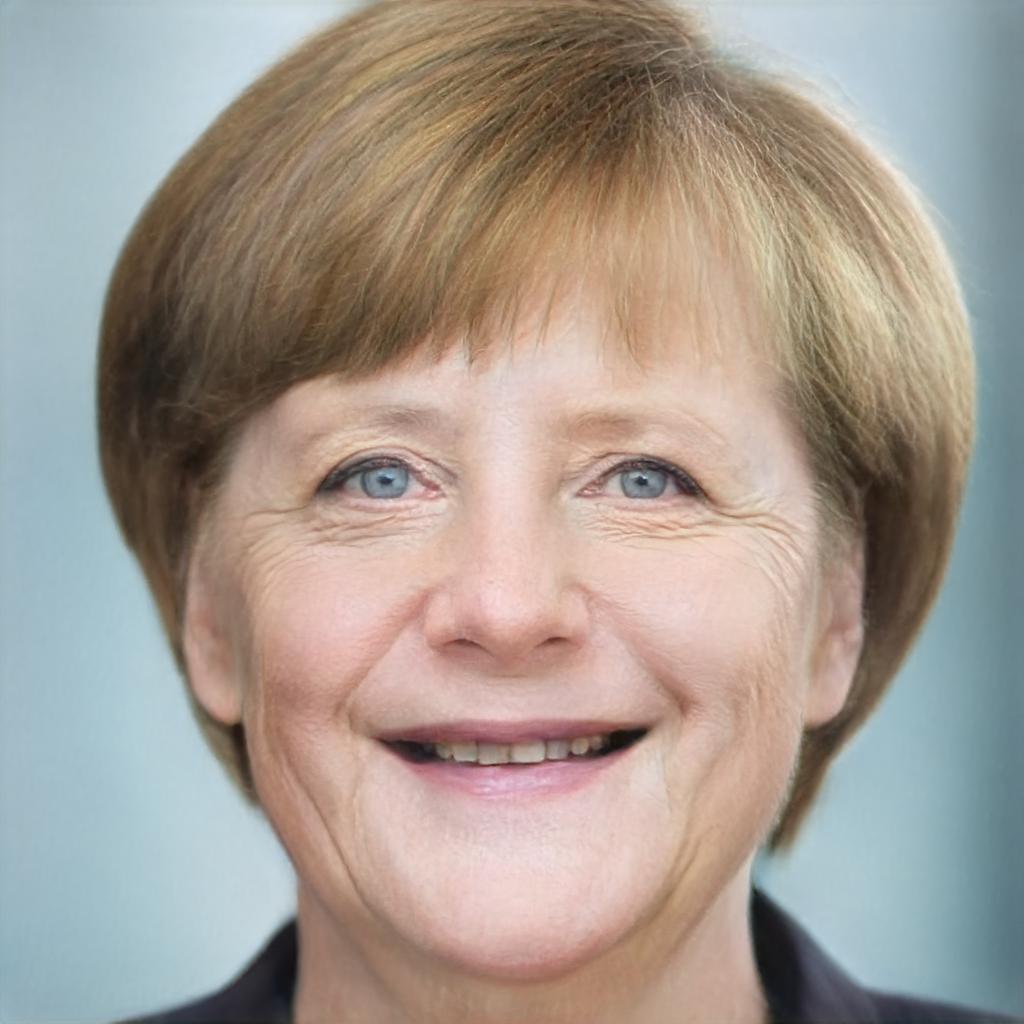}
        		\\
        		& \shortstack{Input (above) \\ \noindent\rule{2cm}{0.4pt} \\ Metric (below)} & \shortstack{FFHQ- \\ StyleGAN}  & \shortstack{Diff- \\ Augment} & \shortstack{MyStyle \\ (Ours)} \\
        		\midrule
        		& User \% ($\uparrow$) & 2.3 & 10.9 & - \\
                & ID ($\uparrow$) & 0.60 $\pm$ 0.08 & 0.66 $\pm$ 0.05 & \textbf{0.74 $\pm$ 0.04} \\
                \bottomrule
        	\end{tabular}
        \end{subfigure}%
    
	\end{center}
	\caption{
	Comparing editing performance with priors of different generators -- FFHQ-StyleGAN, tuned with DiffAugment, tuned with MyStyle (ours). ID score is reported for head pose.
	User study values reflect the percentages of responses in which the compared method was preferred over MyStyle.
    \camera{Top to bottom - \copyright Austen Hufford, \copyright CAS.}
	}
	\label{fig:editing_united_comparison}
\end{figure}

A sample of qualitative results are portrayed in \figref{fig:editing_united_comparison}.
One can observe the identity drift caused by editing head pose with baseline methods. The same phenomenon, while more nuanced, exists when adding a smile.
The domain prior edit affects the smile in a disentangled manner and maintains Angela Merkel's appearance elsewhere. Therefore, previous works have referred to such results \cite{roich2021pivotal, alaluf2021restyle} as successfully preserving identity. 
However, observing real images of Merkel smiling (see \figref{fig:merkel_smile_ref}), one can clearly notice that the smile added by the baselines is uncharacteristic to Merkel and therefore is not fully identity preserving. This is not surprising as the notion of a smile was learned from thousands of different individuals. Conversely, our prior has learned Merkel's smile and is thus more identity preserving.

We next quantitatively evaluate the performance of the methods for pose editing, for two individuals - Barack Obama and Joe Biden.
For every input and method, we create a large gallery of edited images with head pose varying between $(-40,40)$ degrees ~\cite{zhou2020whenet}. We then sample 5 equally spaced images. Similarly to previous experiments, ID metric and users' preference are reported in \figref{fig:editing_united_comparison}. As can be seen, our results are more identity preserving and strongly preferred by users.

\ifcamera\else
    \begin{figure}
	\centering
	\setlength{\tabcolsep}{1.5pt}
	\setlength{\imwidth}{0.32\linewidth}
	
	\begin{tabular}{ccc}
		\includegraphics[width=\imwidth]{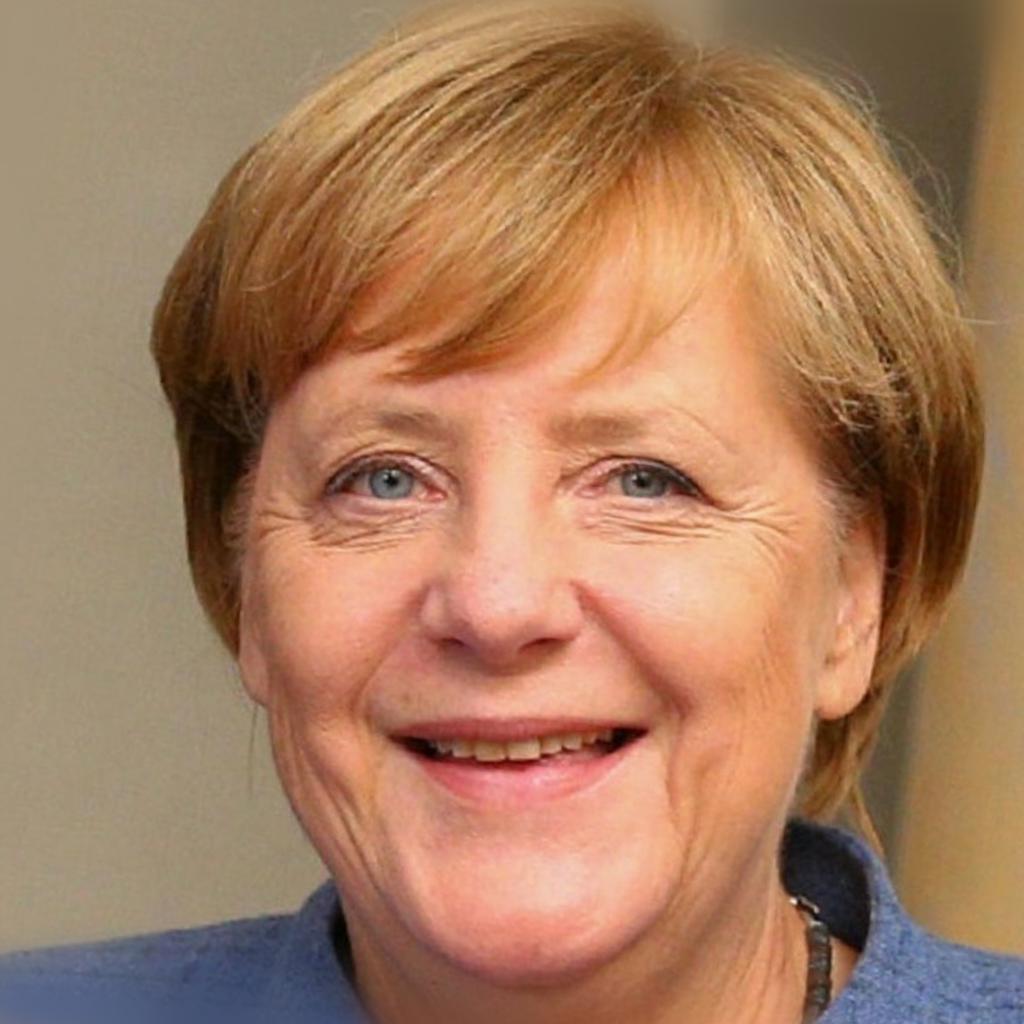} &
		\includegraphics[width=\imwidth]{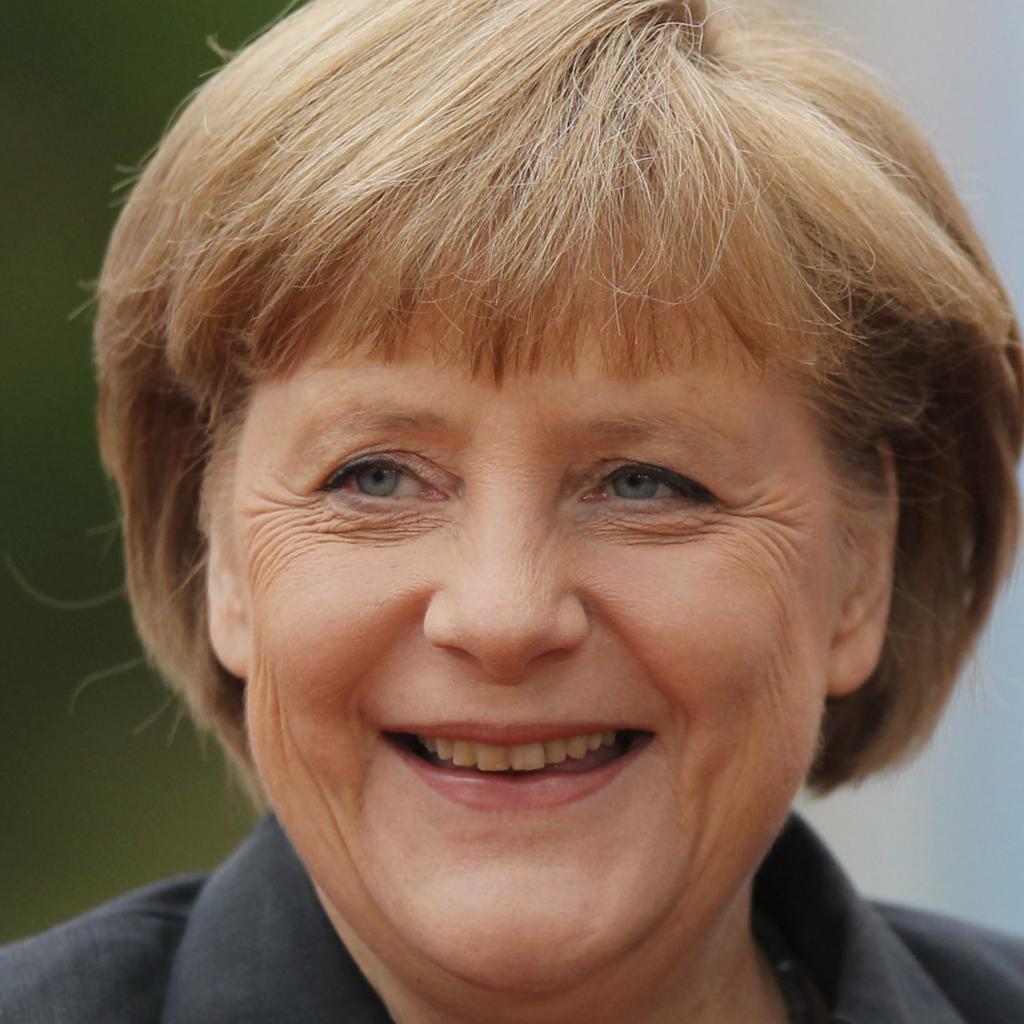} &
		\includegraphics[width=\imwidth]{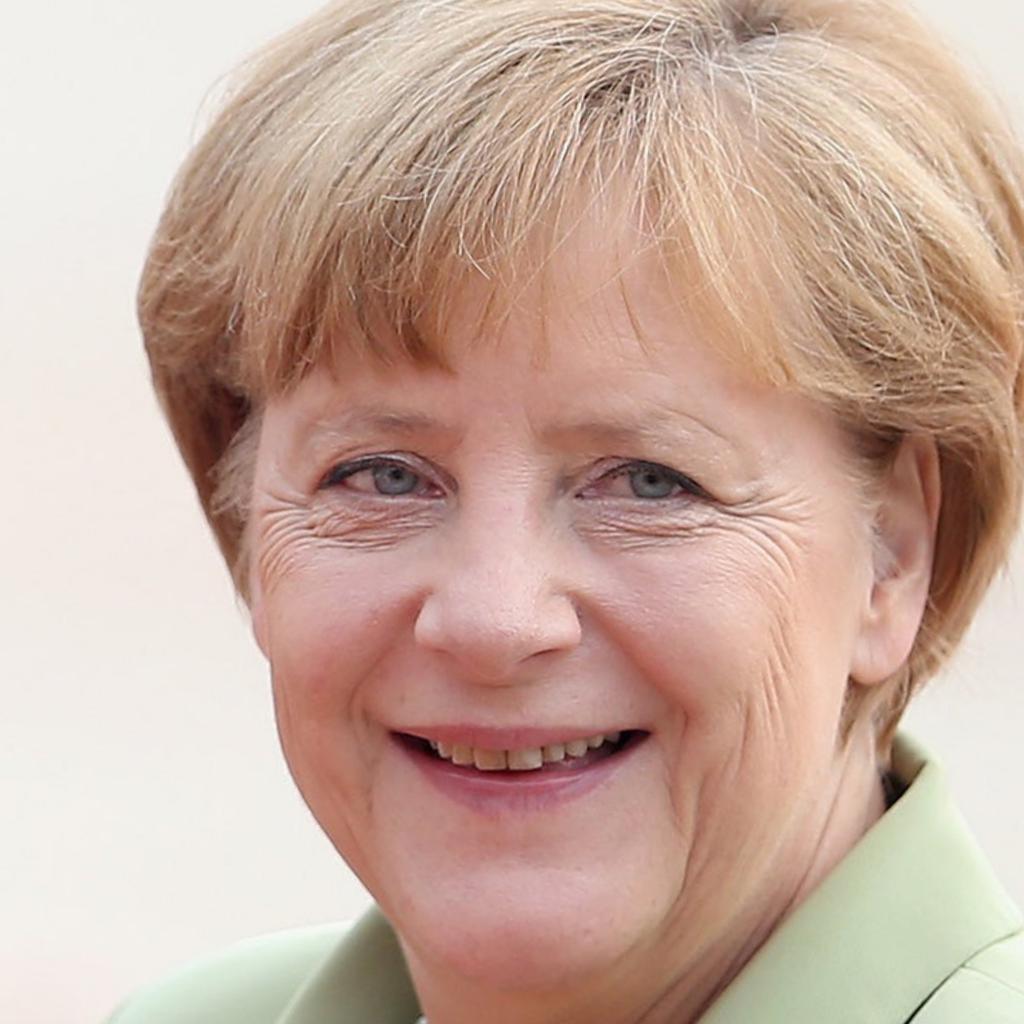}
	\end{tabular}
	\caption{
	    A sample of images of Angela Merkel smiling. 
	    Compare Merkel's real smile with the different editing results appearing in \figref{fig:editing_united_comparison}.
	}
	\label{fig:merkel_smile_ref}
\end{figure}

\fi
\section{Conclusions}

We have presented a new problem: forming a personalized prior, and introduced a method to achieve it through a few-shot tuning method. Our method takes a small sized reference set of photos, and learns a personalized prior represented in a subspace of StyleGAN latent space. We demonstrated that latent codes within that subspace represent images portraying the individual's unique identity.
We then developed methods to leverage the personalized prior for popular, ill-posed downstream tasks -- image enhancement and semantic editing. Our results were shown to be effective at solving the tasks while crucially preserving the identity of the individual, significantly improving upon state-of-the-art baselines. 

\begin{figure}
    \centering
    \includegraphics[width=\linewidth]{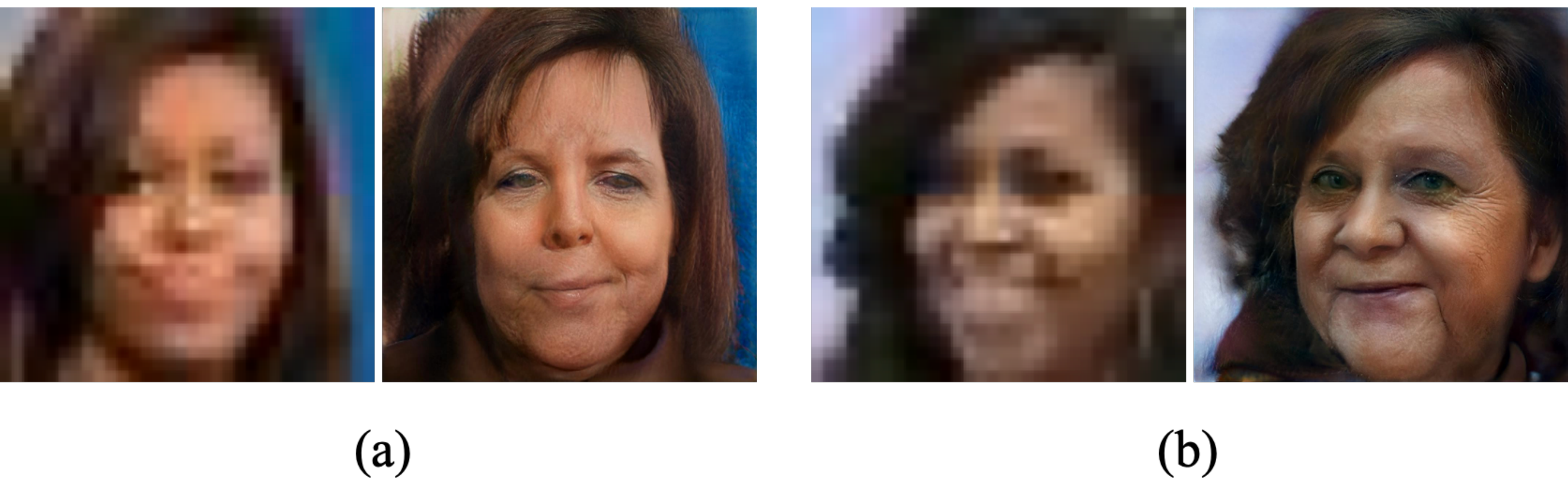}
    \caption{Failure case - enhancement of degraded images of Michelle Obama (left) using a prior learned on Angela Merkel's reference set. It can be seen that due to the mismatch between the key facial characteristics of the person depicted in the reference set and the attributes of the person in the degraded image, the output images (right) contain visual artifacts and depict a person that does not adequately resemble any of the individuals.}
    \label{fig:failure_deep_fake}
\end{figure}

Our method, however, has a few limitations. Using a generative prior, whether personalized or not, to represent images that are out of the training set's distribution remains challenging, and requires a compromise between quality and fidelity. Common examples of such out-of-distribution images are faces in extreme poses or faces that are partially occluded by accessories.
Another example, of specific importance to this work, is the case in which the input image depicts a different person than that captured by the prior. In this case, the results may not adequately resemble any of the individuals, as demonstrated in \figref{fig:failure_deep_fake}. 

\begin{figure}
	\centering
	\includegraphics[width=\linewidth]{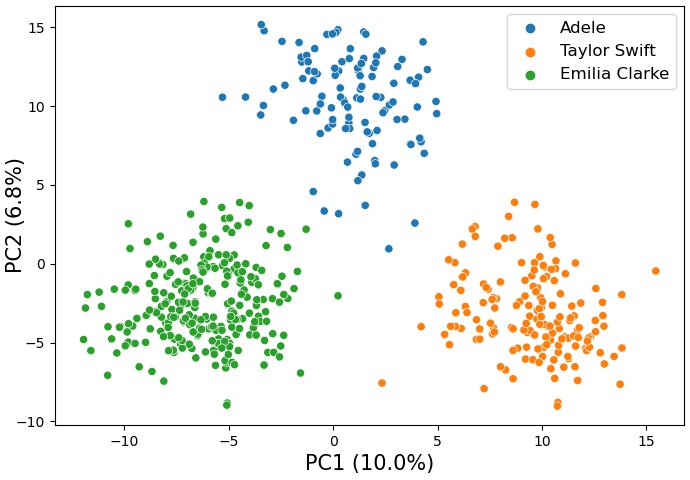}
	\caption{
	Identity-based clustering in latent space. We perform PCA (two components) on the anchors of three different, relatively similar, individuals. As can be seen, the anchors are strongly grouped according to identity. This is a strong indication that one can form multiple disjoint personalized priors within the same generator.}
	\label{fig:anchors_pca}
\end{figure}

As future work, it could be fruitful to model multiple individuals in a single generator (see \figref{fig:anchors_pca}) and to use a dedicated axis to model variations in appearance that vary through time, such as hairstyles. 
We also believe that such restricted sub-spaces, can be effective for more tasks, beyond identity preservation, and can help direct or regularize various tasks, such as articulation of human poses, expressions of face, or shapes of chairs.

\begin{acks}
We thank Rinon Gal, Or Patashnik, Amit Attia, Yuval Alaluf, Assaf Shocher, and Michael Rubinstein for reviewing early drafts and suggesting improvements. We thank Yogev Nitzan for his encouragement and Wei Xiong for help running ComodGAN baselines. This work was partially supported by the Israeli Science Foundation (3441/21, 2492/20).
\end{acks}

\vfill\eject
\pagebreak

{\small
\bibliographystyle{ACM-Reference-Format}
\bibliography{main}
}

\appendix
\section{Personalized Dataset}
\label{sec:dataset}

\subsection{Dataset Curation}
\label{sec:dataset_curation}
We collected a new dataset that contains sets of face images of celebrities. For each celebrity, we first curated a random set of images with resolution greater than $1024 \times 1024$. 
We first performed automatic filtering to remove images that were in gray scale, had low quality or the featured face was too small. 
We also used a face recognition network \cite{deng2019arcface} to remove images of undesired individuals.
We then used Amazon Mechanical Turk to further filter  images. Workers were instructed to flag images that do not match the identity of the wanted celebrity, images that contain occlusions of the face area, images with low resolution and watermarked images. 
Furthermore, to remove duplicated face images we filtered out images whose features extracted by arcface had a cosine similarity greater than 0.9.
Finally, the faces were aligned by the alignment process presented by Karras \etal \shortcite{karras2017progressive} and separated into training sets and test sets for each celebrity. The number of images that were included in the test set and the training set of each celebrity presented in \tblref{tab:dataset_stats}.

\begin{table}[tbh!]
    \begin{center}
        \caption{The sizes of the training and test sets of our dataset.}
        \label{tab:dataset_stats}
        \begin{tabular}{lcc}
            \toprule
            Celebrity &  Training set size & Test set size \\
            \midrule
            Adele & 109 & 8\\
            Angela Merkel & 145 & 10\\
            Barack Obama & 192 & 13 \\
            Dwayne Johnson & 97 & 11 \\
            Emilia Clarke & 258 & 11 \\
            Jeff Bezos & 114 & 3\\
            Joe Biden & 206 & 13\\
            Kamala Harris & 110 & 7 \\
            Lady Gaga & 133 & 6 \\
            Michelle Obama & 279 & 9\\
            Oprah Winfrey & 135 & 9\\
            Taylor Swift & 158 & 11\\
            Xi Jinping & 92 & 15 \\
            \midrule
            \bottomrule
        \end{tabular}
    \end{center}
\end{table}

\subsection{Considerations of Data Collection}
Images of different individuals may contain different attributes (\eg eyeglasses) and variations (\eg hair styles). %
Although our method has shown robustness to a variety of datasets in all experiments, we explicitly note that there exists datasets for which our method would function improperly or not ideally. Like any data-driven solution, our prior is bound to what it sees during training. 
Consider for example, if one were to use a dataset with poor variation, such as one composed of sequential video frames. In this case, the personalized prior would have an incredibly restricted understanding of the individual's appearance. One cannot expect to perform inpainting, super-resolution and editing with such a prior, as the fidelity would inevitable be poor.

Another case to consider is that of specific images from the dataset being "bad``. In theory, many forms of images can damage the prior. In fact, the purpose of the manual and automatic data filtering is to prevent the inclusion of such images. Beyond filtering mistakes, undesired images can still find their way into the dataset in the form of non-genuine images of the individual (\eg edited via Photoshop), images in which the identity is obscure (\eg wearing heavy make up, extreme head pose), etc.
We empirically observed our method was successful on the collected datasets, although they are not guaranteed to be sterile, and might contain undesired images. 

We believe that extending our filtering with further criteria could benefit the method. A specifically interesting avenue of future investigation is filtering related to StyleGAN. For example, one could filter out images that are not well reconstructed during inversion or tuning. A similar approach was recently proposed for training GANs \cite{mokady2022selfdistilled}.

\section{Additional Details}

\subsection{Sampling From $\mathcal{P}_{0}$}
\label{subsec:sampling}
Commonly, sampling from generative models is trivial as the probability density over the latent space is known. Conversely, $\mathcal{P}_{0}$ is defined as the area enclosed by a convex hull. We therefore require a sampling strategy to sample from $\mathcal{P}_{0}$. We propose a simple protocol where we uniformly sample a random vector from $\mathcal{A}_{0}$, in which there are three non-zero entries. This is trivially done by sampling three anchors and three scalars from $[0,1)$ which are then normalized. From that vector, we obtain the latent code in $\mathcal{P}_{0}$.
We note that there exists a large body of works on sampling from convex bodies~\cite{hit-and-run}, which may produce superior results. However, this investigation is out of the scope of this paper.

\subsection{Projection-based Enhancement}
\label{subsec:domain_enhancement_bg}
As discussed in \secref{subsec:applying}, our enhancement method in an adaption of an approach used by previous works \cite{menon2020pulse, luo2020time}. In sake of completeness, we next lay out the method used by these works.

As input, these works take a degraded image $I$ and access to a differentiable function simulating the degradation, $\phi$. They then devise methods to find the latent code from which StyleGAN produces an image, that after degradation with $\phi$ reconstructs $I$, most accurately. Formally, they seek for
\begin{equation}
\label{eq:normal_projection}
    w^* = \argmin_w \mathcal{L}((\phi \circ G)(w) , I),
\end{equation}
where $\mathcal{L}$ is some reconstruction loss. 
To ensure the high fidelity of the projection, such works use the \wplus space \cite{abdal2019image2stylegan} which is virtually infinitely expressive. But for that reason, they also need to regularize the $w^+$ code, preventing the simple solution of degradation appearing in $G(w)$. Finally, they take the enhanced output image to be $I_e = G(w^*)$. 

\subsection{Implementation Details}
All projection experiments, including inversion, inpainting and super-resolution inherit the setting and hyperparameters from StyleGAN2 projection code \cite{karras2019style}. This includes the optimizer, number of steps, learning rate schedulers, etc. 
Similarly, the tuning process inherits its hyperparameters from PTI \cite{roich2021pivotal}.
For both tuning our model and projecting into it, we observe no need for early-stopping. On the contrary, all projections to DiffAugment were stopped after 200 iterations. We observe that more iterations lead to significant degradation in their results.

Hyperparameters values used in experiments -- $s=100$ in \eqnref{eq:softplus} of the main paper, $\lambda_{L_2} = 1$ in \eqnref{eq:training_loss} of the main paper, $\lambda_{d-reg} = 10$ and applied only on the first 12 layers in \eqnref{eq:delta_reg} of the main paper.

Tuning our model takes roughly 48 seconds per anchor on a single V100 GPU. This translates to roughly 80 minutes for 100 anchors.

Previous methods performing optimization in latent space \cite{abdal2019image2stylegan} have found it beneficial to initialize the optimization to the mean code in latent space -- $\bar{w}$. Similarly, we initialize our optimization to the center of \p. This corresponds to initializing $\vect{\alpha}$ to a vector whose all components are equal.

For DiffAugment synthesis in \secref{subsec:synth_comparison}, we truncate the sampled latent code with $\psi=0.7$~\cite{brock2018large}, which improves its results significantly.

\subsection{Editing - Projecting back to \pbeta}
\label{subsec:editing_projection}
Continuing from \secref{subsec:editing}, we next elaborate on some considerations of choosing whether to stop or continue editing once the allowed dilation is reached.
Continuing to traverse the linear direction causes the latent codes to drift further away from \pbeta. As we demonstrated in \secref{sec:dch_analysis}, this causes the personalization to vanish and is thus undesired.

The user is presented with a choice. One option is to simply stop at the allowed dilation. However, the user might not be satisfied with the extent of allowed editing. The alternative is to continue editing but project the latent code back to \pbeta, using a simple convex optimization problem. We note that the projected latent code no longer follows the linear direction $proj_{\textit{V}}(\vect{n})$. Therefore, it might degrade the disentanglement of editing by affecting other properties. 
In practice, since $\vect{n}$ is itself not perfectly disentangled, we found that the degradation of disentanglement caused by projection is acceptable.

\begin{figure}
    \centering
    \includegraphics[width=\linewidth]{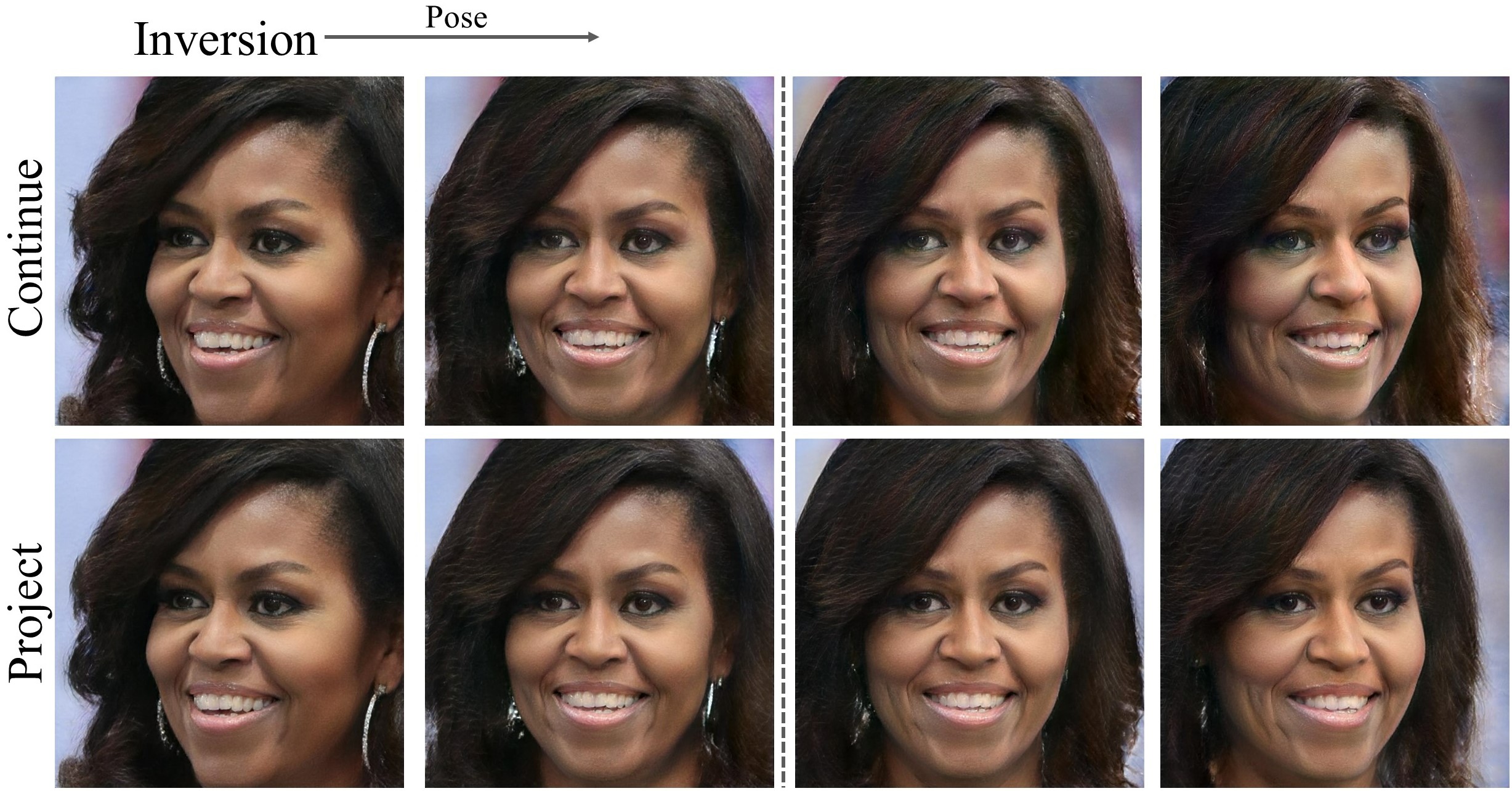}
    \caption{Editing head pose using two different strategies. Top row continues editing along the linear direction. Bottom row, projects the edited latent code back to \pbeta if the latent code has traversed past $\beta$-dilation, which is represented with a dashed line. As can be seen, simply continuing to edit causes artifacts as a result of the vanishing prior. Stopping at the allowed dilation assures high quality and identity preserving image, but the extent of editing might not be sufficient. Projecting back to \pbeta allows to continue editing while enjoying the good nature of the prior.}
    \label{fig:editing_projection}
\end{figure}

A sample of qualitative results is provide in \figref{fig:editing_projection}. As can be seen, following the projection strategy, the user gained the ability to edit the head pose to a greater extent while maintaining high visual quality and personalization. The inversion in this experiment is obtained using PTI \cite{roich2021pivotal} and the allowed dilation is $\beta=0.02$.

\subsection{Anchors' Linear Independence}
In several occasions throughout the paper, we have implicitly assumed that the obtained anchors are linearly independent. We first note that in practice, this was the case in all of our experiments. 
However, this might not always be the case, specifically as $N$ increases and becomes close to $k$ or surpasses it. 
Linearly dependent anchors have no strong impact on our method. One can drop the anchors that are internal to the convex hull and obtain a linearly independent set spanning the same \p. Doing so is mostly useful as a means to reduce the dimension of \aspace and for projecting to the span of the anchors. Synthesis and projection may seamlessly operate with a dependent set of anchors.

\section{Additional Experiments}
\label{sec:more_ablation}

\subsection{Further Analysis of Identity Preservation During Interpolations}
\label{subsec:supp_dch_analysis}

\begin{figure}
    \centering
    \includegraphics[width=\linewidth]{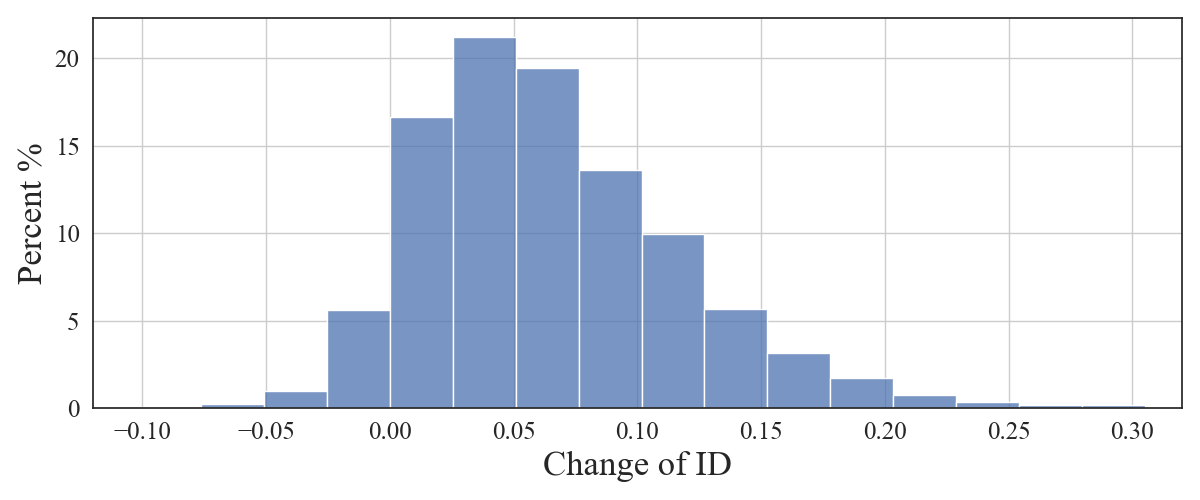}
    \caption{Distribution of change in ID score during interpolations, with respect to the minimal anchor's ID. One can observe that most latent codes (93.3\%) exhibit an improvement in ID. The maximal decrease is 0.1 and only 0.2\% exhibit a decrease of more than 0.05, which is arguably negligible. We conclude that observing a decrease in identity preservation during interpolating is rare and insignificant.}
    \label{fig:supp_change_of_id}
\end{figure}

\begin{figure}
    \centering
    \begin{subfigure}{\linewidth}
    \includegraphics[width=\linewidth]{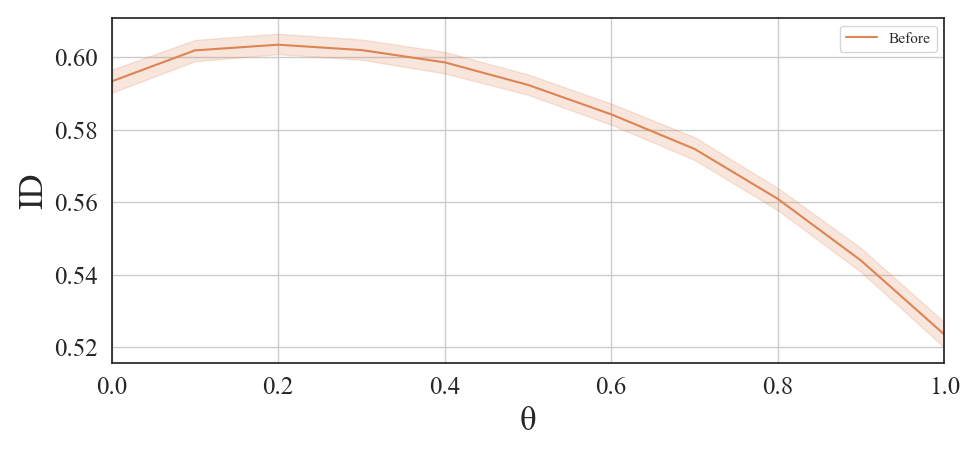}
    \end{subfigure}
    \\
    \begin{subfigure}{\linewidth}
    \includegraphics[width=\linewidth]{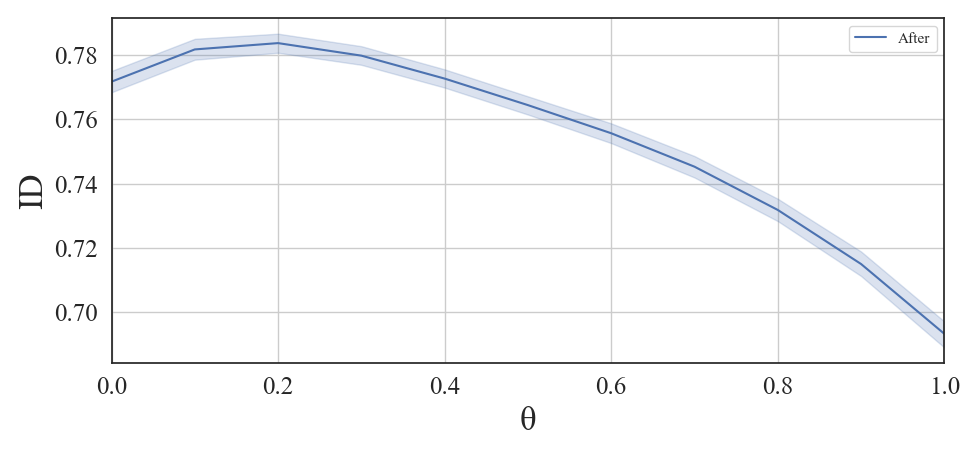}
    \end{subfigure}
    
    \caption{Evaluating ID, while interpolation from a more identity preserving anchor ($\theta=0$) to a less identity preserving anchor ($\theta=1$). Evaluation is performed in the generator {\color{orange} before} and {\color{blue} after} tuning. Perhaps surprisingly, for a substantial duration of interpolation ($\theta < 0.4$) ID increases despite interpolating towards a less identity preserving anchor. This phenomenon is observed both before and after tuning, and is hence probably unrelated to tuning.
    }
    \label{fig:supp_inter_to_better}
\end{figure}

\secref{sec:dch_analysis} and \figref{fig:analysis} of the main paper indicate that as a result of tuning, interpolations between anchor pairs are identity preserving. This was demonstrated using the mean and standard deviation of ID along the interpolation. \Ie the results thus indicate what happens in the "average case``, but not under extreme circumstances.

We next perform additional, more detailed, experimentation under the same setting. The reference for evaluating interpolation are the anchor endpoints. Ideally, we would like interpolated codes to be at least as ID preserving as the less ID preserving anchor.
We thus compute the difference between the ID score of every interpolated image, from its less ID preserving anchor. The distribution of this difference is portrayed in \figref{fig:supp_change_of_id}.

As can be seen, only a relatively small percentage ($6.7\%$) of interpolated latents have a smaller ID score then both their anchors. 
One should also consider the magnitude in which the ID score lessens. We find that the maximal decrease is $0.1$ while only $0.2 \%$ exhibit a decrease of more than $0.05$. Such differences are arguably negligible, and well within the margin of error face recognition networks make for the same individual. In fact, the mean ID score of Obama's test set with respect to the same train set is $0.73$.

Perhaps surprisingly, $46.7 \%$ of the interpolated images have ID scores that are greater than both anchor endpoints. The commonness of this phenomenon also makes it observable in the mean trend of data, displayed in \figref{fig:analysis} of the main paper.
To better understand this phenomenon we perform another experiment. We sort all the interpolations such that the more identity preserving anchor of the two is the beginning point of traversal ($\theta = 0$).

In \figref{fig:supp_inter_to_better}, we report the mean ID preservation along the interpolation in the generators before and after tuning. We find that, both before and after tuning, for a substantial duration of the interpolation, the ID scores improve. This is arguably surprising, as the interpolation is headed towards a less identity preserving anchor. We propose a theory for this phenomenon. Regardless of the personalization of each anchor, the interpolation aids in smoothly averaging out extreme properties. If only one of the anchor images depict an extreme expression or illumination, it would gradually vanish throughout the interpolation. Such averaging yields more conservative looking face images and would therefore be associated with greater ID scores.

The remaining $46.6 \%$, exhibit an ID score which is between the two anchors.

\subsection{Additional analysis of the prior-expressiveness tradeoff}
\label{subsec:beta_tradeoff_appendix}

\begin{figure}
	\centering
	\setlength{\tabcolsep}{1pt}
	\setlength{\imwidth}{0.22\linewidth}

    \begin{tabular}{*6c}
		& & Input & $\beta = 0$ & $\beta = 0.02$ & $\beta$ unbounded \\ 
		
		\multirow{2}{1.5em}[1.3ex]{\rotatebox[origin=c]{90}{Inversion}} &
		\rotatebox{90}{\phantom{kk} Typical} &
        \includegraphics[width=\imwidth]{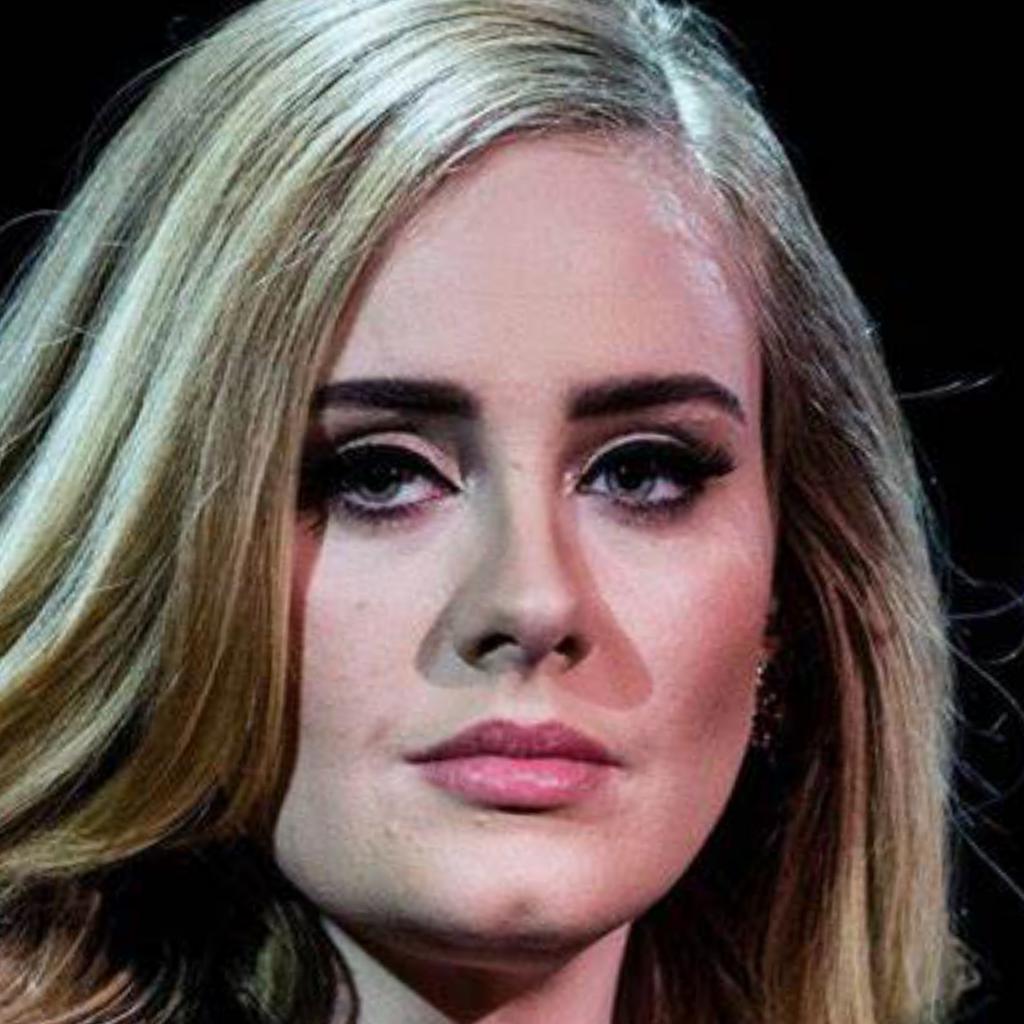} &
		\includegraphics[width=\imwidth]{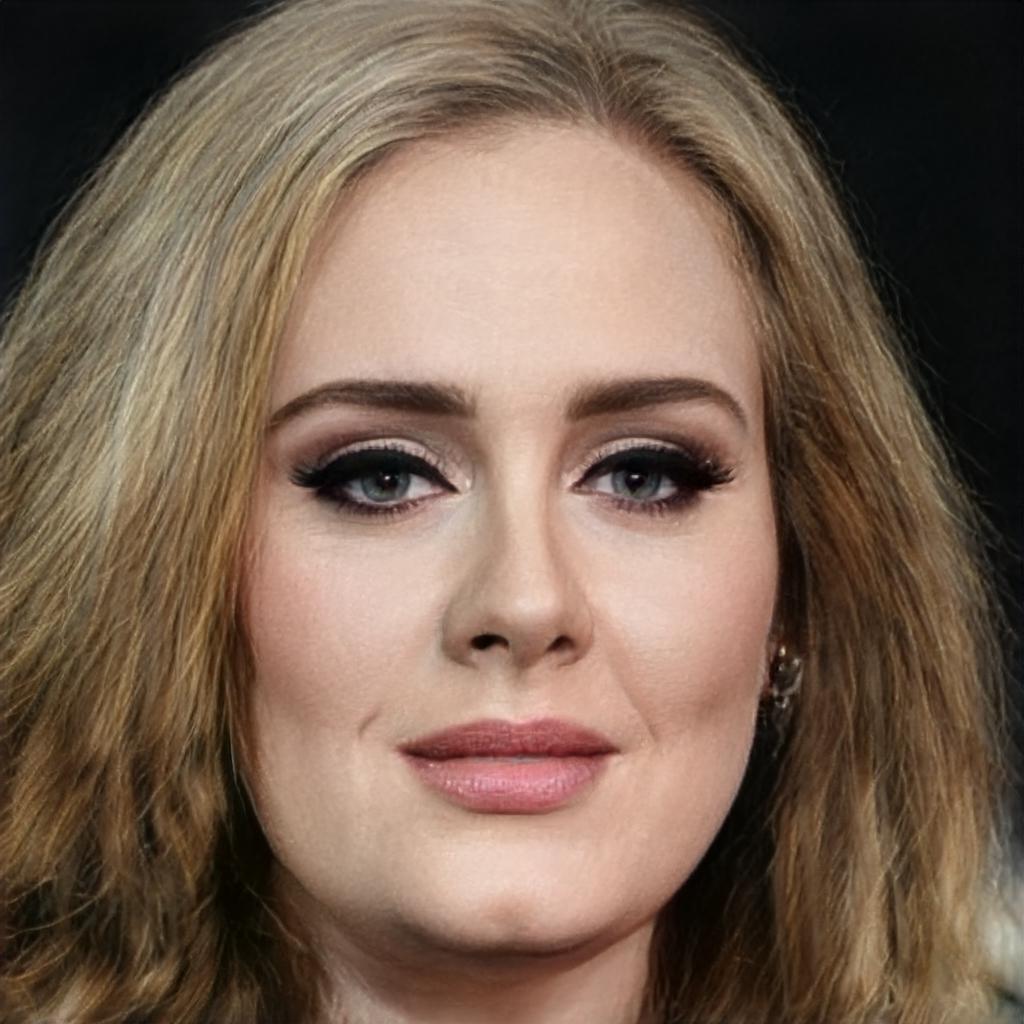} &
		\includegraphics[width=\imwidth]{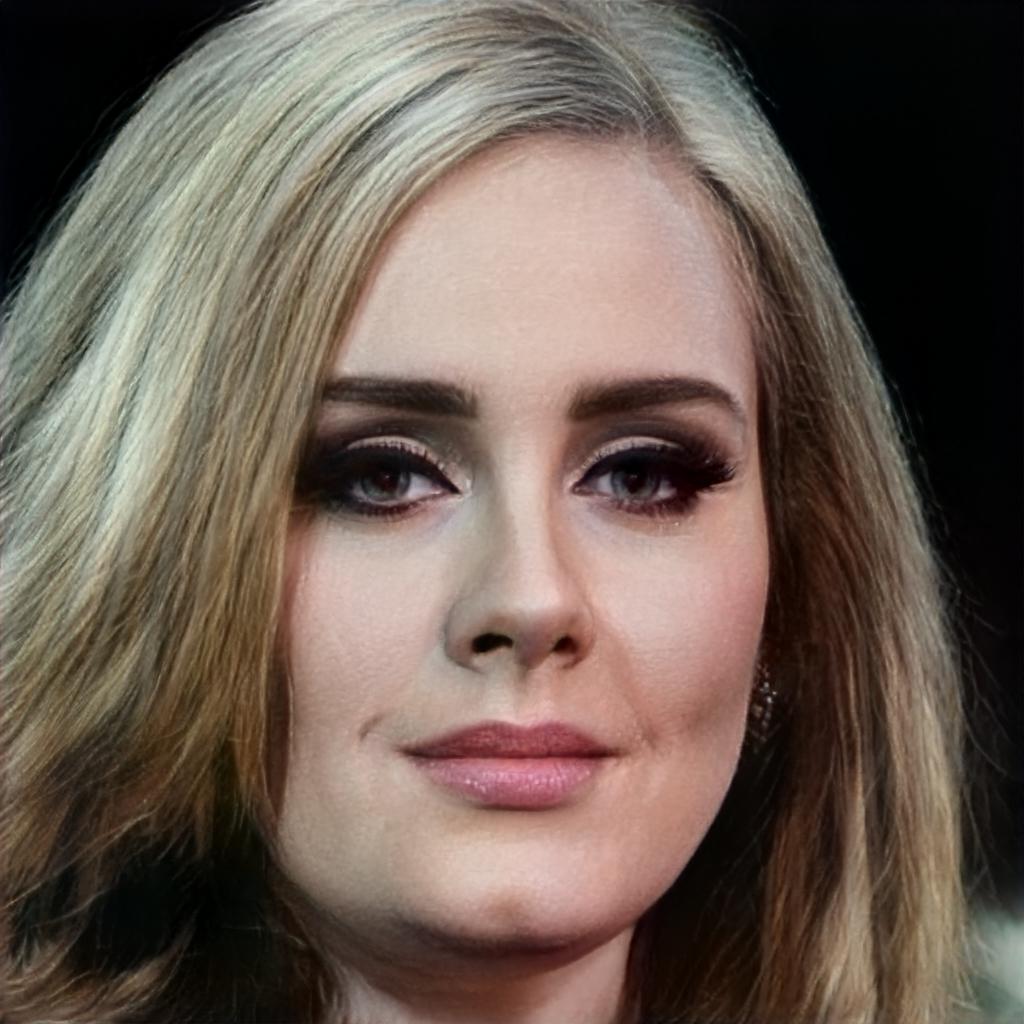} &
		\includegraphics[width=\imwidth]{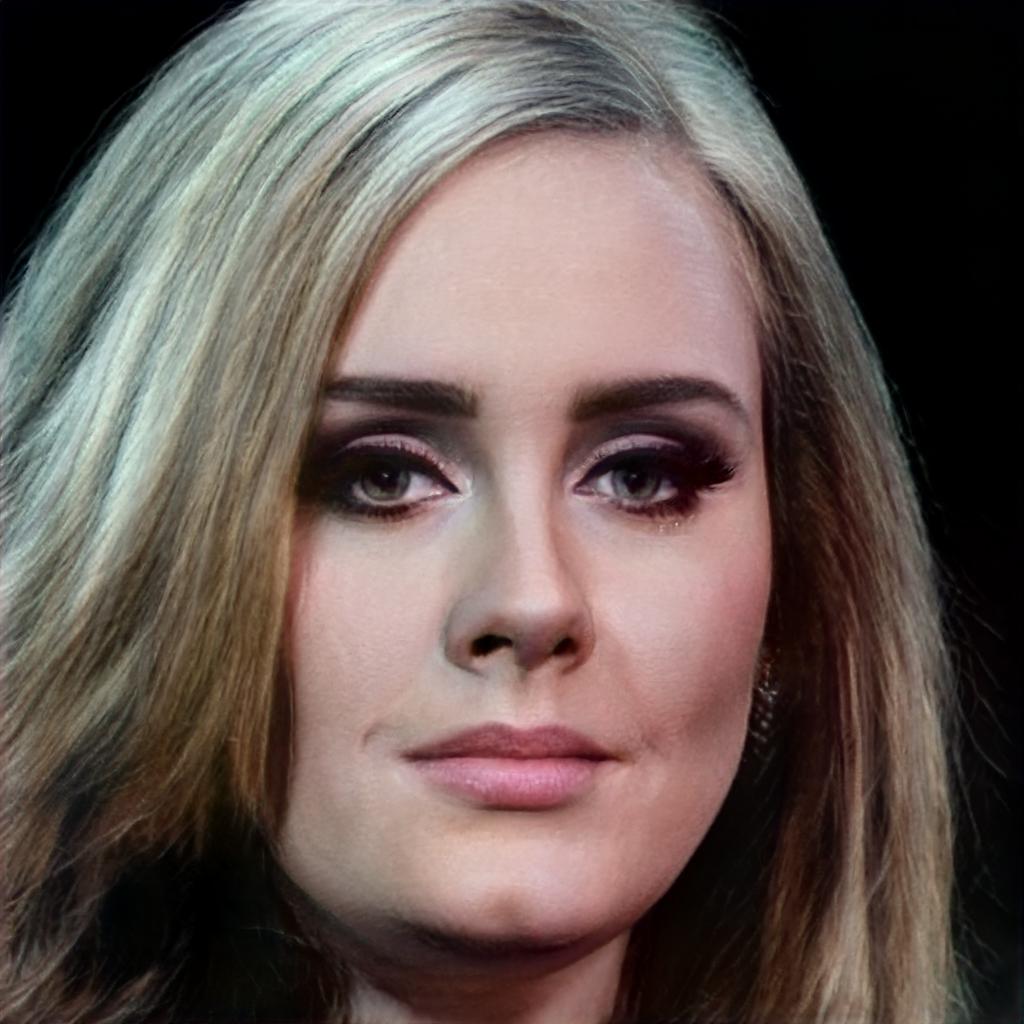}
        
        \\
        &
        \rotatebox{90}{\phantom{kk} Atypical} &
		\includegraphics[width=\imwidth]{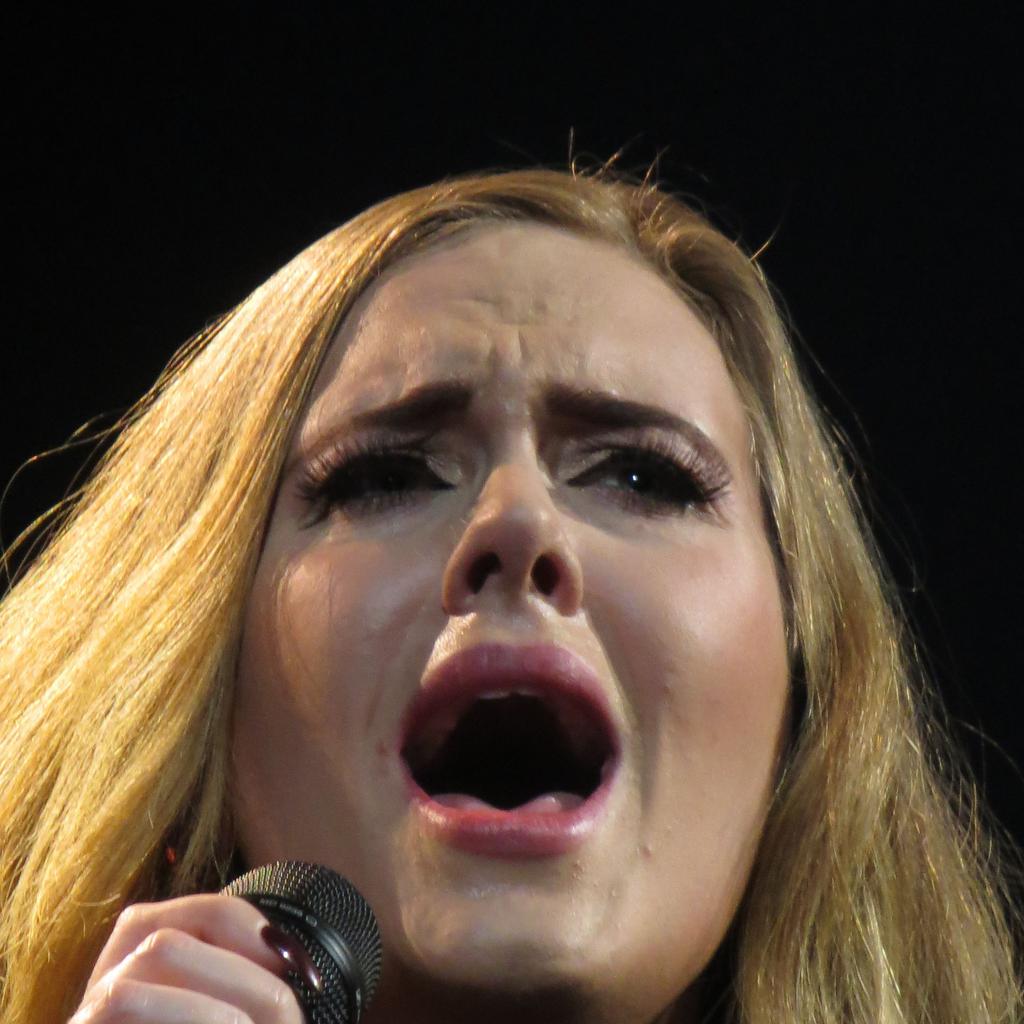} &
		\includegraphics[width=\imwidth]{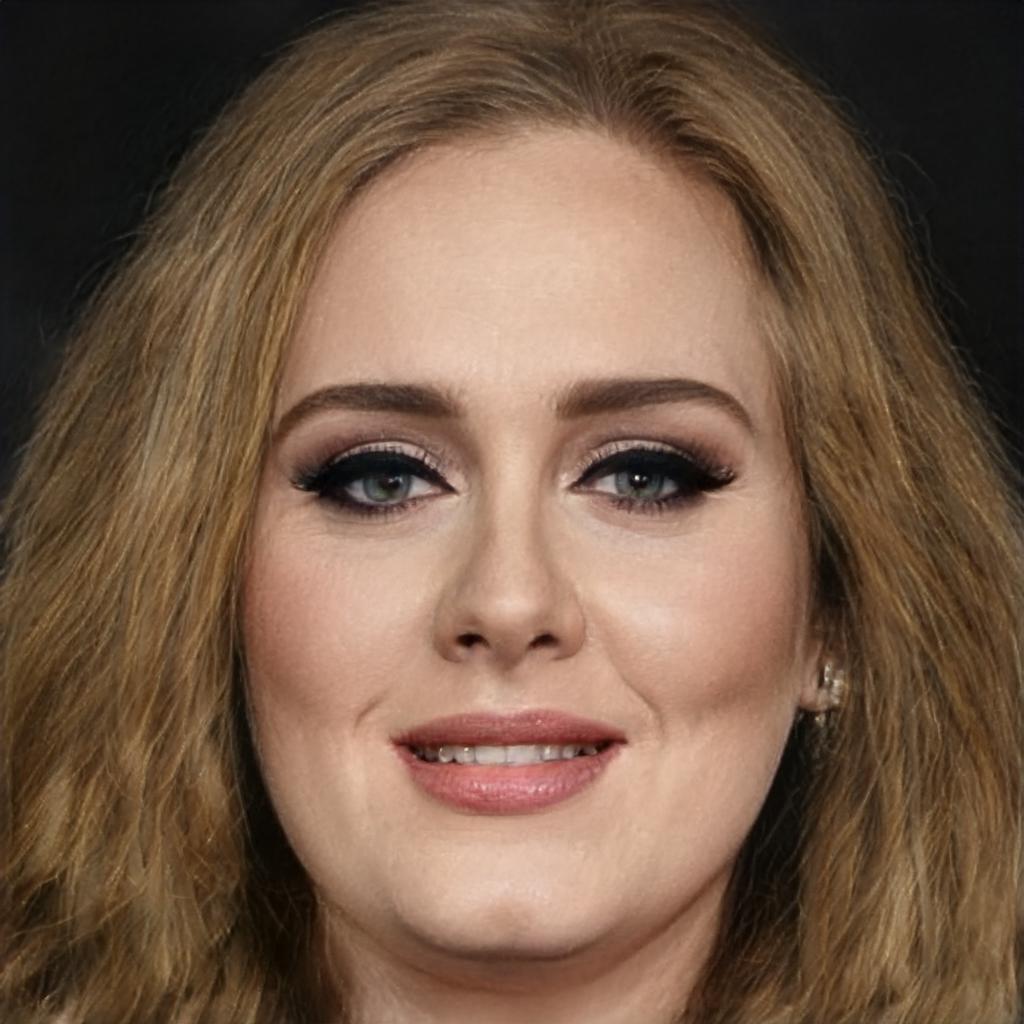} &
		\includegraphics[width=\imwidth]{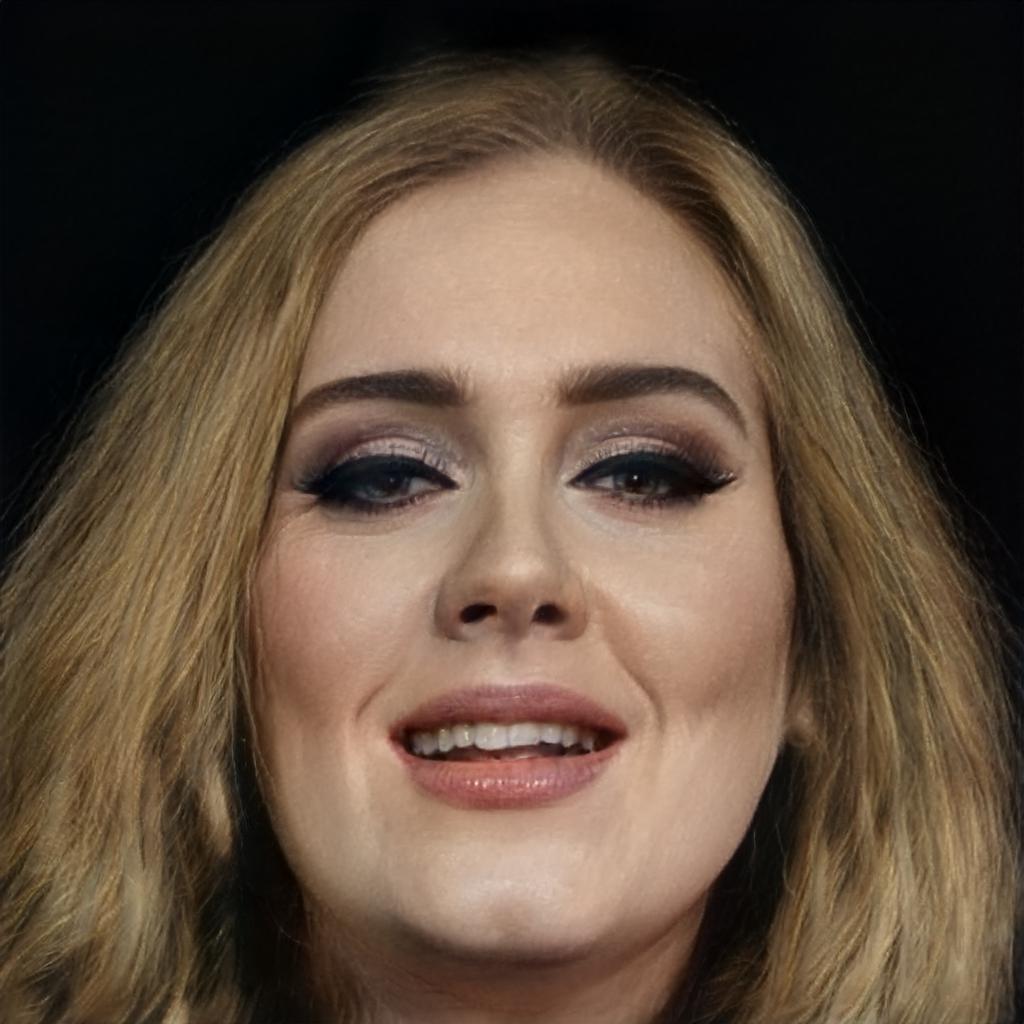} &
		\includegraphics[width=\imwidth]{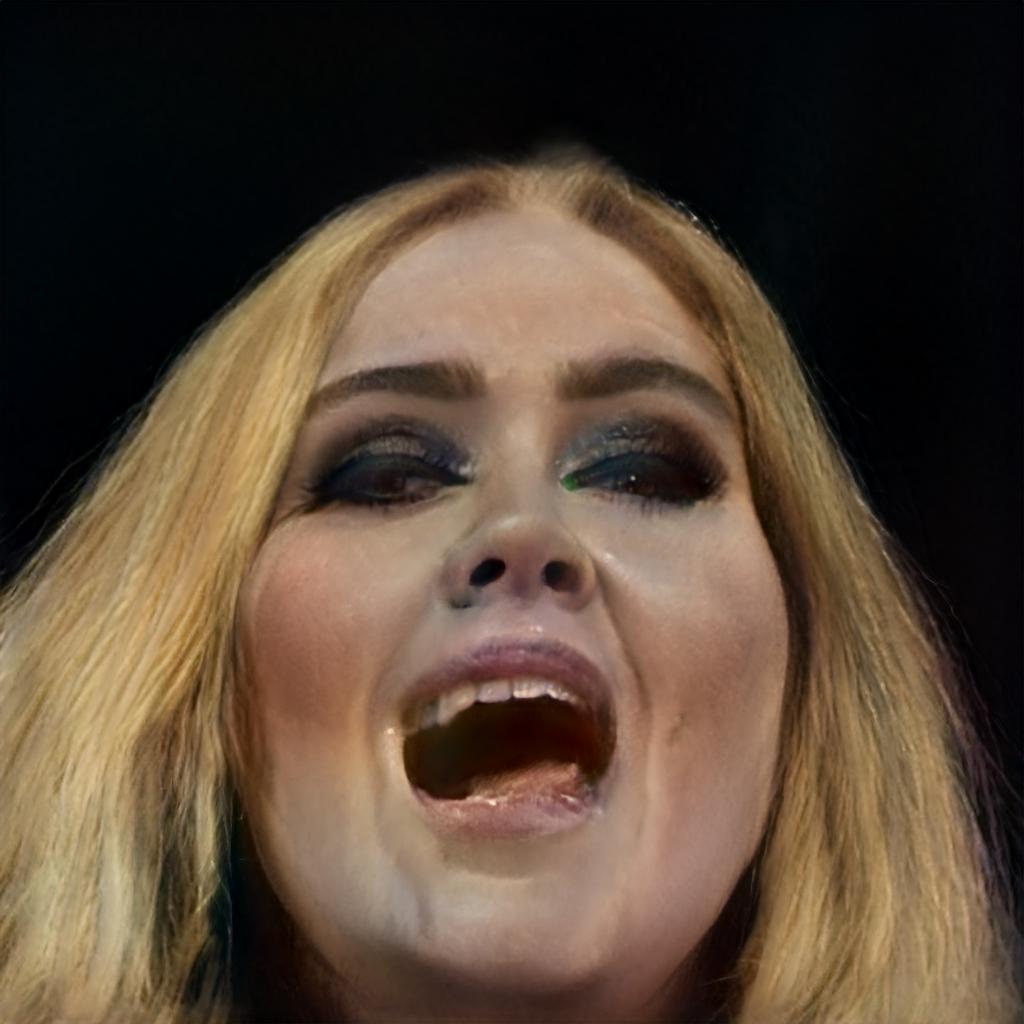}
        
        \\
        
		\midrule
		
		\multirow{2}{1.5em}[5.5ex]{\rotatebox[origin=c]{90}{Super-Resolution}} &
		\rotatebox{90}{\phantom{kk} Typical} &
		\includegraphics[width=\imwidth]{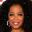} &
		\includegraphics[width=\imwidth]{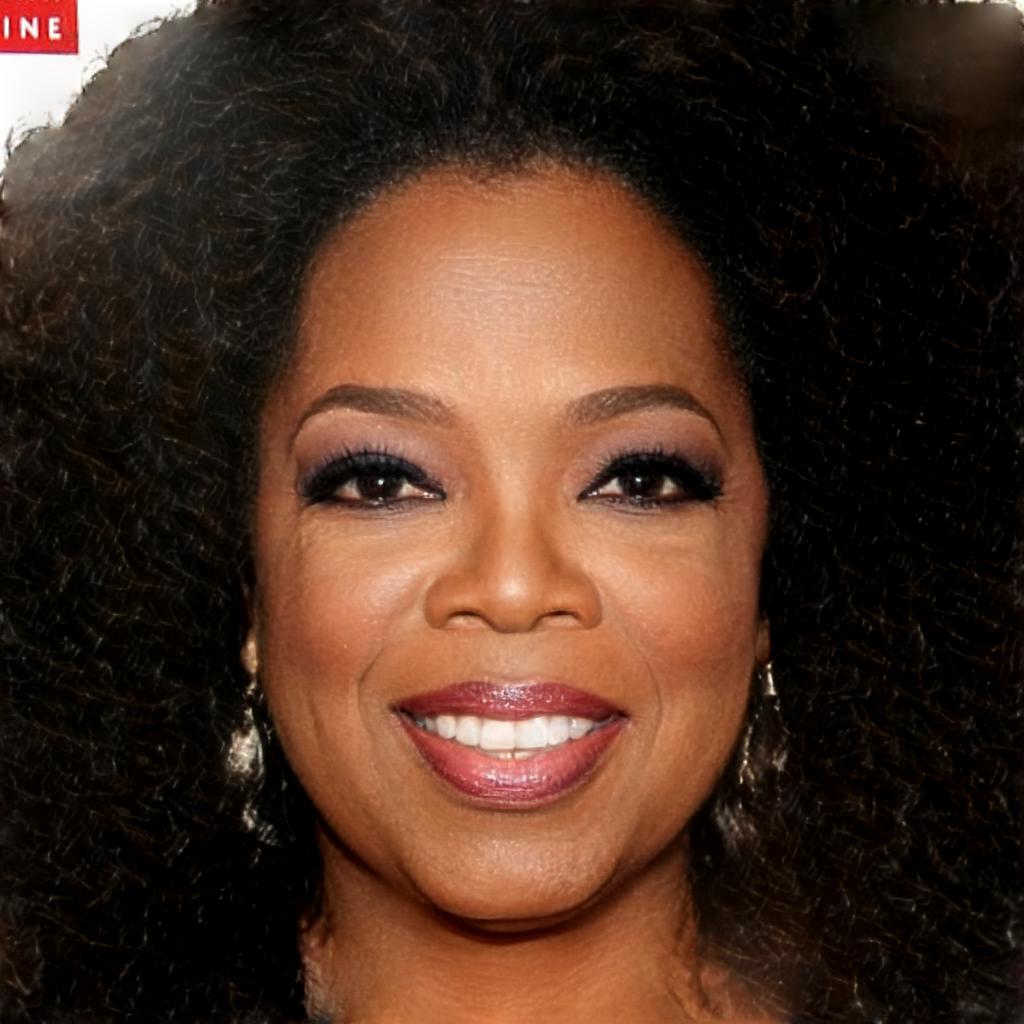} &
		\includegraphics[width=\imwidth]{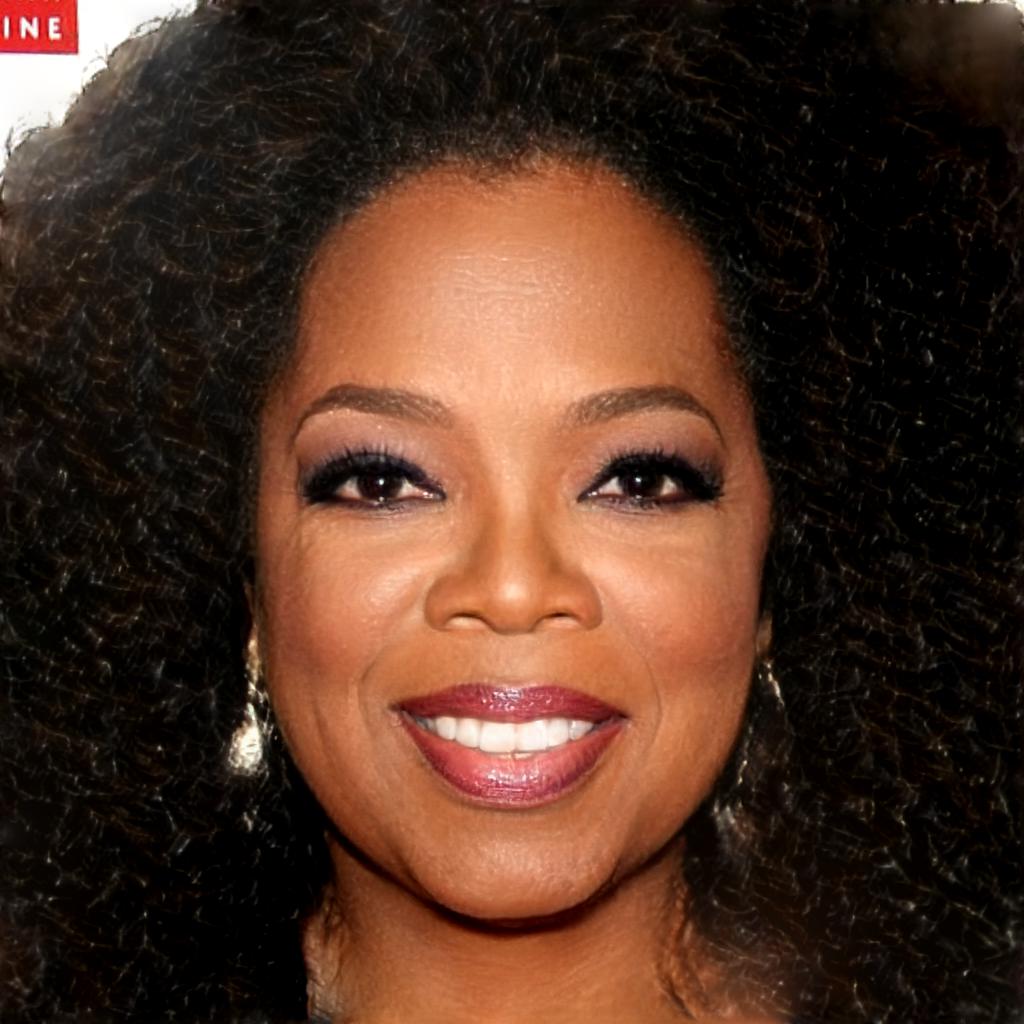} &
		\includegraphics[width=\imwidth]{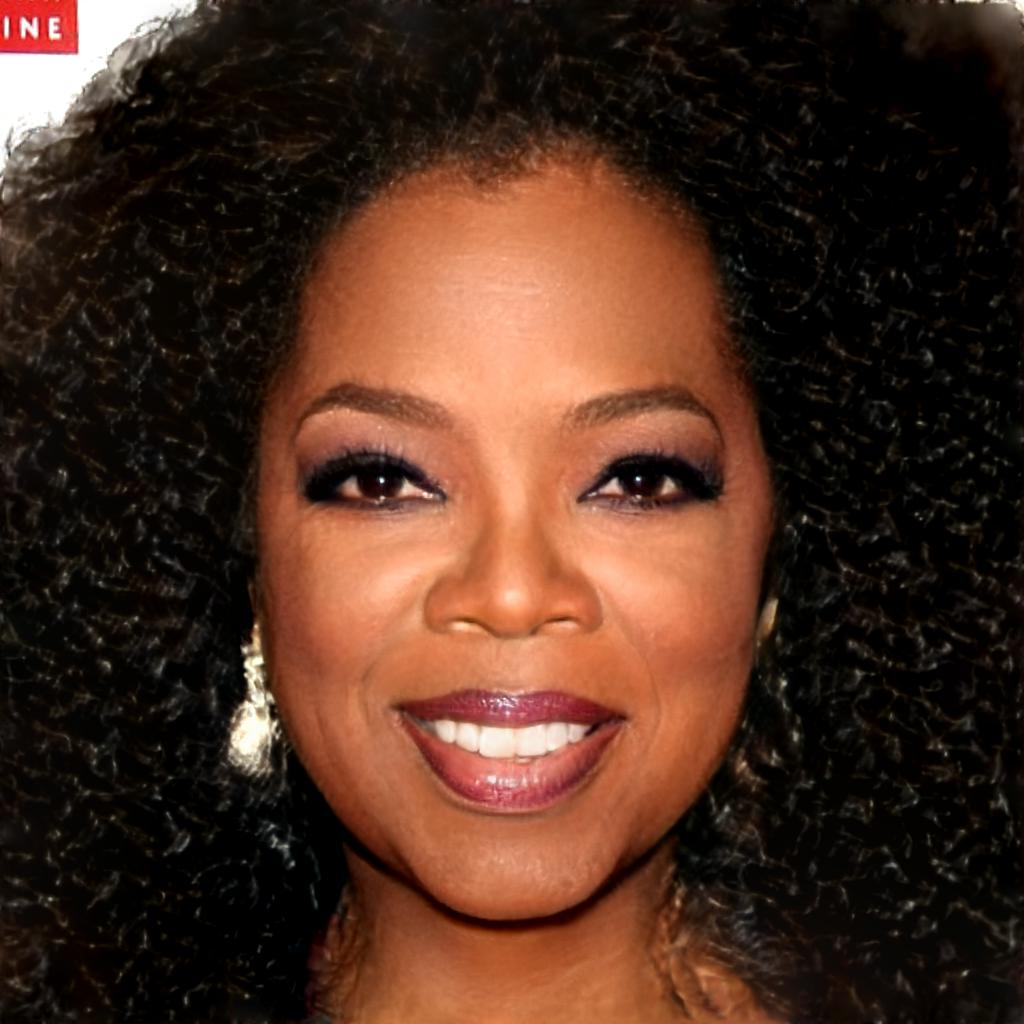}
        \\
        &
        \rotatebox{90}{\phantom{kk} Atypical} &
		\includegraphics[width=\imwidth]{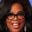} &
		\includegraphics[width=\imwidth]{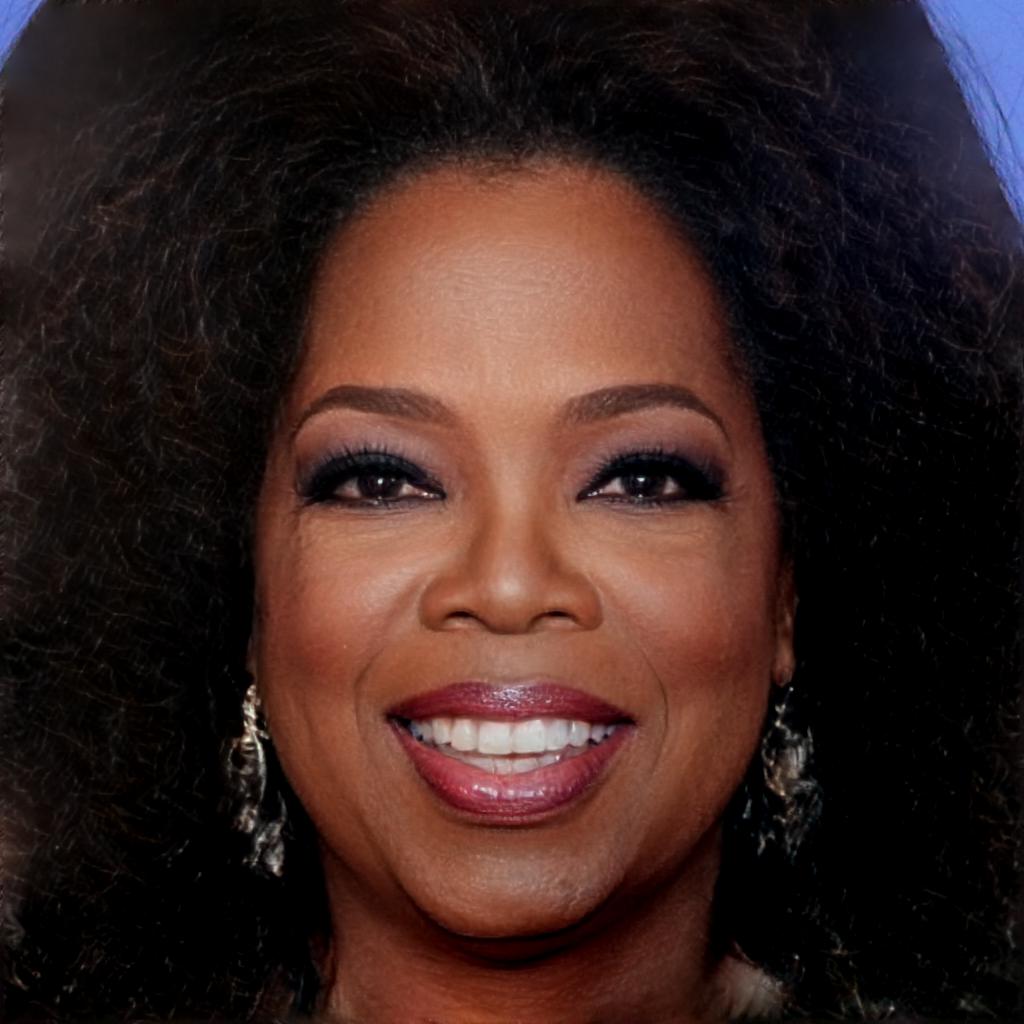} &
		\includegraphics[width=\imwidth]{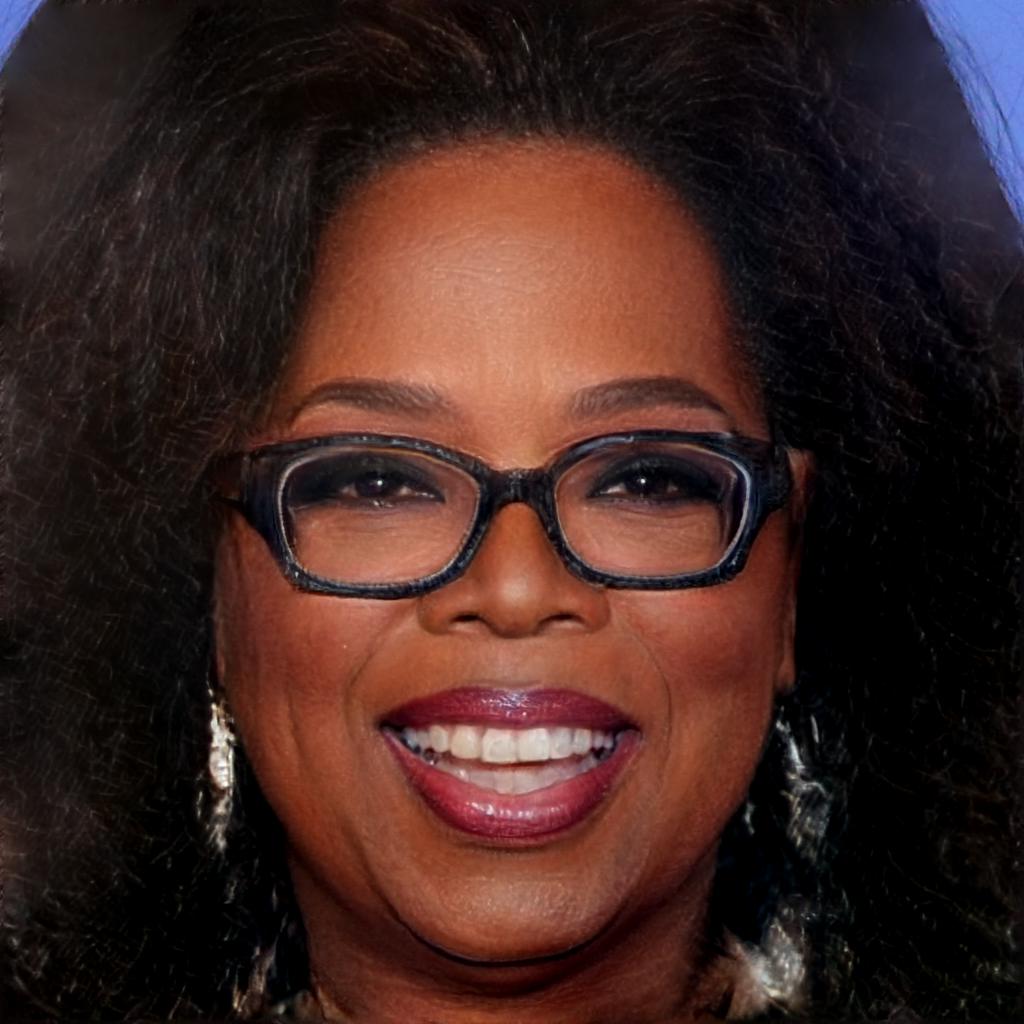} &
		\includegraphics[width=\imwidth]{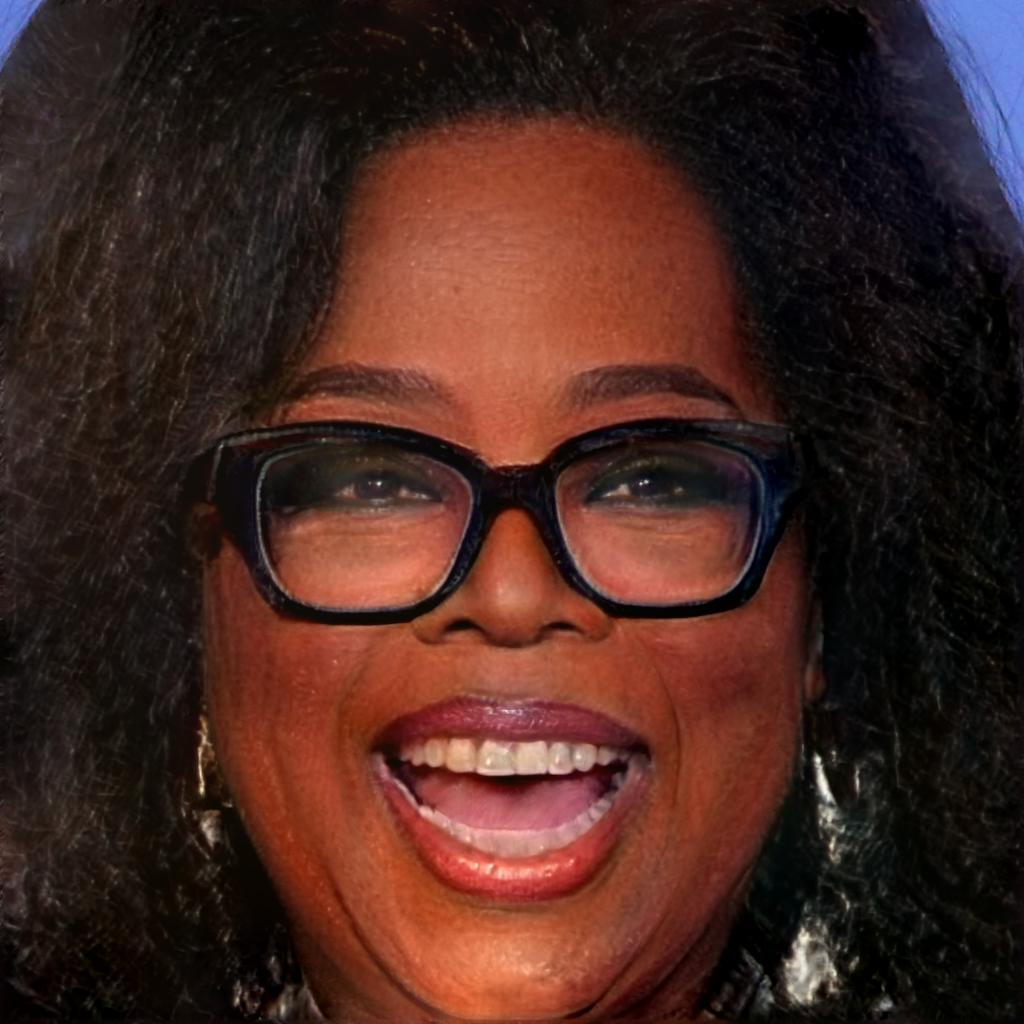}
        \\
        
        \midrule
        
		\multirow{2}{1.5em}[2ex]{\rotatebox{90}{Inpainting}} &
		\rotatebox{90}{\phantom{kk} Typical} &
        \includegraphics[width=\imwidth]{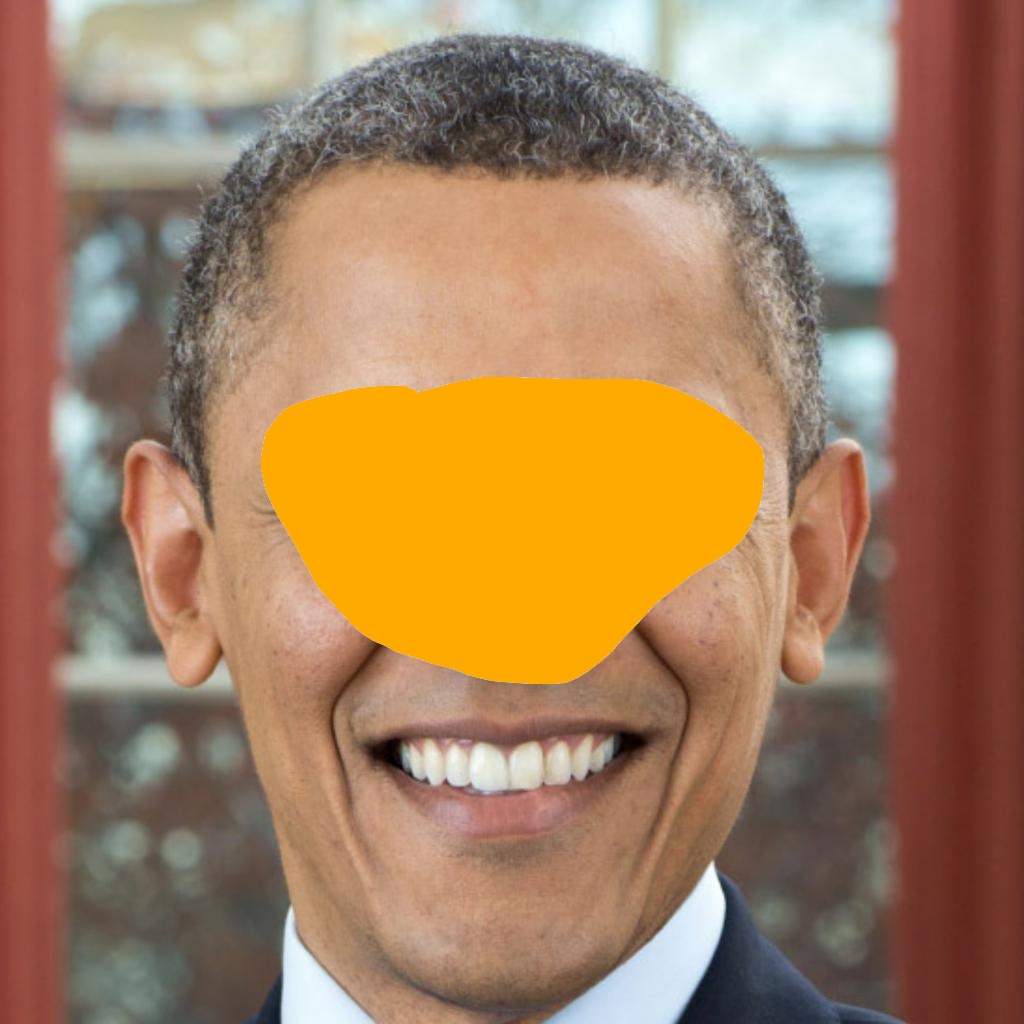} &
		\includegraphics[width=\imwidth]{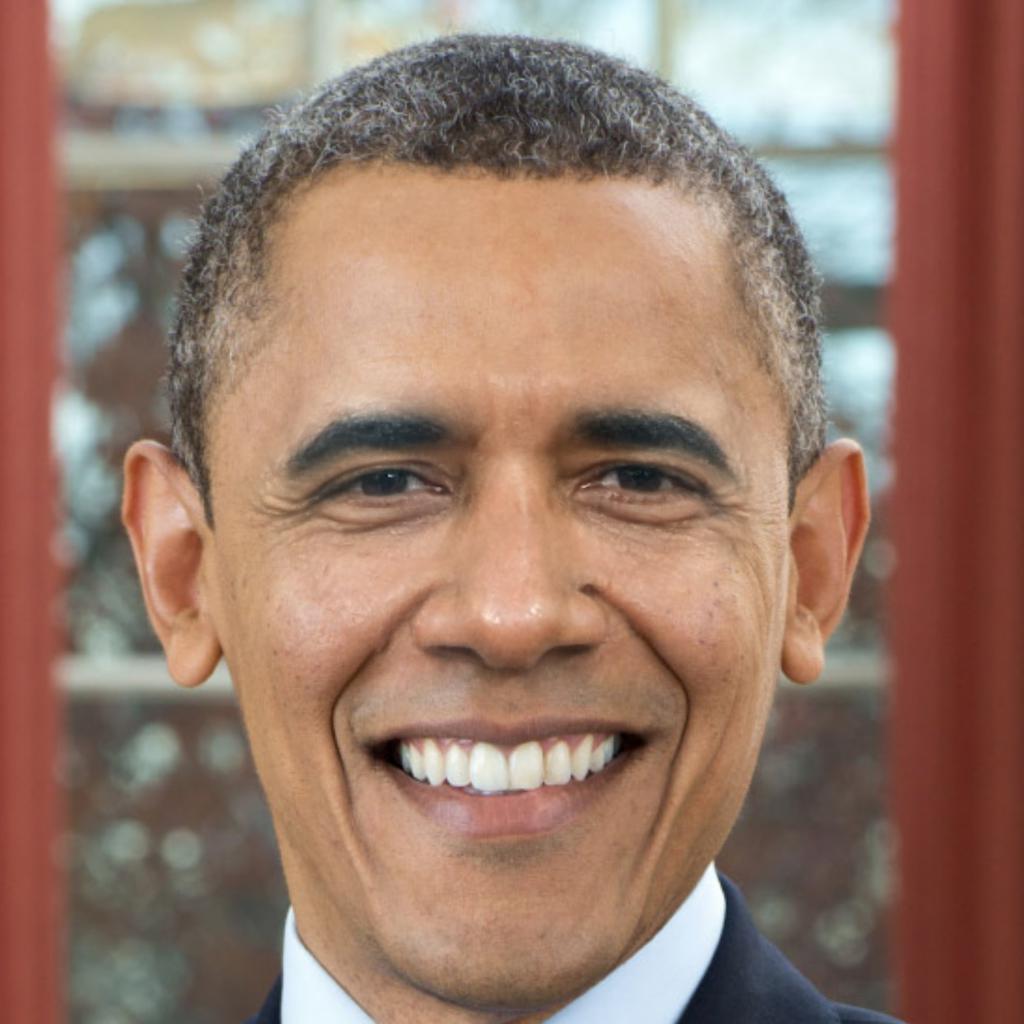} &
		\includegraphics[width=\imwidth]{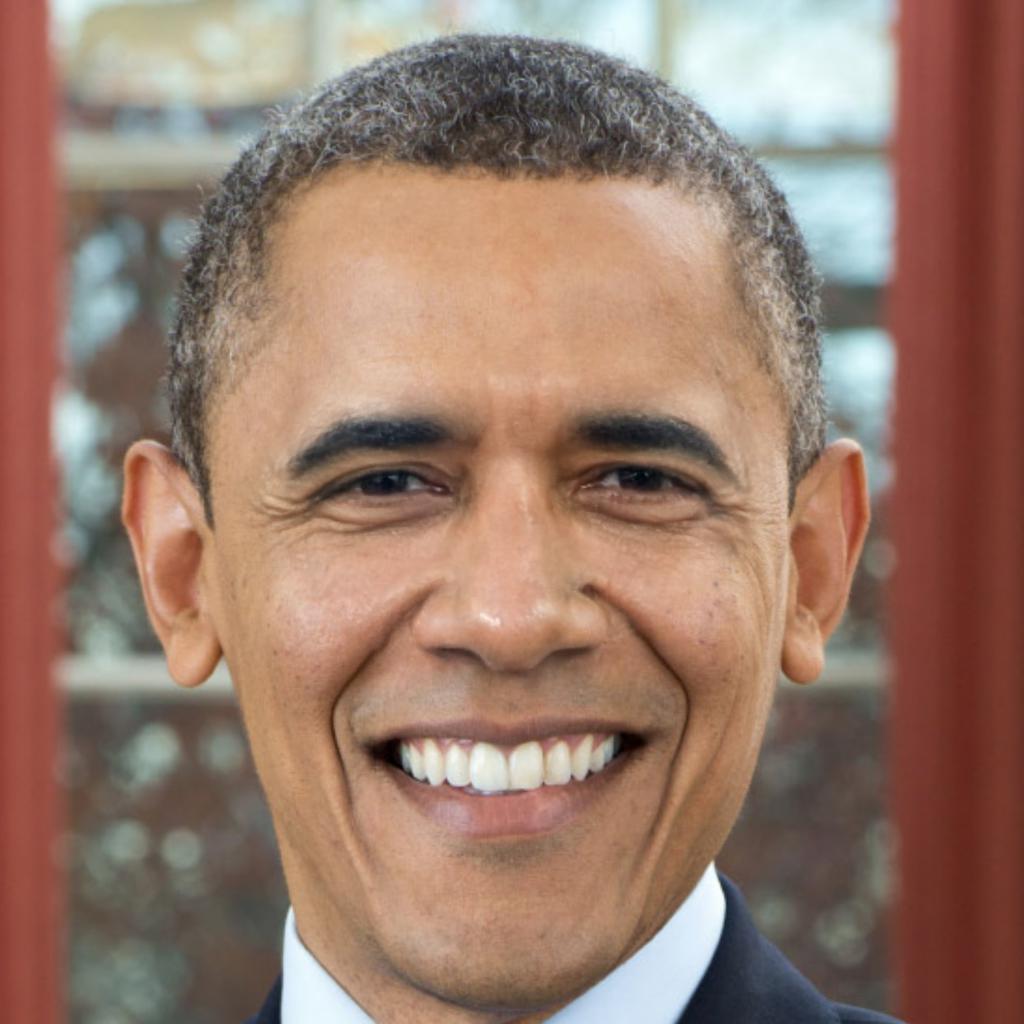} &
		\includegraphics[width=\imwidth]{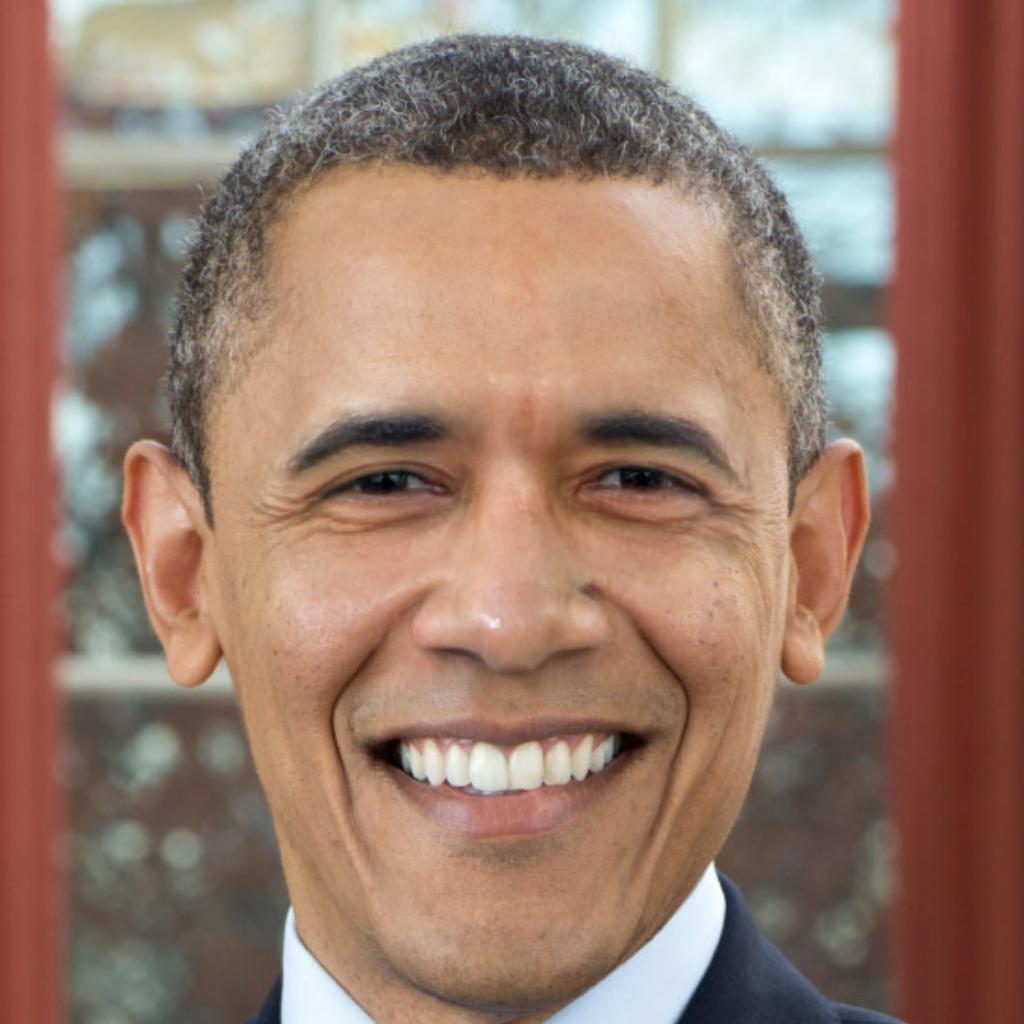}
        \\
        &
        \rotatebox{90}{\phantom{kk} Atypical} &
        \includegraphics[width=\imwidth]{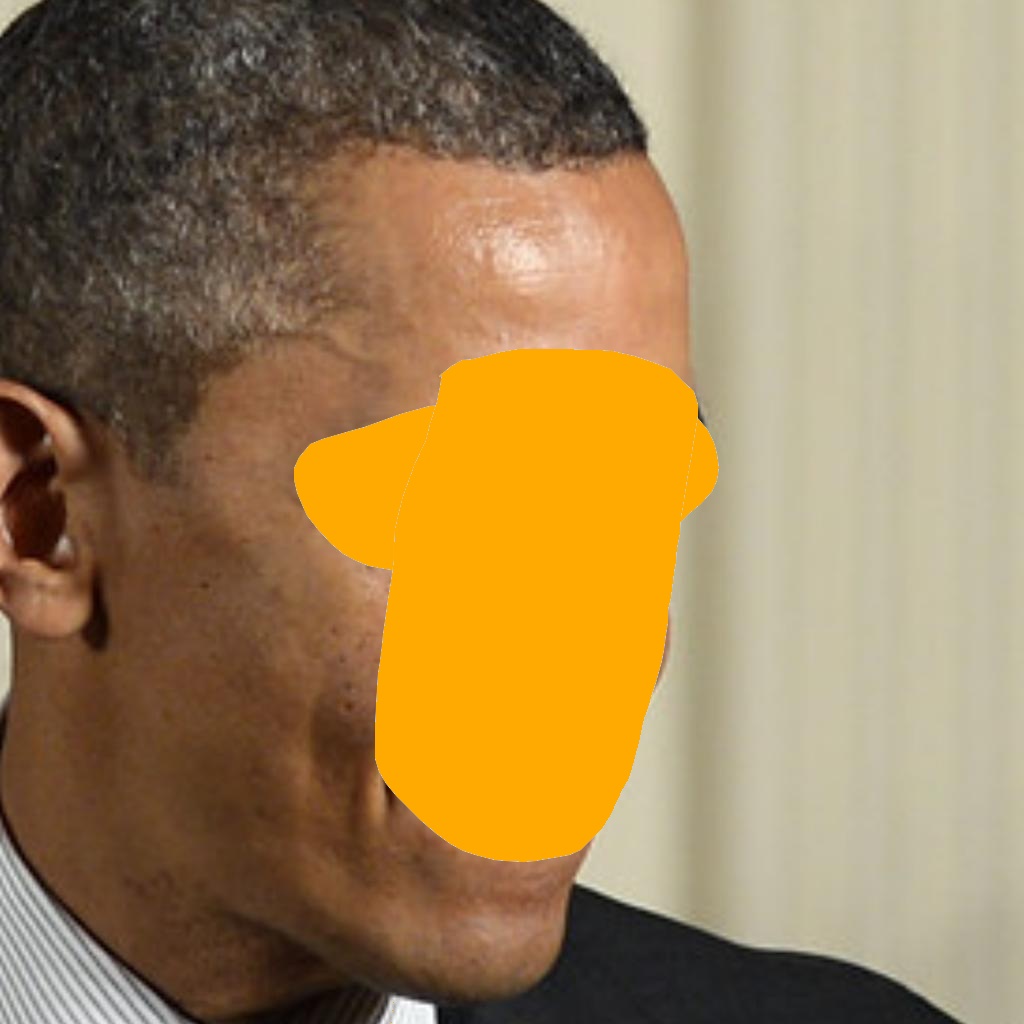} &
		\includegraphics[width=\imwidth]{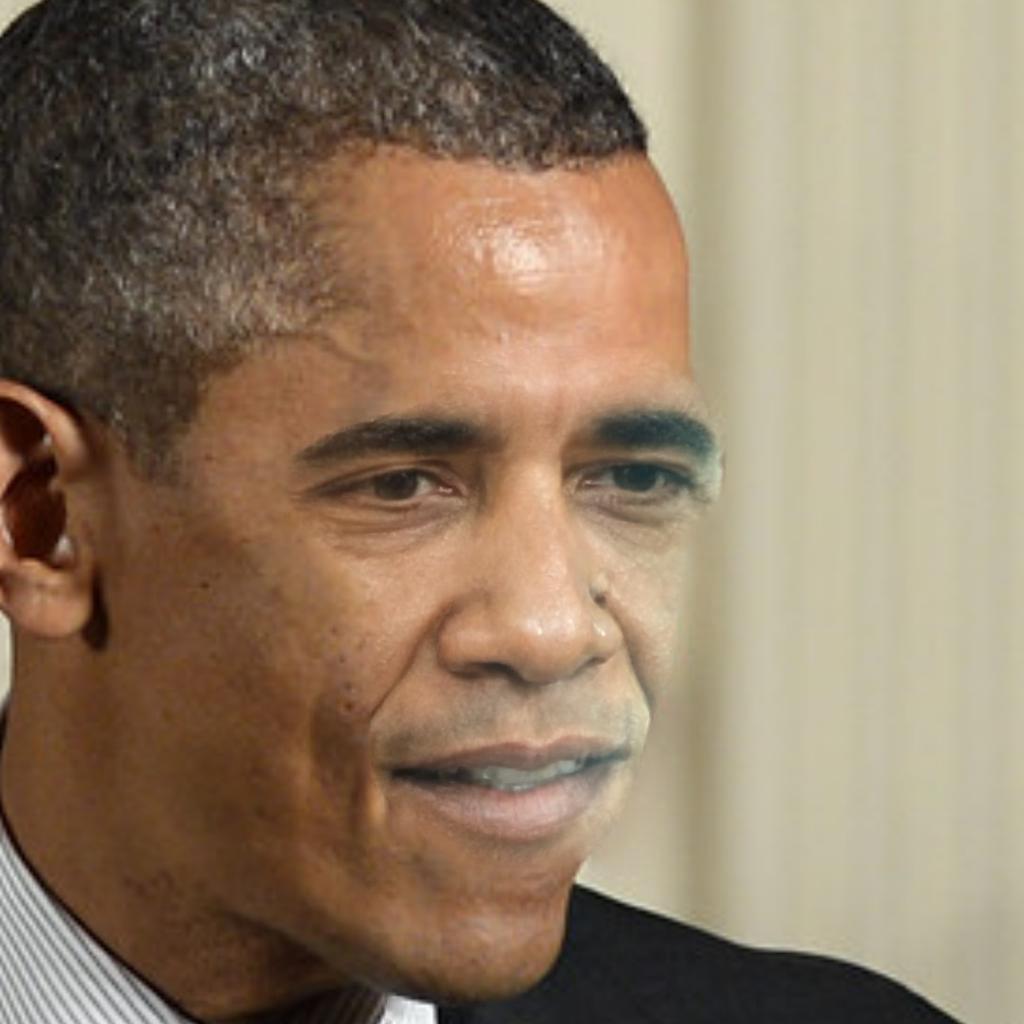} &
		\includegraphics[width=\imwidth]{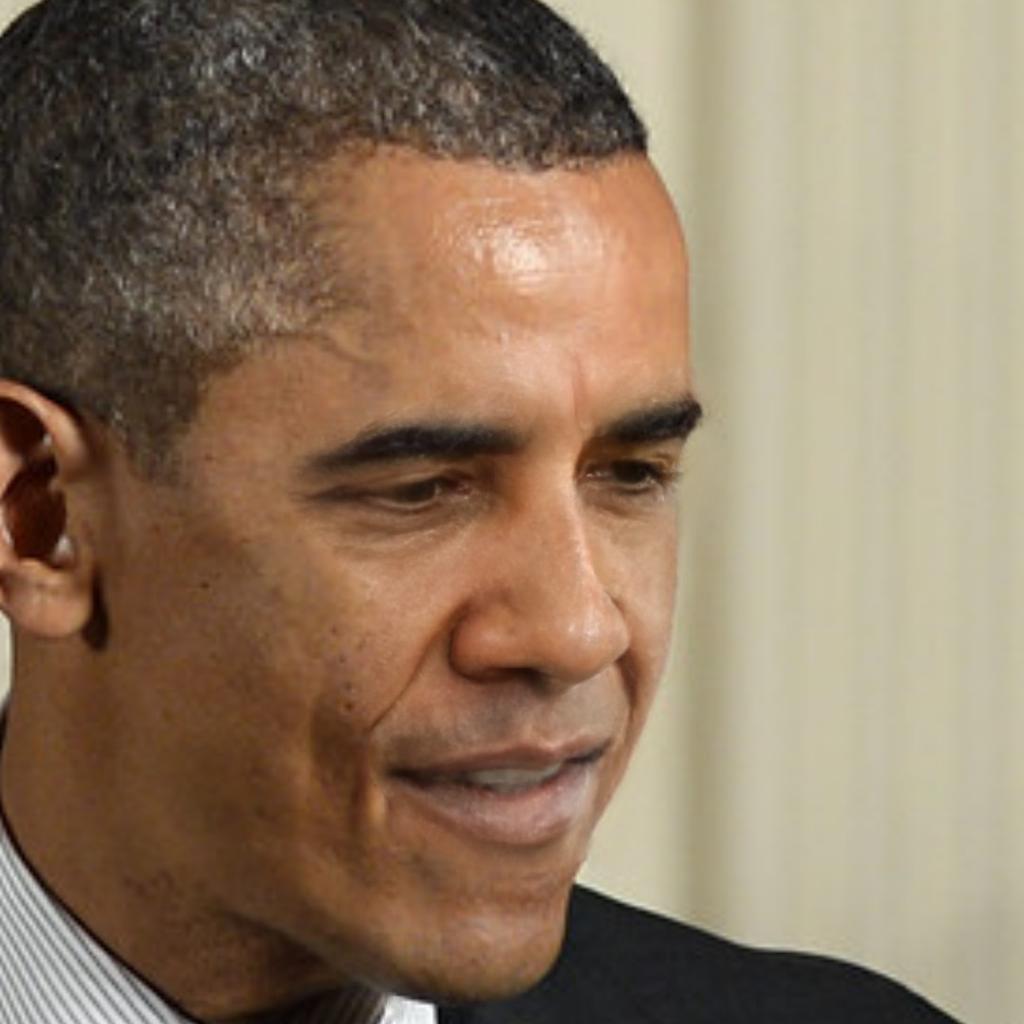} &
		\includegraphics[width=\imwidth]{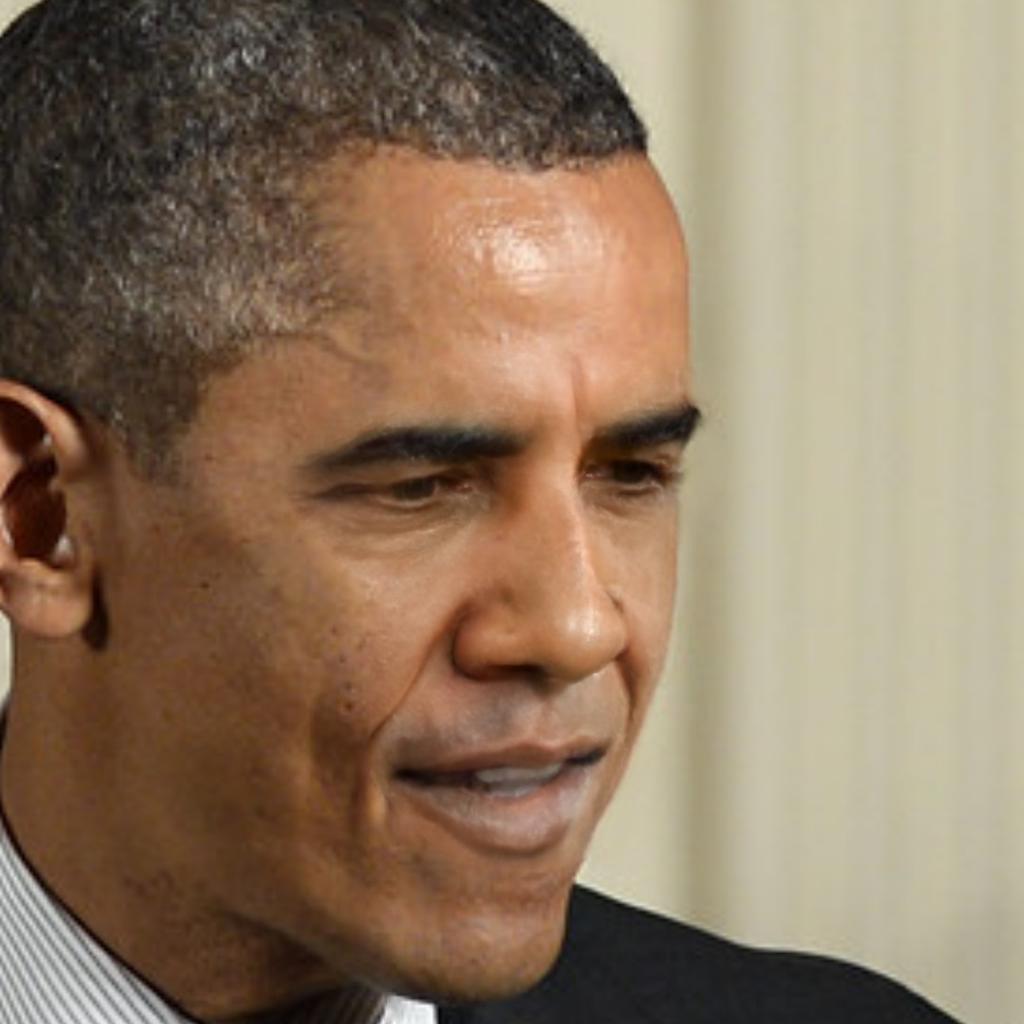}
	\end{tabular}
	\caption{
	    Effect of $\beta$ on projection for inversion, super-resolution and inpainting. The effect is demonstrated for both typical and atypical images for each person and application. It can be seen that $\beta$ controls a tradeoff between the personalized prior and expressiveness. 
	    Results produced with $\beta=0$ are highly characteristic and conservative. These results might be sufficient for some cases (e.g. rows 3, 5) but have poor fidelity in many other. 
	    Increasing the allowed $\beta$, allows greater expressivity and thus better fidelity, but at the cost of weakening the guidance of the prior. At the extreme, $\beta$ is not explicitly bounded (\eqnref{eq:softplus} not used). These results might be viable in some cases (e.g. rows 1, 3, 6) but may also produce results containing artifacts (e.g. rows 2, 4) or slight drift in identity (e.g. row 5). 
	    We find it beneficial in all applications to bound $\beta$ to some positive small value, in order to balance the trade-off.
	}
	\label{fig:ablation_beta}
\end{figure}

\yn{
This section extends the evaluation of the prior-expressiveness presented in \secref{sec:dch_analysis},} by delving deeper into understanding how is the tradeoff expressed for different samples and applications. To this end, we perform super-resolution and inpainting in addition to the previously performed inversion. Qualitative results are provided in \figref{fig:ablation_beta}.

Generalizing the results from the previous experiment, projection to $\mathcal{P}_0^{+}$ exhibits strong ID preservation - yielding a highly characteristic image of the person. On the other hand, it is also strongly conservative, \ie relating to a common and simple appearance of the person -- almost frontal, neutral or smiling expression, simple illumination conditions and no accessories. 
Therefore, typical or more characteristic appearances are better reconstructed than atypical appearances.

In most cases, increasing $\beta$ corresponds to greater expressivity, indicated by better fidelity. 
This is most evident for atypical images. We find that for such images, the optimization in fact converges to latent codes outside the previous smaller dilation. On the other hand, we observe that, "typical`` images that were properly reconstructed with the smaller dilation, converge to a similar dilation and do not leverage the newly enlarged one.
This behavior is intuitive due to the optimization's dynamic.
As a reminder -- $\beta$ values serve as upper-bounds in \eqnref{eq:softplus} of the main paper, and projection is initialized to the center of $\mathcal{P}_0^{+}$.
Therefore, the optimization tends to converge at a smaller dilation when possible.
While increasing $\beta$ improves fidelity, we also observe the deterioration of quality, manifested through artifacts and slight drifts to identity. Once more this is true mostly for projections that resort to a greater effective dilation.

We conclude the $\beta$ controls the prior-expressiveness tradeoff for image enhancement applications as well. Additionally, we observe that the fidelity obtained with a specific $\beta$ mostly depends on how typical is the input image. Typical and atypical images might require different $\beta$ values to obtain satisfactory results, and the meaning of satisfactory would depend on the user's preference for the balance between expressiveness and faithfulness to the prior.

In our experiments, we use the minimal $\beta$ that was able to obtain fair fidelity for most images. We note that this simple guideline depends on the application. For example, when solving inpainting, only a small portion of the generated image is used by blending it to the input image. Therefore, obtaining fair fidelity is possible with smaller $\beta$, compared to the one needed for fair fidelity in inversion and super-resolution.

\begin{figure}
    \centering
	\setlength{\tabcolsep}{1pt}
	\setlength{\imwidth}{0.33\linewidth}
	\begin{subfigure}{\linewidth}
	    \begin{tabular}{ccc}
	         Real & Before Tuning & After Tuning \\
	         \includegraphics[width=\imwidth]{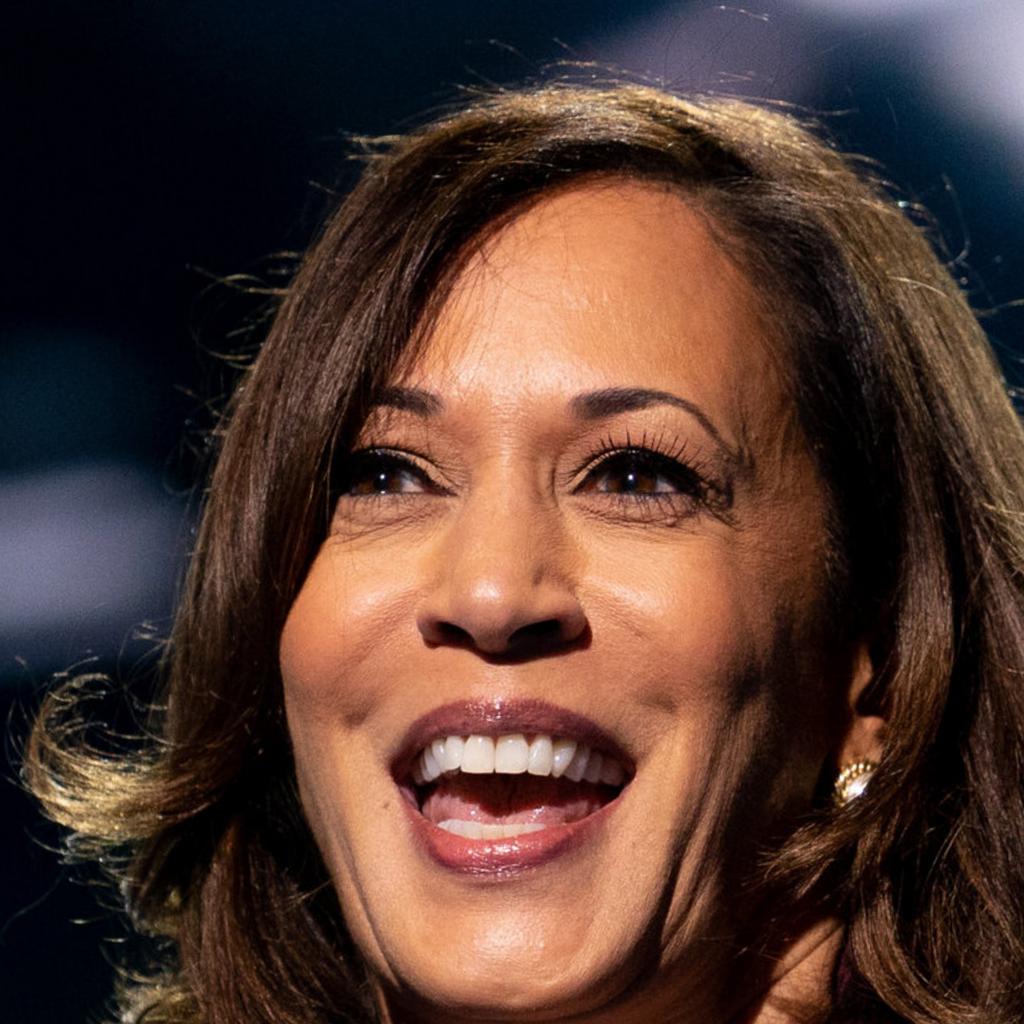} &
	         \includegraphics[width=\imwidth]{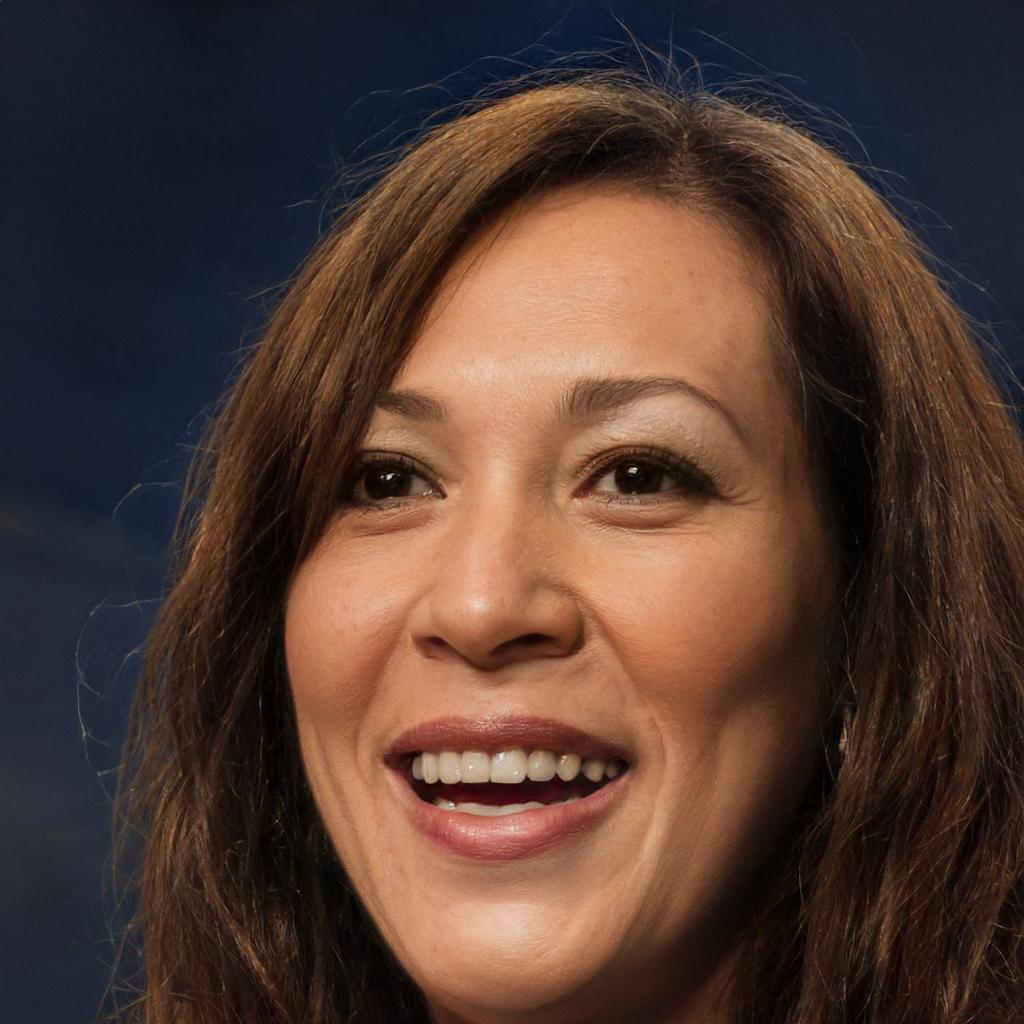} & 
	         \includegraphics[width=\imwidth]{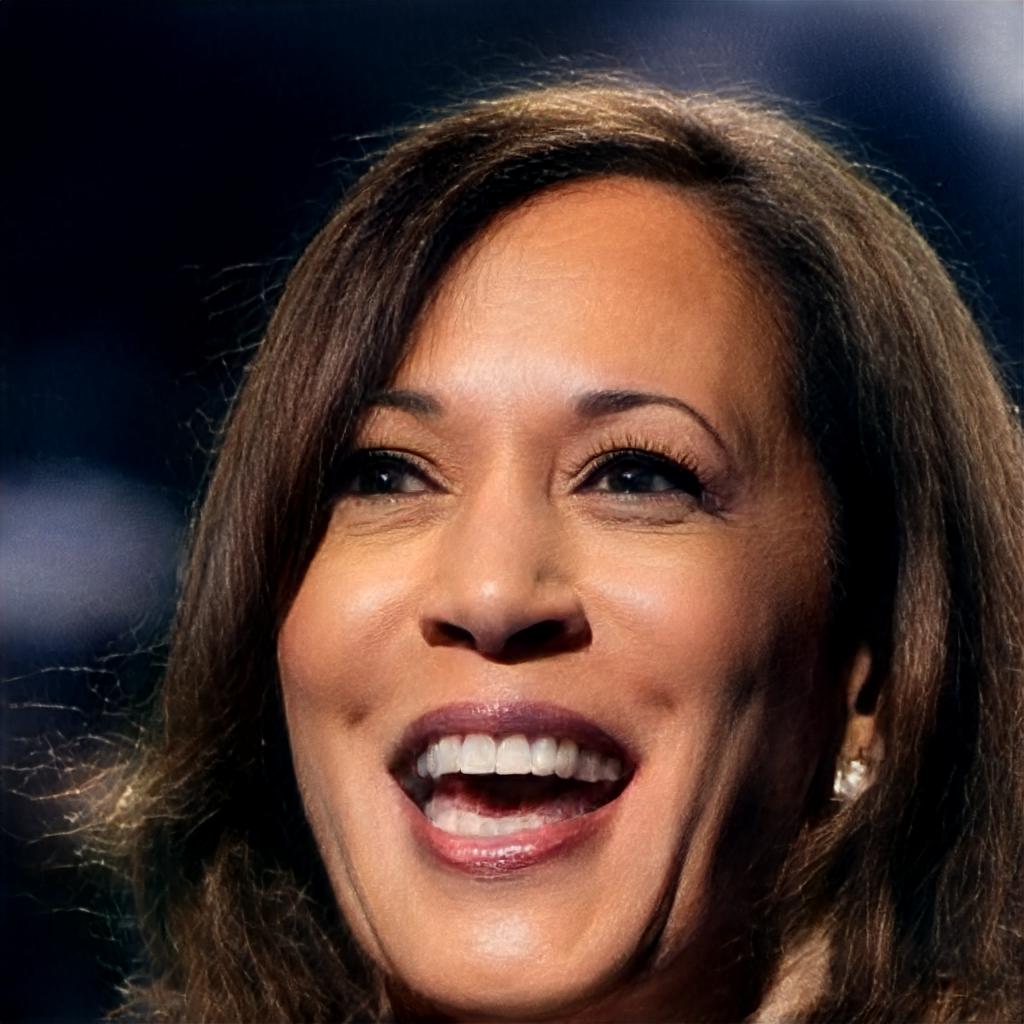} \\
	         
	         \includegraphics[width=\imwidth]{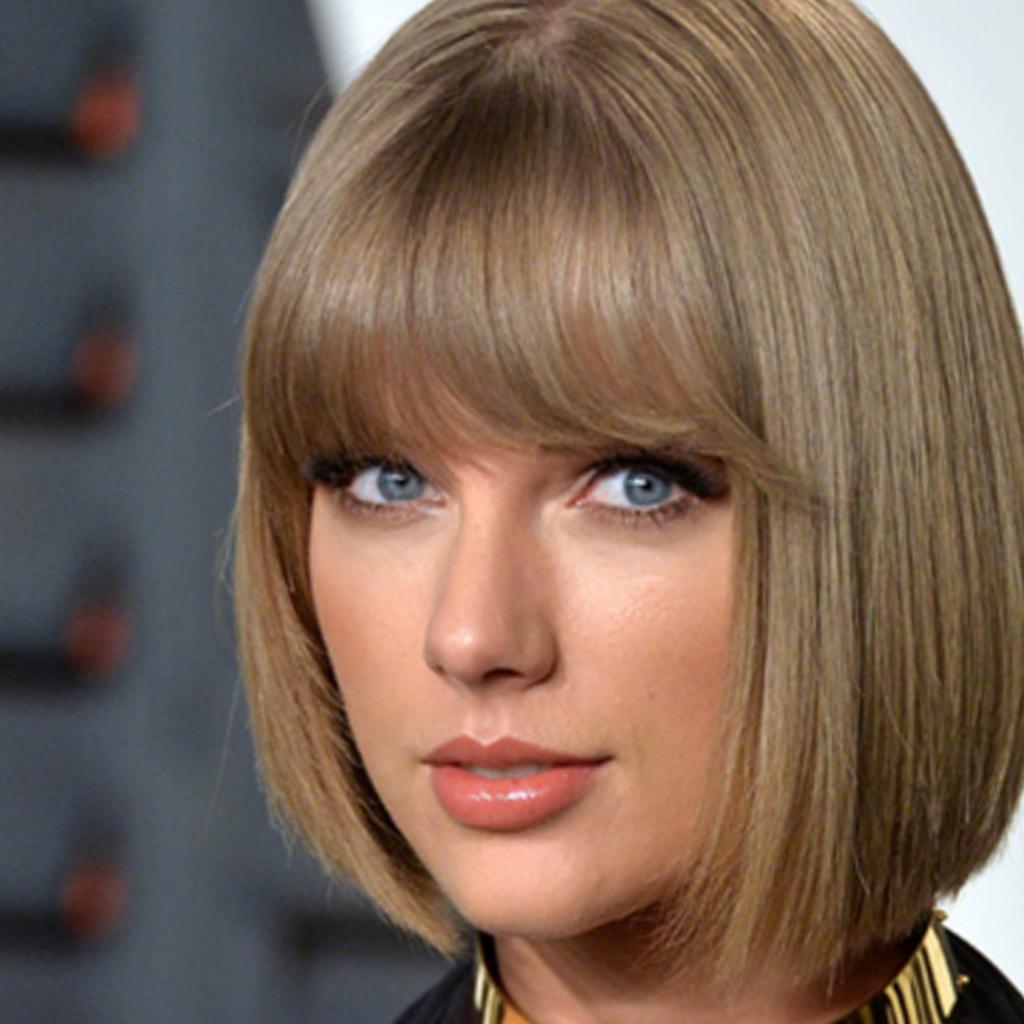} &
	         \includegraphics[width=\imwidth]{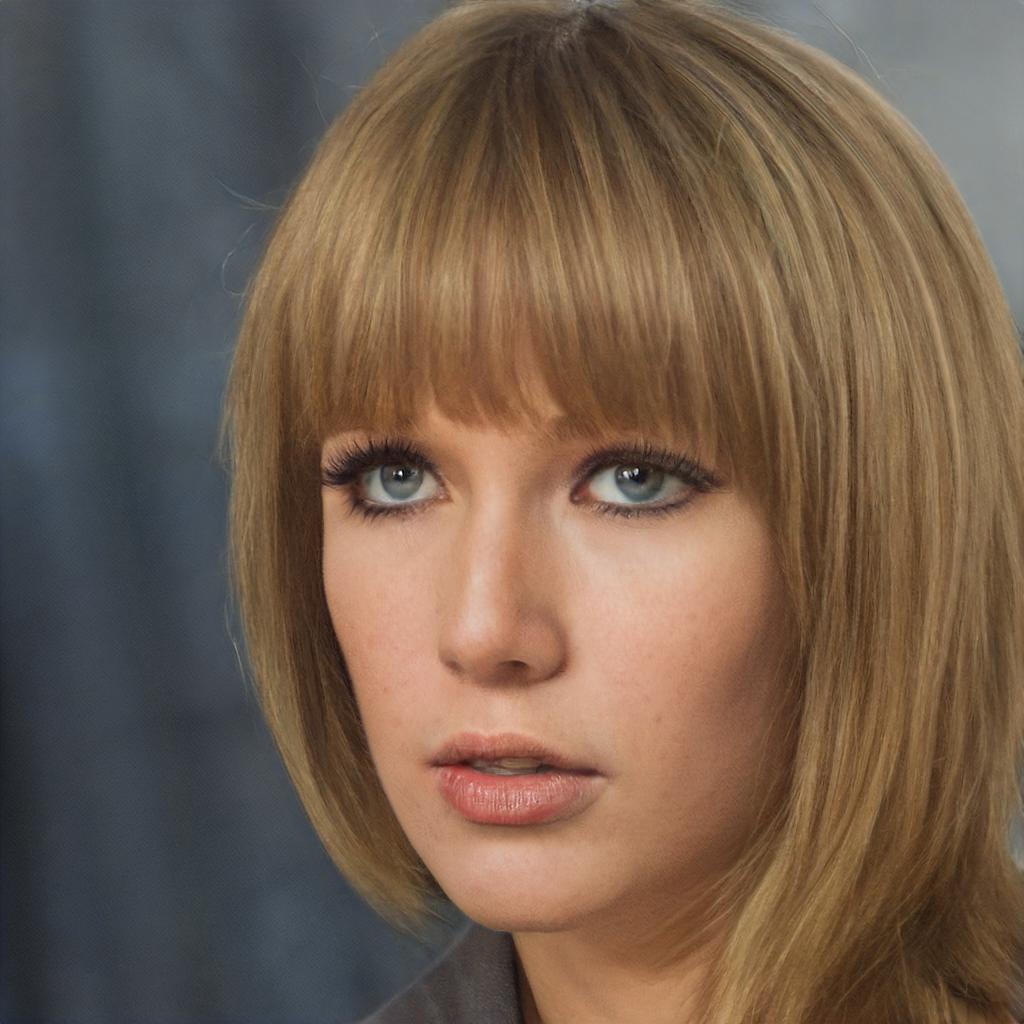} & 
	         \includegraphics[width=\imwidth]{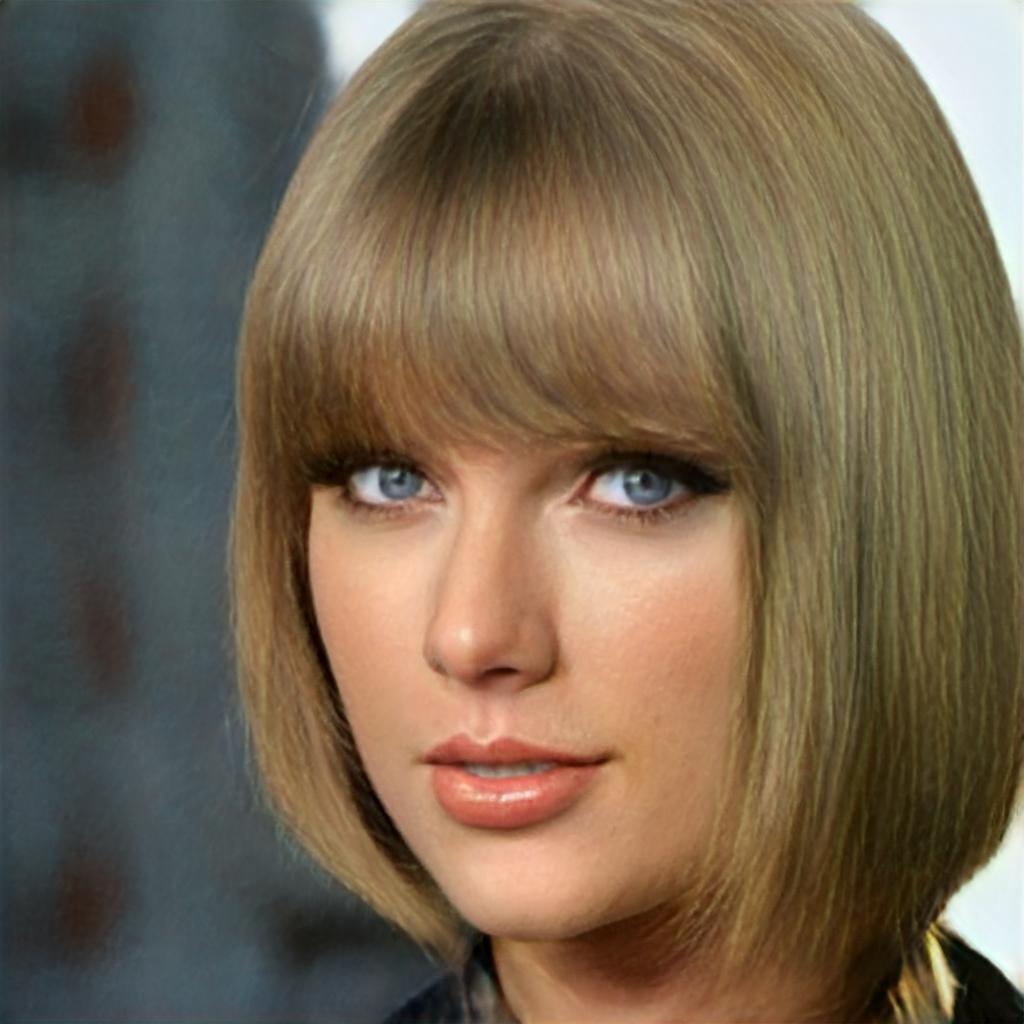} \\
	         
	         \includegraphics[width=\imwidth]{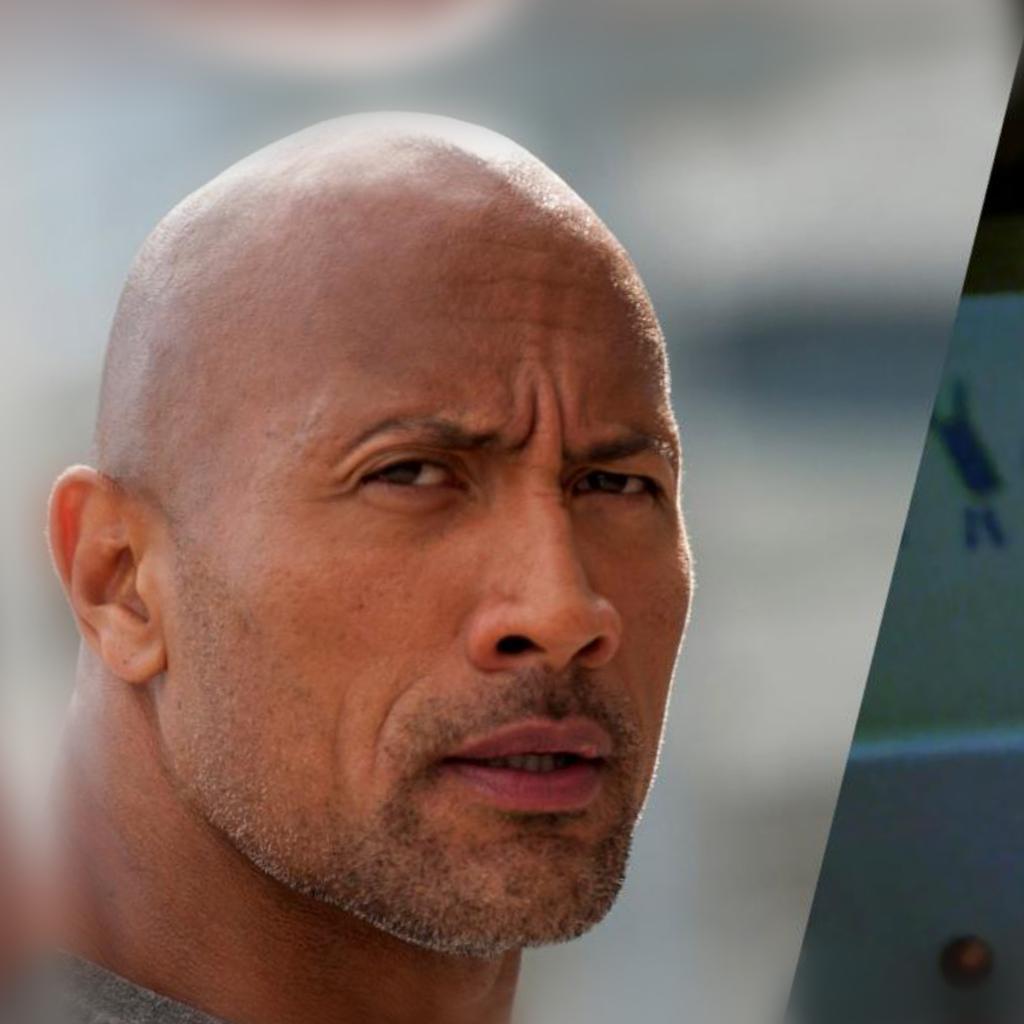} &
	         \includegraphics[width=\imwidth]{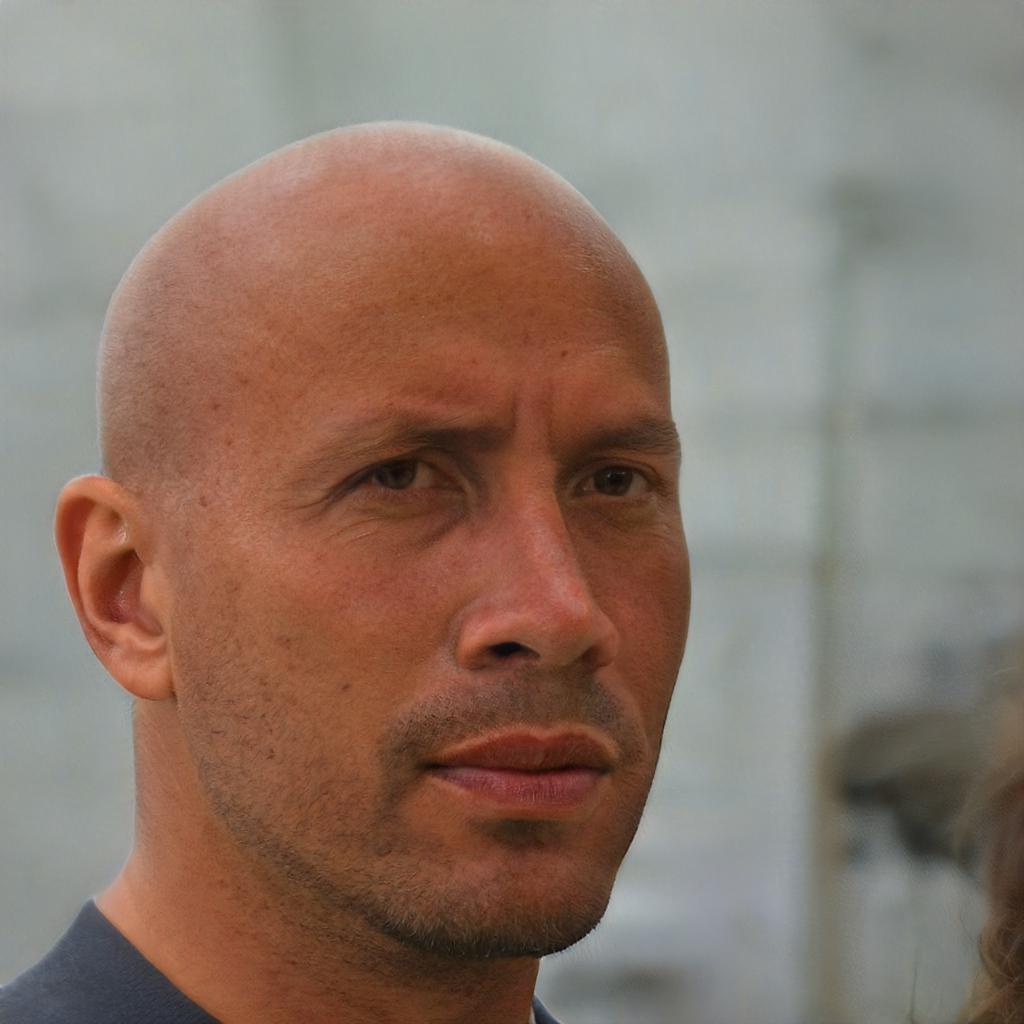} & 
	         \includegraphics[width=\imwidth]{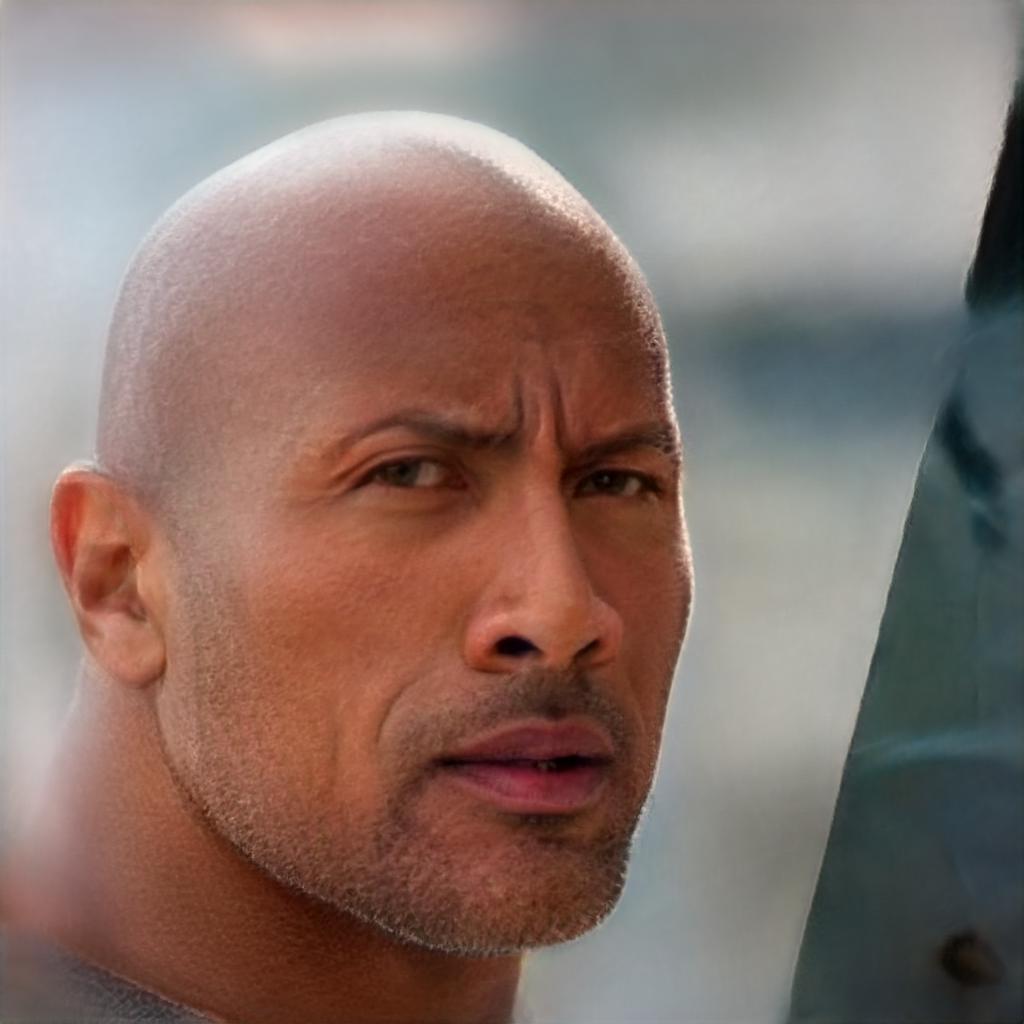} \\
	    \end{tabular}
	\end{subfigure}
	
	\begin{subtable}{\linewidth}
	    \centering
	    \begin{tabular}{l >{\centering\arraybackslash}p{3cm} >{\centering\arraybackslash}p{3cm} >{\centering\arraybackslash}p{2.5cm}}
	    \toprule
         Metric & Before Tuning & After Tuning  \\
         \midrule
         LPIPS Error ($\downarrow$) & 0.37 $\pm$ 0.04 & \textbf{0.22 $\pm$ 0.02} \\
         ID Similarity ($\uparrow$) &  0.56 $\pm$ 0.07 & \textbf{0.92 $\pm$ 0.02} \\
         \bottomrule
	    \end{tabular}    
	\end{subtable}

    \caption{Quantitative and qualitative inspection of the set's reconstruction from the anchors, before and after tuning. Quantitative metrics measure 1-to-1 reconstruction between the corresponding generated and real images, using arcface for ID \cite{deng2019arcface} and LPIPS \cite{zhang2018unreasonable} to extract image features. }
    \label{fig:supp_anchor_reconstruction}
\end{figure}

\subsection{Reconstruction of Personalized Set}
Our tuning procedure optimizes the generator to reconstruct a personalized reference set. This training objective is not the actual goal we aim to achieve, but rather a proxy to forming the personalized prior.
Nevertheless, as can be seen in \figref{fig:supp_anchor_reconstruction}, the tuning does reconstruct the reference set.

\subsection{Latent Spaces of $G_p$}
\label{subsec:latent_spaces_comparison}
\begin{figure}
	\centering
	\setlength{\tabcolsep}{1pt}
    
	\setlength{\imwidth}{0.2\linewidth}
	\begin{tabular}{*5c}
		
		Input & \personw & \wpplus & \pbeta & \pbetaplus 
		\\
        \includegraphics[width=\imwidth]{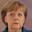} &
		\includegraphics[width=\imwidth]{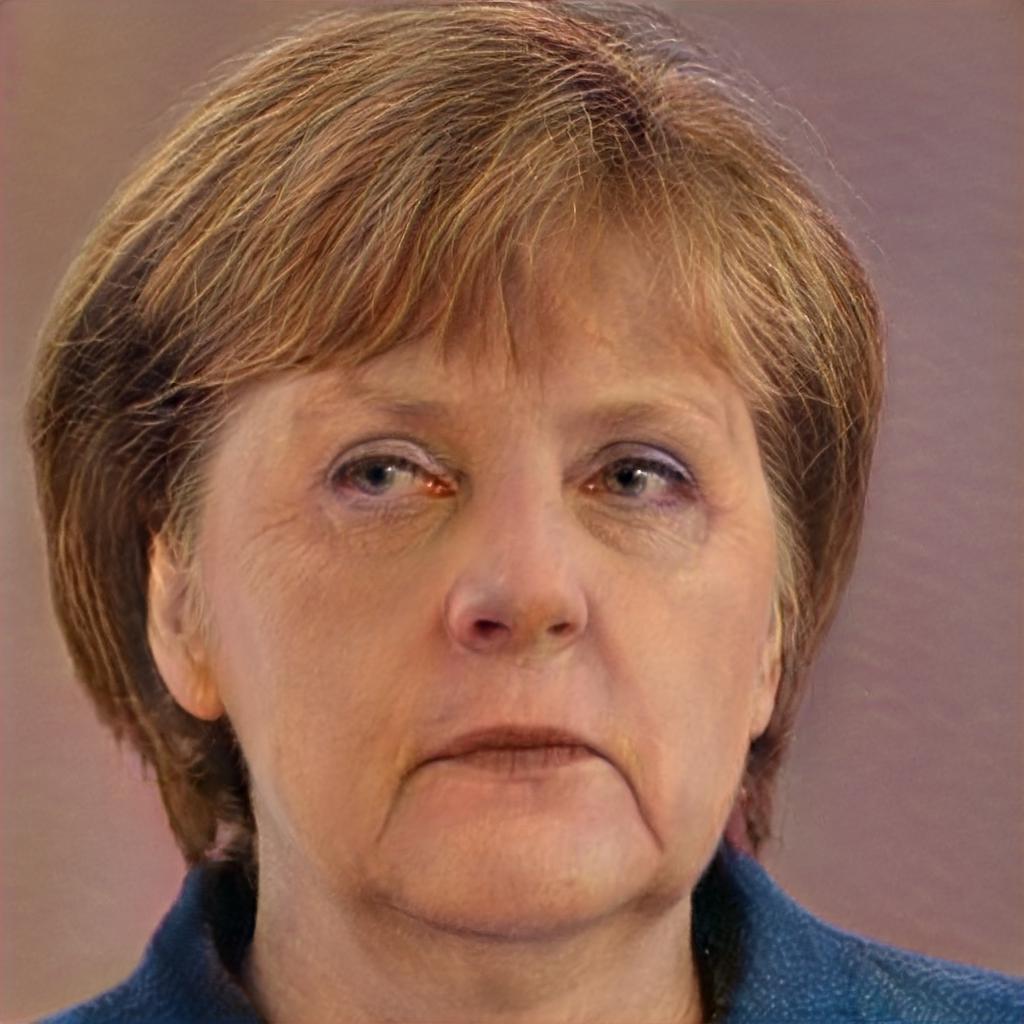} &
		\includegraphics[width=\imwidth]{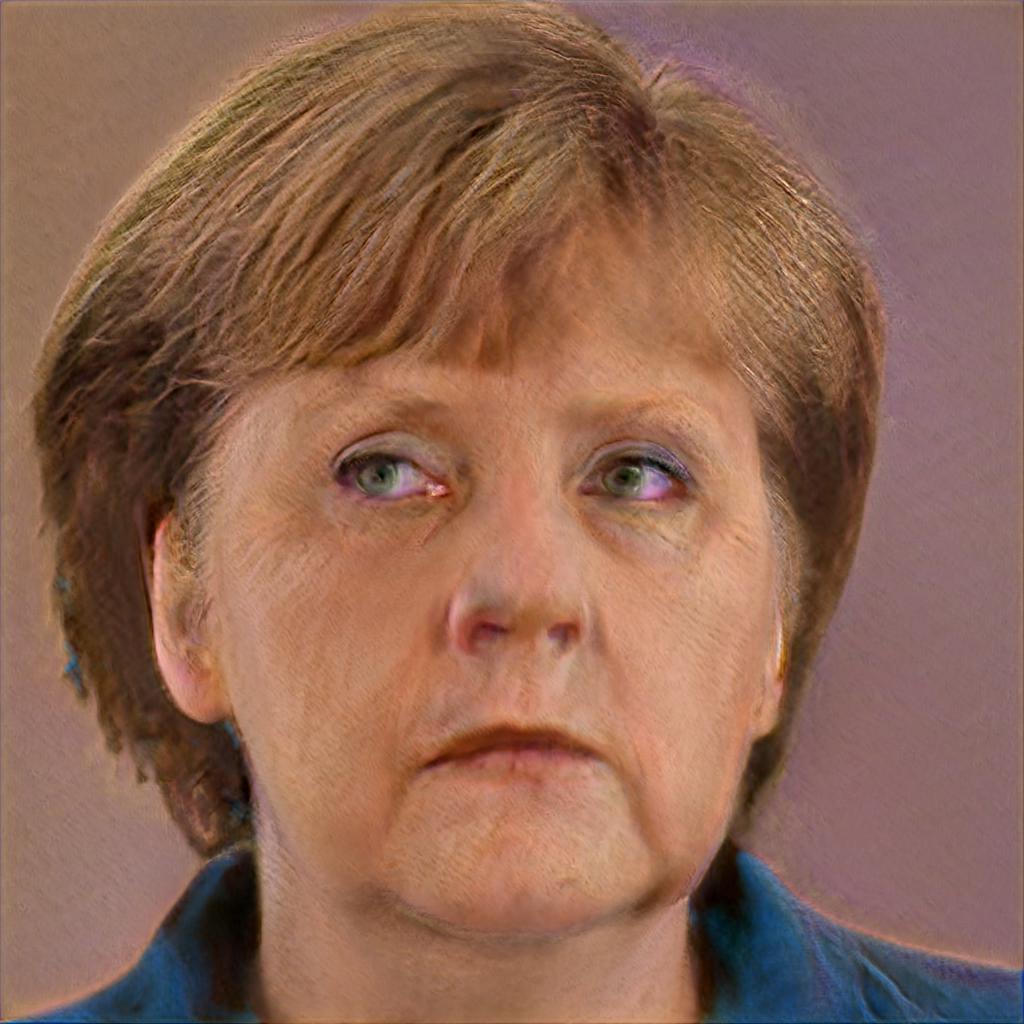} &
		\includegraphics[width=\imwidth]{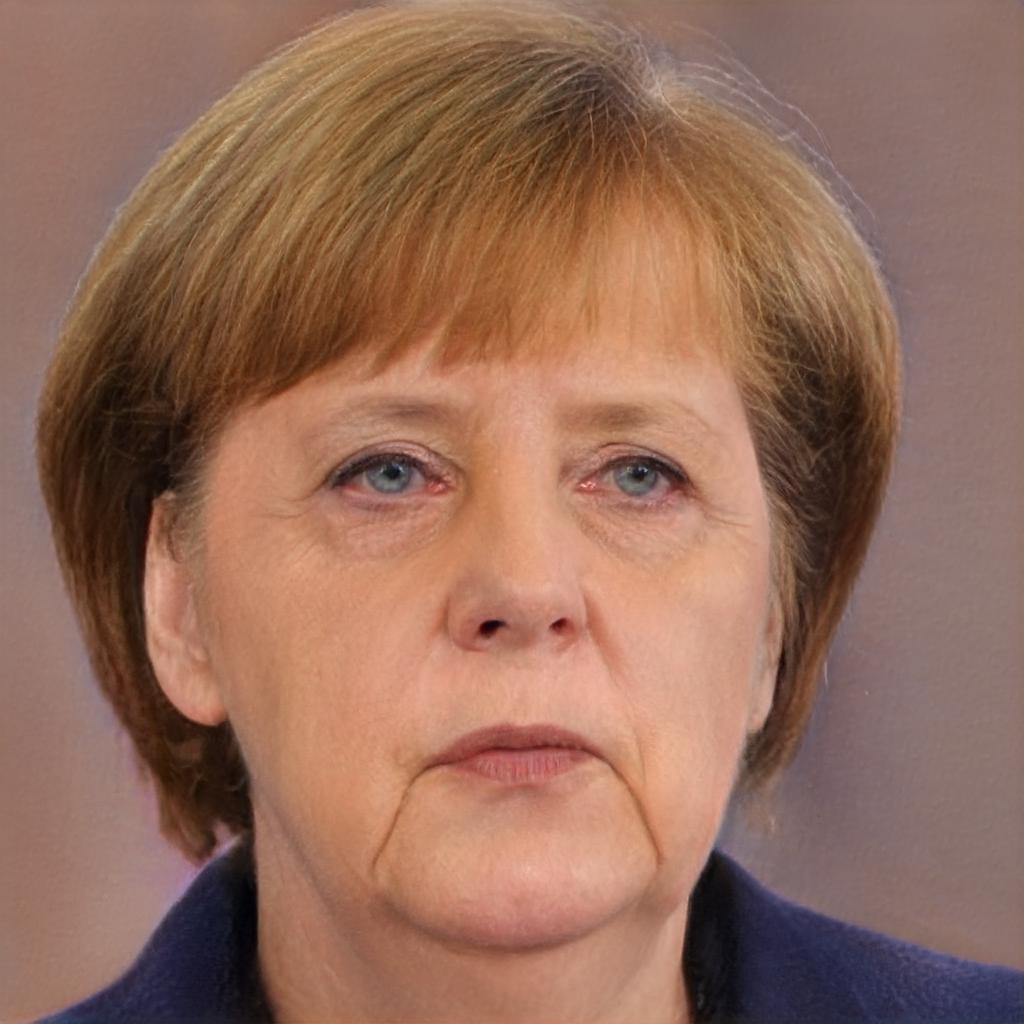} &
        \includegraphics[width=\imwidth]{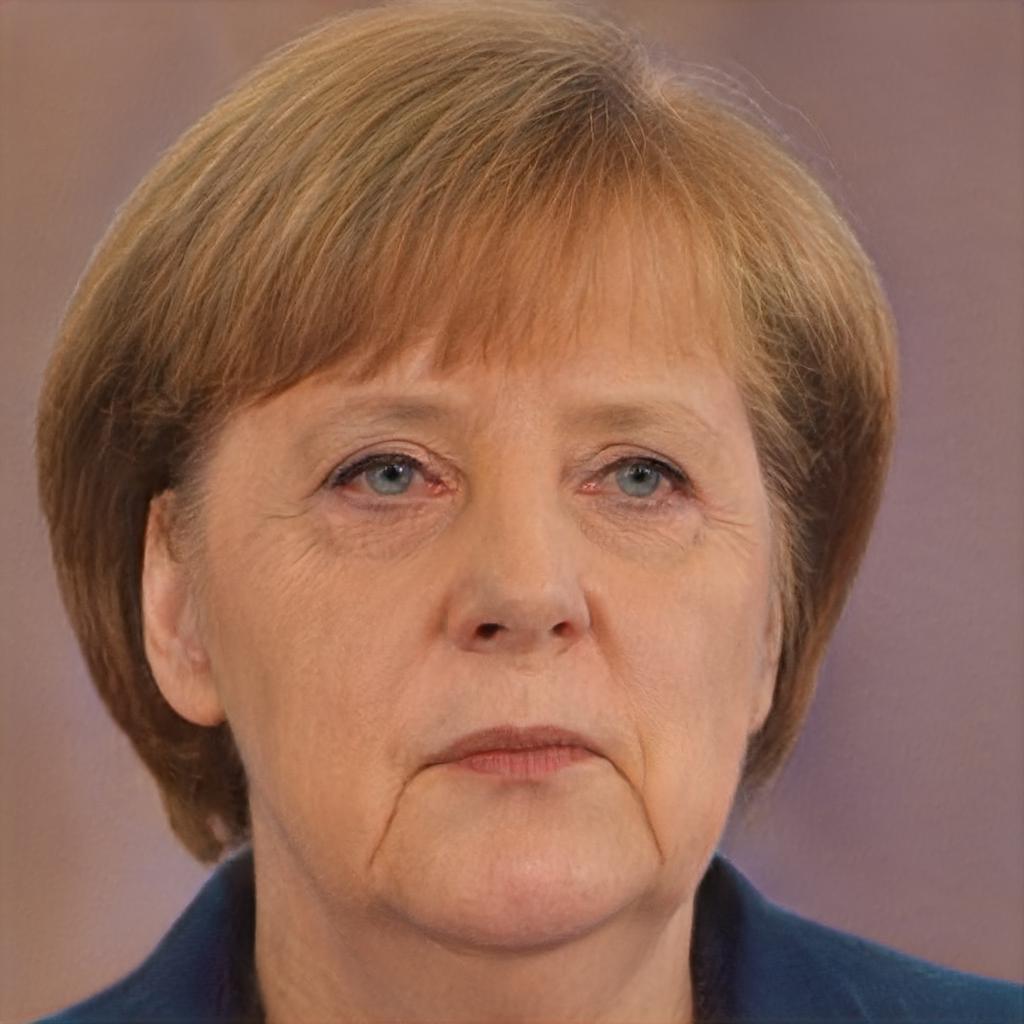}
        
        \\

        \includegraphics[width=\imwidth]{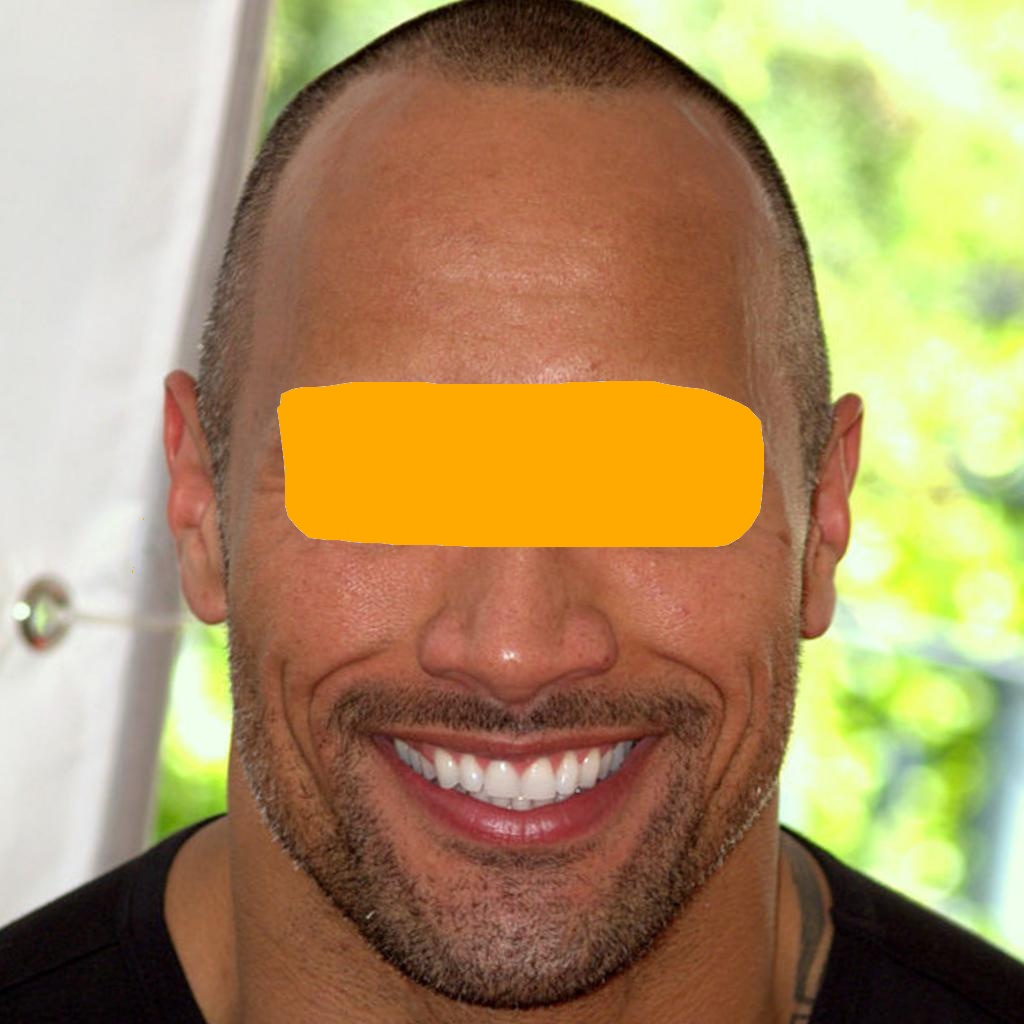} &
		\includegraphics[width=\imwidth]{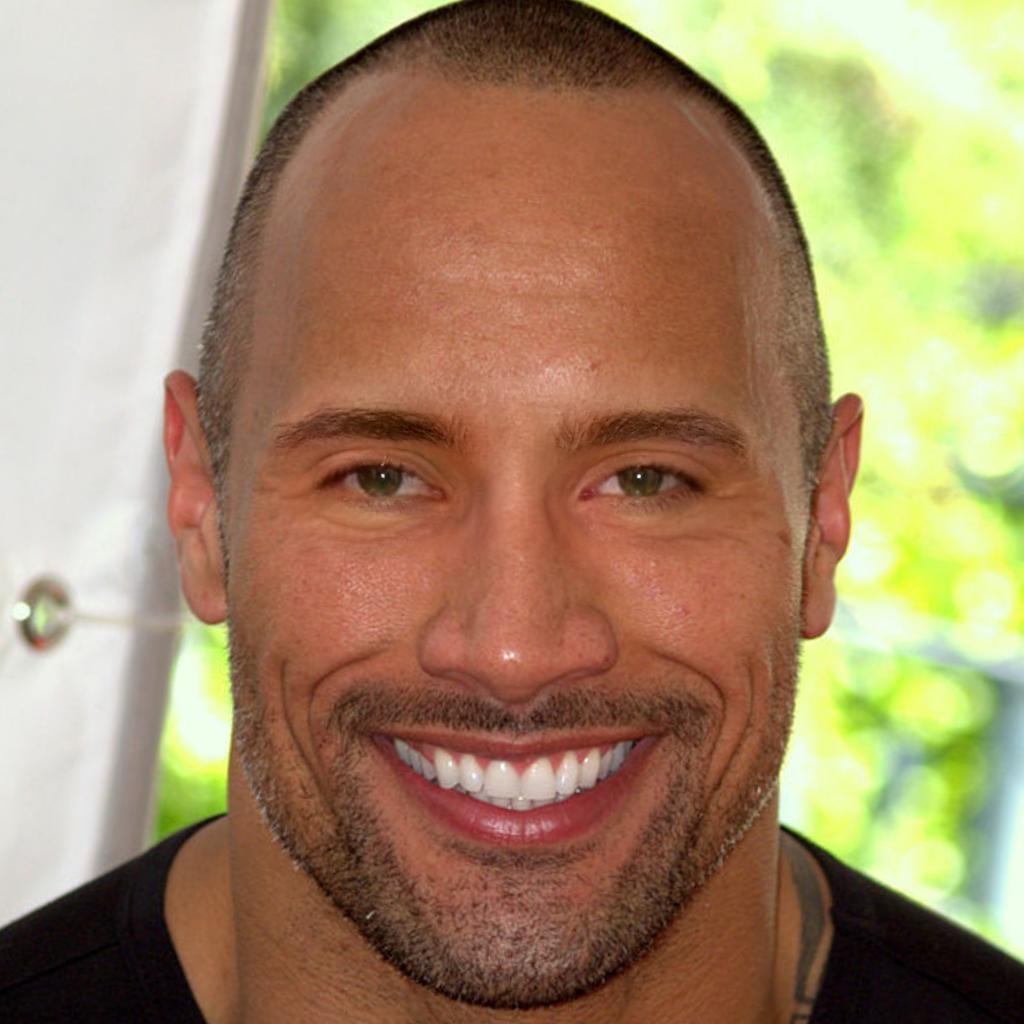} &
		\includegraphics[width=\imwidth]{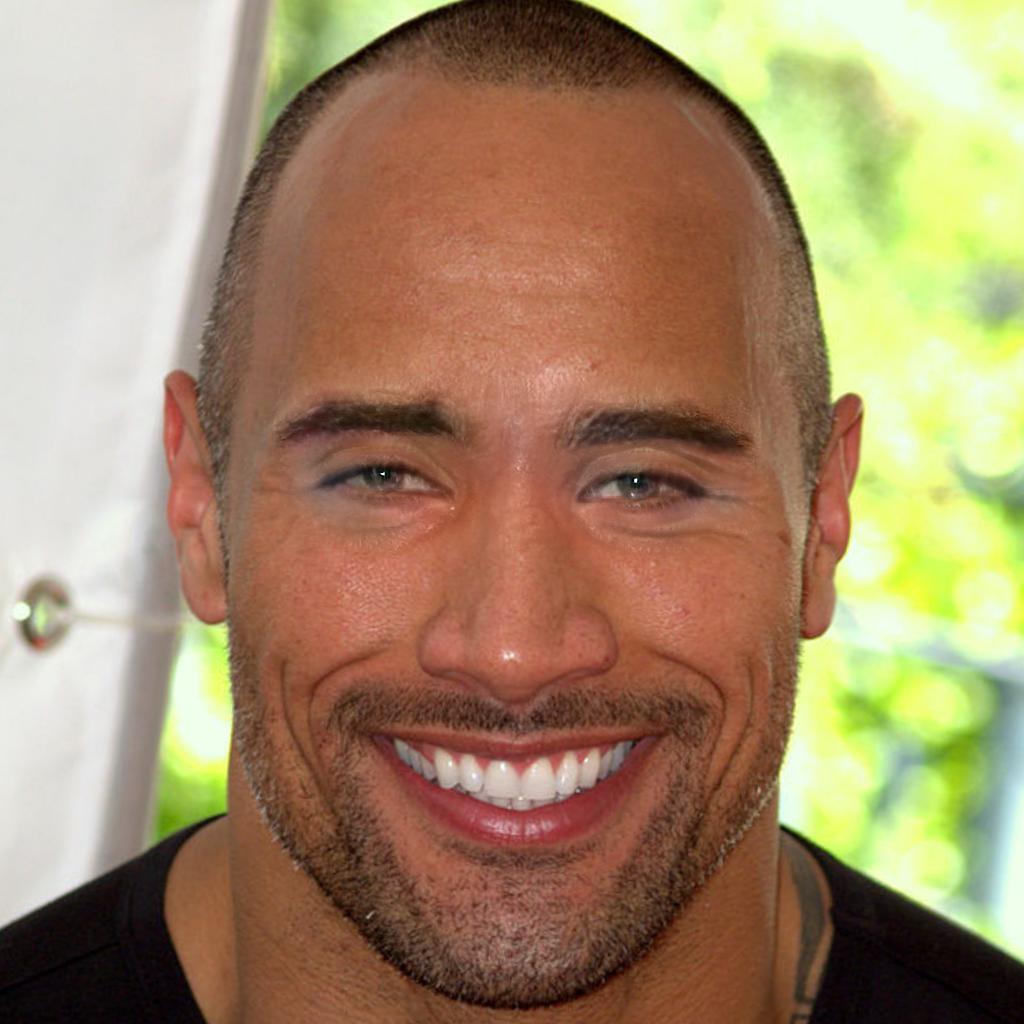} &
		\includegraphics[width=\imwidth]{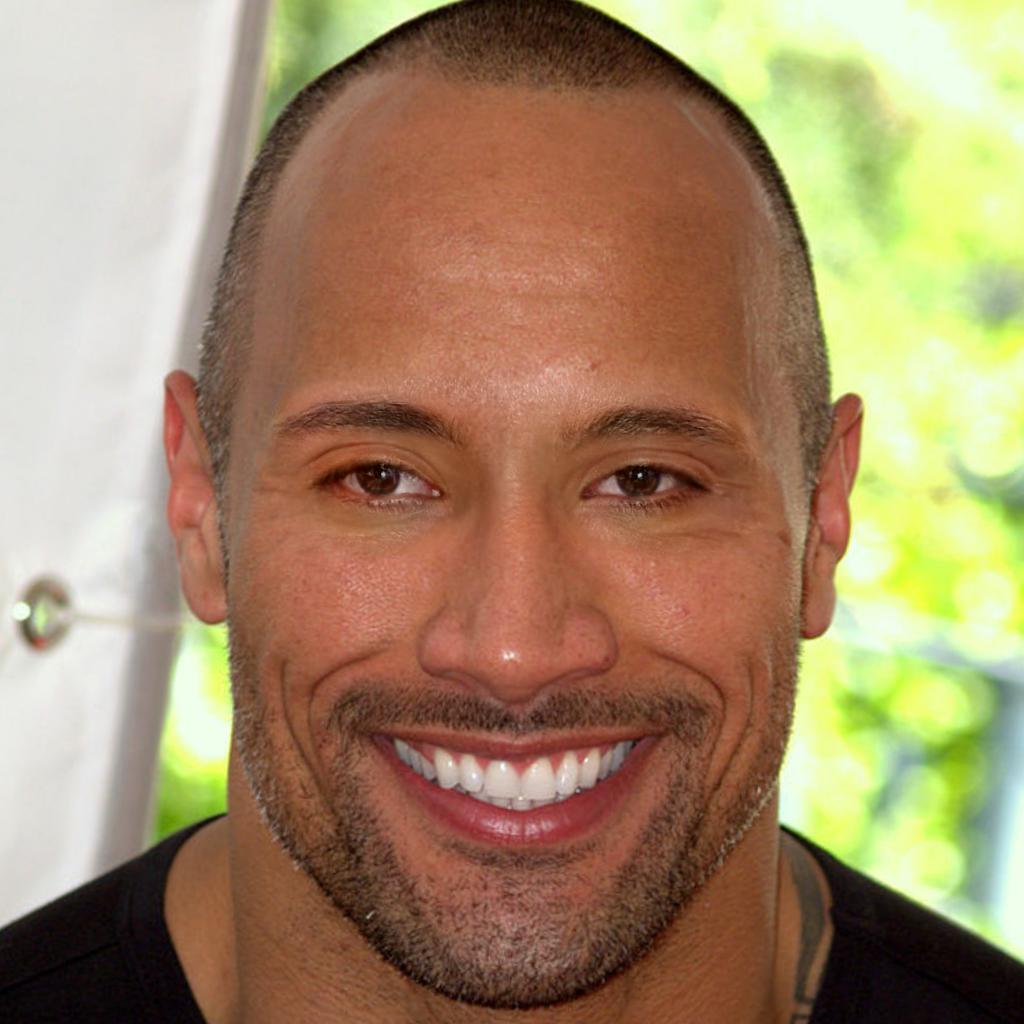} &
        \includegraphics[width=\imwidth]{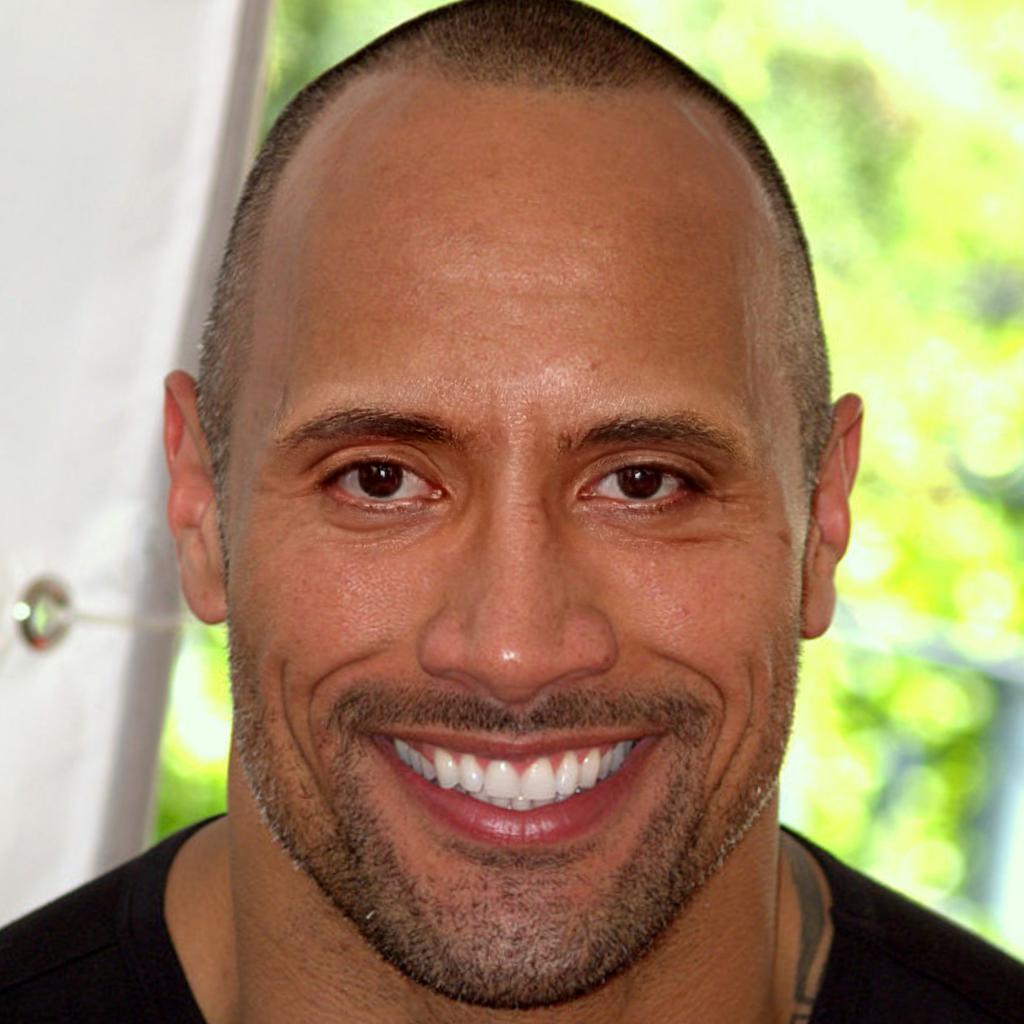}
        
	\end{tabular}
	
	\caption{
	We demonstrate the effect of projecting degraded input images into different latent spaces for image enhancement.
	We find that projecting to \personw and \wpplus yields non-personalized and low quality results. Projection to \pbetaplus is superior to \pbeta in terms of personalization and fidelity.
	}
	\label{fig:ablation_latent_space}
\end{figure}

We next evaluate the effect the choice of latent space has on image enhancement results.
As demonstrated in \figref{fig:ablation_latent_space}, projecting to \pbetaplus is favorable for image enhancement. Enhancement results in \personw and \wpplus are of low quality and not personalized. This support our previous findings that the space has not been personalized uniformly and additionally indicates that the projection does not converge to the personalized space without explicit regularization. 

We additionally find that for image enhancement, \pbetaplus is superior to \pbeta in two manners. First, it provides more expressive power, as \wplus does with respect to \w (see first row in \figref{fig:ablation_latent_space}). Second, we find that \pbeta sometimes produces slightly less identity preserving results than \pbetaplus (see last row in \figref{fig:ablation_latent_space}). This is initially surprising as \w is known to provide more reliable prior than \wplus, however with \pbeta and \pbetaplus the opposite appears to be true. We empirically find that results from \pbeta often fully leverage the allowed dilation and arrive at exactly $\beta$, while results from \pbetaplus often converge at a smaller dilation. We speculate that results in \pbeta stray to further dilation to mitigate the limited expressiveness.

\subsection{Nearest-Neighbor Experiments}
\begin{figure*}
    \centering
    \setlength{\imwidth}{0.1\linewidth}
    \setlength{\tabcolsep}{1pt}
    \begin{subfigure}[b]{0.2\linewidth}
        \begin{tabular}{*3c}
             MyStyle & NN \\
             \includegraphics[width=\imwidth]{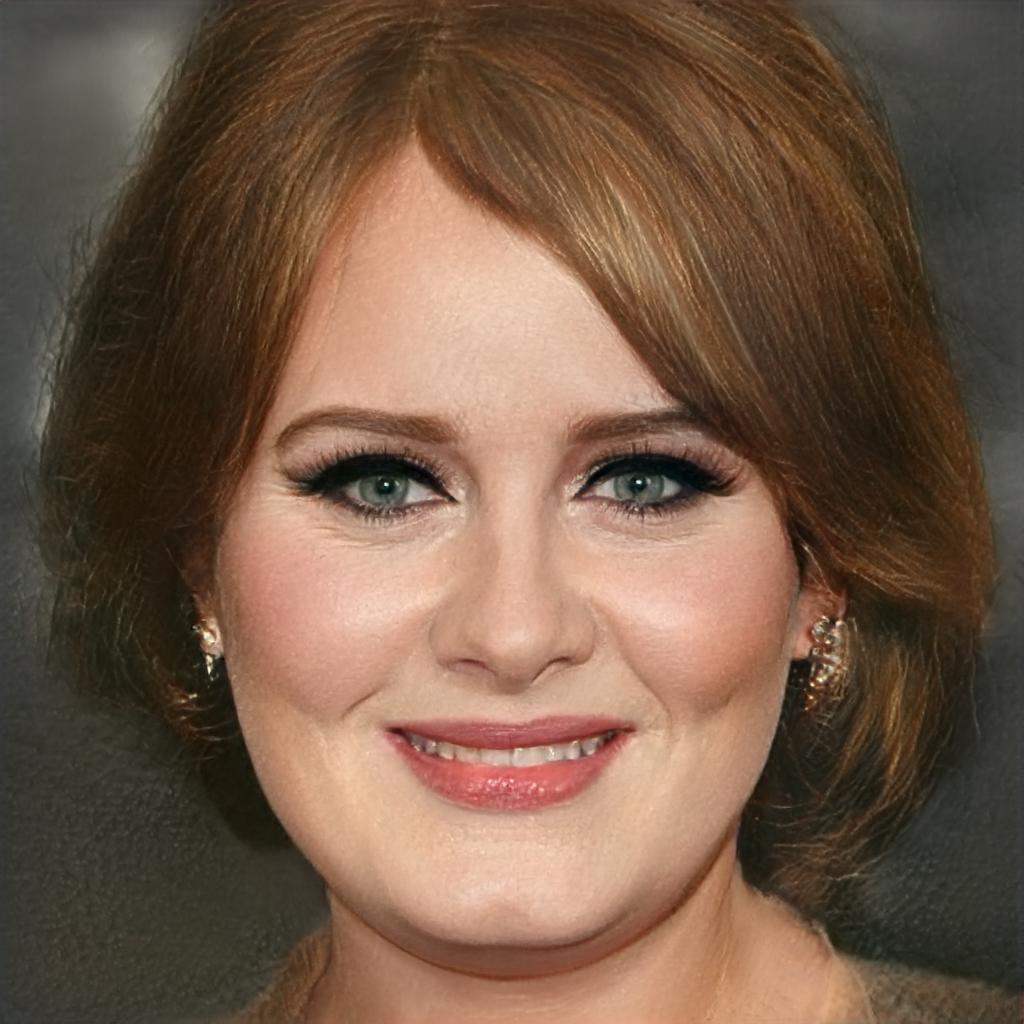} &
             \includegraphics[width=\imwidth]{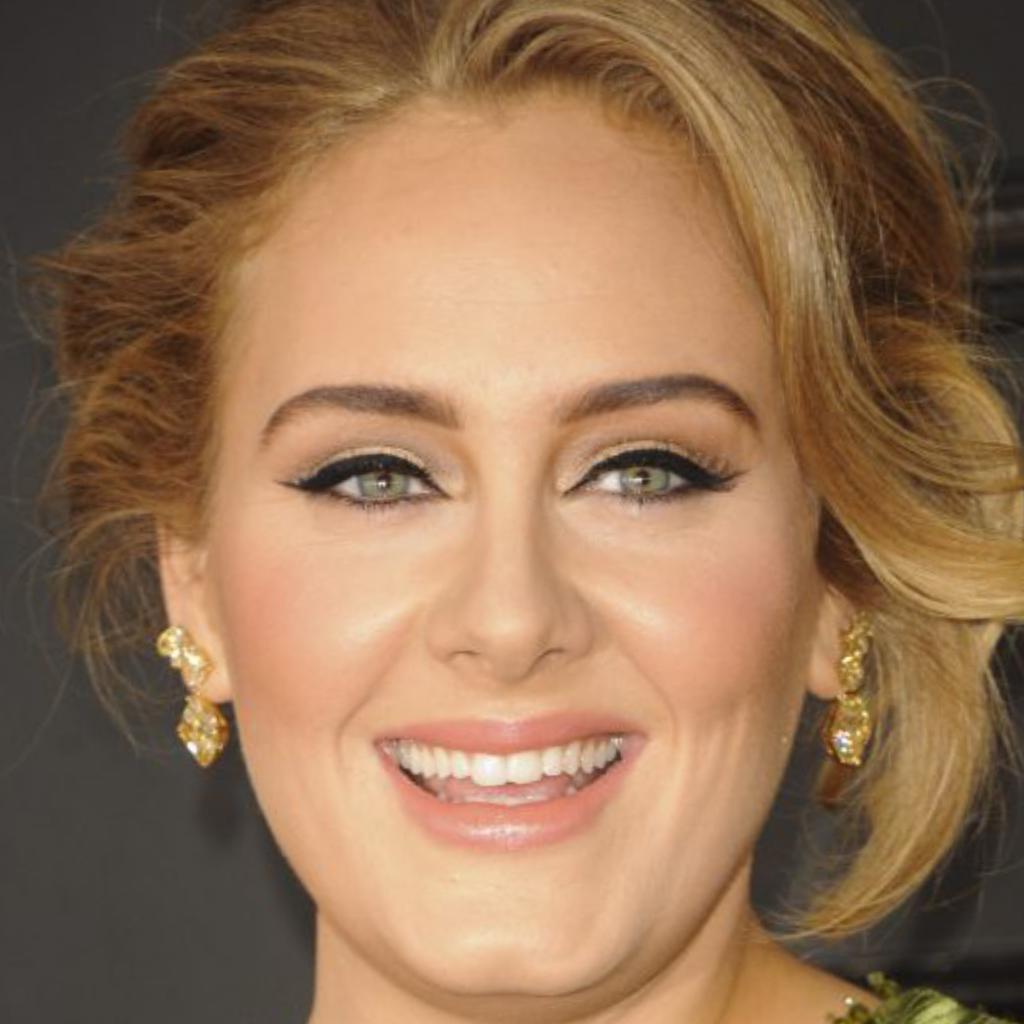} \\
             
             \includegraphics[width=\imwidth]{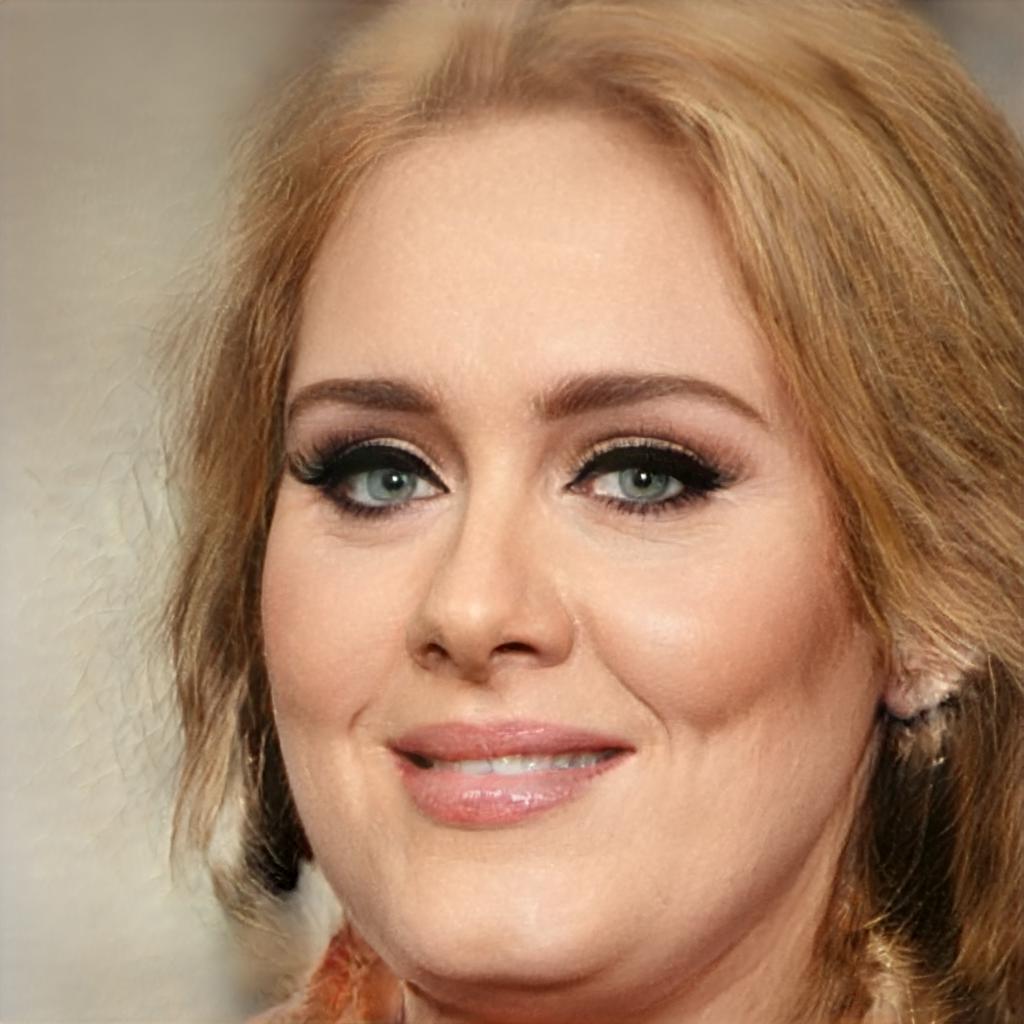} &
             \includegraphics[width=\imwidth]{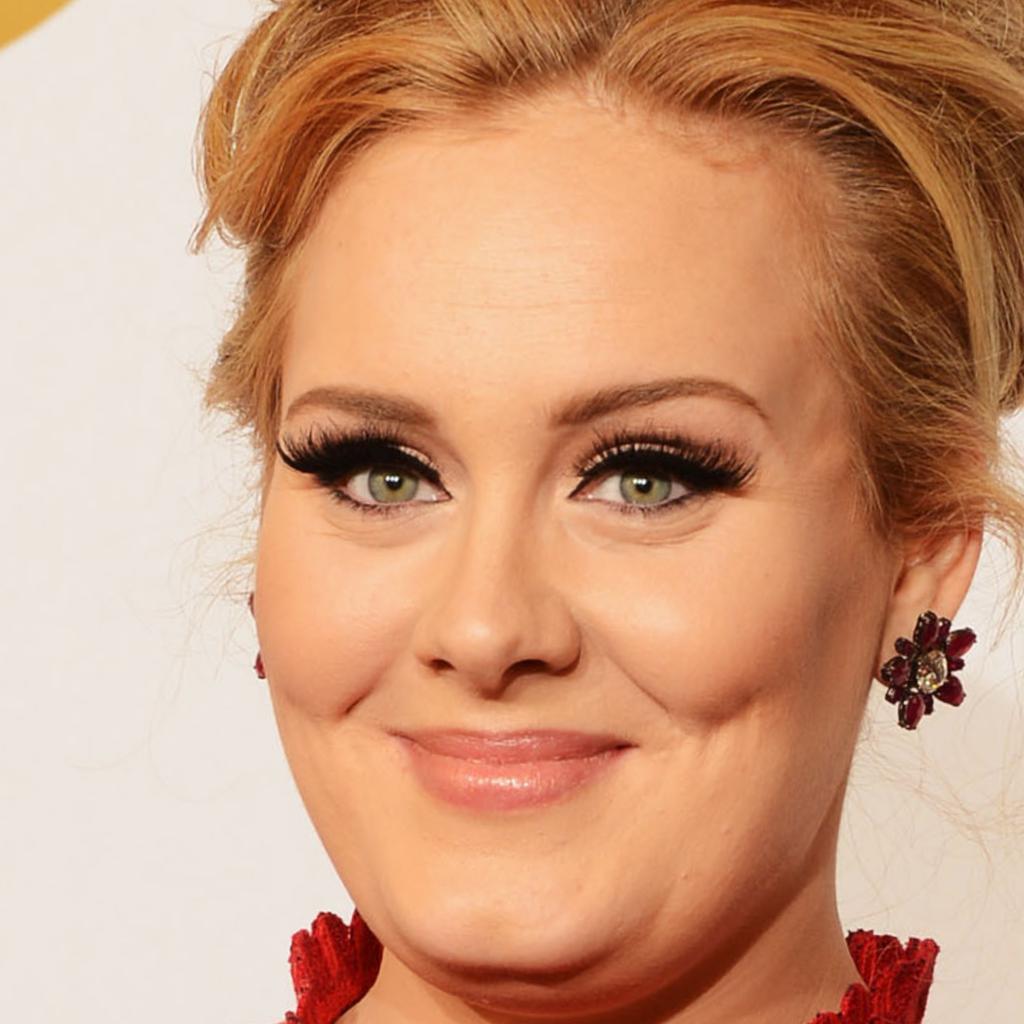} \\
             
             \includegraphics[width=\imwidth]{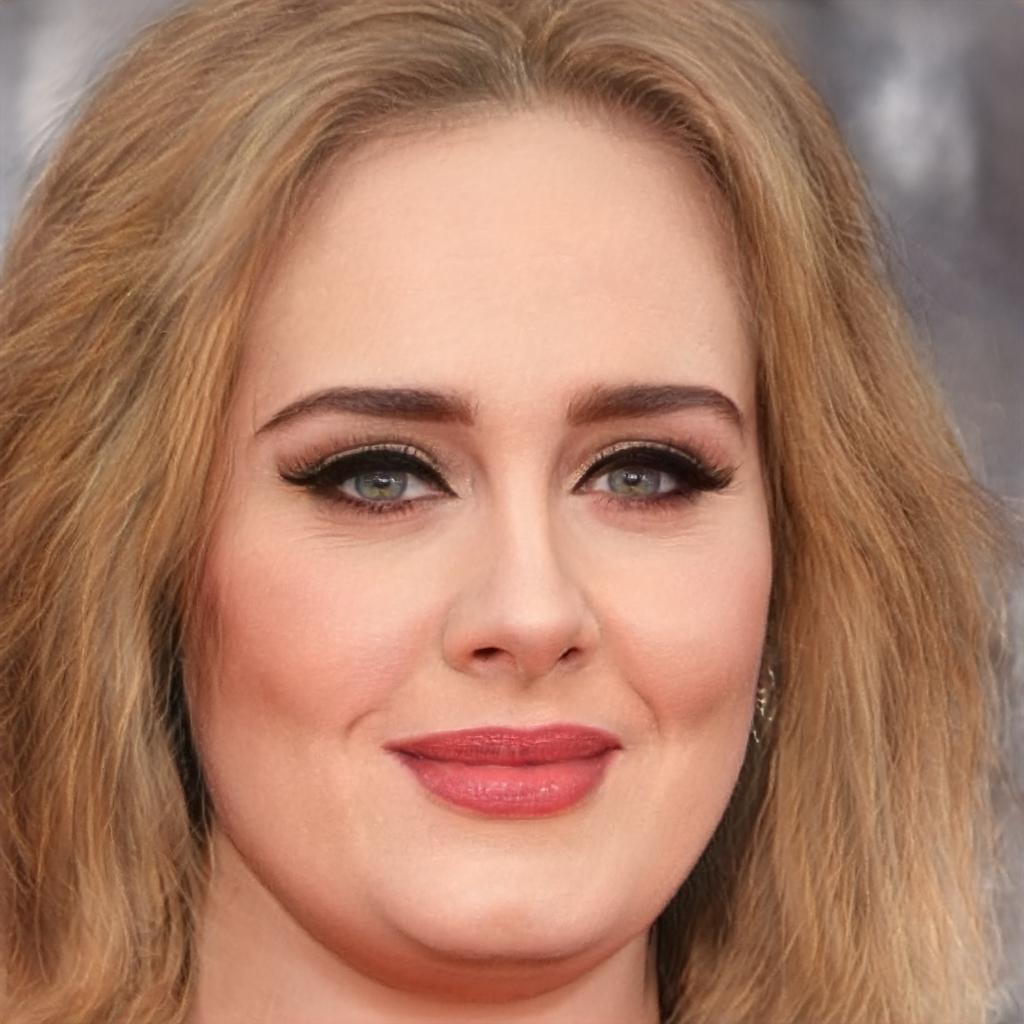} & 
             \includegraphics[width=\imwidth]{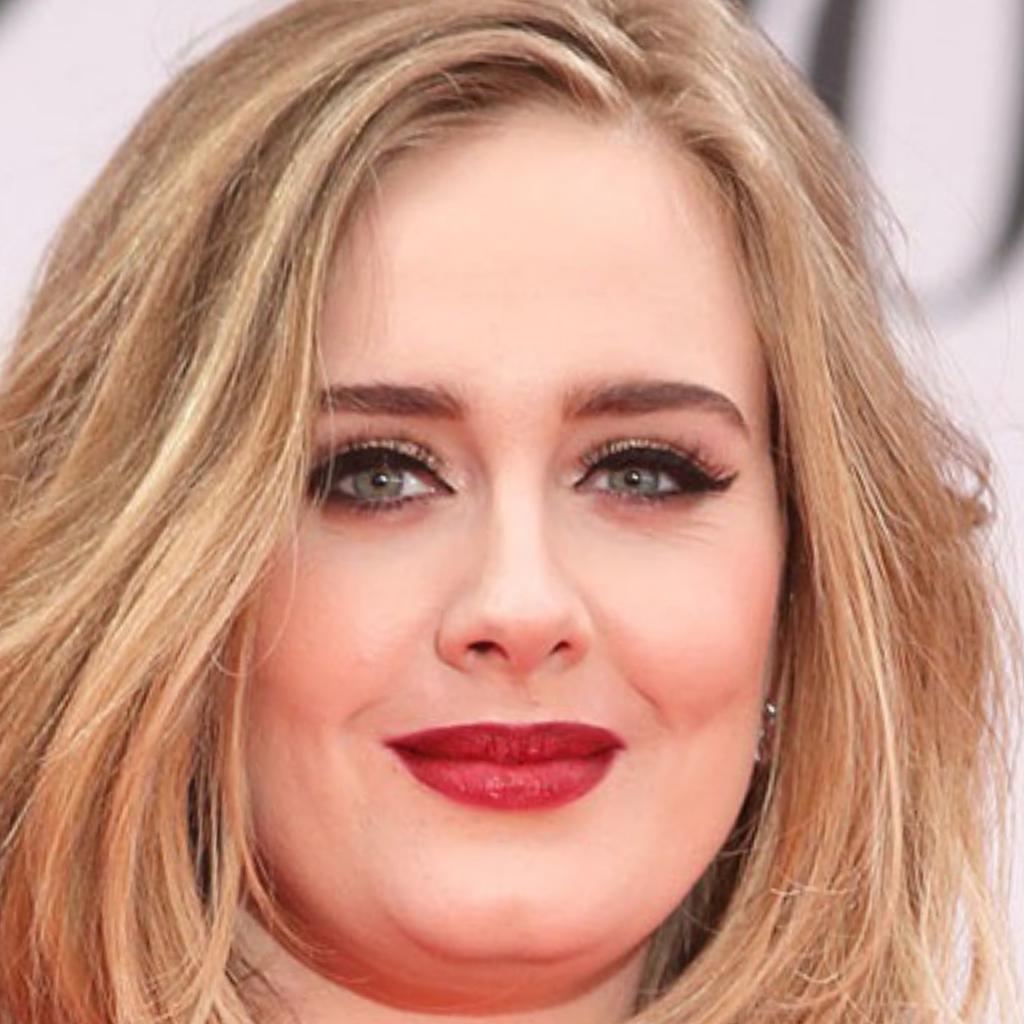} \\
             
             \includegraphics[width=\imwidth]{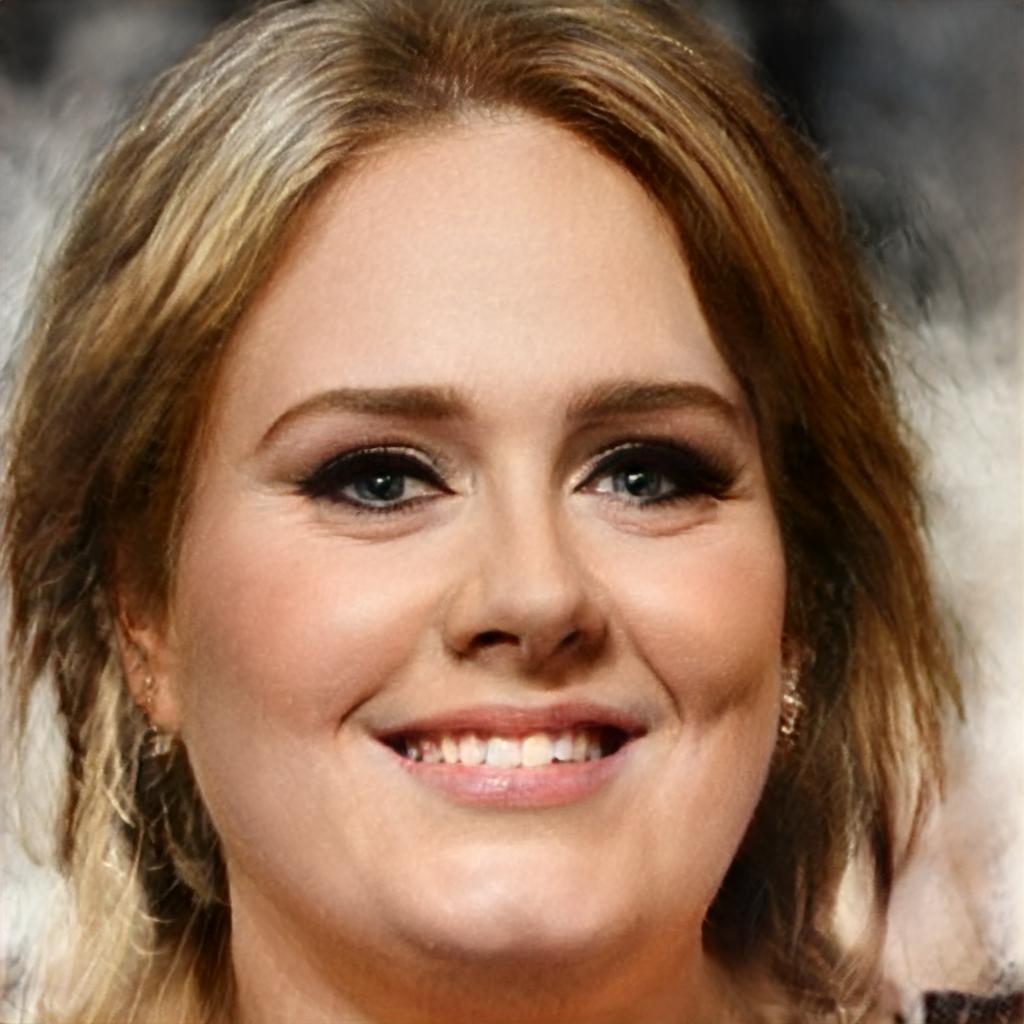} & 
             \includegraphics[width=\imwidth]{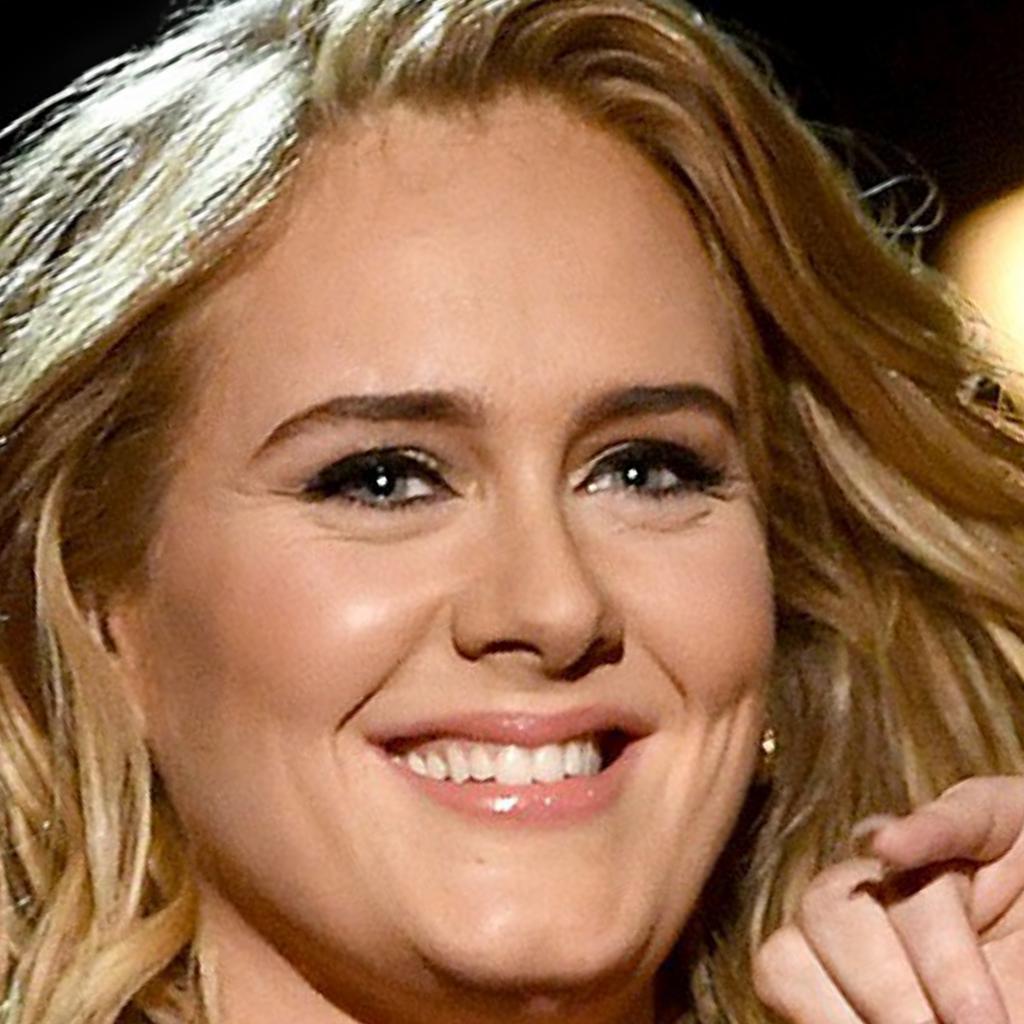} \\
        \end{tabular}
    \caption{Synthesis}
    \end{subfigure} 
    \qquad
    \begin{subfigure}[b]{0.3\linewidth}
        \begin{tabular}{*3c}
             Input & MyStyle & NN \\
             \includegraphics[width=\imwidth]{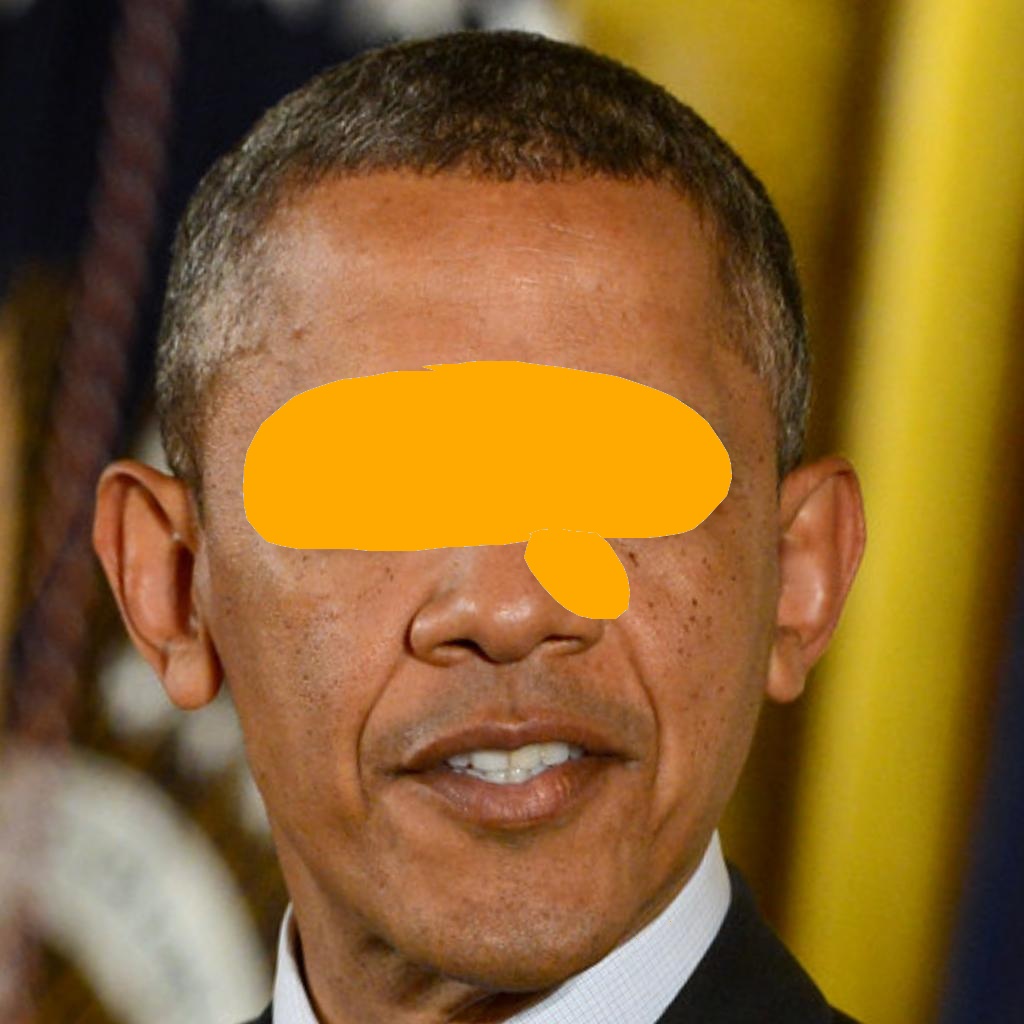} &
             \includegraphics[width=\imwidth]{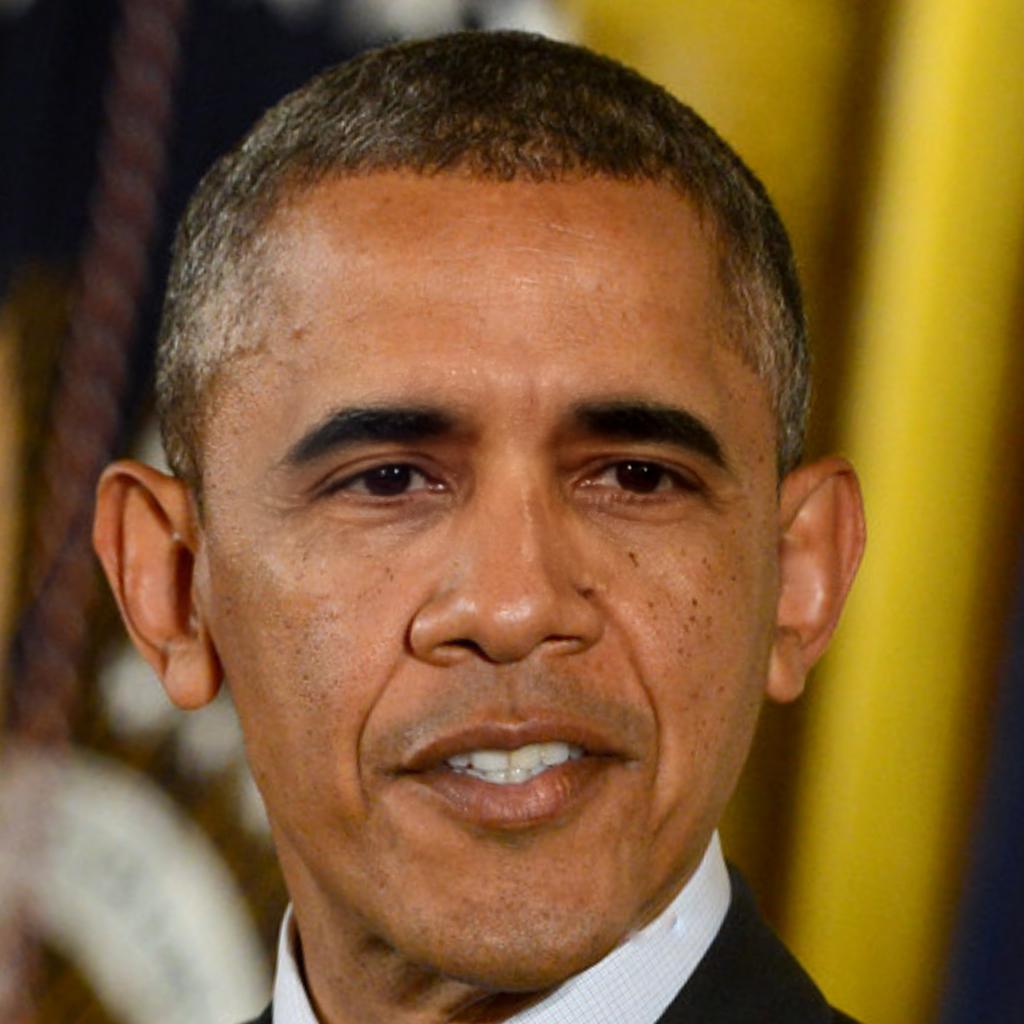} &
             \includegraphics[width=\imwidth]{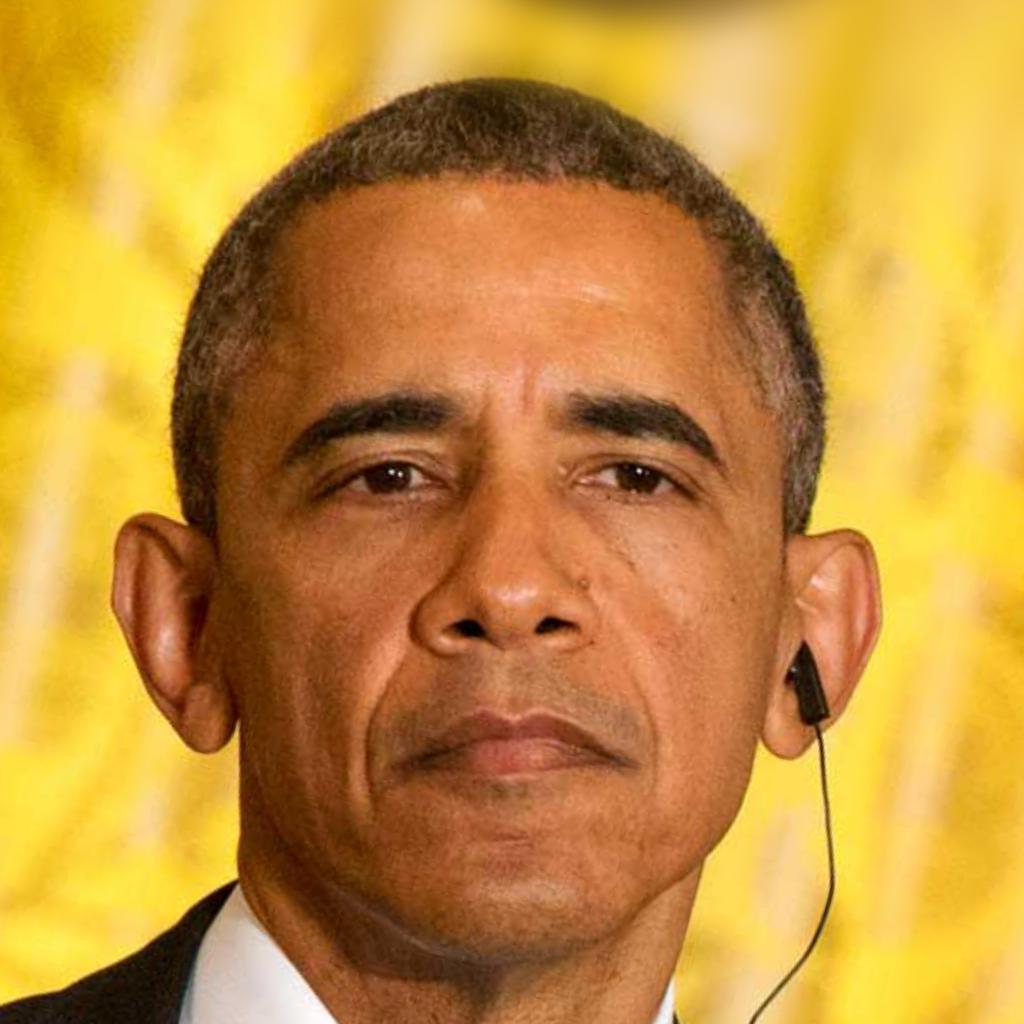} \\
             
             \includegraphics[width=\imwidth]{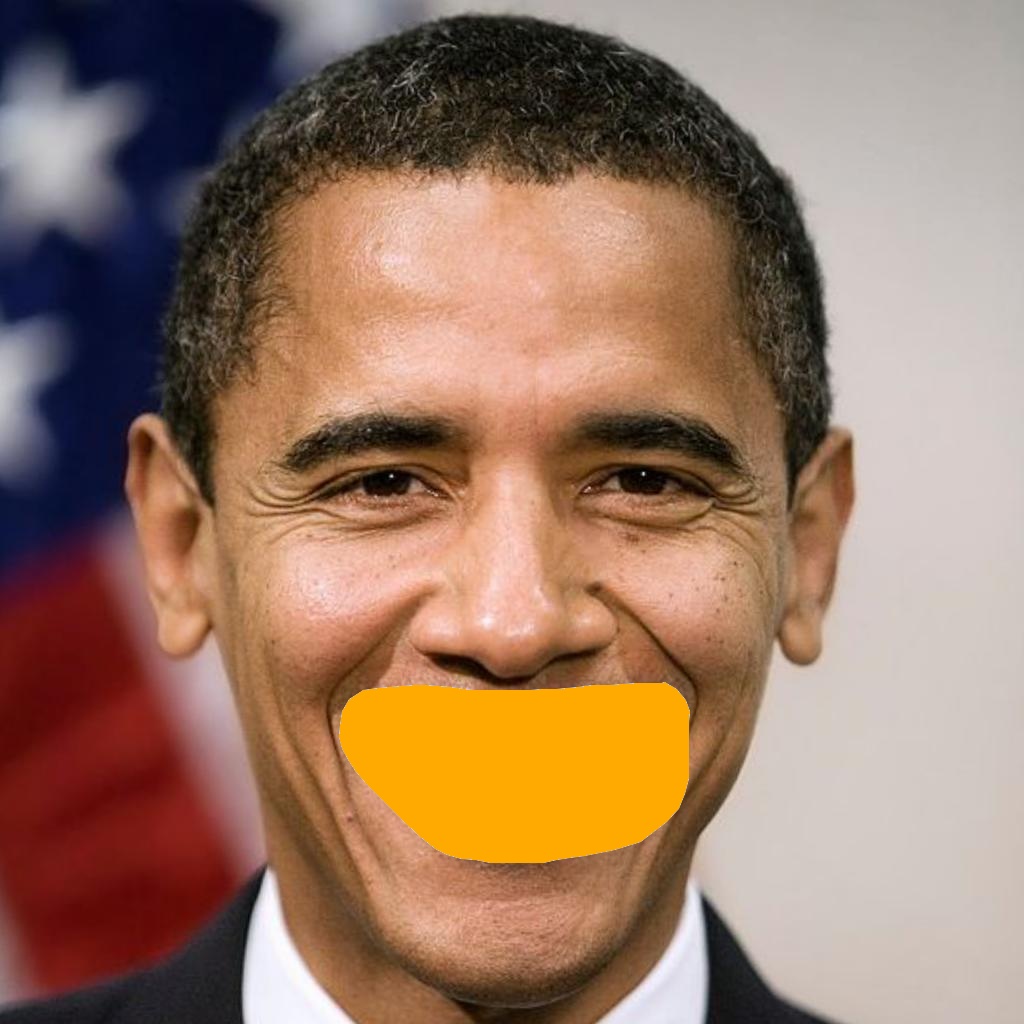} &
             \includegraphics[width=\imwidth]{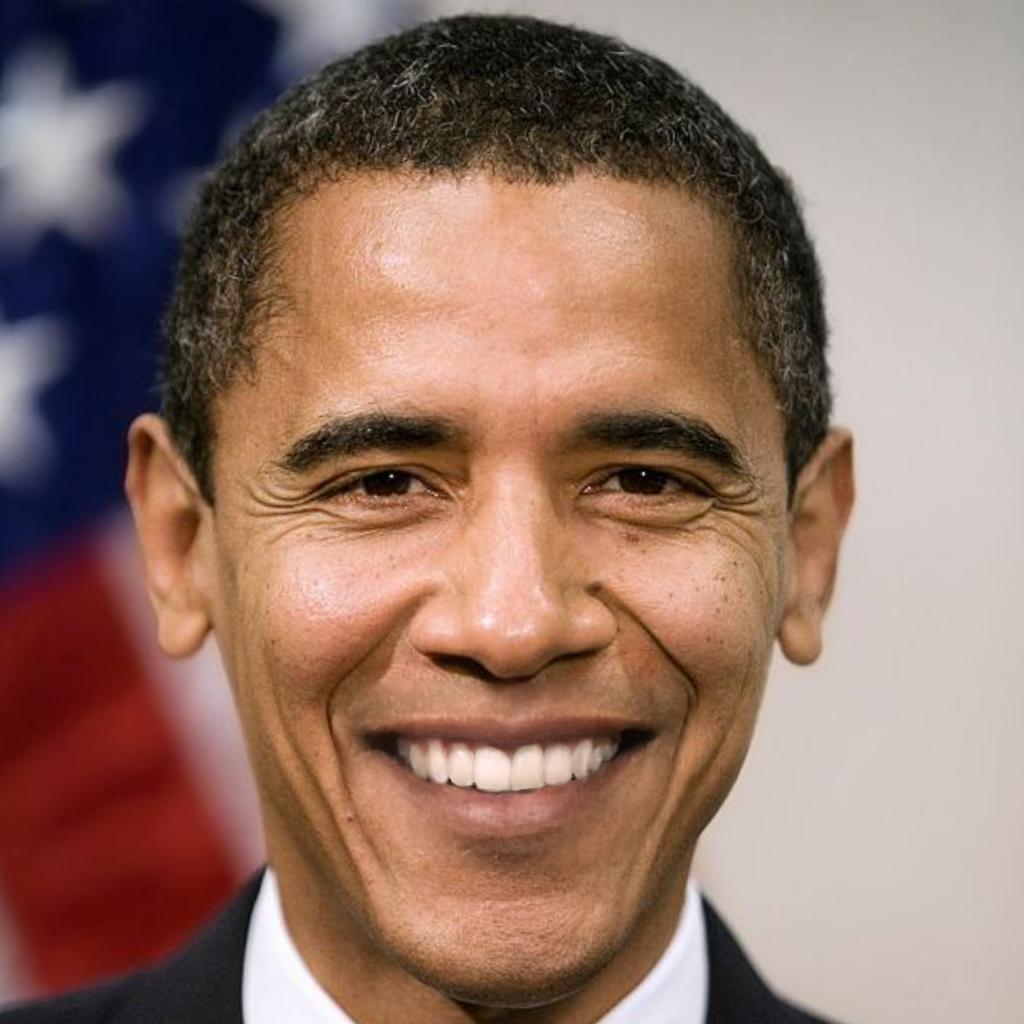} &
             \includegraphics[width=\imwidth]{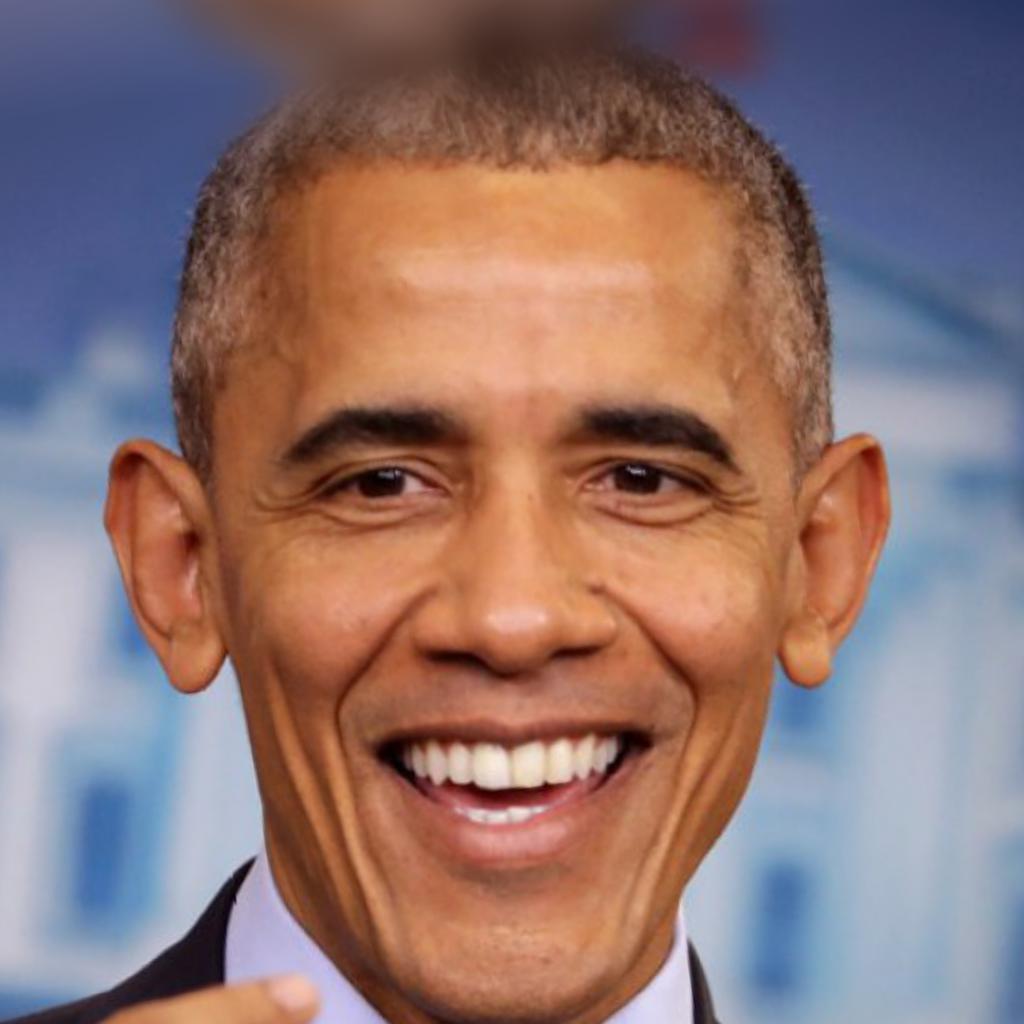} \\
             
             \includegraphics[width=\imwidth]{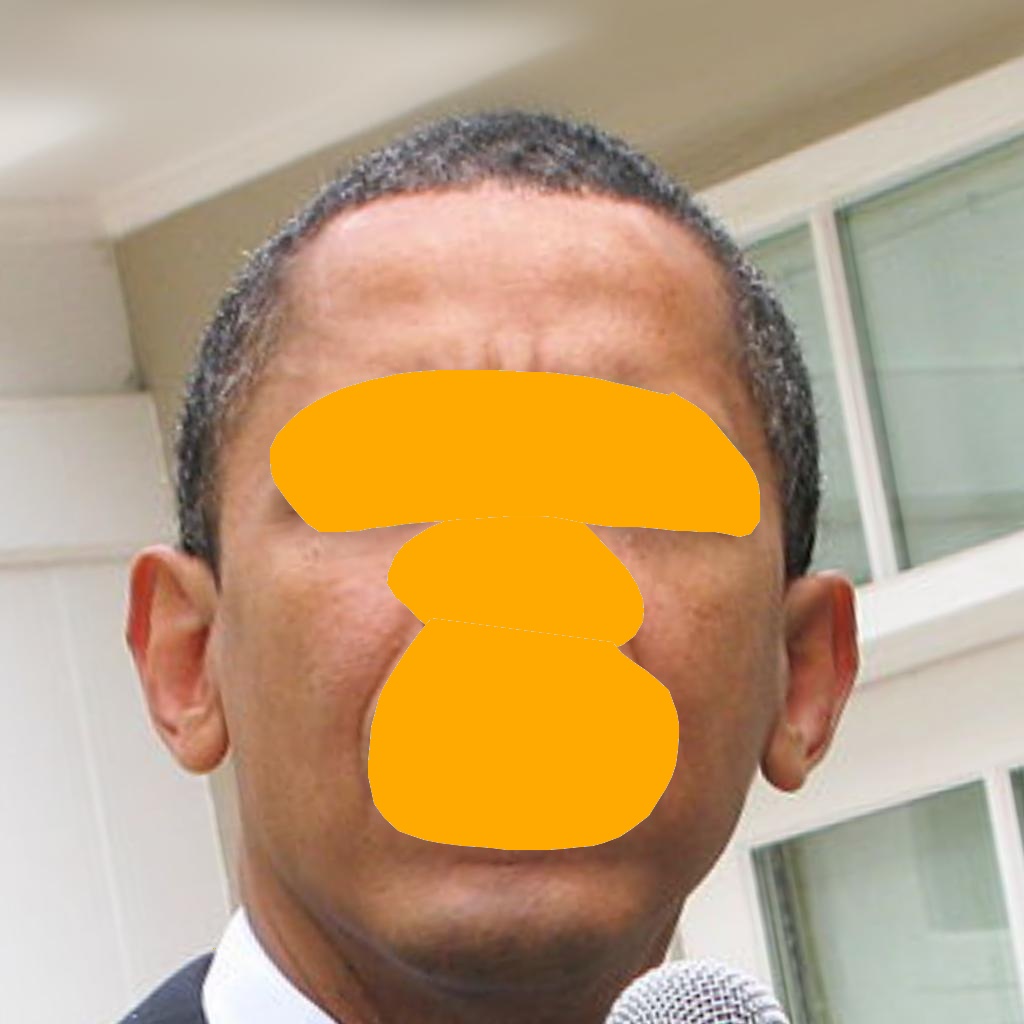} &
             \includegraphics[width=\imwidth]{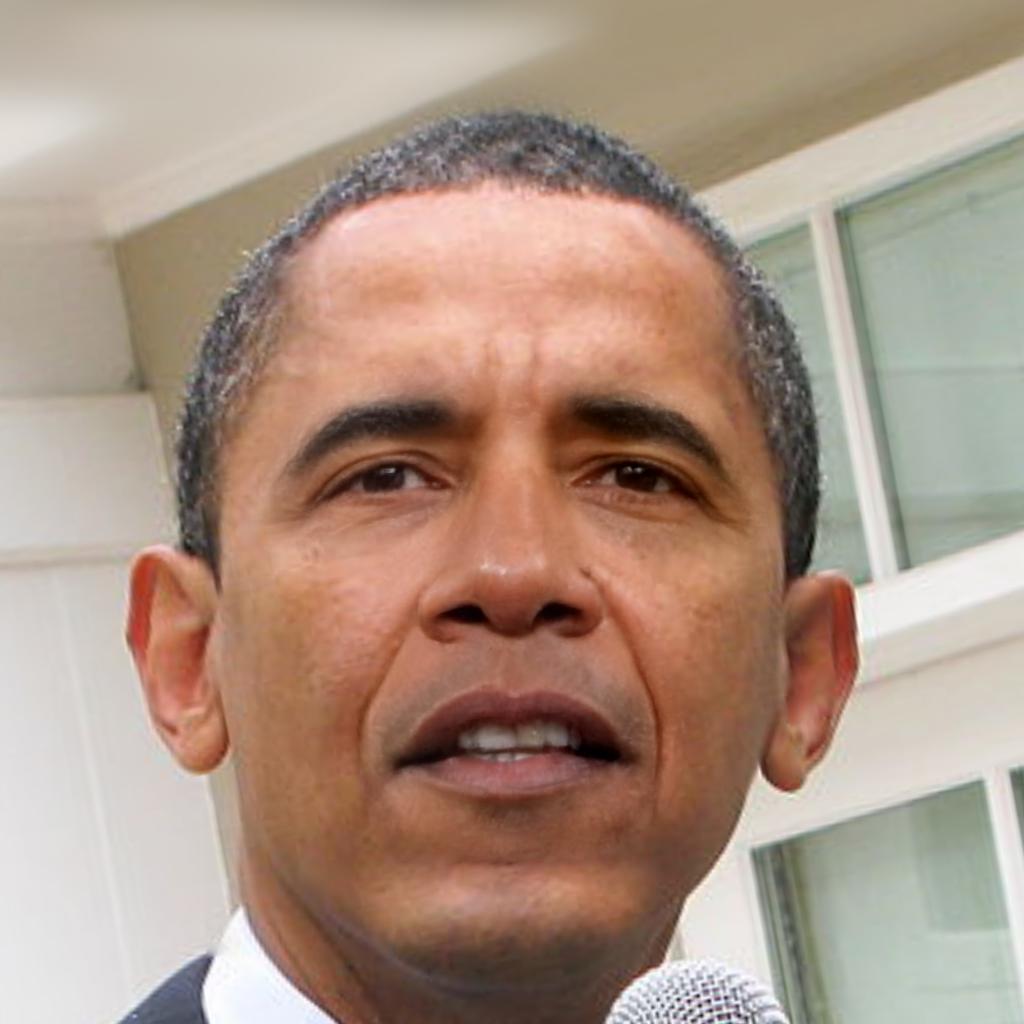} &
             \includegraphics[width=\imwidth]{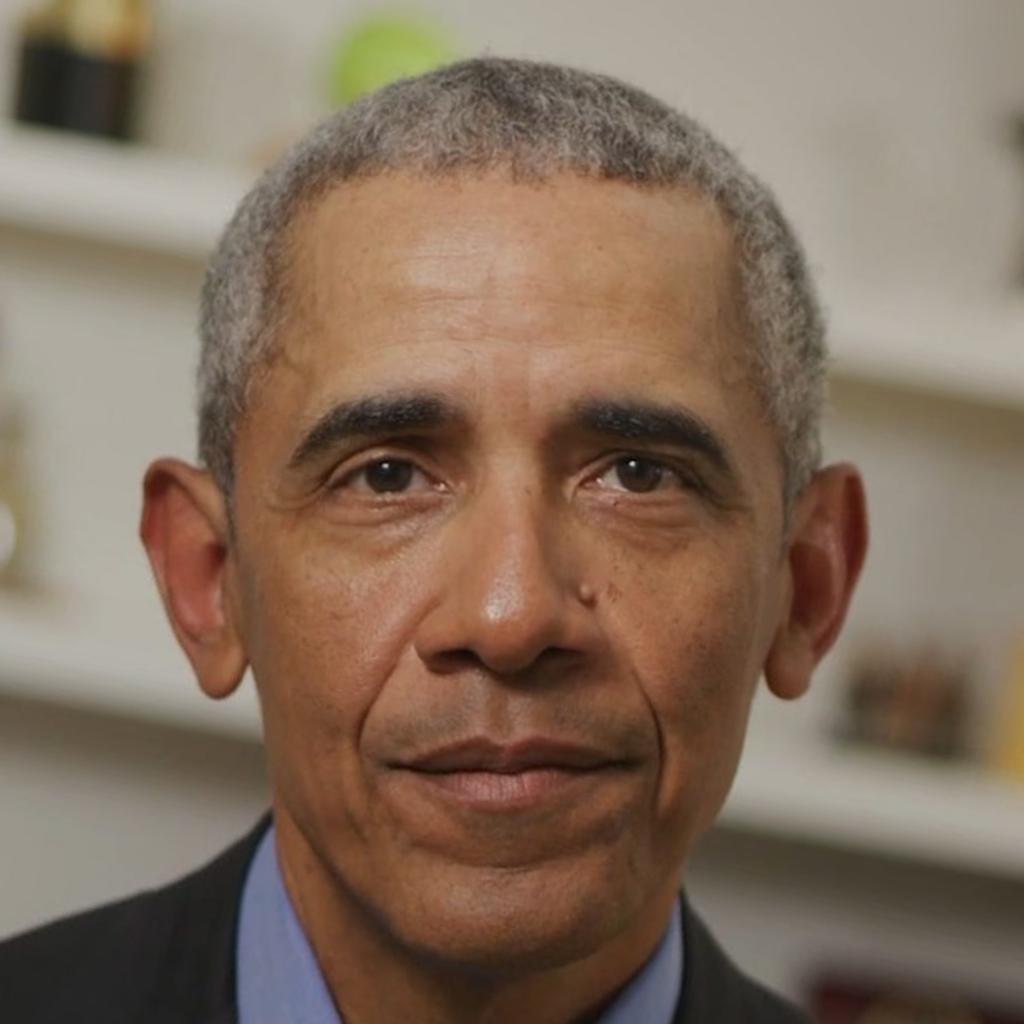} \\
             
             \includegraphics[width=\imwidth]{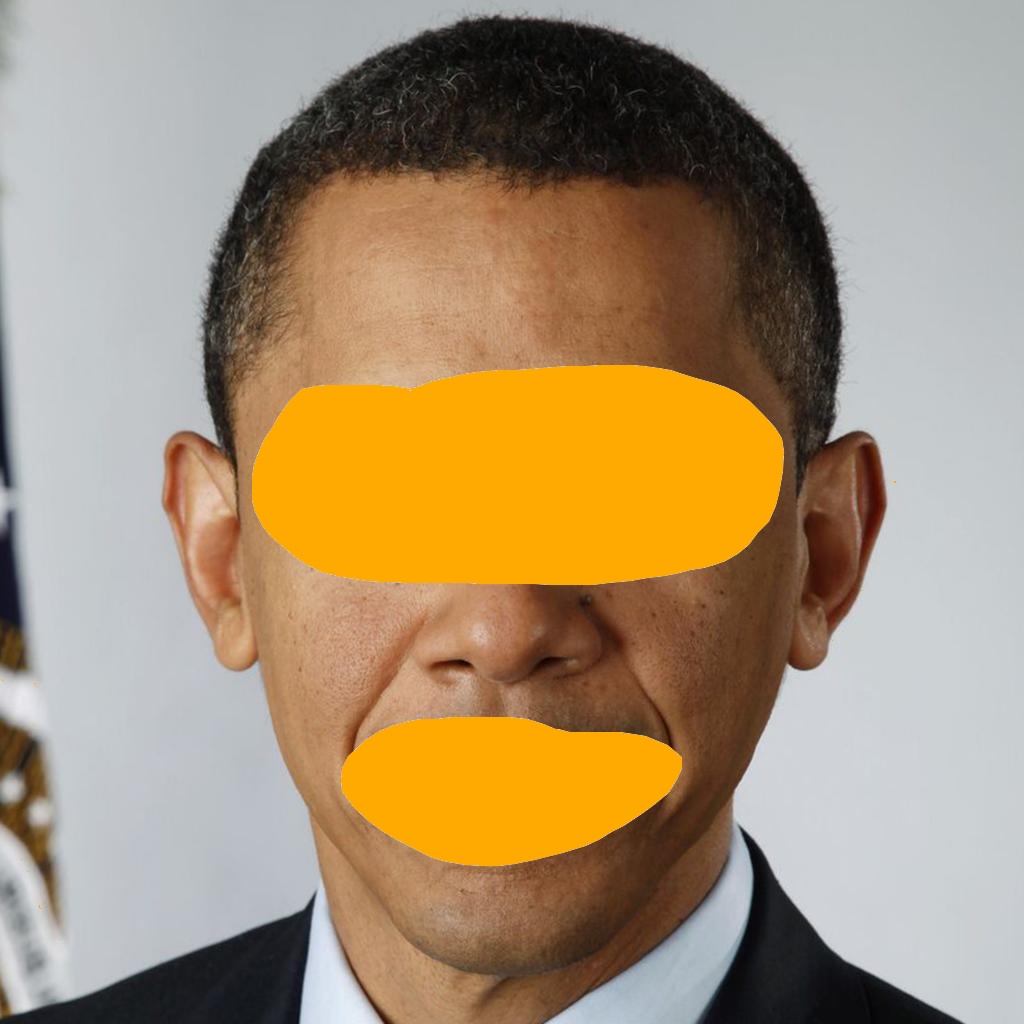} &
             \includegraphics[width=\imwidth]{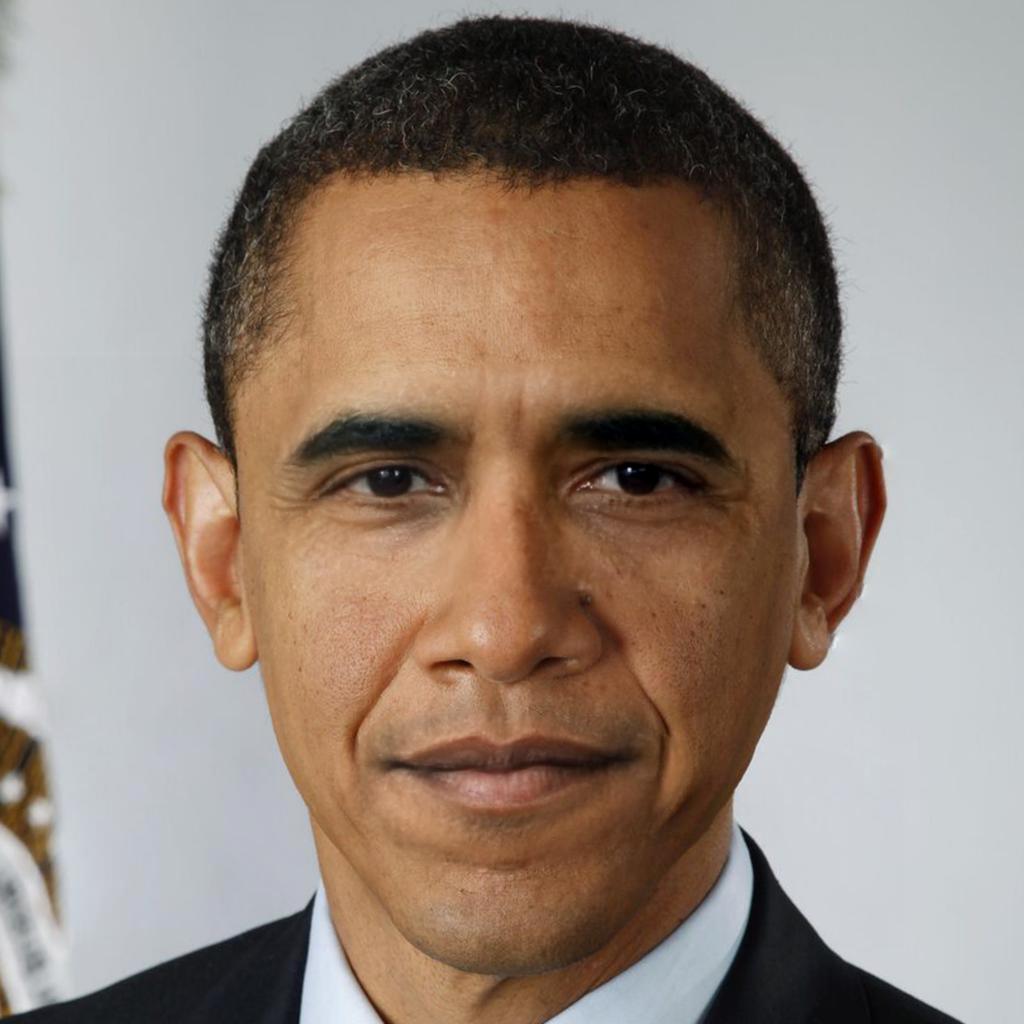} &
             \includegraphics[width=\imwidth]{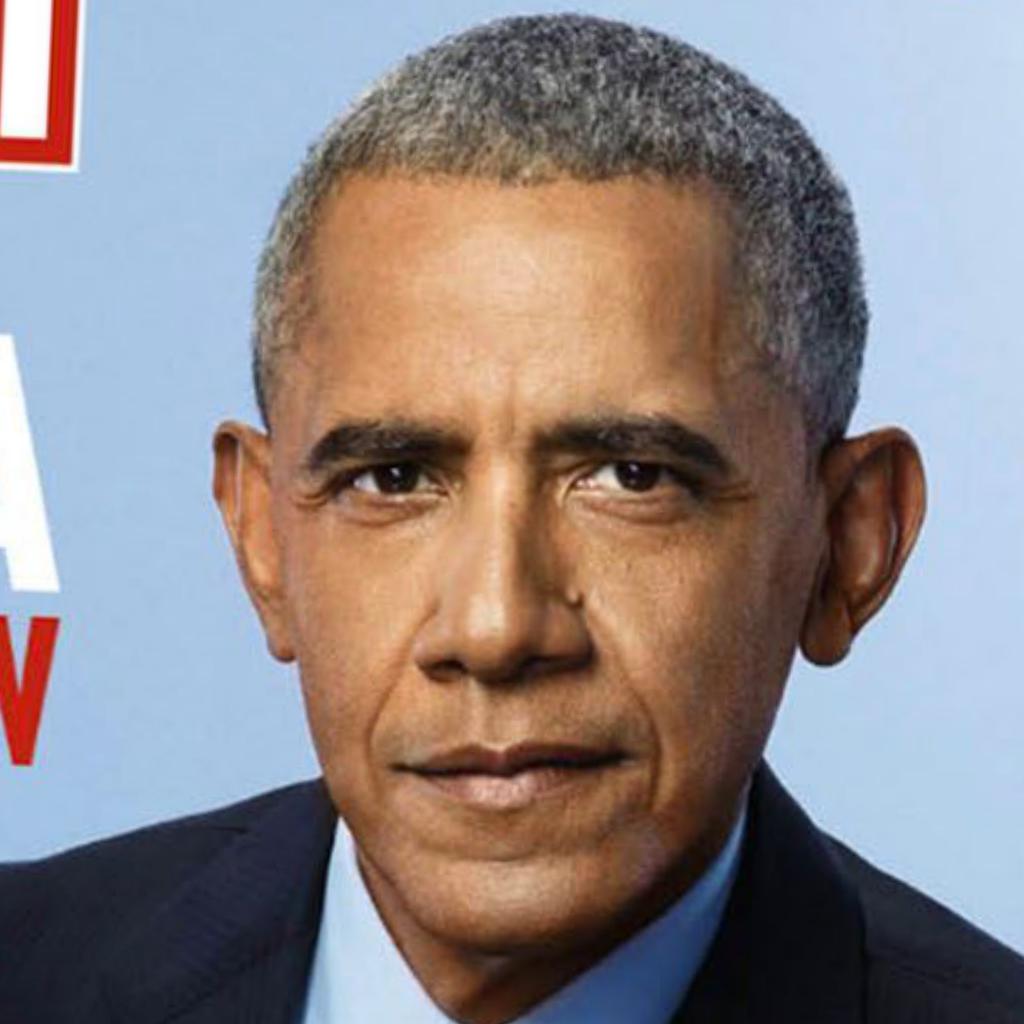}
        \end{tabular}
    \caption{Inpainting}
    \end{subfigure} 
    \qquad
    \begin{subfigure}[b]{0.3\linewidth}
        \begin{tabular}{*3c}
             Input & MyStyle & NN \\
             \includegraphics[width=\imwidth]{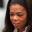} &
             \includegraphics[width=\imwidth]{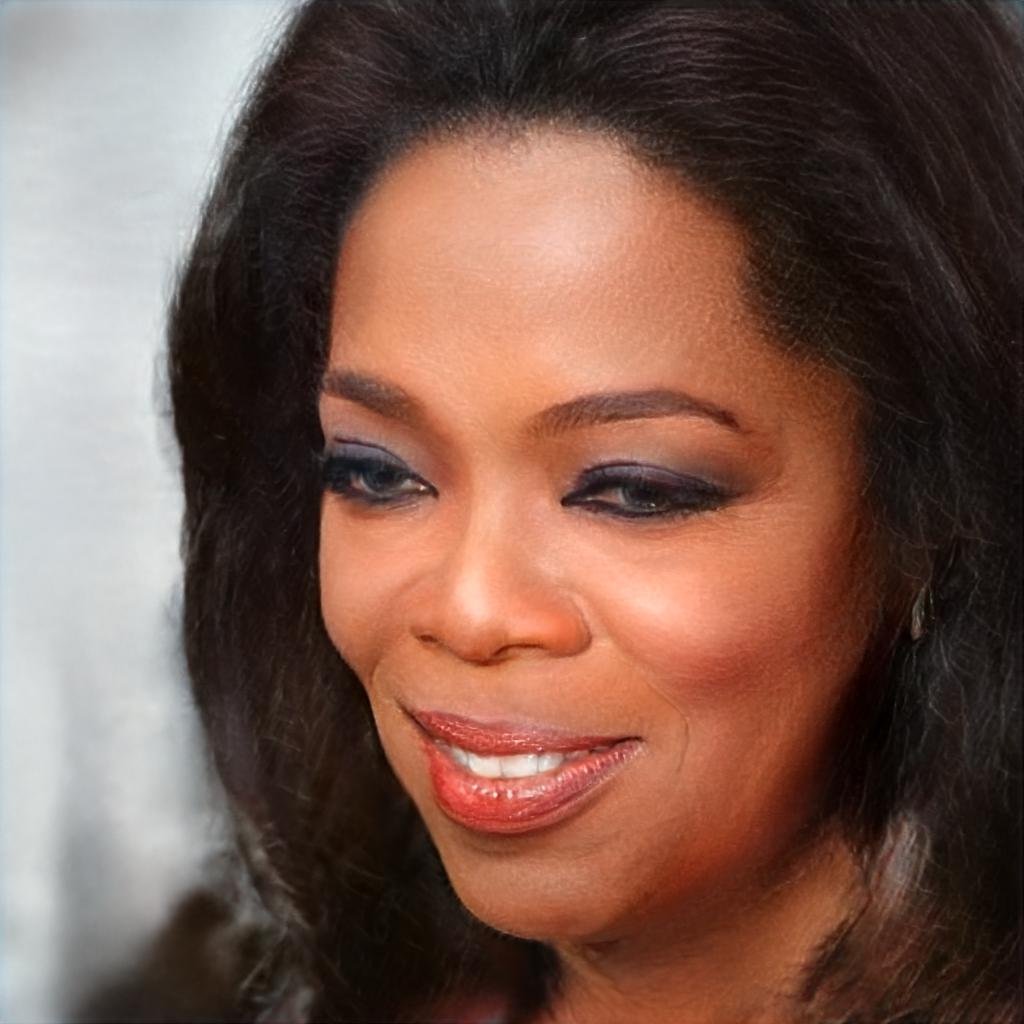} &
             \includegraphics[width=\imwidth]{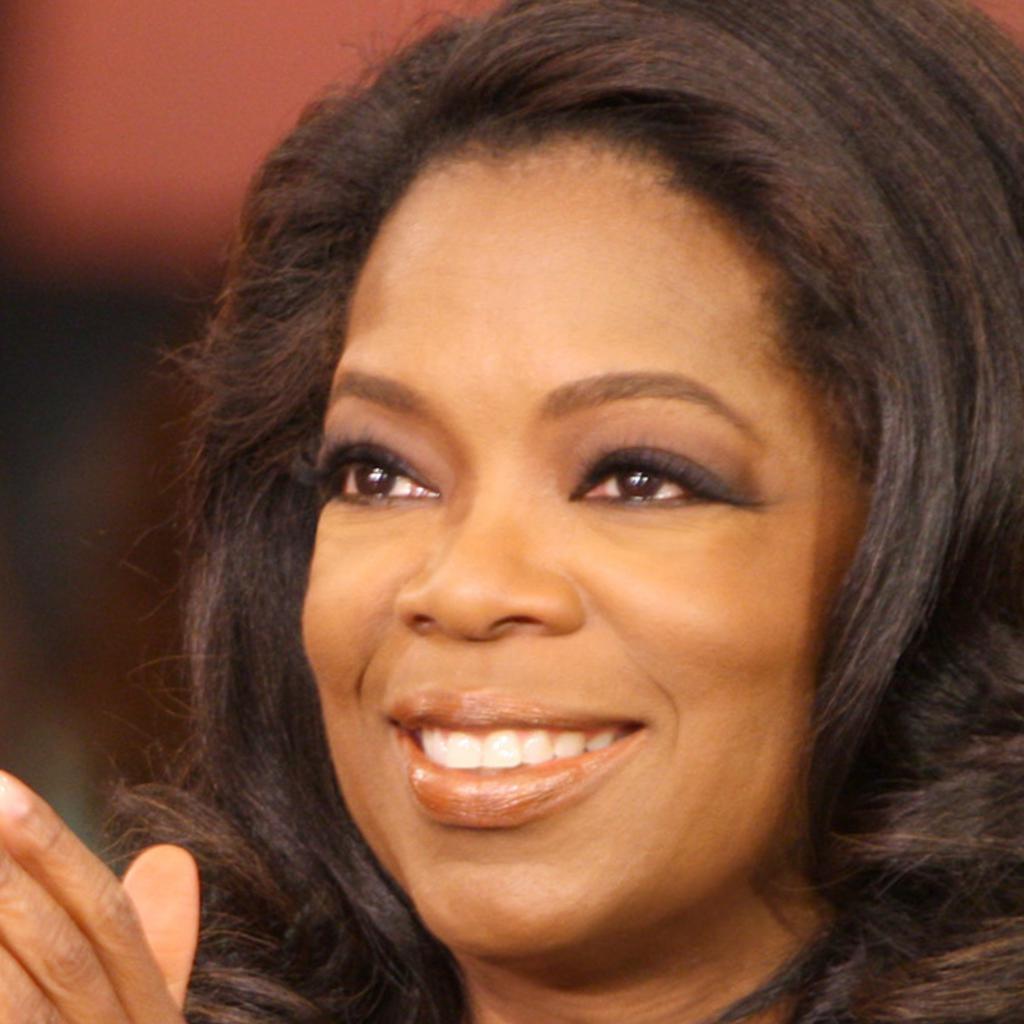} \\
             
             \includegraphics[width=\imwidth]{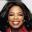} &
             \includegraphics[width=\imwidth]{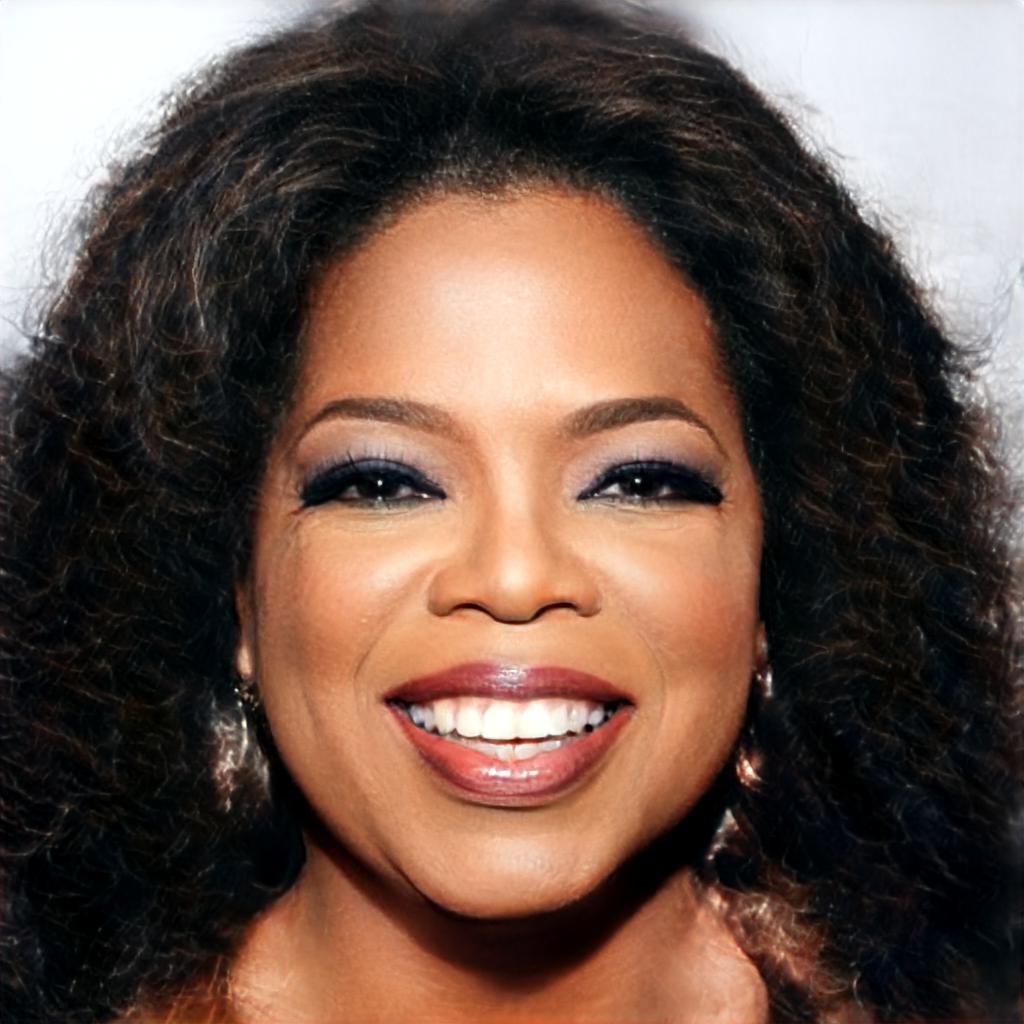} &
             \includegraphics[width=\imwidth]{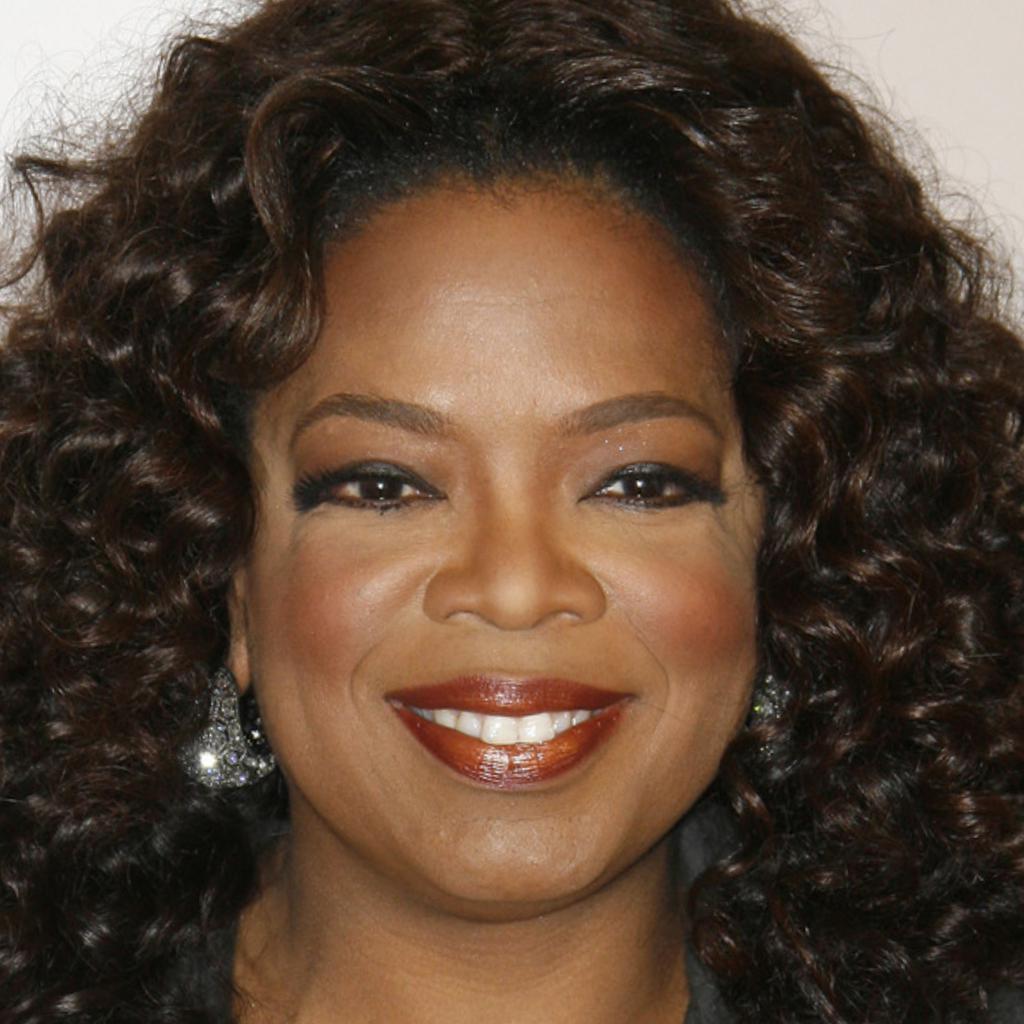} \\
             
             \includegraphics[width=\imwidth]{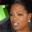} &
             \includegraphics[width=\imwidth]{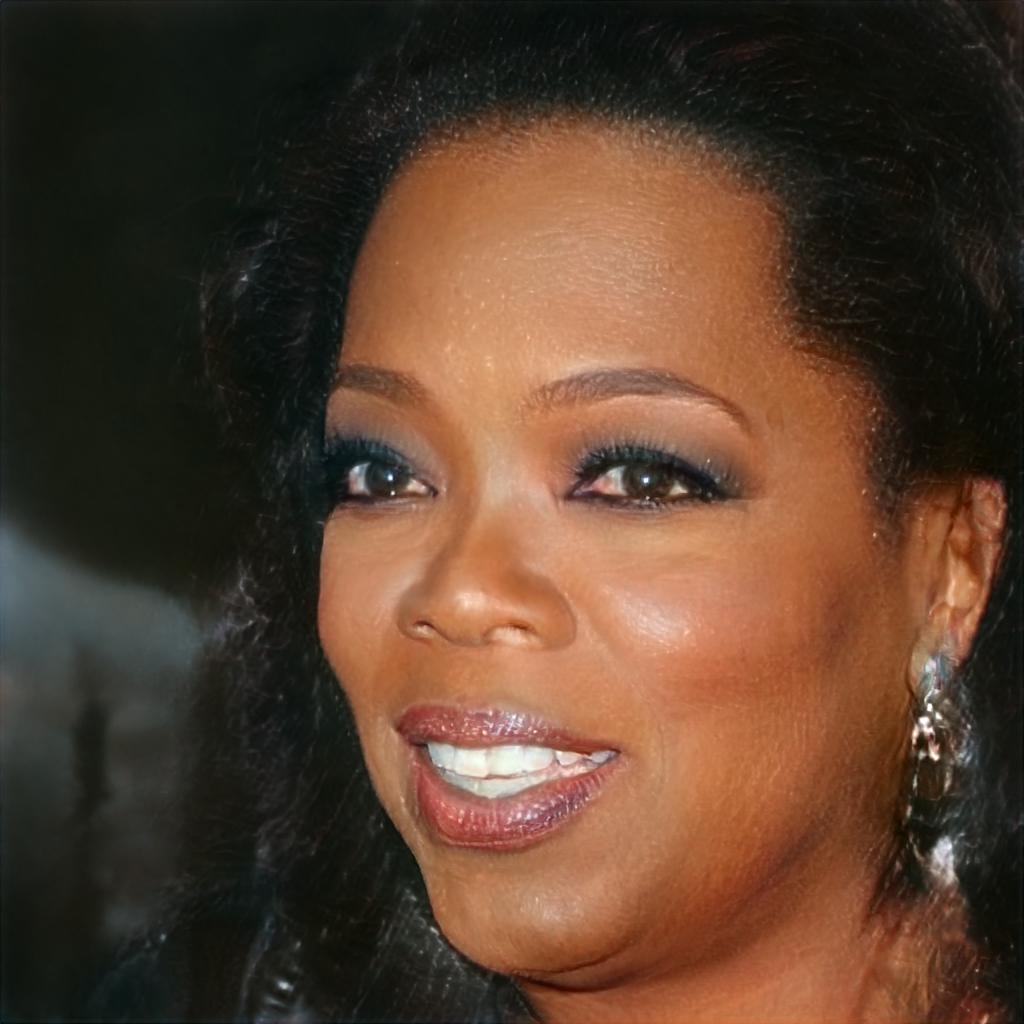} &
             \includegraphics[width=\imwidth]{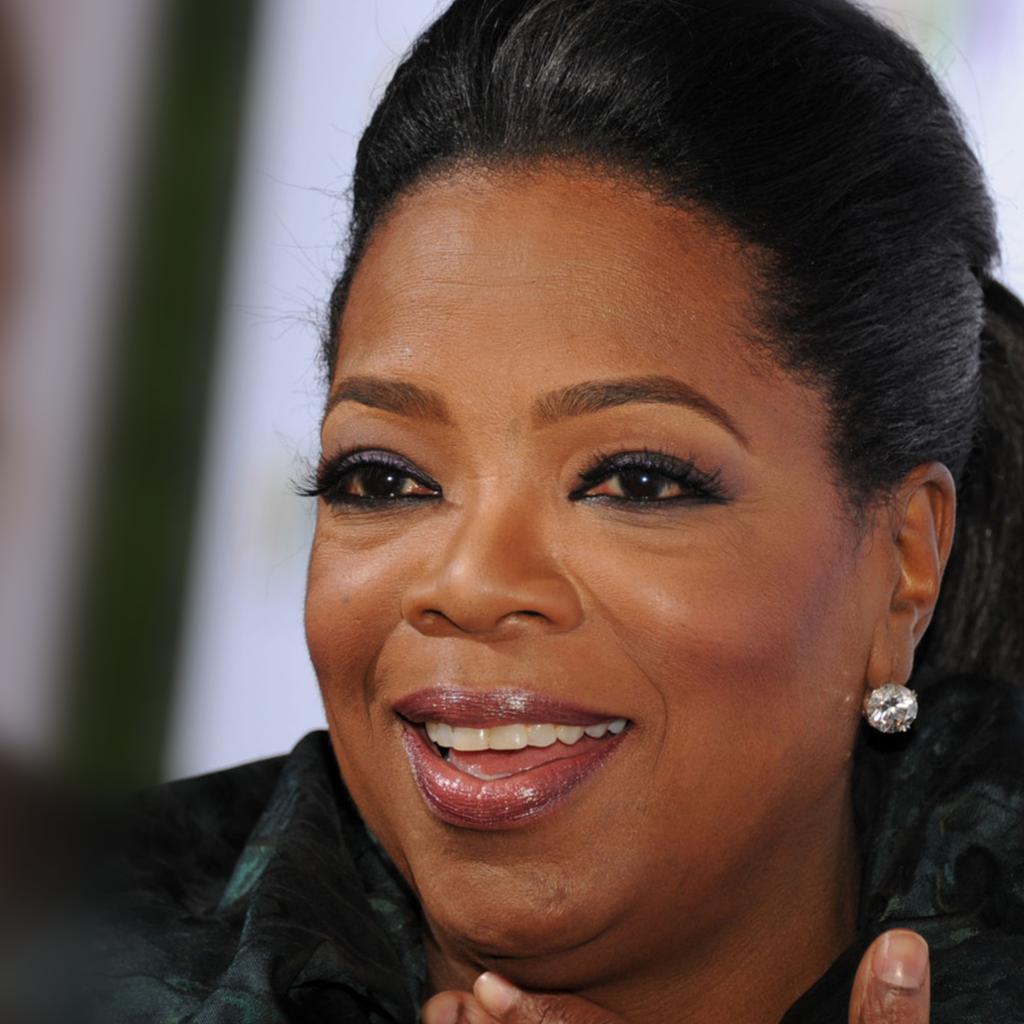} \\
             
             \includegraphics[width=\imwidth]{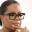} &
             \includegraphics[width=\imwidth]{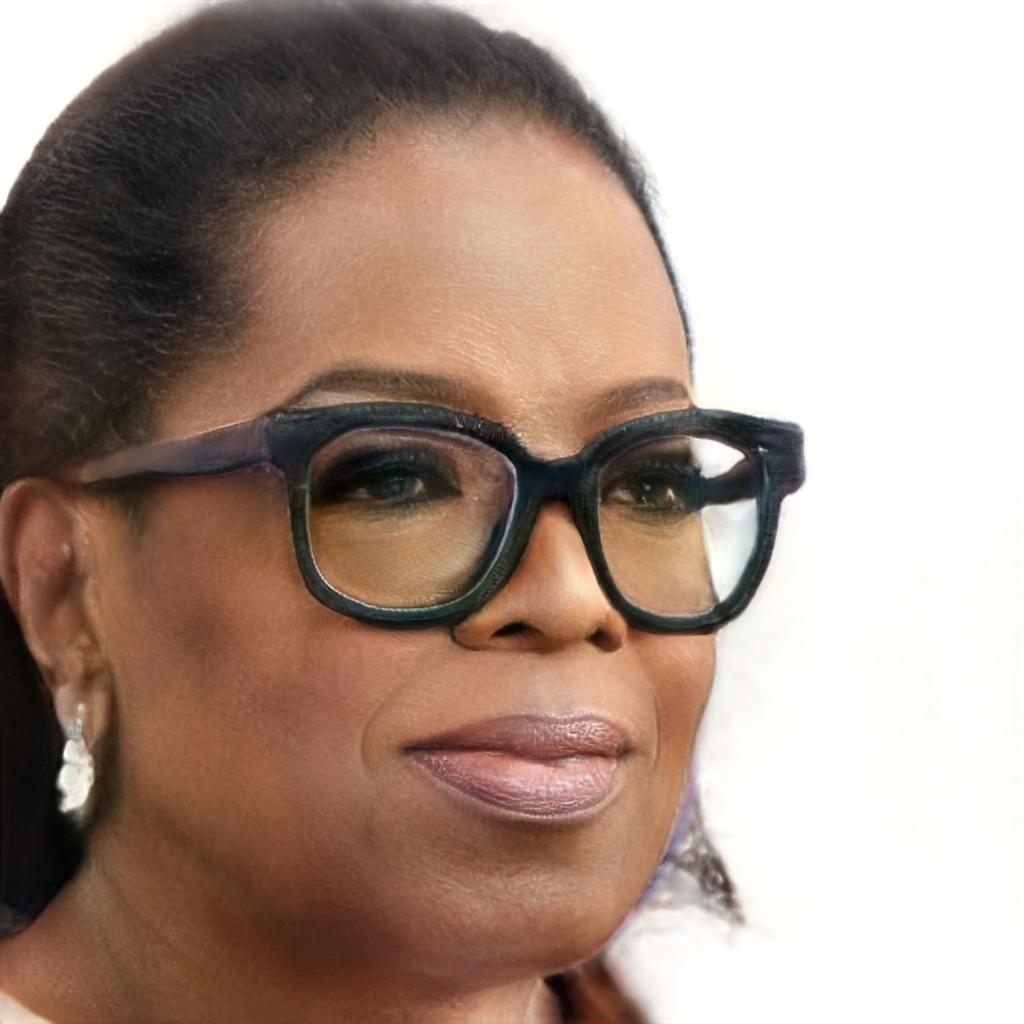} &
             \includegraphics[width=\imwidth]{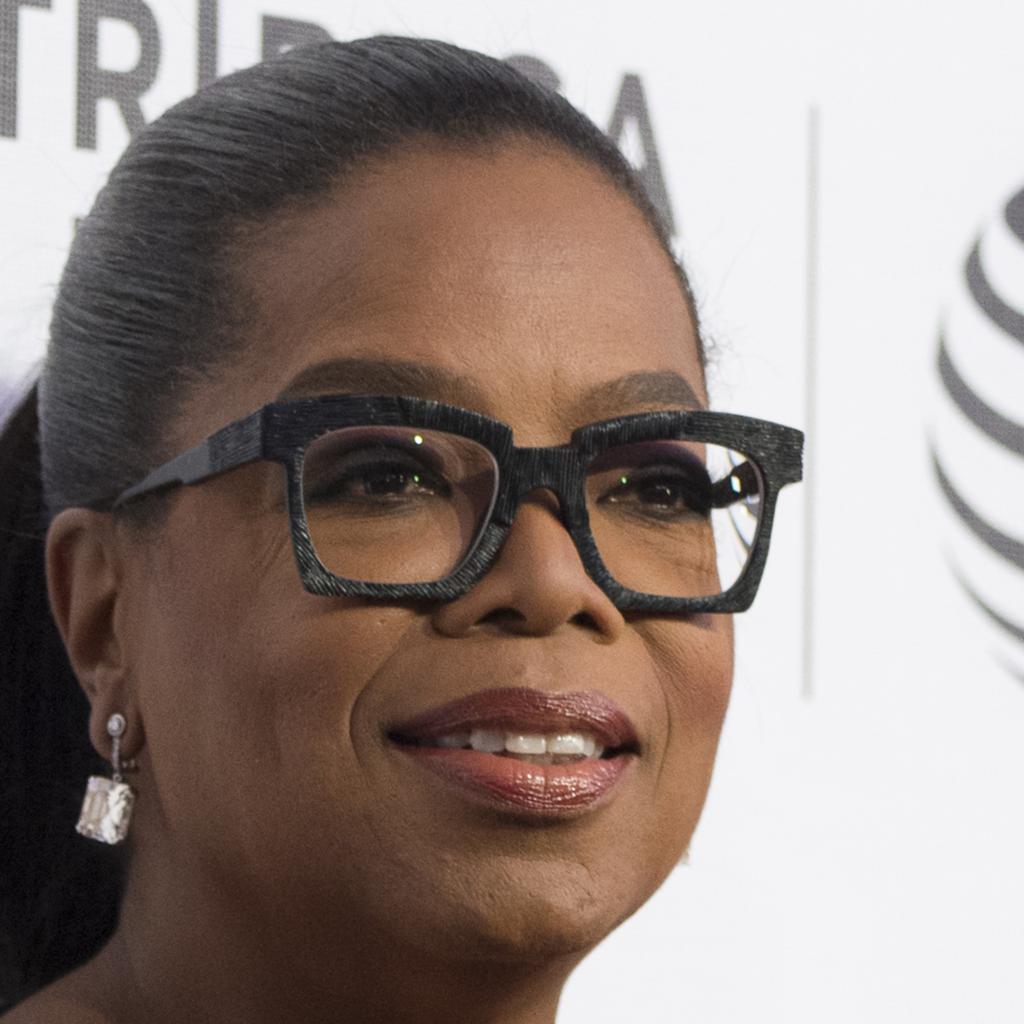}
        \end{tabular}
    \caption{Super-Resolution}
    \end{subfigure}
    
    \caption{Nearest-Neighbor experiments. We present images generated by our models for the tasks of synthesis, inpainting and super-resolution along side their respective LPIPS nearest-neighbors (marked NN). For image enhancement results, the inputs are also displayed.
    As can be seen, in no case are the results obtained by our method merely duplicates of their nearest neighbor in the reference set.}
    \label{fig:applications_nn_comp}
\end{figure*}
We next demonstrate that the outputs of our method are not merely duplicates of the reference set. In \figref{fig:applications_nn_comp}, we display inpainting, super-resolution and synthesis results along with their LPIPS \cite{zhang2018unreasonable} nearest-neighbor from the reference set. As can be seen, our result resembles the nearest-neighbor which is expected but is never a simple duplicate.

\subsection{Reference set for ID metric}
In various experiments, we used an ID metric to validate identity preservation.
The reference set used in all experiments, was the training set or a variation of it. We next explain the rationale of this aware choice and discuss an alternative.

In the context of generative models, \textit{quality} refers to being faithful to the training distribution. The identity of an individual is the most evident shared property in our training set. Therefore, quality and identity preservation are highly correlated. Following the standard practice for quality metrics (e.g FID \cite{heusel2017gans}), we choose to report identity preservation with respect to the training set. 
The only variation to the reference set is in the experiment described in \secref{sec:dch_analysis}. Since that experiment involves traversal from and to anchors, we removed the images corresponding to the anchors from the reference set. This was done in order to avoid the impression that identity preservation is only with respect to the anchors used for traversal.

\begin{figure}
    \centering
    \begin{subfigure}{\linewidth}
        \includegraphics[width=\linewidth]{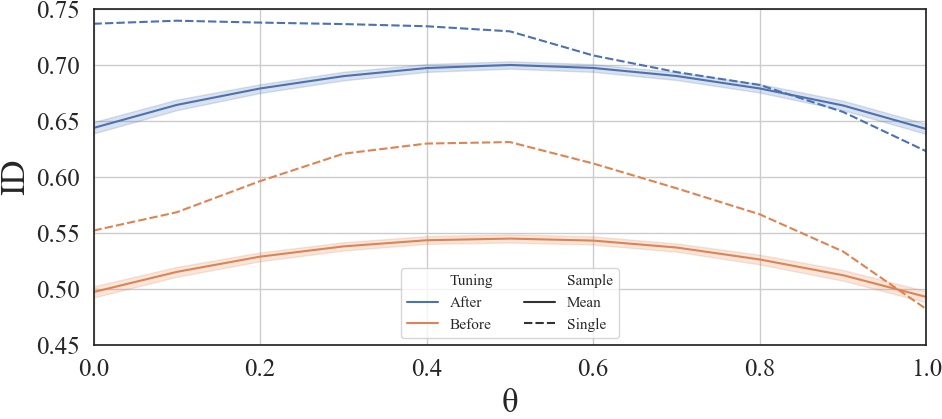}
        \caption{Identity preservation along interpolations. Compare to \figref{fig:analysis} of the main paper.}
        \label{fig:supp_ID_validation_interpolation}
    \end{subfigure}
    \\
    \vspace{3mm}
    \begin{subtable}{\linewidth}
        \centering
        \begin{tabular}{lc}
        \toprule
         Method &  ID ($\uparrow)$ \\
        \midrule
         GPEN & 0.56 $\pm$ 0.07 \\
         GPEN+FT & 0.67 $\pm$ 0.07 \\
         DiffAugment & 0.67 $\pm$ 0.05 \\
         MyStyle (Ours) & \textbf{0.75 $\pm$ 0.03} \\
        \bottomrule
        \end{tabular}
    \caption{Identity preservation for super-resolution. Compare to \figref{fig:enhancement_united_compare} of the main paper.}
    \end{subtable}
    \caption{Repeating two experiments from the main paper, but reporting ID preservation with respect to a held-out test set. As can be seen, the conclusion from the experiment does not change for either experiment.}
    \label{fig:supp_ID_validation}
\end{figure}

A possible alternative would be to report the ID metric with respect to an unseen validation set. While we believe the comparison to the training set is better motivated, we repeat several experiments using the validation set to mitigate the suspicion regarding the source of our success in identity preservation. Results are reported in \figref{fig:supp_ID_validation}. 

As can be seen, the conclusions from both experiments are still valid when compared to the reference set. Additionally, note that in \figref{fig:supp_ID_validation_interpolation}, the ID preservation of the generators before and after tuning has similarly decreased. The "before`` generator hasn't seen either of the train or test set. As our ID metric measures the maximum similarity to a set, increasing the set size leads to an increase in metric. We therefore speculate that the decrease of ID for the generator after tuning could be mostly explained by this behavior.

\subsection{Predetermined vs Trained Anchors}
When adapting the pre-trained generator $G_d$, we minimize the reconstruction loss on a set of images $\{x_i\}_{i=1}^N$ with their corresponding predetermined anchors $\{w_i\}_{i=1}^N$. We might ask, \emph{do we need to predetermine the anchors?}

\begin{figure}
	\centering
	\setlength{\tabcolsep}{1pt}
	\setlength{\imwidth}{\linewidth}
	\begin{tabular}{c}
	
	\includegraphics[width=\imwidth]{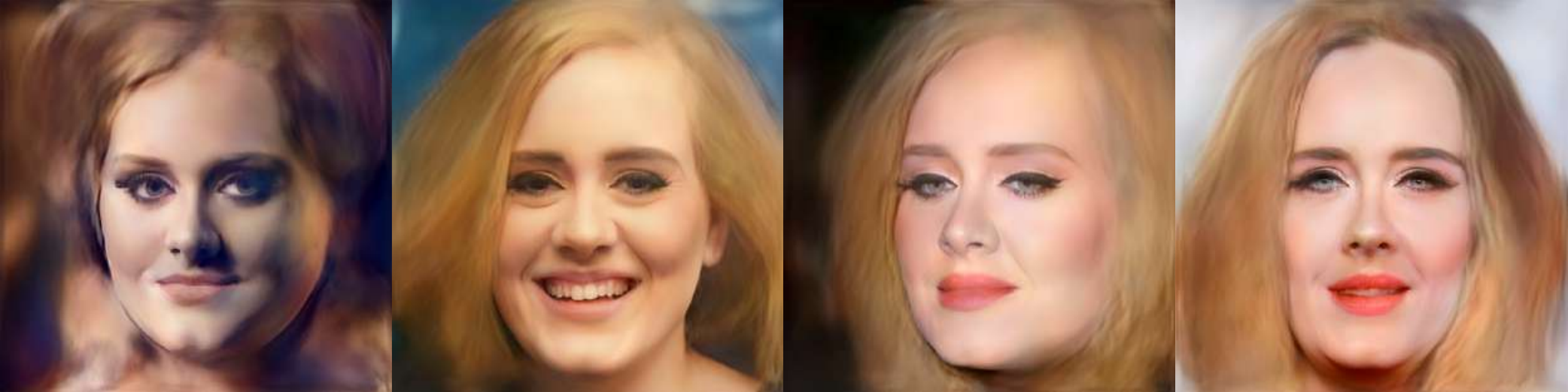} \\
	\includegraphics[width=\imwidth]{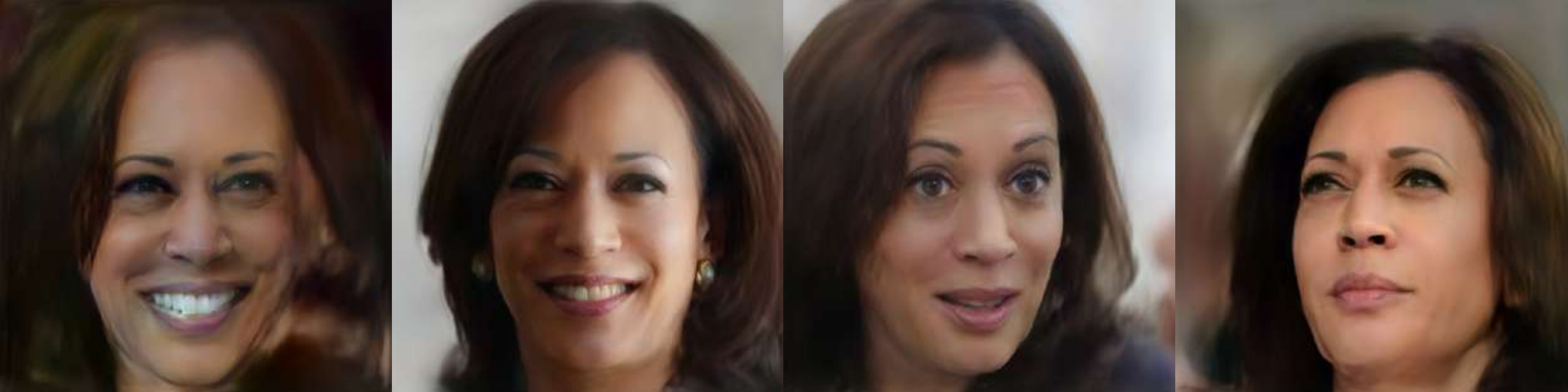}
	\\
	\includegraphics[width=\imwidth]{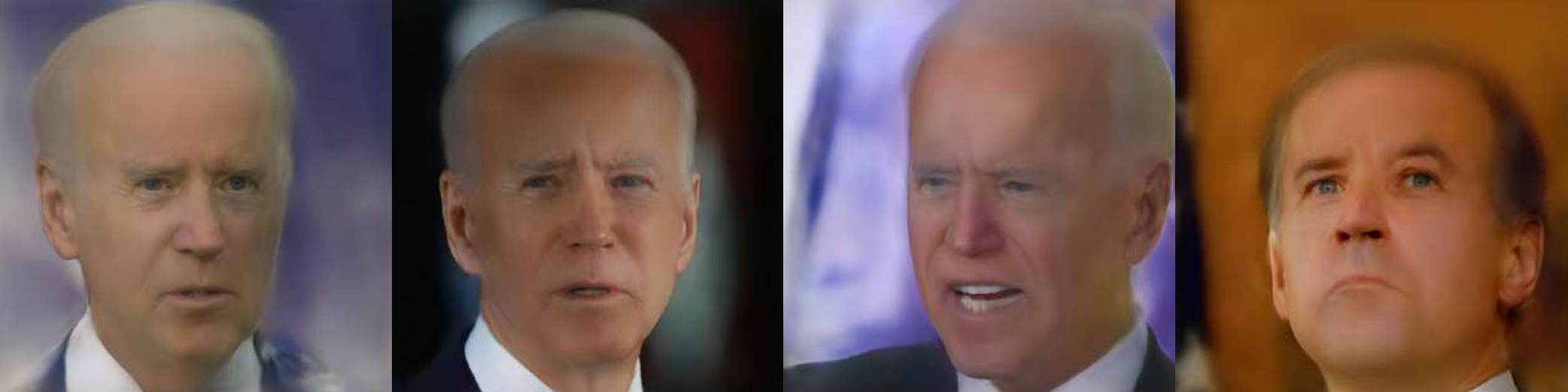}

	\end{tabular}
	\caption{Random images synthesized by adapting the generator using GLO \cite{bojanowski2017optimizing}. The synthesized images are blurry, less realistic and exhibit subtle distortions in the key-facial characteristics of the person, in contrast to our results in \figref{fig:synthesis_united_comparison} of the main paper.}
	\label{fig:glo_results}
	\vspace{-2mm}
\end{figure}

If we instead let anchors train together with the generator, the resulting process is similar to generative latent optimization (GLO) \cite{bojanowski2017optimizing}. 
We next evaluate the effect of replacing our tuning approach with GLO. 
\figref{fig:glo_results} presents random images synthesized with the obtained GLO generator for three individuals - Adele, Kamala Harris and Joe Biden. As can be seen, the synthesized images are less realistic and exhibit considerable artifacts and blurriness as well as distortions in key-facial characteristics of the person. To conclude, optimizing the anchors along with generator causes inferior results to those obtained with predetermined anchors as demonstrated in \figref{fig:synthesis_united_comparison} of the main paper.

\end{document}